\documentclass[11pt,onecolumn,journal]{IEEEtran}

\usepackage{mathtools}
\mathtoolsset{showonlyrefs}
\usepackage[below]{placeins}
\usepackage[all,cmtip]{xy}
\usepackage{a4wide}
\usepackage{amsmath, amsthm, amssymb}
\usepackage[latin1]{inputenc}
\usepackage{graphicx}
\usepackage[hang]{subfigure}
\usepackage{multirow}
\usepackage{color}
\usepackage{array}
\usepackage[ruled,linesnumbered]{algorithm2e}
\usepackage{enumerate}
\usepackage{cite}

\usepackage{setspace}

\newtheorem{theorem}{Theorem}[section]

\newtheorem{proposition}[theorem]{Proposition}
\newtheorem{lemma}[theorem]{Lemma}
\newtheorem{remark}[theorem]{Remark}
\newtheorem{example}[theorem]{Example}
\newtheorem{corollary}[theorem]{Corollary}

\providecommand{\norm}[1]{\lVert#1\rVert}
\providecommand{\abs}[1]{\lvert#1\rvert}

\def\ed{ \end{document} }

\newcommand{\ww}{27mm}
\newcommand{\hh}{20mm}
\newcommand{\hhh}{30mm}
\newcommand{\hhhh}{25mm}
\newcommand{\wwww}{30mm}

\ifCLASSINFOpdf
\else
\fi
\begin{document}
%
\title{A Three-stage Approach for Segmenting Degraded Color Images: Smoothing, Lifting and Thresholding (SLaT)}
%
%
%

\author{Xiaohao Cai,
 Raymond Chan,
 Mila Nikolova, ~\IEEEmembership{Seinor Member,~IEEE,}
 and Tieyong Zeng
\thanks{X. Cai is with the Department of Plant Sciences, and Department of Applied
 Mathematics and Theoretical Physics (DAMTP), University of Cambridge, CB2 3EA, Cambridge, UK. 
His work is partially supported by Welcome Trust and KAUST Award No. KUK-I1-007-43. Email: xc274@cam.ac.uk.}
\thanks{R. Chan is with the Department of Mathematics, The Chinese University of Hong Kong,
 Shatin, Hong Kong.  His work is partially supported by HKRGC GRF Grant No. CUHK400412, CUHK300614,
 CRF Grant No. CUHK2/CRF/11G, AoE Grant AoE/M-05/12, CUHK DAG No. 4053007, and FIS Grant No. 1907303. Email: rchan@math.cuhk.edu.hk.}
\thanks{M. Nikolova is with the CNRS, ENS Cachan, 61 Avenue du President Wilson, F-94230 Cachan, France.
	  Her work  is partially supported by HKRGC GRF Grant No. CUHK300614. Email: nikolova@cmla.ens-cachan.fr.}
\thanks{T. Zeng is with the Department of Mathematics, Hong Kong Baptist University,
 Kowloon Tong, Hong Kong.  His work is partially supported by
 NSFC 11271049, RGC 211911, 12302714 and RFGs of HKBU. Email: zeng@hkbu.edu.hk.}
}

\maketitle

\begin{abstract}

In this paper, we propose a SLaT (Smoothing, Lifting and Thresholding) method with three stages for multiphase segmentation of color images corrupted by different degradations: noise, information loss, and blur. At the first stage, a convex variant of the Mumford-Shah model is applied to each channel to obtain a smooth image. We show that the model has unique solution under the different degradations. In order to properly handle the color information, the second stage is dimension lifting where we consider a new vector-valued image composed of the restored image and its transform in the secondary color space with additional information. This ensures that even if the first color space has highly correlated channels, we can still have enough information to give good segmentation results. In the last stage, we apply multichannel thresholding to the combined vector-valued image to find the segmentation. The number of phases is only required in the last stage, so users can choose or change it all without the need of solving the previous stages again. Experiments demonstrate that our SLaT method gives excellent results in terms of segmentation quality and CPU time in comparison with other state-of-the-art segmentation methods.
\end{abstract}


%
\IEEEpeerreviewmaketitle


\section{Introduction}
%
%
%
%


\IEEEPARstart{I}MAGE segmentation is a fundamental and challenging task in image processing and computer vision. It can serve as a preliminary step for object recognition and interpretation, to mention a few.
The goal of image segmentation is to group parts of the given image with similar
characteristics together. These characteristics include, for example, edges,
intensities, colors, and textures.
For a human observer, image segmentation seems obvious, but consensus between different
observers is seldom found. The problem is much more difficult to be solved by a computer.
A nice overview of region-based and edge-based segmentation methods is given in \cite{CRD07}.
In our work we investigate the image segmentation problem for color images
corrupted by different types of degradations: noise, information loss and blur.

Let $\Omega \subset \mathbb{R}^2$ be a bounded open connected set,
and $f: \Omega \rightarrow \mathbb{R}^d$ with $d\ge 1$ be a given vector-valued image. For example, $d = 1$ for gray-scale images and $d=3$ for the usual RGB (red-green-blue) color images.
One has $d>3$ in many cases such as in hyperspectral imaging \cite{PBB09} or in medical imaging
\cite{T08}. In this paper, we are mainly concerned
with color images (i.e. $d=3$) though our approach can be extended to
higher-dimensional vector-valued images. Without loss of generality, we restrict the range of $f$ to
$[0,1]^3$ and hence $f\in L^{\infty}(\Omega)$.

In literature, various studies have been carried out and many techniques
have been considered for image segmentation \cite{shi2000malik,Arbelaez14pami,CV01,grady2006,TurboPixels2009,LNZS10,SW14,tai2007jia,zhang2011jia}. For gray-scale images, i.e. $d=1$, Mumford and Shah proposed in \cite{MS85, MS89} an energy minimization problem for image segmentation which
finds optimal piecewise smooth approximations. More precisely,
this problem was formulated in \cite{MS89} as
\begin{equation}\label{ms}
E_{{\rm MS}}(g, \Gamma):=\frac{\lambda}{2}\int_{\Omega}(f-g)^2 dx+ \frac{\mu}{2}
\int_{\Omega\setminus\Gamma}|\nabla g|^2 dx+ {\rm Length}(\Gamma),
\end{equation}
where $\lambda$ and $\mu$ are positive parameters, and
$g:\Omega\rightarrow \mathbb{R}$ is continuous in
$\Omega\setminus\Gamma$ but may be discontinuous across the sought-after boundary
$\Gamma$. Here, the length of $\Gamma$ can be written as $\mathcal{H}^{1}(\Gamma)$,
the 1-dimensional Hausdorff measure in $\mathbb{R}^2$.
Model \eqref{ms} has attractive properties even though finding a globally optimal
solution remains an open problem and it is an active area of research.
A recent overview can be found in \cite{MS_Handbook_15}.

For image segmentation,
the Chan-Vese model \cite{CV01} pioneered a simplification of functional (\ref{ms})
where $\Gamma$ partitions the image domain into two constant segments and thus
$\nabla g = 0$ on $\Omega \setminus \Gamma$. More generally, for $K$ constant
regions $\Omega_i$, $i\in\{1,\ldots,K\}$, the
multiphase piecewise constant Mumford-Shah model \cite{VC02} reads as
\begin{equation}
 E_{{\rm PCMS}}\left(\left\{\Omega_i, c_i\right\}_{i=1}^K\right)
=\frac{\lambda}{2}\sum_{i=1}^K\int_{\Omega_i}(f-c_i)^2 dx +
\frac12\sum_{i=1}^K \mathrm{Per}( \Omega_i),
\label{pcms}
\end{equation}
where $\mathrm{Per}( \Omega_i)$ is the perimeter of $\Omega_i$ in $\Omega$,
all $\Omega_i$'s are pairwise disjoint and $\Omega = \bigcup_{i=1}^K\Omega_i$.
The Chan-Vese model where $K=2$ in \eqref{pcms} has many applications
for two-phase image segmentation.
Model \eqref{pcms} is a nonconvex problem, so
the obtained solutions are in general local minimizers.
To overcome the problem,
convex relaxation approaches \cite{BEVTO07,CEN06,PCCB09, ChambolleCremersPock12},
graph cut method \cite{GA08} and fuzzy membership functions \cite{LNZS10} were proposed.

After \cite{CV01}, many approaches decompose the segmentation process into several steps and here we give a brief overview of recent work in this direction. The paper \cite{KT09} performs a simultaneous segmentation of
the input image into arbitrarily many pieces
using a modified version of model \eqref{ms} and the final segmented image results from a stopping rule using a multigrid approach. In \cite{CCBR13}, an initial hierarchy of regions is obtained by greedy iterative region merging using model \eqref{pcms};
the final segmentation is obtained by thresholding this hierarchy using hypothesis testing.
The paper \cite{BenninghoffGarcke14} first determines homogeneous regions in the noisy image with a special emphasis
on topological changes; then each region is restored using model \eqref{ms}.
Further multistage methods extending model \eqref{pcms} can be found in
\cite{TaiZhangShen13} where wavelet frames were used, and in \cite{CS13}
which is based on iterative thresholding of the minimizer of the ROF functional \cite{ROF92},
just to cite a few. In the discrete setting, the piecewise constant Mumford-Shah model \eqref{pcms}
amounts to the classical Potts model \cite{Potts52}.
The use of this kind of functionals for image segmentation was pioneered by Geman and Geman in \cite{Geman84}.
In \cite{SW14}, a coupled Potts model is used for direct partitioning of images using a convergent
minimization scheme.

In \cite{CCZ13}, a conceptually different two-stage method for the segmentation
of gray-scale images was proposed.
In the first stage, a smoothed solution $g$ is extracted from the given image $f$
by minimizing a non-tight convexification of the Mumford-Shah model (\ref{ms}).
The segmented image is obtained in the second stage by applying a thresholding technique to $g$.
This approach was extended in \cite{CYZ13} to images corrupted by Poisson and Gamma noises.
Since the basic concept of our method in this paper is similar, we will give more details
on \cite{CCZ13,CYZ13} in Section \ref{rev-2-stage}.

Extending or conceiving segmentation methods for color images is not a simple task since one needs to discriminate segments
with respect to both luminance and chrominance information. The two-phase Chan-Vese model \cite{CV01}
was generalized to deal with vector-valued images in \cite{CSV00} by combining the information in the different channels using the data fidelity term.
Many methods are applied in the usual RGB 
color space \cite{CSV00,JungKangShen07,CRD07,KT09,PCCB09,SW14,C15}, among others.
It is often mentioned that the RGB color space is not well adapted to segmentation because for real-world images the R, G and B channels can be highly correlated.
In \cite{RGZ08}, RGB images are transformed into {HSI} (hue, saturation, and intensity) color space in order to perform segmentation. In \cite{BenninghoffGarcke14} a general segmentation approach is developed for gray-value images
and further extended to color images in the RGB, the HSV (hue, saturation, and value) and the CB (chromaticity-brightness) color spaces.
However, a study on this point in \cite{Paschos01} shows that the Lab
(perceived lightness, red-green and yellow-blue) color space defined by the CIE (Commission Internationale de l'Eclairage)
is better adapted for color image segmentation than the RGB and the HSI color spaces.
In \cite{CCBR13} RGB input images are first converted to Lab space.
In \cite{WTMC15} color features are described using the Lab color space and texture using histograms in RGB space.

A careful examination of the methods that transform a given RGB image to another color space (HSI, CB, Lab, ...)
before performing the segmentation task shows that these algorithms are always applied only to noise-free
RGB images (though these images unavoidably contain quantization and compression noise).
For instance, this is the case of \cite{RGZ08,CCBR13,BenninghoffGarcke14,WTMC15}, among others.
One of the main reasons is that if the input RGB image is degraded, the degradation would be hard to
control after a transformation to another color space \cite{Paschos01}.

Our goal is to develop an image segmentation method that has the following properties:
\begin{enumerate}[\rm(a)]
\item
work on vector-valued (color) images possibly corrupted with noise, blur and missing data;
\item initialization independent and non-supervised (the number of segments is not fixed in advance);
\item avoid dealing with nonconvex and combinatorial methods for the sake of stability;
\item no need to solve the whole problem again when the number of segments required is changed;
\item take into account perceptual edges between colors and between intensities so as to detect vector-valued
objects with edges and also objects without edges.
\end{enumerate}

\begin{figure*}[!htb]
\begin{center}
\begin{tabular}{ccc}
\includegraphics[width=\wwww, height=\hhhh]{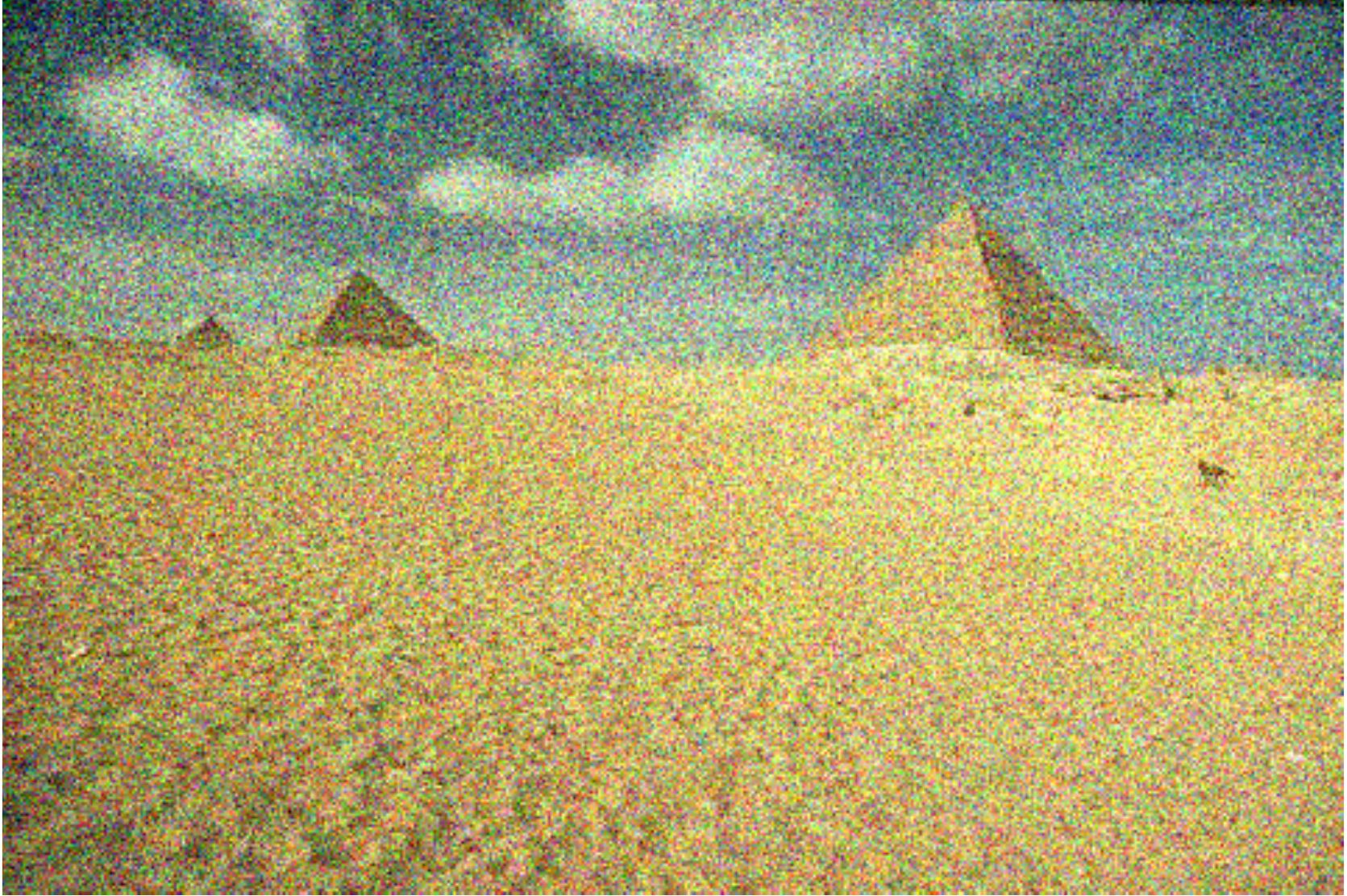} &
\includegraphics[width=\wwww, height=\hhhh]{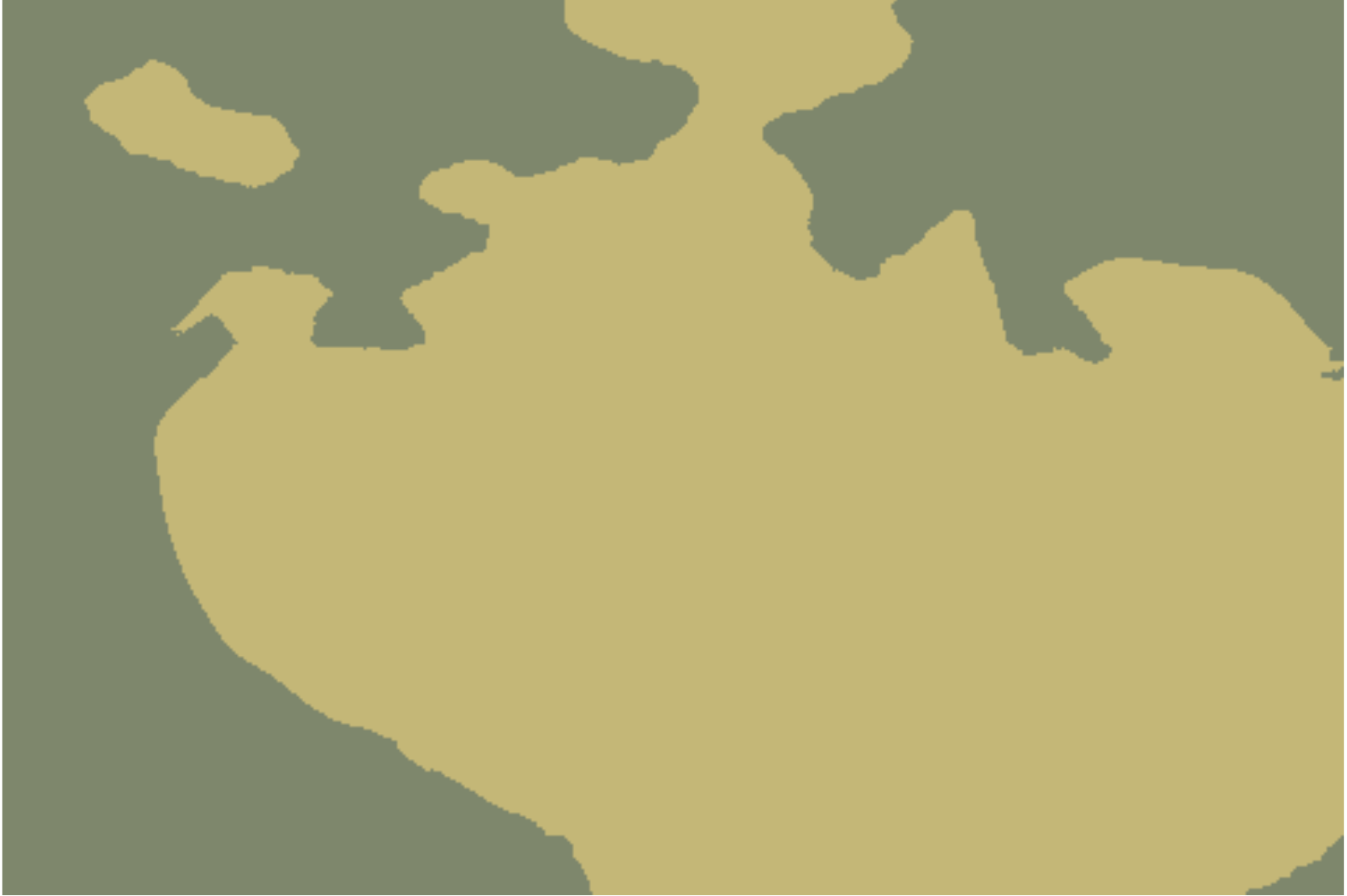} &
\includegraphics[width=\wwww, height=\hhhh]{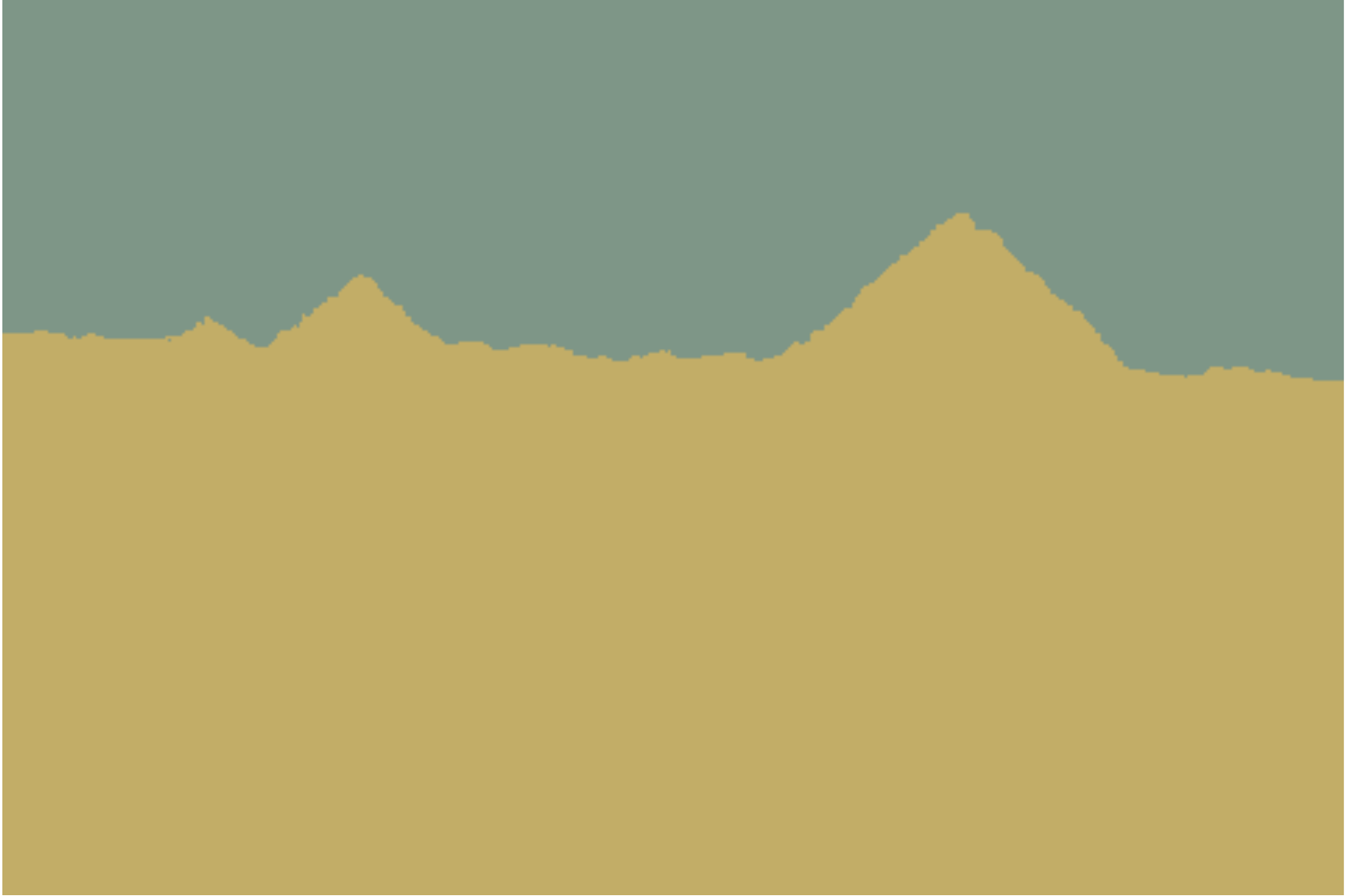} \\
(a) Given noisy image & (b) Using only RGB space &(c) Using RGB+Lab space \end{tabular}
\end{center}
\caption{Segmentation results for a noisy image (a) without the dimension lifting in Stage 2 (b)
and with Stage 2 (c).
}\label{flowers-diff-color-space-result}
\end{figure*}

{\it Contributions.} The main contribution of this paper is to propose a segmentation method having all these properties.
Goals (a)--(d) lead us to explore possible extensions of the methods \cite{CCZ13,CYZ13} to vector-valued (color) images.
Goal (e) requires finding a way to use information from perceptual color spaces
even though our input images are corrupted; see goal (a).
Let ${\cal V}_1$ and ${\cal V}_2$ be two color spaces.
Our method has the following 3 steps:
\begin{enumerate}
\item[1)]
Let the given degraded image be in ${\cal V}_1$.
The convex variational model \cite{CCZ13,CYZ13} is applied in parallel to each channel of ${\cal V}_1$. This yields a restored smooth image.
We show that the model has unique solution.
\item[2)]
The second stage consists of color dimension lifting: we transform the smooth color image obtained at Stage 1
to a secondary color space ${\cal V}_2$ that provides us with complementary information.
Then we combine these images as a
new vector-valued image composed of all the channels from color spaces
${\cal V}_1$ and ${\cal V}_2$.
\item[3)]
According to the desired number of phases $K$, we
apply a multichannel thresholding to the combined ${\cal V}_1$-${\cal V}_2$ image to
obtain a segmented image.
\end{enumerate}
We call our method ``SLaT" for Smoothing, Lifting and Thresholding.
Unlike the methods that perform segmentation in a different color space like
\cite{MVP03,RGZ08,CCBR13,BenninghoffGarcke14,WTMC15}, we can deal with degraded images
thanks to Stage 1 which yields a smooth image that we can transform to another color space.
We will fix ${\cal V}_1$ to be the RGB color space since one usually has RGB color images.
We use the Lab color space \cite{Gevers12} as the secondary color space ${\cal V}_2$ since it
 is often recommended for color segmentation
\cite{CRD07,CCBR13,Paschos01}.
The crucial importance of the dimension lifting Stage 2 is illustrated in Fig. \ref{flowers-diff-color-space-result}
which shows the results without Stage 2, i.e. ${\cal V}_2=\emptyset$ (middle) or with Stage 2 (right).
To the best of our knowledge, it is the first time that two color spaces
are used jointly in variational methods for segmentation.
This provides us with additional information
on the color image so that in all cases we can obtain very good segmentation results.
The number of phases $K$ is needed only in Stage 3.
Its value can reasonably be selected based on the RGB image obtained at Stage 1.

Extensive numerical tests on synthetic and real-world images have shown that
our method outperforms state-of-the-art variational segmentation methods like
\cite{LNZS10,PCCB09,SW14} in terms of
segmentation quality, speed and parallelism of the algorithm,
and the ability to segment images corrupted by different kind of degradations.

{\it Outline.}
 In Section \ref{rev-2-stage},
we briefly review the models in \cite{CCZ13,CYZ13}. Our SLaT segmentation method
is presented in Section \ref{sec:model}.
In Section \ref{sec:experiments}, we provide
experimental results on synthetic and real-world images.
Conclusion remarks are given in Section~\ref{sec:conclusions}.

\section{Review of the Two-stage Segmentation Methods in \cite{CCZ13,CYZ13}}
\label{rev-2-stage}
The methods in \cite{CCZ13,CYZ13} for the segmentation of gray-scale images
are motivated by the observation that one can obtain a good segmentation by properly thresholding
a smooth approximation of the given image. Thus in their first stage,  these methods  solve
a minimization problem of the form
\begin{equation}
\label{CaiModel}
	\inf_{g\in W^{1,2}(\Omega)} \left\{\frac{\lambda}{2}\int_\Omega \Phi(f, g) dx 	
	+\frac{\mu}{2}\int_\Omega \abs{\nabla g}^2 dx + \int_\Omega \abs{\nabla g}dx	 \right\},
\end{equation}
where  $\Phi(f, g)$ is the data fidelity term, $\mu$ and $\lambda$ are positive
parameters. We note that the model (\ref{CaiModel}) is a convex non-tight
relaxation of the Mumford-Shah model in \eqref{ms}.
Paper \cite{CCZ13} considers
$\Phi(f, g) = (f-{\cal A} g)^2$ where $\mathcal{A}$ is a given blurring operator;
when $f$ is degraded by Poisson or Gamma noise,
the statistically justified choice $\Phi(f, g) = {\cal A}g - f \log( {\cal A}g)$ is used in \cite{CYZ13}.
Under a weak assumption, the functional in \eqref{CaiModel} has a unique minimizer,
say $\bar{g}$, which is a smooth approximation of $f$.
The second stage is to use
the K-means algorithm \cite{KMNPSW02} to determine the thresholds
for segmentation.

These methods
have important advantages: they can
segment degraded images
and the minimizer $\bar g$ is unique.
Further, the segmentation stage being independent from the optimization problem  (\ref{CaiModel}),
one can change the number of phases $K$ without solving (\ref{CaiModel}) again.

\section{SLaT: Our Segmentation Method for Color Images}\label{sec:model}
Let $f = (f_1, \ldots, f_d )$ be a given color image with channels
$f_i: \Omega \rightarrow \mathbb{R}, i=1, \cdots, d$.
For $f$ an RGB image, $d=3$.
This given image $f$ is typically a blurred and noisy version of
an original unknown image. It can also be incomplete: we
denote by $\Omega_0^i$ the open nonempty subset of $\Omega$
where the given $f_i$ is known for channel $i$. Our SLaT segmentation
method consists of three stages described next.

\subsection{First Stage: Recovery of a Smooth Image }
First, we restore each channel $f_i$ of $f$ separately by minimizing the functional $E$ below
\begin{equation}\label{model-1st-extend-n}
E(g_i)=\frac{\lambda}{2} \int_{\Omega}\omega_i\cdot \Phi(f_i, g_i) dx
+ \frac{\mu}{2} \int_{\Omega}|\nabla g_i|^2 dx
+  \int_{\Omega} |\nabla g_i| dx, \quad i=1, \ldots, d,
\end{equation}
where
$|\cdot| $ stands for Euclidian norm and
 $\omega_i(\cdot)$ is the characteristic function of $\Omega_0^i$, i.e.
\begin{equation} \label{omega}
\omega_i(x) =
\begin{cases}
1, & x \in \Omega_0^i, \\
0, & x \in \Omega\setminus \Omega_0^i.
\end{cases}
\end{equation}
For $\Phi$ in \eqref{model-1st-extend-n} we consider the following options:
\begin{enumerate}
 \item\label{L2} $\Phi(f, g) = (f-{\cal A} g)^2$, the usual choice;
 \item\label{Poi} $\Phi(f, g) = {\cal A}g - f \log ({\cal A}g)$ if data are corrupted by Poisson noise.
\end{enumerate}
The restoration step amounts to the first stage in \cite{CCZ13,CYZ13}
in the simple case when $d=1$
and $\Omega_0^1= \Omega$. Theorem \ref{thm-eu} below proves the existence and the uniqueness of
the minimizer of \eqref{model-1st-extend-n}. In view of \eqref{model-1st-extend-n}
and \eqref{omega}, we define the linear operator $(\omega_i{\cal A})$ by
\begin{equation}\label{no}
(\omega_i {\cal A}) : u(x) \in L^2(\Omega) \mapsto
\omega_i(x) ({\cal A}u)(x) \in L^2(\Omega).
\end{equation}
\begin{theorem} \label{thm-eu}
Let $\Omega$ be a bounded connected open subset of $\mathbb{R}^2$ with a Lipschitz boundary.
Let ${\cal A}:L^2(\Omega)\rightarrow L^2(\Omega)$ be bounded and linear.
For $i\in\{1,\ldots,d\}$, assume that $f_i \in L^2(\Omega)$ and that ${\rm Ker} (\omega_i {\cal A} )\bigcap {\rm Ker} (\nabla) = \{0 \}$
where ${\rm Ker} $ stands for null-space. Then (\ref{model-1st-extend-n}) with either
$\Phi(f_i, g_i) = (f_i-{\cal A} g_i)^2$ or $\Phi(f_i, g_i) = {\cal A}g_i - f_i \log ({\cal A}g_i)$
has a unique minimizer $\bar g_i \in W^{1,2}(\Omega)$.
\end{theorem}

The proof is outlined in Appendix I.
The condition ${\rm Ker} (\omega_i {\cal A} )\bigcap {\rm Ker} (\nabla) = \{0 \}$
is mild---it means that ${\rm Ker} (\omega_i {\cal A} )$ does not contain constant images.

{\it The discrete model.}
In the discrete setting, $\Omega$ is an array of pixels, say of size $M\times N$, and our model \eqref{model-1st-extend-n} reads as
\begin{equation}\label{x}
 E(g_i)=\frac{\lambda}{2} \Psi(f_i, g_i) +\frac{\mu}{2}\|\nabla g_i\|_F^2+\|\nabla g_i\|_{2,1},
 \quad i=1, \ldots, d.
\end{equation}
Here
\begin{equation*}
\Psi(f_i,g_i):=\sum_{j\in\Omega} \left(\omega_i \cdot (f_i-{\cal A} g_i)^2\right)_j \ {\rm or} \
\Psi(f_i,g_i):=\sum_{j\in\Omega} \left(\omega_i \cdot \big( {\cal A}g_i - f_i \log ({\cal A}g_i)\big)\right)_j.
\end{equation*}

The operator $\nabla =(\nabla _x,\nabla_y )$ is discretized using backward differences with Neumann boundary conditions. Further,
$\| \cdot \|_F^2$ is the Frobenius norm, so 
$$
\|\nabla g_i\|_F^2 = \sum_{j\in\Omega}\big((\nabla_x g_i)_j^2+(\nabla_y g_i)_j^2\big),
$$
and $\|\nabla g_i\|_{2,1}$ is the usual discretization of the TV semi-norm given by
\[ 
\|\nabla g_i\|_{2,1} =\sum_{j\in\Omega}\sqrt{(\nabla_x g_i)_j^2+(\nabla_y g_i)_j^2}.\]
For each $i$, the minimizer $\bar{g}_i$ can be computed easily using different methods,
for example the primal-dual algorithm \cite{CP11,CLO13}, alternating direction method with multipliers (ADMM) \cite{BPCPE10}, or the split-Bregman algorithm \cite{GO09}.
Then we rescale each $\bar g_i$ onto $[0,1]$ to obtain
$\{\bar g_i\}_{i=1}^d\in[0,1]^d$.

\subsection{Second Stage: Dimension Lifting with Secondary Color Space}\label{dim}

\begin{figure*}[!htb]
\begin{center}
\begin{tabular}{ccc}
\includegraphics[width=\wwww, height=\hhhh]{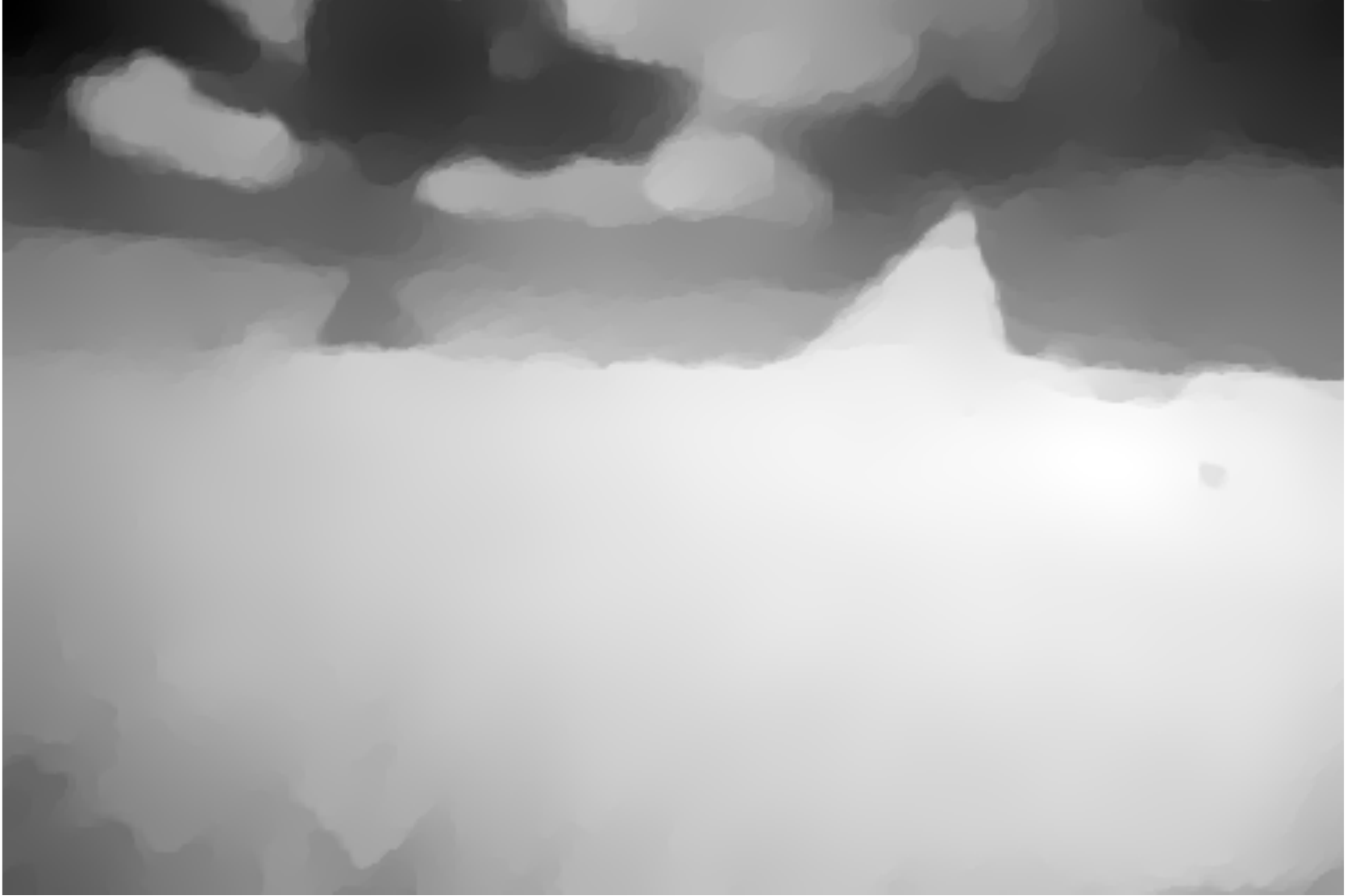} &
\includegraphics[width=\wwww, height=\hhhh]{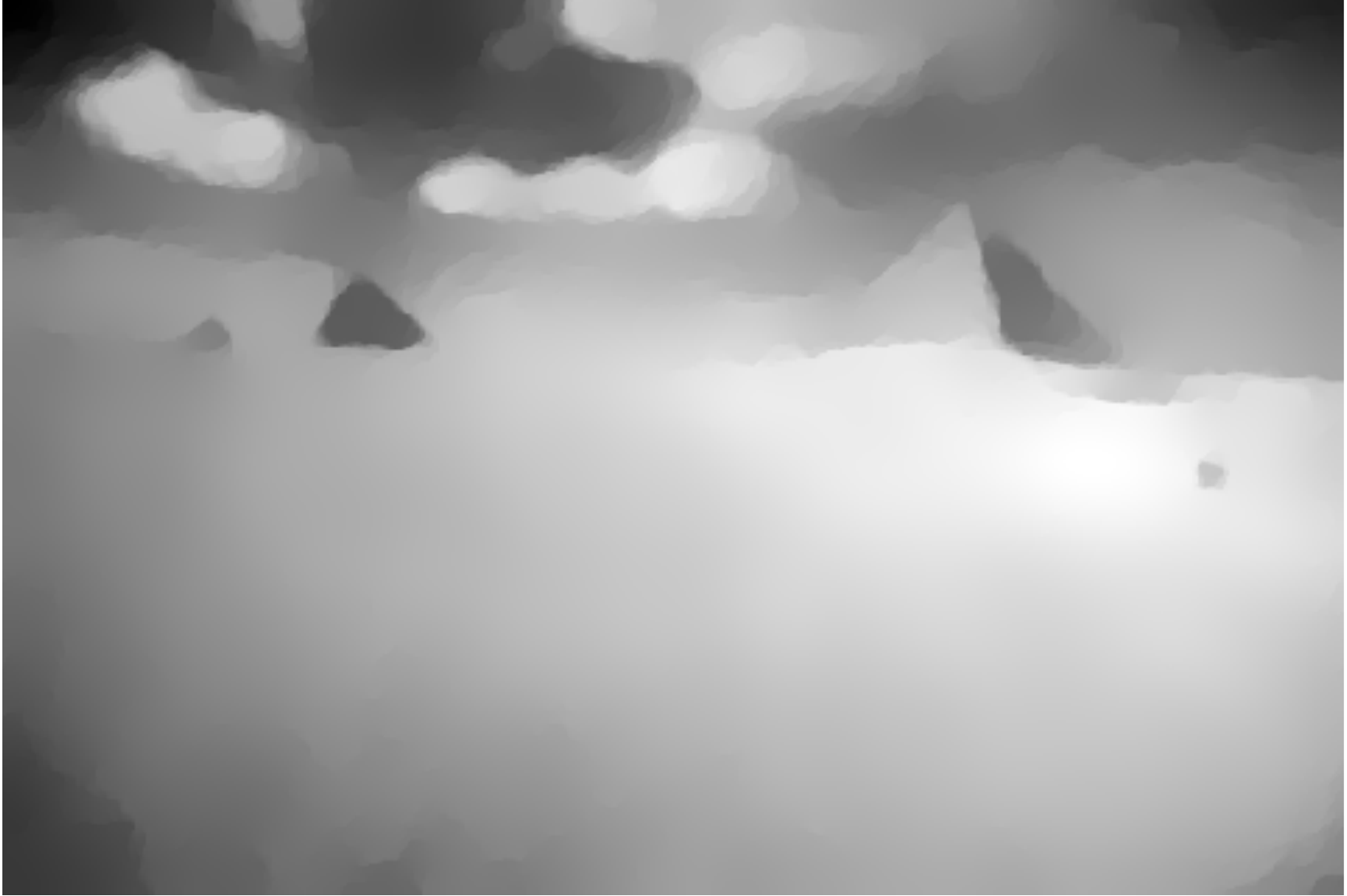} &
\includegraphics[width=\wwww, height=\hhhh]{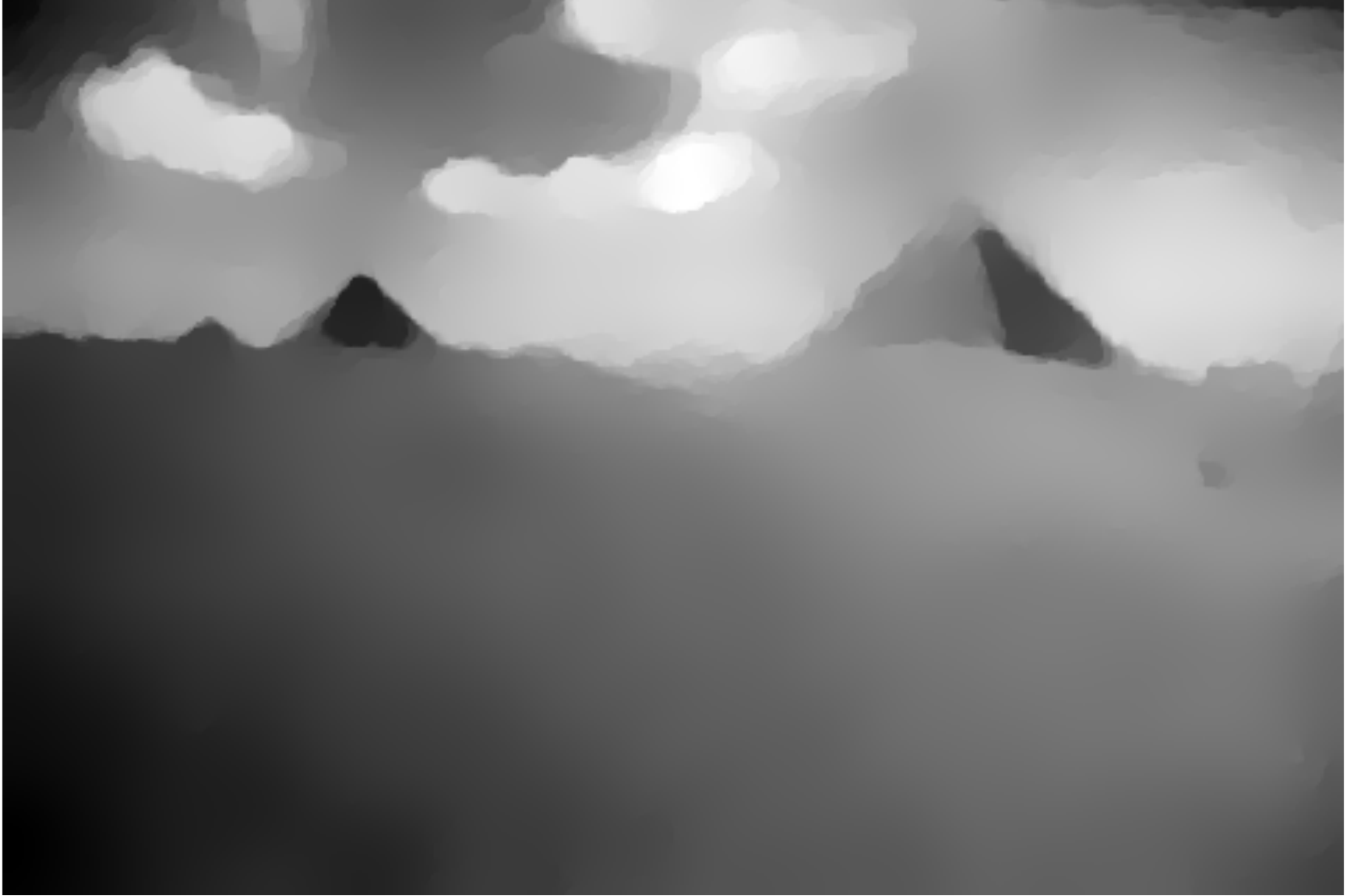} \\
(a) {R} channel $\bar g_1$ &(b) { G} channel $\bar g_2$ & (c) { B} channel $\bar g_3$ \\
 \includegraphics[width=\wwww, height=\hhhh]{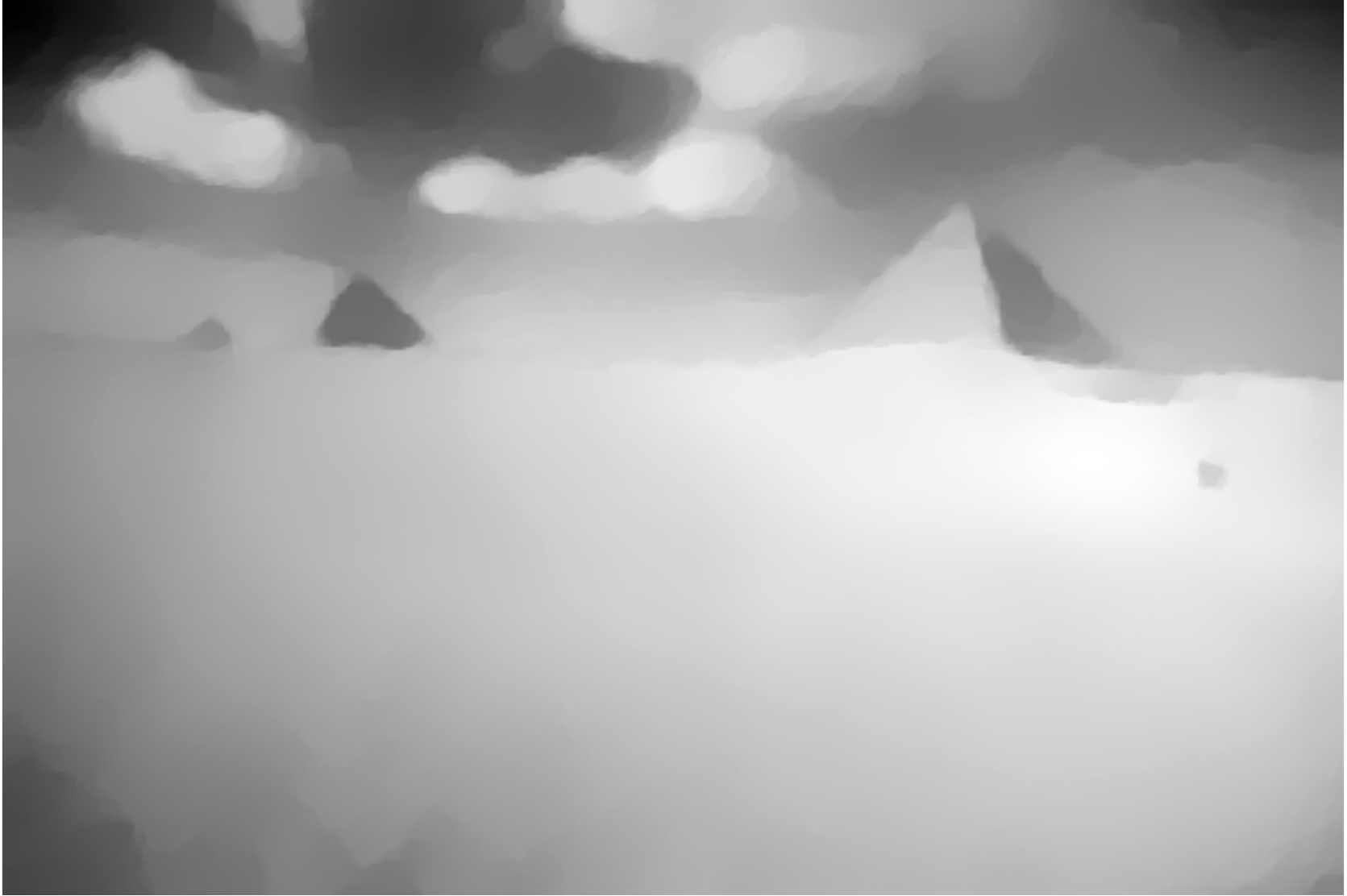} &
\includegraphics[width=\wwww, height=\hhhh]{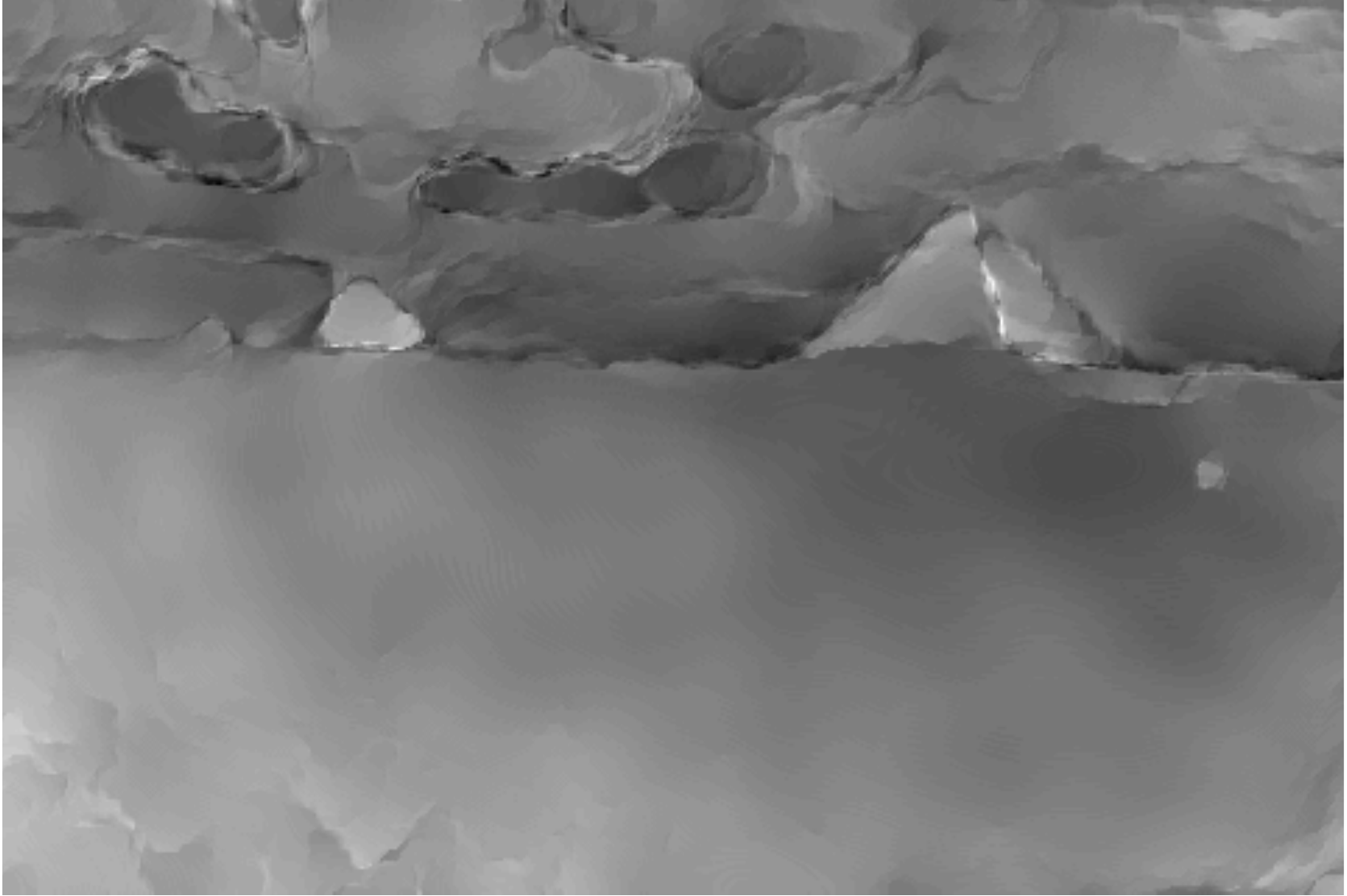} &
\includegraphics[width=\wwww, height=\hhhh]{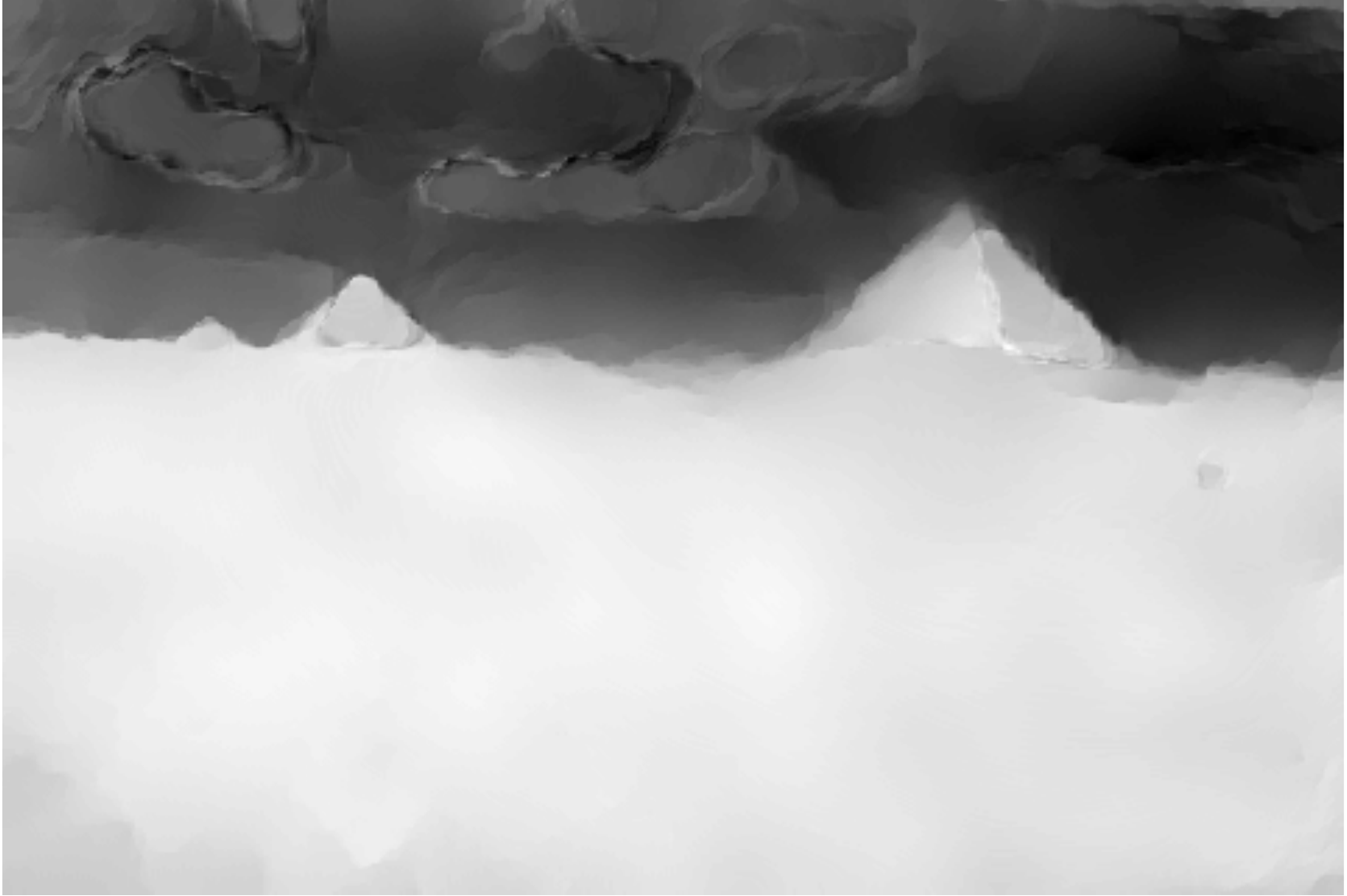} \\
 (d) {L} channel $\bar g_1^t$ & (e) {a} channel $\bar g_2^t$ & (f) { b} channel $\bar g_3^t$
\end{tabular}
\end{center}
\caption{Channels comparison 
for the restored (smoothed) $\bar g$ in Stage 1 used in
Fig. \ref{flowers-diff-color-space-result}.
(a)--(c): the {R}, {G} and {B} channels of $\bar g$;
 (d)--(f): the {L}, {a} and { b} channels of $\bar g^t$ -- the
Lab transform of $\bar g$. Both $\bar g$ and $\bar g^t$ were used to obtain the
result in Fig. \ref{flowers-diff-color-space-result} (c).
}\label{color-space-rgb-lab}
\end{figure*}

For the ease of presentation, in the following, we assume ${\cal V}_1$
 is the RGB color space.
The goal in color segmentation is to recover segments
both in the luminance and in the chromaticity of the image.
It is well known that the R, G and B channels can be highly correlated.
For instance, the R, G and B channels of the output of Stage 1 for the noisy pyramid image in Fig.~\ref{flowers-diff-color-space-result} are depicted in Fig.~\ref{color-space-rgb-lab} (a)--(c).
One can hardly expect to make a meaningful segmentation based on these channels---see the result in Fig.~\ref{flowers-diff-color-space-result} (b), as well as Fig.~\ref{twophase-color-pyramid} where
other contemporary methods are compared.

Stage 1 provides us with a restored smooth image $\bar{g}$.
In Stage 2, we perform dimension lifting
in order to acquire additional information on $\bar g$ from a different color space
that will help the segmentation in Stage 3. The choice is delicate.
Popular choices of less-correlated color spaces include HSV, HSI, CB and Lab, as
described in the Introduction.
The Lab color space was created by the CIE
with the aim to be perceptually
uniform \cite{Gevers12} in the sense that the numerical difference between two colors is
proportional to perceived color difference.
This is an important property for color image segmentation, see e.g. \cite{CRD07,CCBR13,Paschos01}.
For this reason in the following we use the Lab as the additional color space.
Here the L channel correlates to perceived lightness, while the a and b channels correlate
approximately with red-green and yellow-blue, respectively.
As an example we show in Fig.~\ref{color-space-rgb-lab} (d)--(f) the L, a and b channels
of the smooth $\bar{g}$ in Stage 1 for the noisy pyramid in Fig. \ref{flowers-diff-color-space-result} (a).
The result in Fig. \ref{flowers-diff-color-space-result} (c) have shown that this additional color
space helps the segmentation significantly.

Let $\bar g'$ denote Lab transform of $\bar g$.
In order to compare $\bar g'$ with $\bar g\in[0,1]^3$, we
rescale on [0,1]
the channels of $\bar g'$ which yields an image denoted by $\bar g^t\in[0,1]^3$.
By stacking together $\bar g$ and $\bar g^t$ we obtain a new vector-valued image
$\bar g^*$ with $2d=6$ channels: 
\begin{equation} \nonumber
\bar{g}^*: = (\bar{g}_1, \bar{g}_2, \bar{g}_3, \bar{g}^t_1, \bar{g}^t_2, \bar{g}^t_3).
\end{equation}
Our segmentation in Stage 3 is done on this $\bar{g}^*$.

\subsection{Third Stage: Thresholding}\label{third}
Given the vector-valued image $\bar{g}^*\in[0,1]^{2d} $ for $d=3$ from Stage 2 we want now
to segment it into $K$ segments.
Here we design a properly adapted strategy to partition vector-valued images into $K$ segments.
It is based on
the $K$-means algorithm \cite{KMNPSW02} because of its simplicity and good asymptotic properties. According to the value of $K$, the algorithm clusters all points of $\{\bar g^*(x)~:~x\in\Omega\}$ into $K$ Voronoi-shaped cells, say $ {\Sigma}_1 \cup {\Sigma}_2 \cdots \cup {\Sigma}_K=\Omega $.
Then we compute the mean vector $c_k\in\mathbb{R}^{6}$ on each cell ${\Sigma}_k$ by
\begin{equation}\label{ck}
{c}_k = \frac{\int_{{\Sigma}_k} \bar g^* dx} {\int_{{\Sigma}_k} dx},
\quad k=1, \ldots, K.
\end{equation}
We recall that each entry $c_k[i]$ for $i=1,\cdots,6$ is a value belonging to $\{\mathrm{R,G,B,L,a,b}\}$, respectively.
Using $\{{c}_k\}_{k=1}^K$, we separate $\bar g^*$ into $K$ phases by
\begin{equation} \label{seg-vector}
\Omega_k := \Big \{x\in\Omega~:~ \|\bar g^*(x) - {c}_k \|_2 = \min_{1\leq j\leq K} \|\bar g^*(x) - {c}_j\|_2 \Big\}, \quad k=1, \ldots, K.
\end{equation}
It is easy to verify that $\{\Omega_k\}_{k=1}^K$ are disjoint
and that $\bigcup_{k=1}^K \Omega_k = \Omega$.
The use of the $\ell_2$ distance here follows from our model \eqref{model-1st-extend-n}
as well as from the properties of the Lab color space \cite{Gevers12,CCBR13}.
Note that in the simple case when $c_k\in \mathbb{R}$ (i.e., $d=1$), the thresholding scheme \eqref{seg-vector}
reduces to the one used in \cite{CCZ13,CYZ13}.

\subsection{The SLaT Algorithm}
We summarize our 3-stage segmentation method for color images
in Algorithm \ref{alg:extend}.
Like the Mumford-Shah model, our model (\ref{model-1st-extend-n})
has two parameters $\lambda$ and $\mu$.
Extensive numerical tests have shown that we can fix $\mu =1$.
We choose $\lambda$ empirically; the method is quite stable with respect to
this choice.

\begin{algorithm}[h]
 \caption{Three-stage Segmentation Method (SLaT) for Color Images}
 \label{alg:extend}

 \KwIn{Given color image $f\in {\cal V}_1$ and color space ${\cal V}_2$.}
 \KwOut{Phases $\Omega_k, k=1, \ldots, K$.}
 Stage one: compute $\bar{g}_i$ the minimizer in \eqref{model-1st-extend-n},
 rescale it on $[0,1]$ for $i = 1, 2, 3$
 and set
 $\bar{g}=(\bar{g}_1, \bar{g}_2, \bar{g}_3)$ in ${\cal V}_1$\;
 Stage two: \\
 \quad compute $\bar g'\in {\cal V}_2$, the transform of $\bar g$ in ${\cal V}_2$, to obtain $\bar{g}^t = (\bar{g}^t_1, \bar{g}^t_2, \bar{g}^t_3)$;\\
 	\quad form $\bar{g}^* = (\bar{g}_1, \bar{g}_2, \bar{g}_3, \bar{g}^t_1, \bar{g}^t_2, \bar{g}^t_3)$\;
 Stage three: \\
 \quad choose $K$, apply the $K$-means algorithm
 to obtain $\{{c}_k\}_{k=1}^K$ in \eqref{ck}
 and find the segments $\Omega_k, k=1, \ldots, K$ using \eqref{seg-vector}.
\end{algorithm}
We emphasize that our method is quite suitable for parallelism since $\{\bar g_i\}_{i=1}^3$ in Stage 1 can be computed in parallel.


\section{Experimental Results}\label{sec:experiments}

In this section, we compare our SLaT method with three state-of-the-art variational color segmentation methods \cite{LNZS10,PCCB09,SW14}. Method \cite{LNZS10} uses fuzzy membership functions to approximate the piecewise constant Mumford-Shah model \eqref{pcms}. Method \cite{PCCB09} uses a primal-dual algorithm to solve a convex relax model of \eqref{pcms} with a fixed code book. Method \cite{SW14} uses an ADMM algorithm to solve the model \eqref{pcms} (without phase number $K$) with Potts priors.
The codes we used are provided by the authors, and the parameters in the codes were chosen by trial and error to give the best results of each method. For our model \eqref{model-1st-extend-n}, we fix $\mu=1$ and only
vary $\lambda$. In the segmented figures below, each phase is
represented by the average intensity of
that phase in the image. All the results were tested on a MacBook with
2.4 GHz processor and 4GB RAM, and {\sc Matlab} R2014a.

\begin{figure*}[!htb]
\begin{center}
\begin{tabular}{ccccc}
\includegraphics[width=\ww , height=\ww]{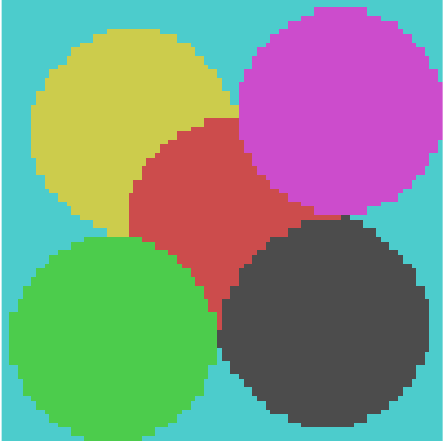} &
\includegraphics[width=\ww, height=\ww]{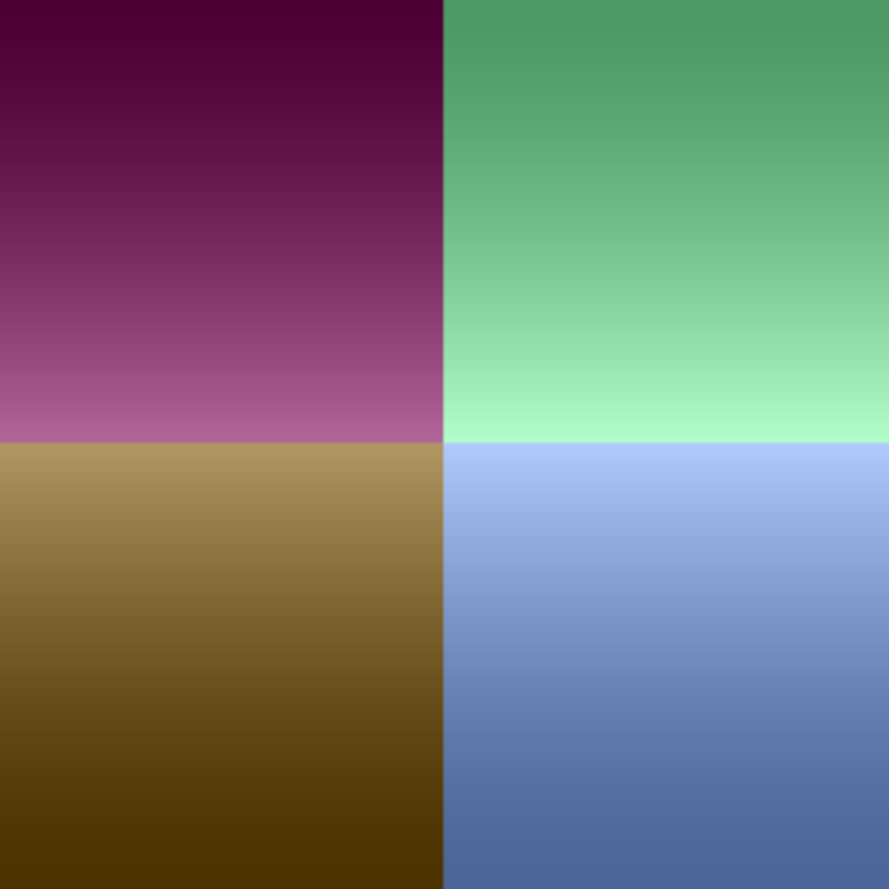} &
\includegraphics[width=\ww, height=\hhh]{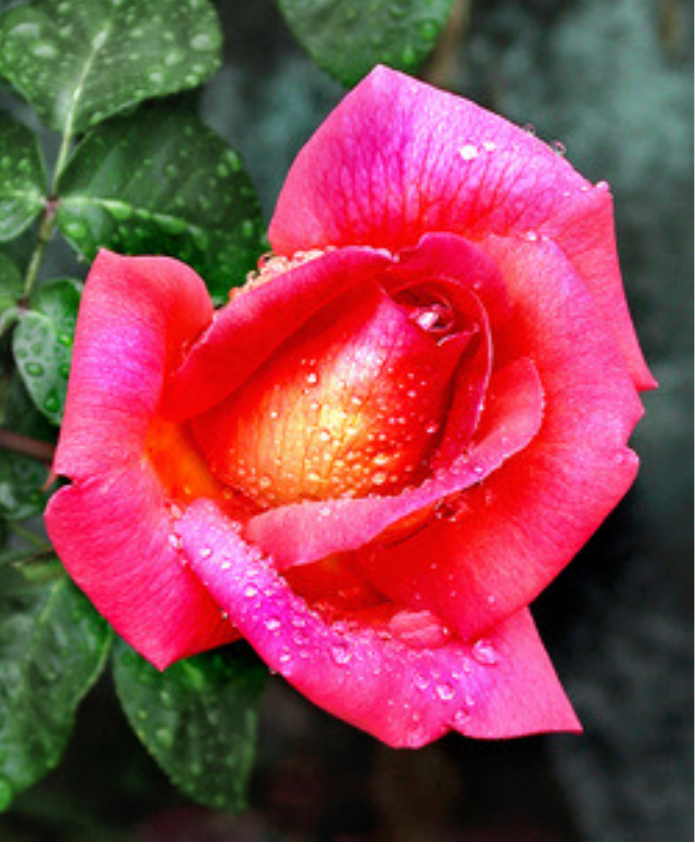} &
\includegraphics[width=\ww , height=\hh]{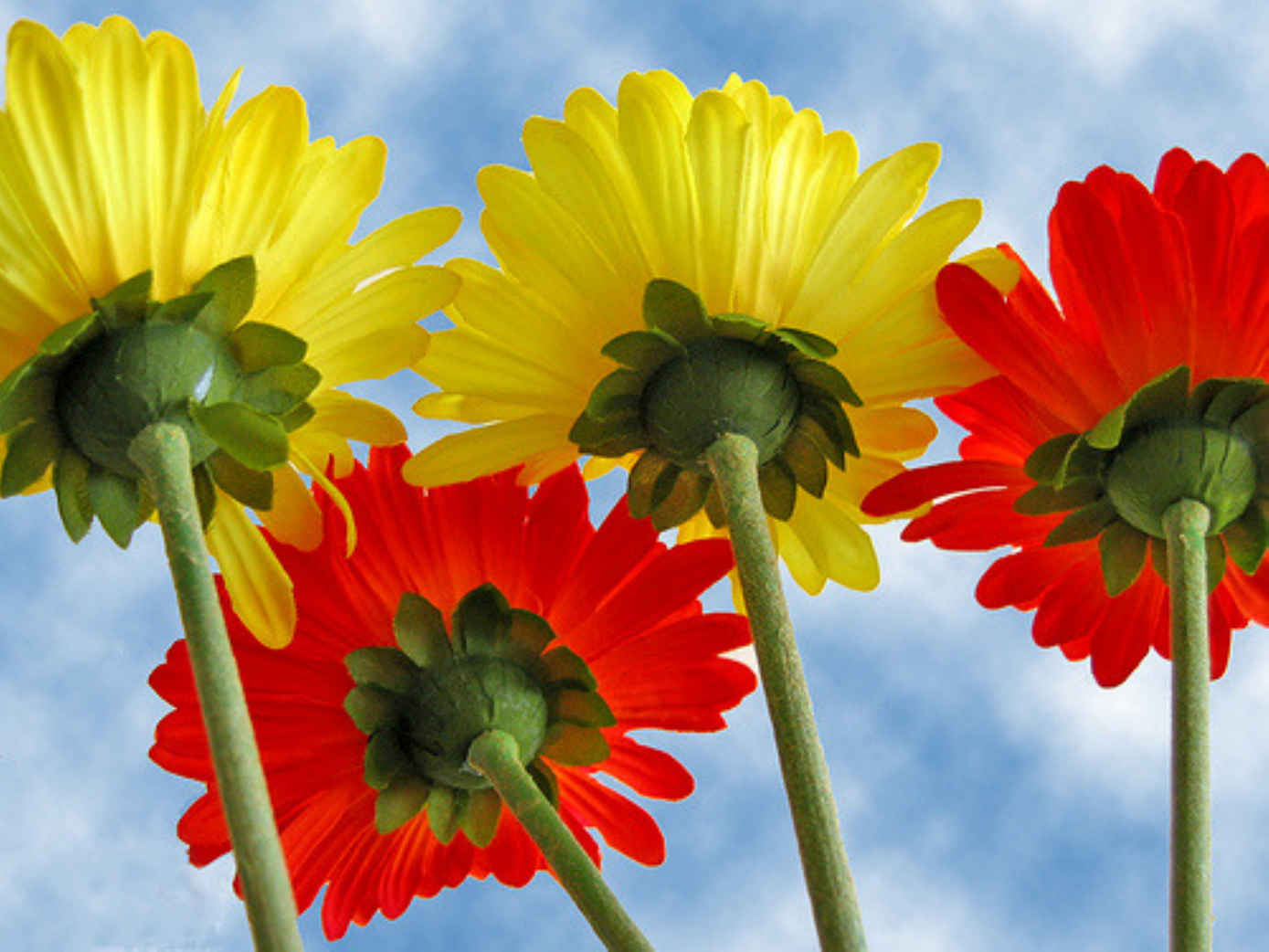} &
\includegraphics[width=\ww , height=\hh]{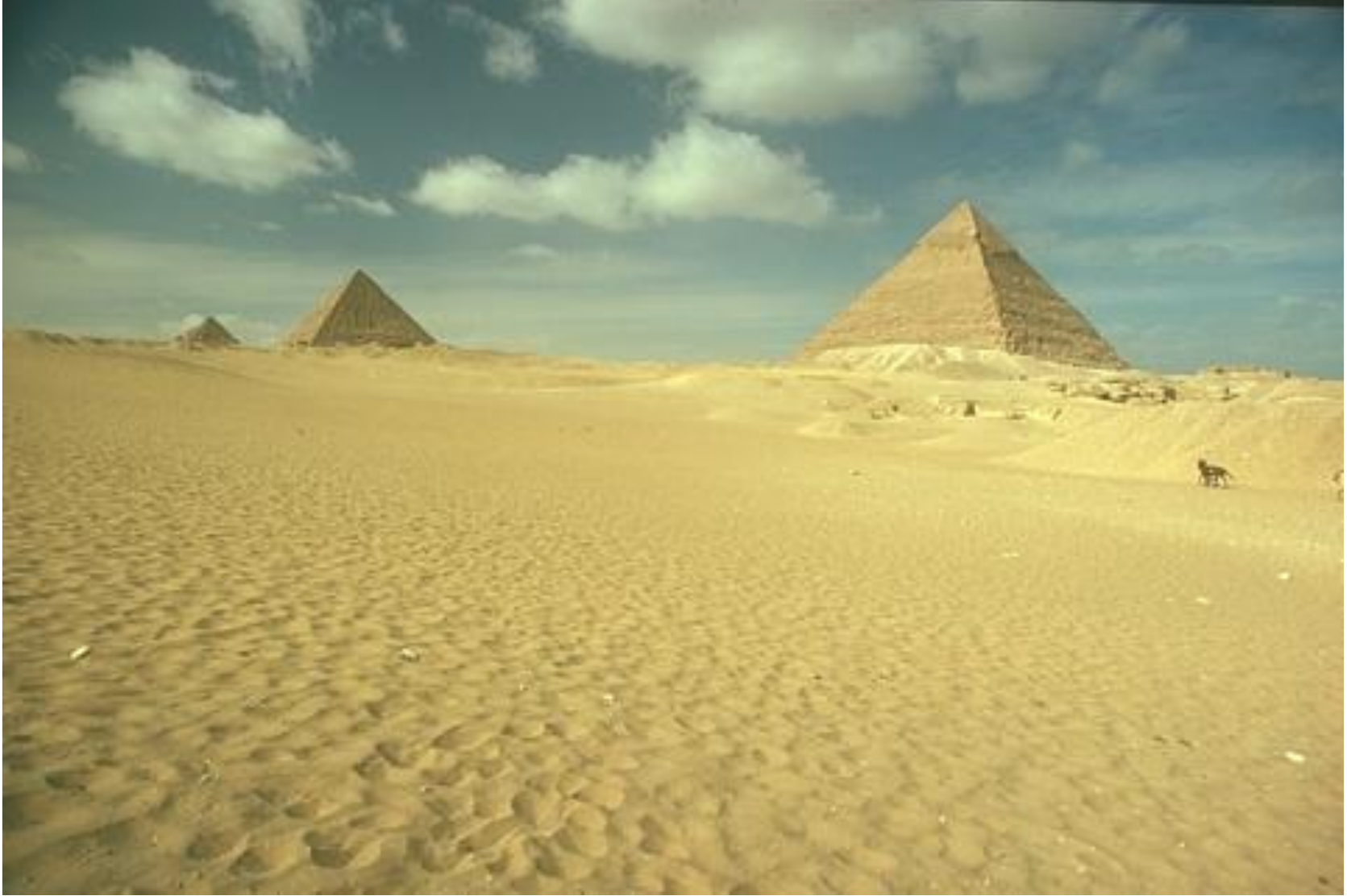} \\
(i) 6-phase & (ii) 4-quadrant & (iii) Rose & (iv) Sunflower & (v) Pyramid \\
\includegraphics[width=\ww , height=\hh]{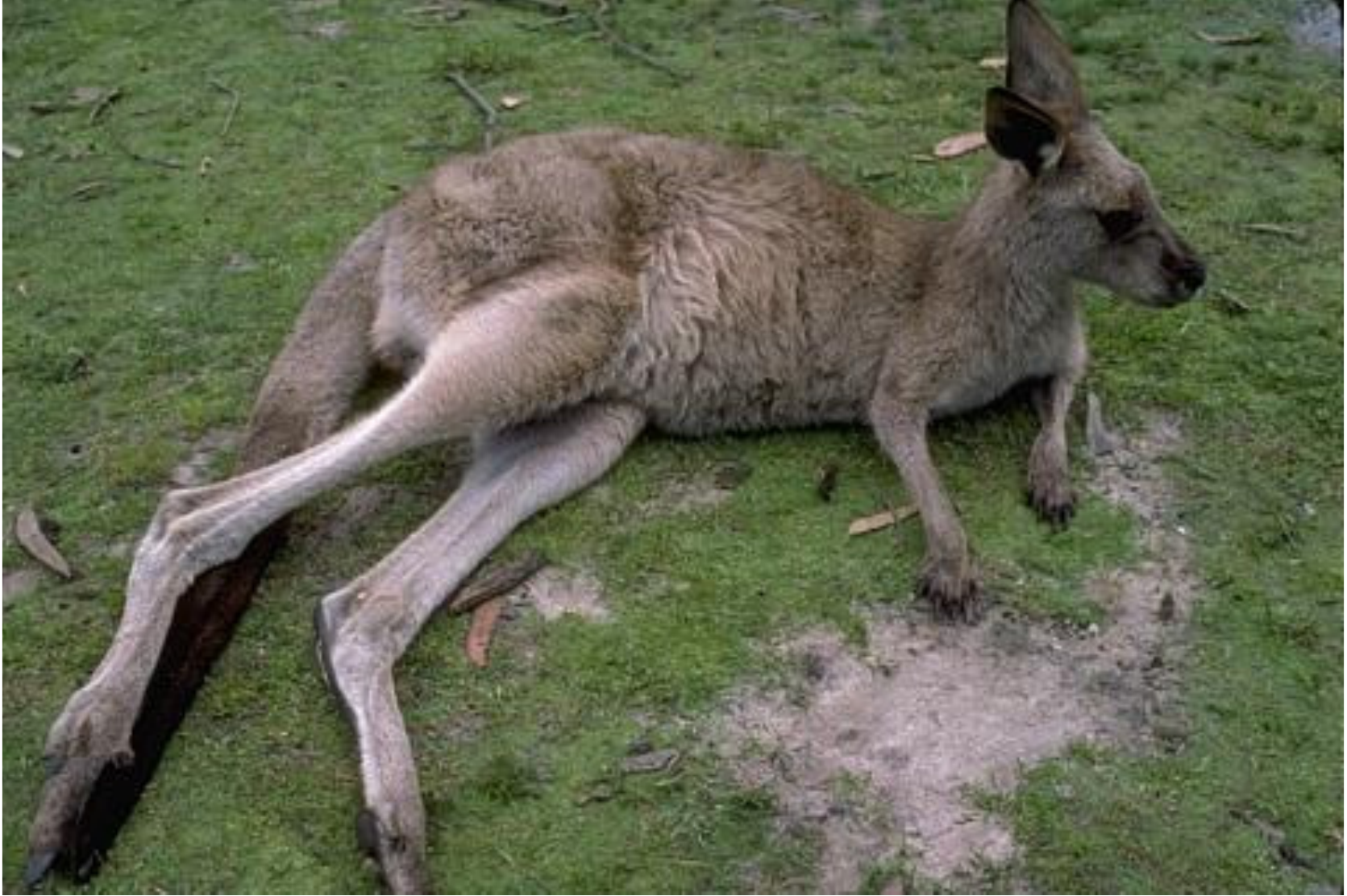} &
\includegraphics[width=\ww , height=\hhh]{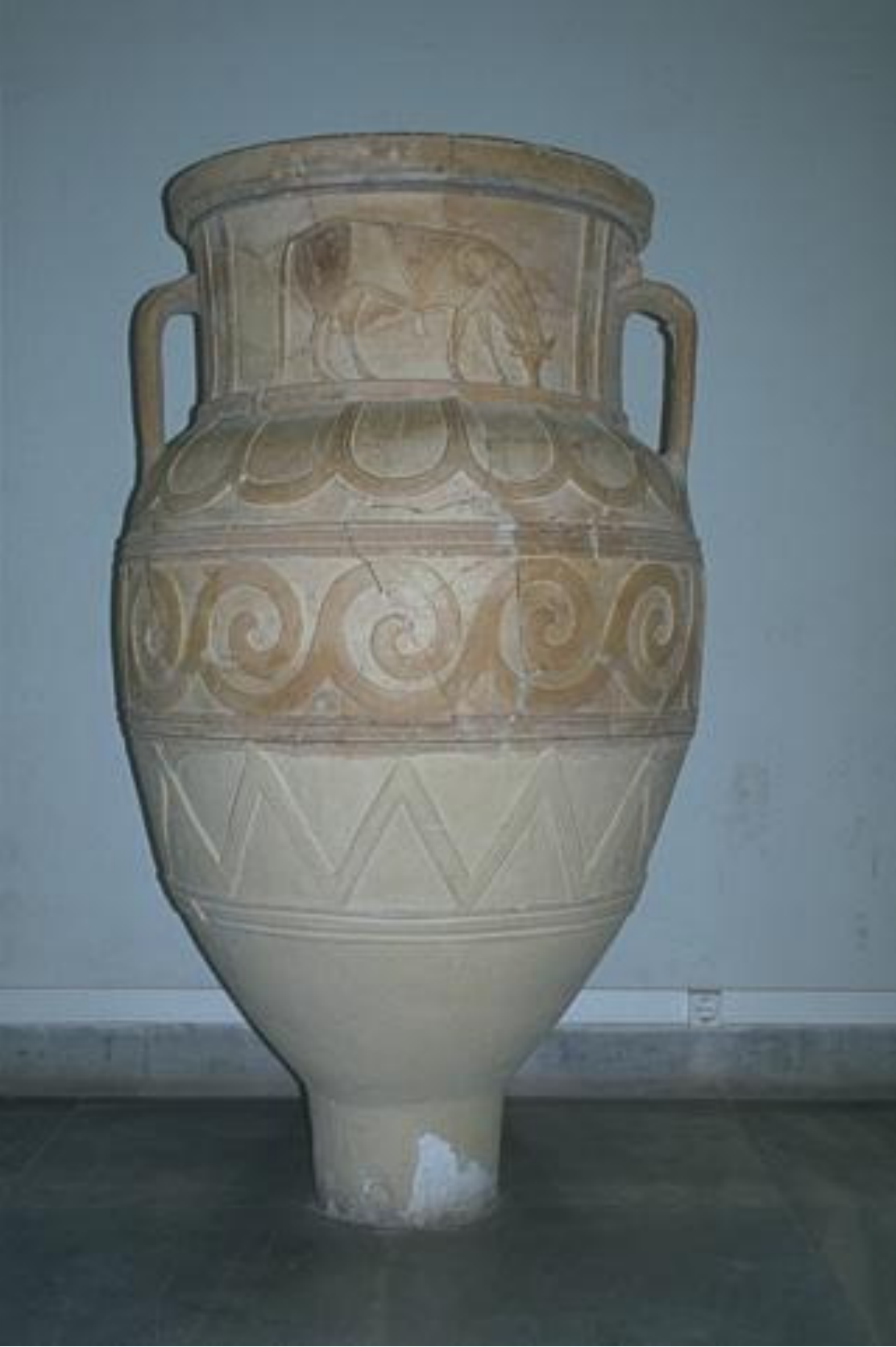} &
\includegraphics[width=\ww , height=\hh]{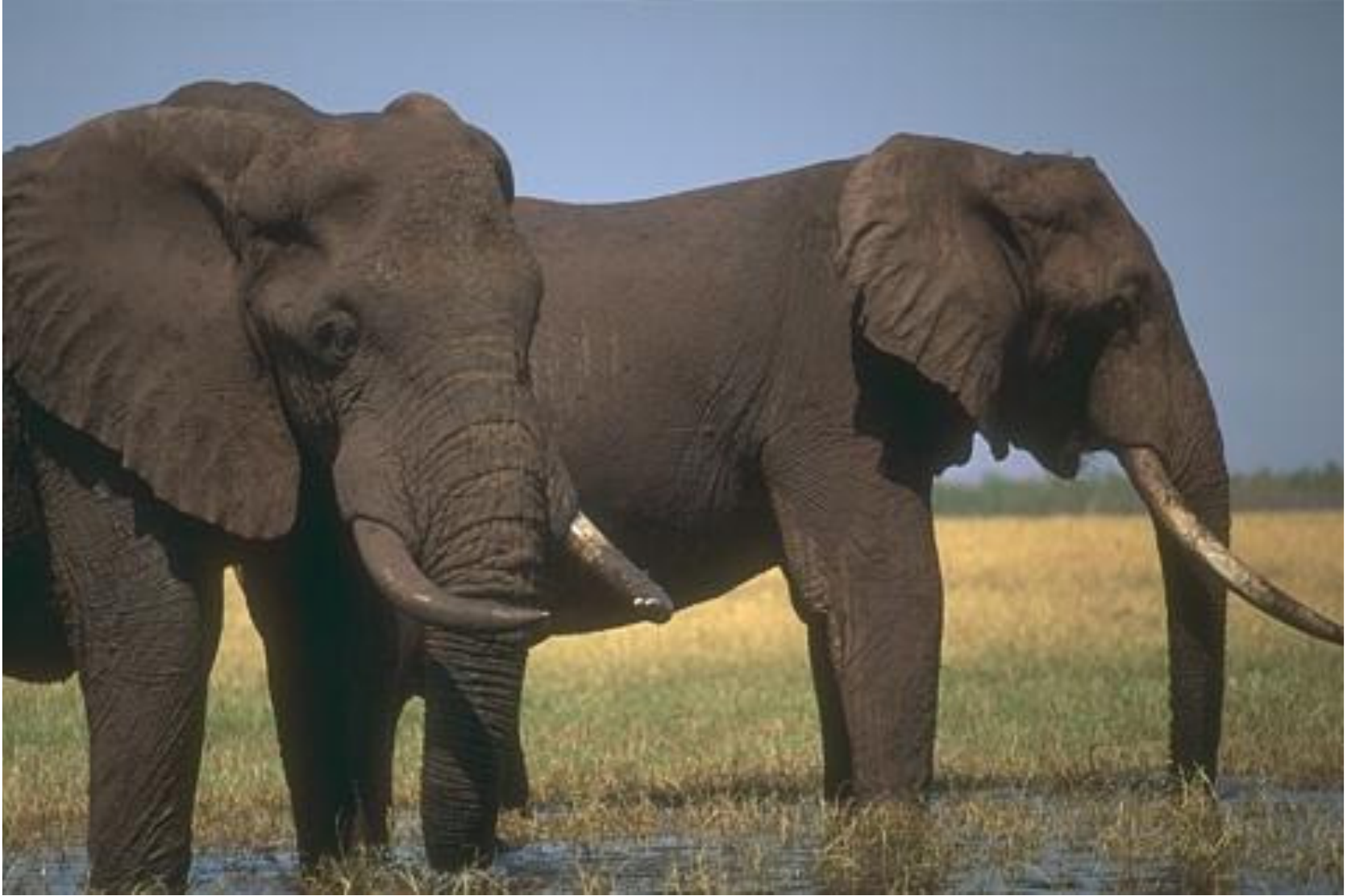} &
\includegraphics[width=\ww , height=\hh]{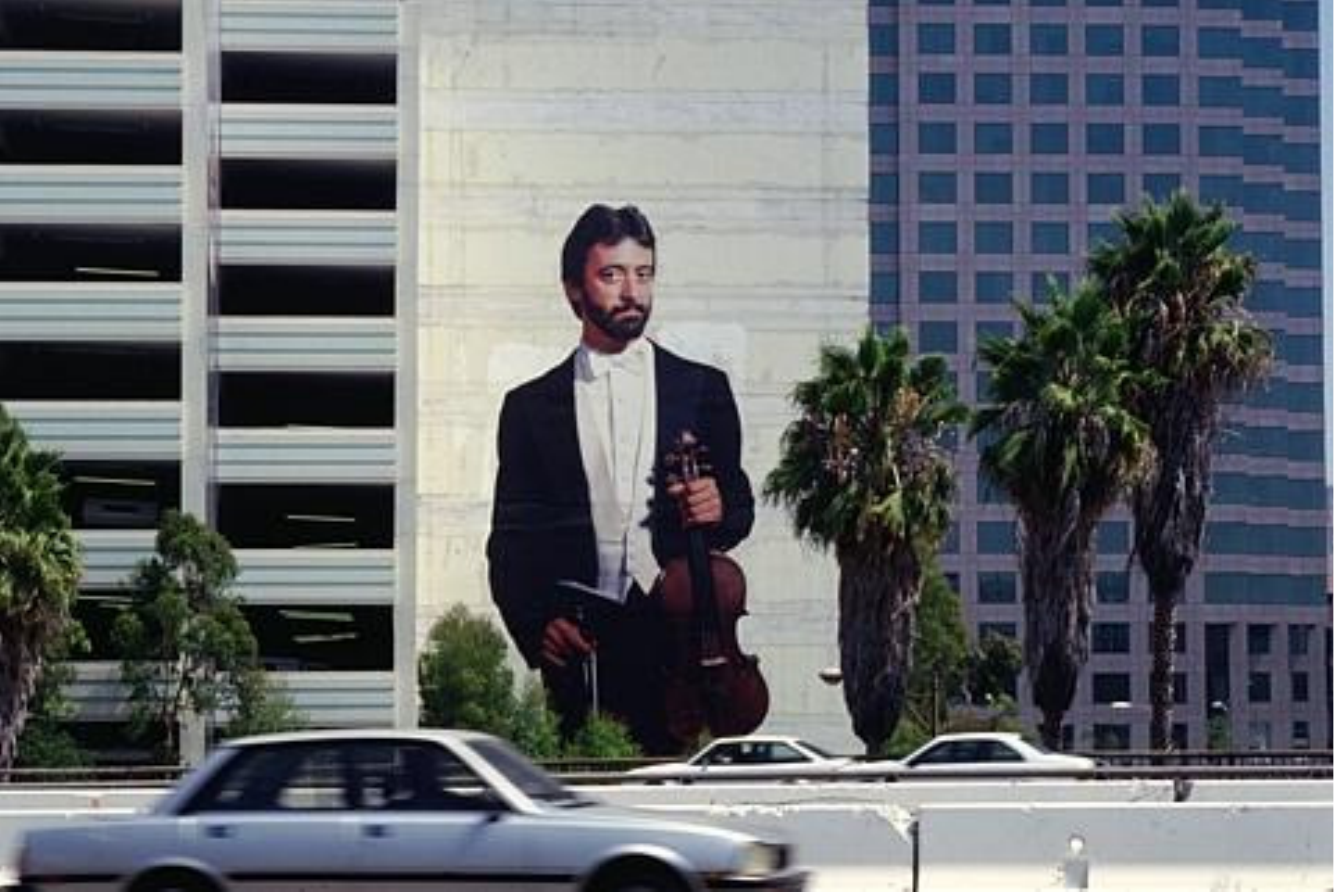} \\
(vi) Kangaroo & (vii) Vase & (viii) Elephant & (ix) Man
\end{tabular}
\end{center}
\caption{Images used in our tests.}
\label{clean_images}
\end{figure*}

We present the tests on two synthesis and seven real-world images
given in Fig. \ref{clean_images}.
The images are all in RGB color space. We considered combinations of
three different forms of image degradation: noise, information loss, and blur.
The Gaussian and Poisson noisy images are all generated using {\sc Matlab}
function {\tt imnoise}. For Gaussian noisy images, the Gaussian noise
we added are all of mean 0 and variance 0.001 or 0.1. To apply the
Poisson noise, we linearly stretch the given image $f$ to [1, 255] first, then
linearly stretch the noisy image back to [0, 1] for testing.
The mean of the Poisson distribution is 10.
For information loss case, we deleted 60\% pixels values randomly. The blur in the
test images were all obtained by a vertical motion-blur with 10 pixels length.

In Stage 1 of our method, the primal-dual algorithm \cite{CP11,CLO13}
and the split-Bregman algorithm \cite{GO09} are adopted to solve \eqref{model-1st-extend-n}
for $\Phi(f, g) = {\cal A}g - f \log ({\cal A}g)$ and $\Phi(f, g) = (f-{\cal A} g)^2$, respectively. We terminate the iterations when
$\frac{\norm{g_i^{(k)}-g_i^{(k+1)}}_2}{\norm{g_i^{(k+1)}}_2} < 10^{-4}$ for $i = 1, 2, 3$
or when the maximum iteration number $200$ is reached.
In Stage 2, the transformation from RGB to Lab color spaces
is implemented by {\sc Matlab} build-in function {\tt makecform('srgb2lab')}.
In Stage 3, given the user defined number of phases $K$,
the thresholds are determined automatically
by {\sc Matlab} K-means function {\tt kmeans}.
Since $\bar{g}^*$ is calculated prior to the choice of $K$,
users can try different $K$ and
segment the image all without re-computing $\bar{g}^*$.

\subsection{Segmentation of Synthetic Images}

\begin{figure*}[!htb]
\begin{center}
\begin{tabular}{ccccc}
\includegraphics[width=\ww, height=\ww]{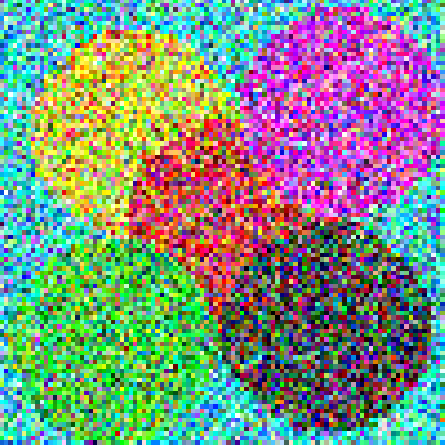} &
\includegraphics[width=\ww, height=\ww]{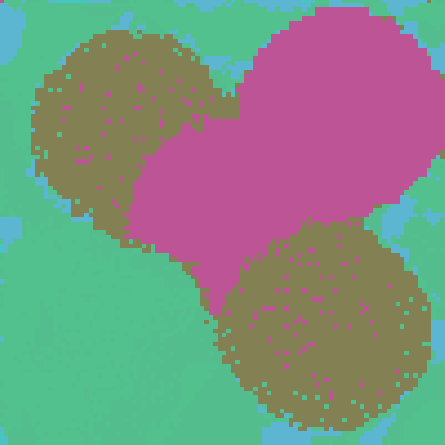} &
\includegraphics[width=\ww, height=\ww]{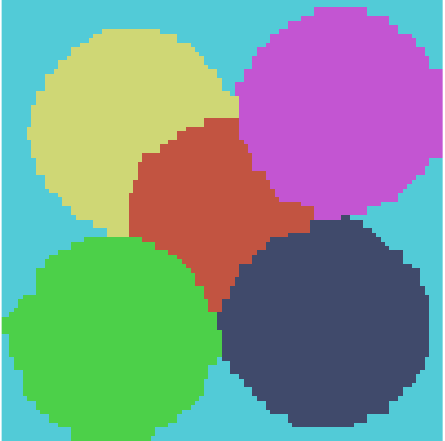} &
\includegraphics[width=\ww, height=\ww]{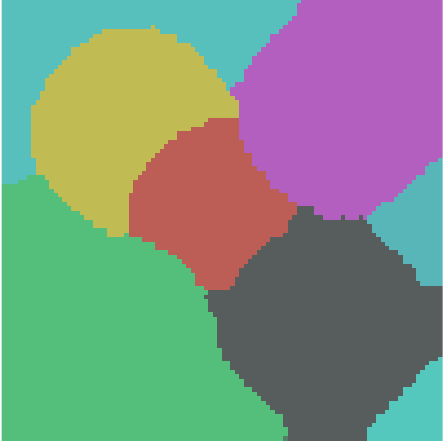} &
\includegraphics[width=\ww, height=\ww]{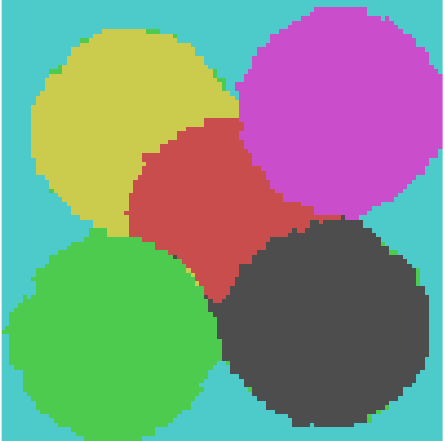} \\
{\small (A) Noisy image} &
{\small (A1) Method \cite{LNZS10}} &
{\small (A2) Method \cite{PCCB09}} &
{\small (A3) Method \cite{SW14}} &
{\small (A4) Ours } \\
\includegraphics[width=\ww, height=\ww]{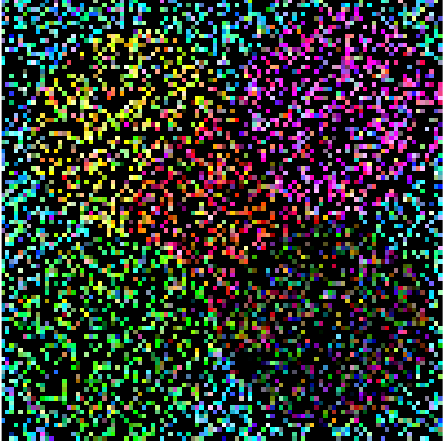} &
\includegraphics[width=\ww, height=\ww]{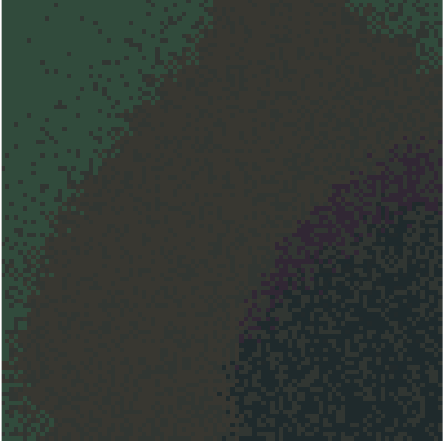} &
\includegraphics[width=\ww, height=\ww]{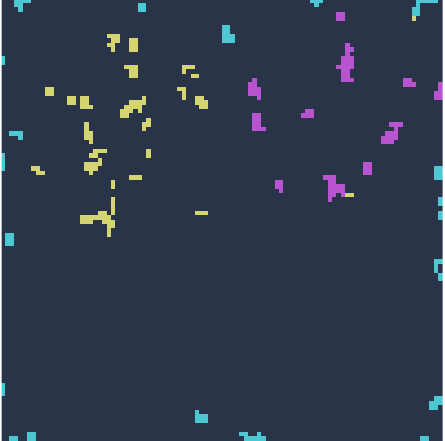} &
\includegraphics[width=\ww, height=\ww]{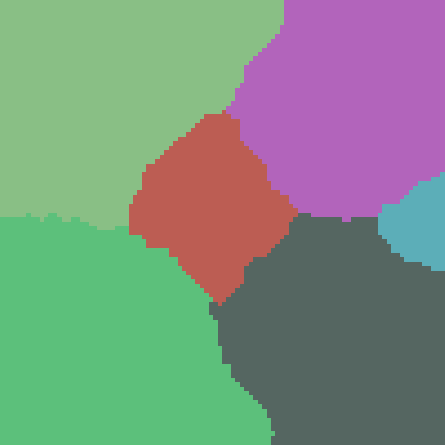} &
\includegraphics[width=\ww, height=\ww]{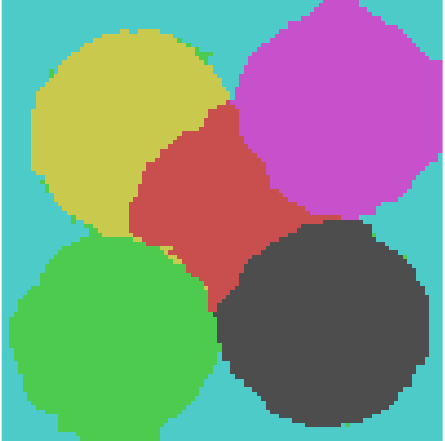} \\
{\small (B) Information} &
{\small (B1) Method \cite{LNZS10}} &
{\small (B2) Method \cite{PCCB09}} &
{\small (B3) Method \cite{SW14}} &
{\small (B4) Ours } \vspace{-0.05in} \\
{\small loss + noise} & & & & \\
\includegraphics[width=\ww, height=\ww]{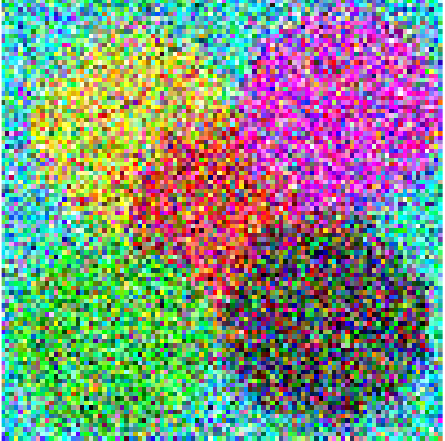} &
\includegraphics[width=\ww, height=\ww]{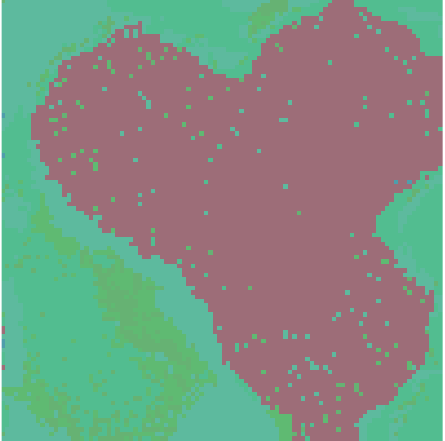} &
\includegraphics[width=\ww, height=\ww]{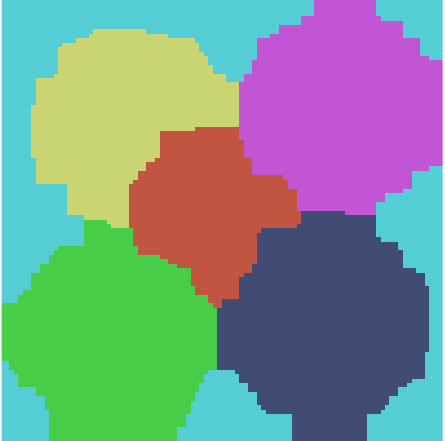} &
\includegraphics[width=\ww, height=\ww]{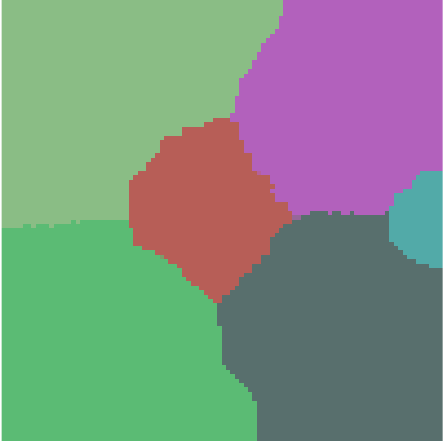} &
\includegraphics[width=\ww, height=\ww]{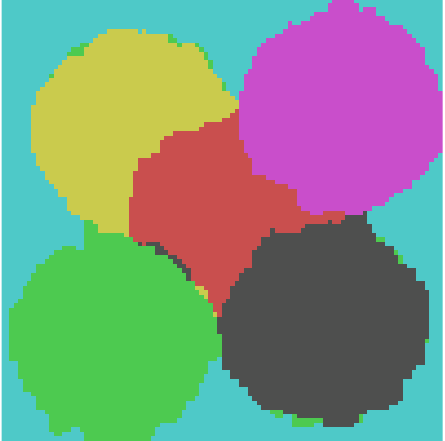} \\
{\small (C) {Blur + noise} } 
&
{\small (C1) Method \cite{LNZS10}} &
{\small (C2) Method \cite{PCCB09}} &
{\small (C3) Method \cite{SW14}} &
{\small (C4) Ours }
\end{tabular}
\end{center}
\caption{Six-phase synthetic image segmentation (size: $100 \times 100$).
(A): Given Gaussian noisy image with mean 0 and variance 0.1;
(B): Given Gaussian noisy image with $60\%$ information loss;
(C): Given blurry image with Gaussian noise; (A1--A4), (B1--B4) and (C1--C4):
Results of methods \cite{LNZS10}, \cite{PCCB09}, \cite{SW14}, and our SLaT on (A), (B) and (C), respectively.
}\label{sixphase-color-syn}
\end{figure*}

\hspace*{5mm}
{\it Example 1. Six-phase segmentation.}
Fig. \ref{sixphase-color-syn} gives the result on a six-phase synthetic image
containing five overlapping circles with different colors. The image is
corrupted by Gaussian noise, information loss, and blur, see Fig. \ref{sixphase-color-syn}
(A), (B), and (C) respectively. From the figures, we see that method Li {\it et al.} \cite{LNZS10}
 and method Storath {\it et al.} \cite{SW14} both fail for the three
experiments while method Pock {\it et al.} \cite {PCCB09} fails for the case of
information lost.

Table \ref{tab:acc-syn} shows the segmentation accuracy by
giving the ratio of the number of correctly segmented pixels to the total
number of pixels. The best ratios are printed in bold face.
From the table, we see that our method gives the highest accuracy
for the case of information loss and blur.
For denoising, method \cite{PCCB09} is 0.02\% better.
Table \ref{tab:time} gives the iteration numbers of each method and
the CPU time cost. We see that our method {outperforms the others compared}. Moreover,
if using parallel technique, the time can be reduced roughly by
a factor of 3.

\begin{figure*}[!htb]
\begin{center}
\begin{tabular}{ccccc}
\includegraphics[width=\ww, height=\ww]{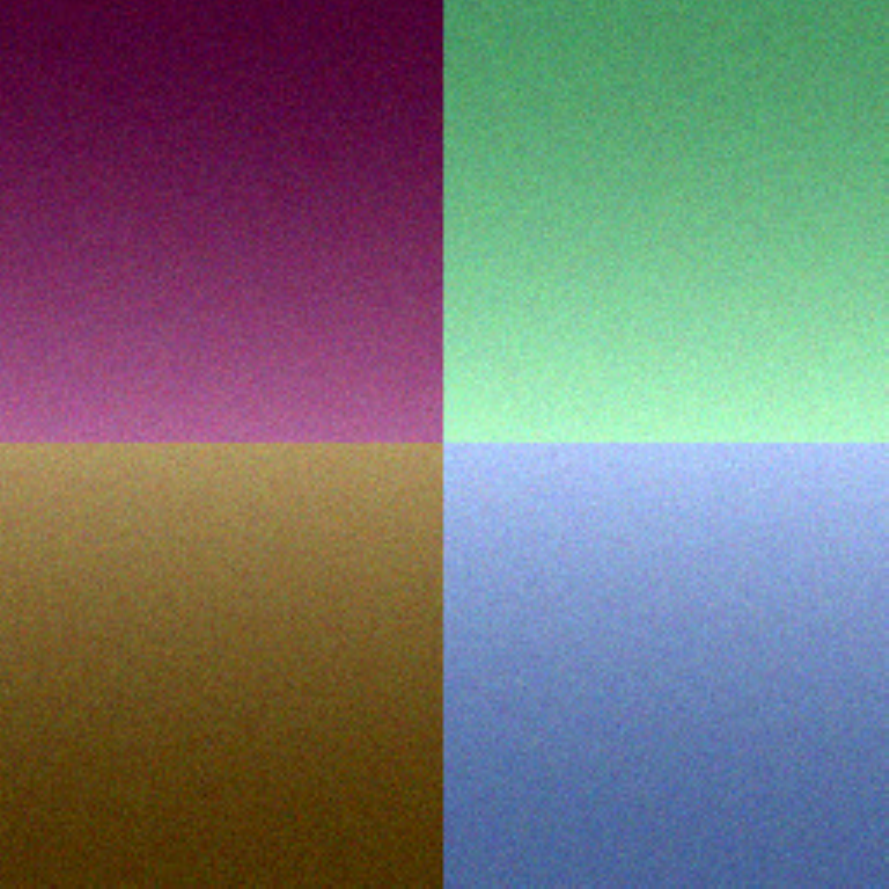} &
\includegraphics[width=\ww, height=\ww]{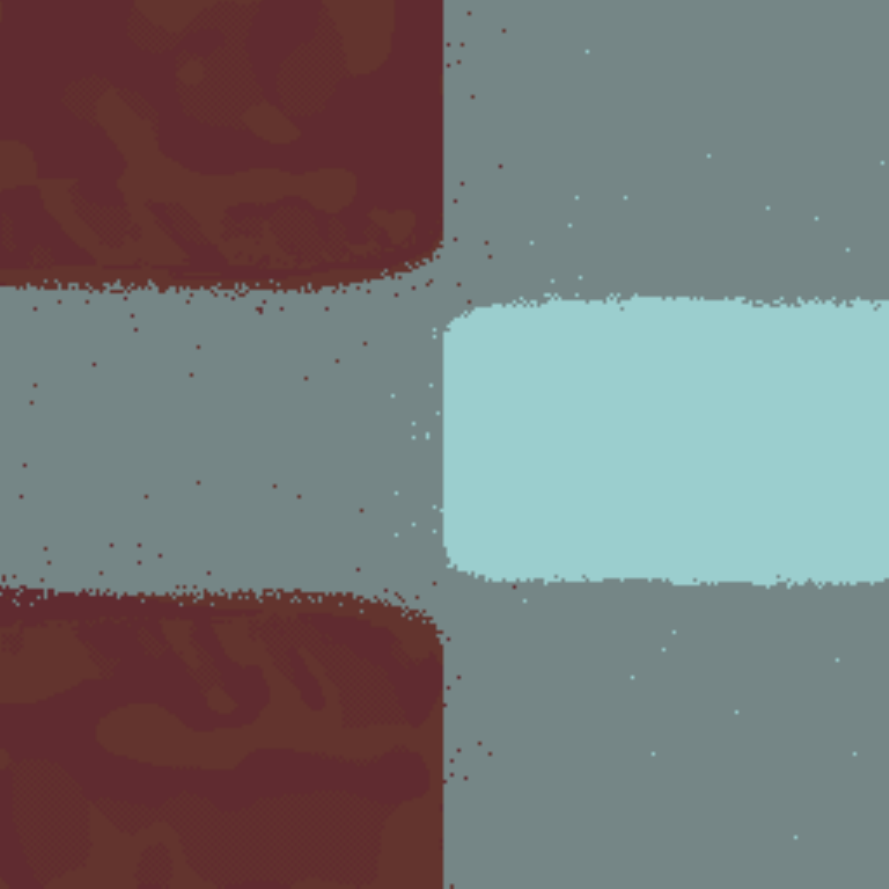} &
\includegraphics[width=\ww, height=\ww]{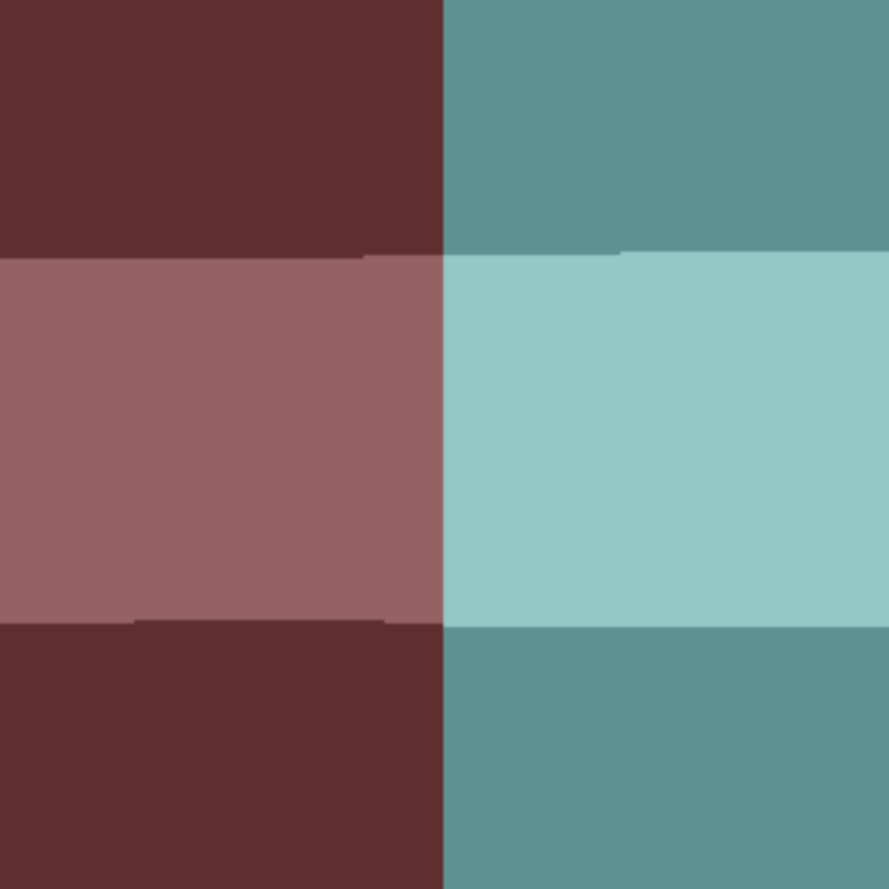} &
\includegraphics[width=\ww, height=\ww]{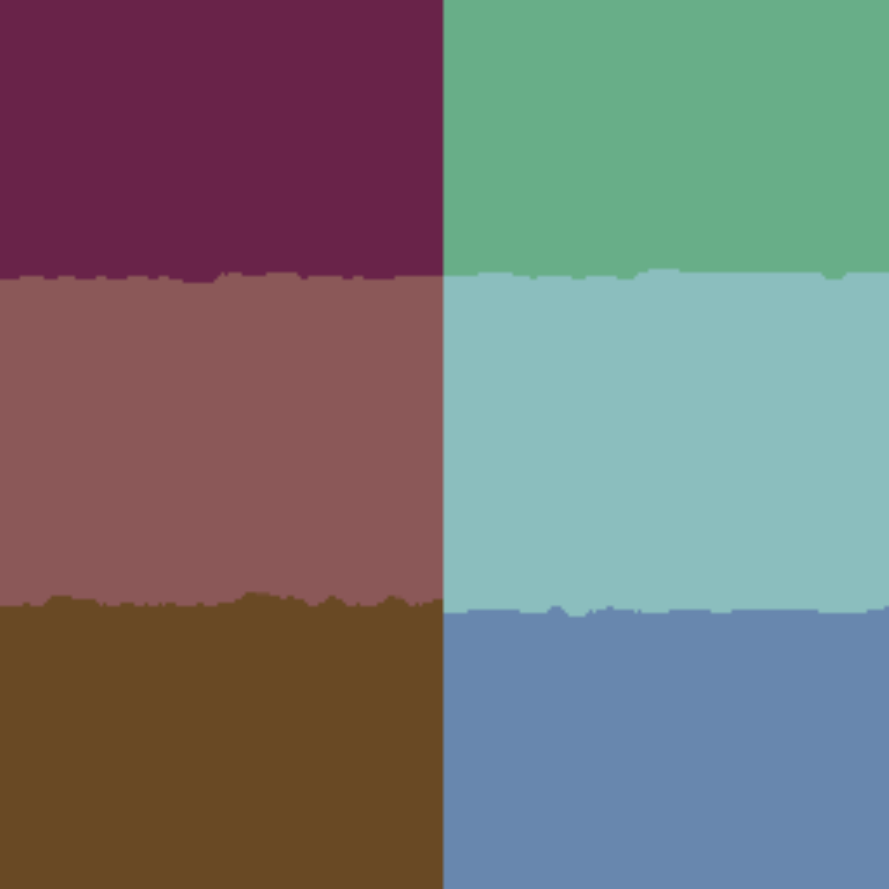} &
\includegraphics[width=\ww, height=\ww]{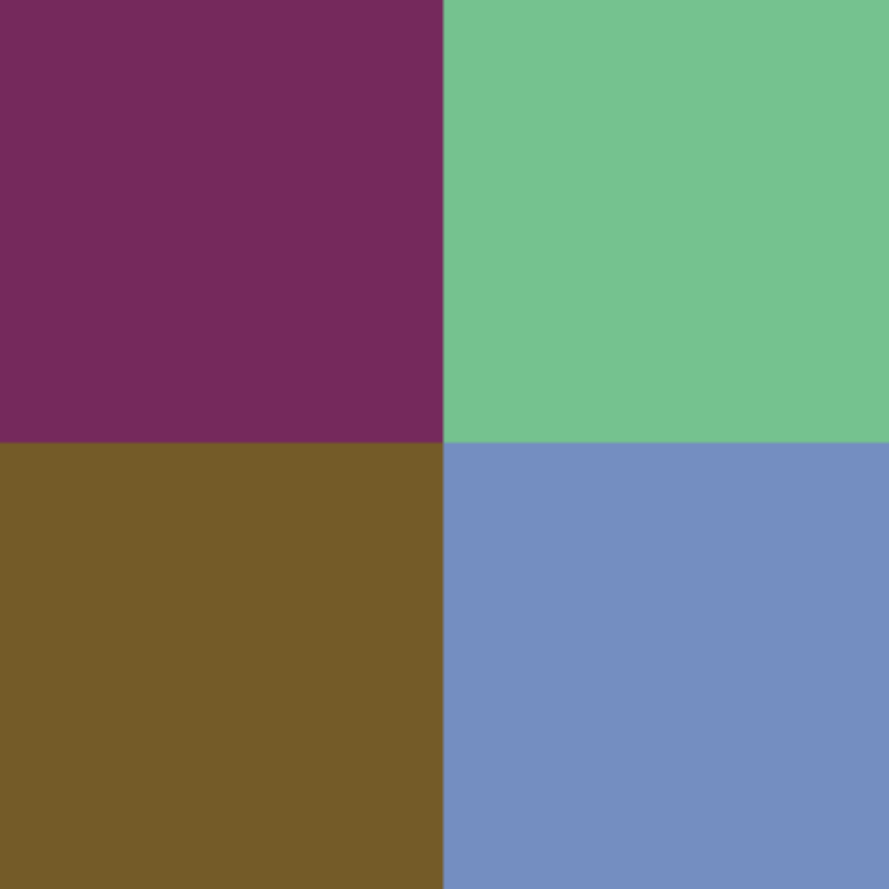} \\
{\small (A) Noisy image} &
{\small (A1) Method \cite{LNZS10}} &
{\small (A2) Method \cite{PCCB09}} &
{\small (A3) Method \cite{SW14}} &
{\small (A4) Ours } \\
\includegraphics[width=\ww, height=\ww]{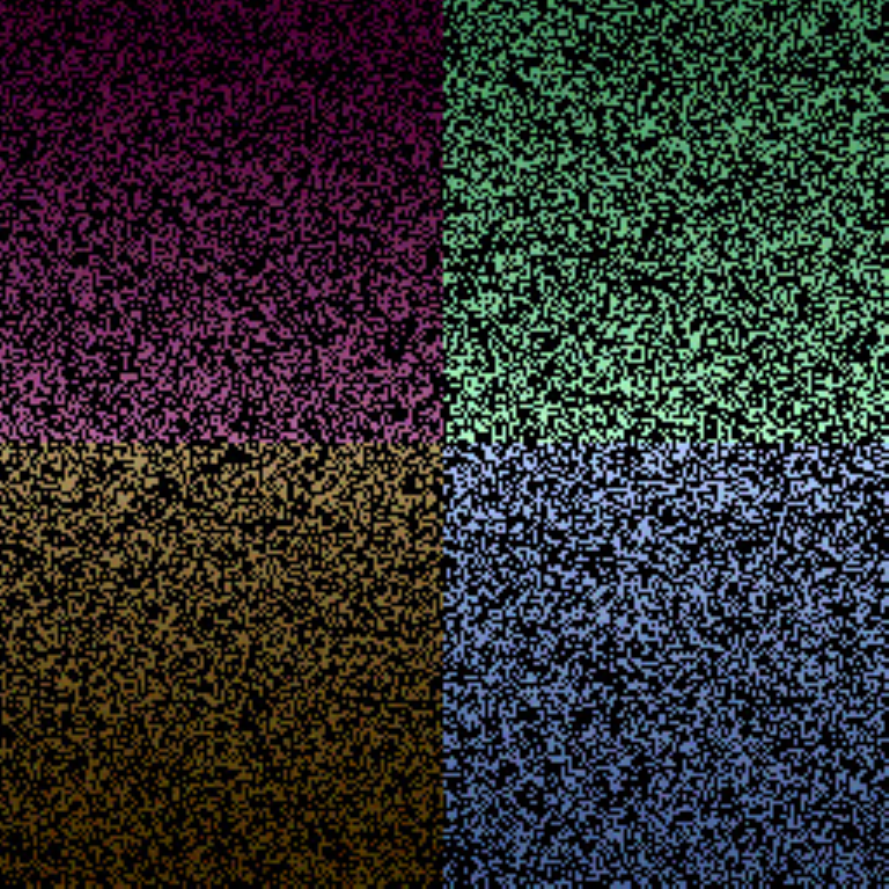} &
\includegraphics[width=\ww, height=\ww]{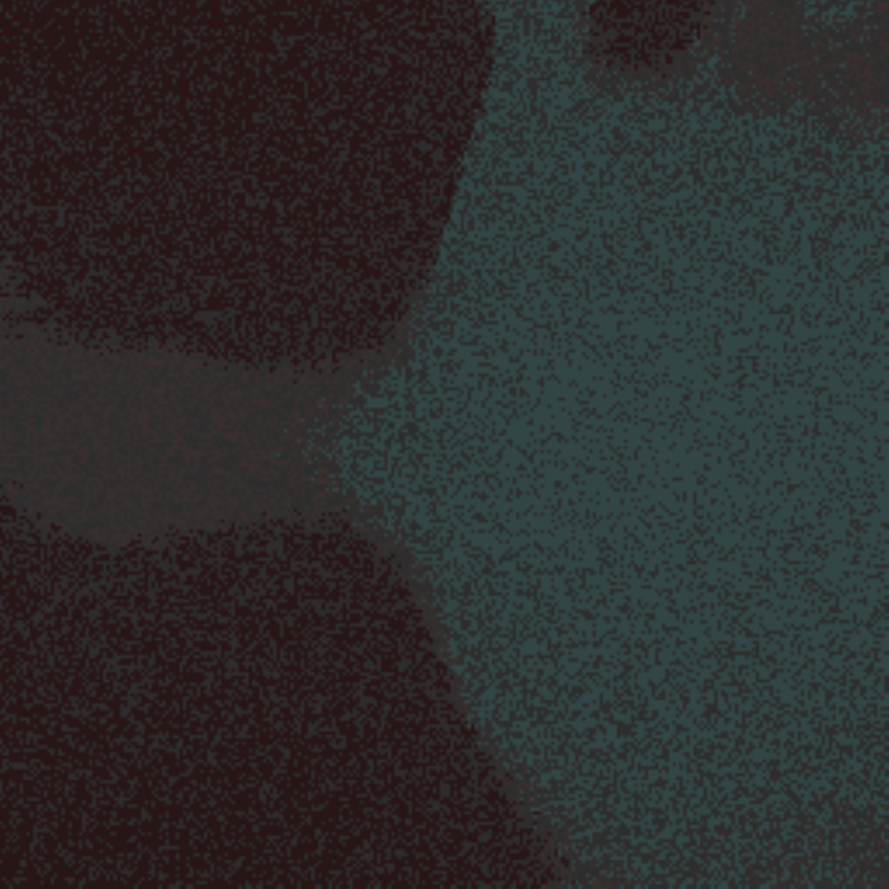} &
\includegraphics[width=\ww, height=\ww]{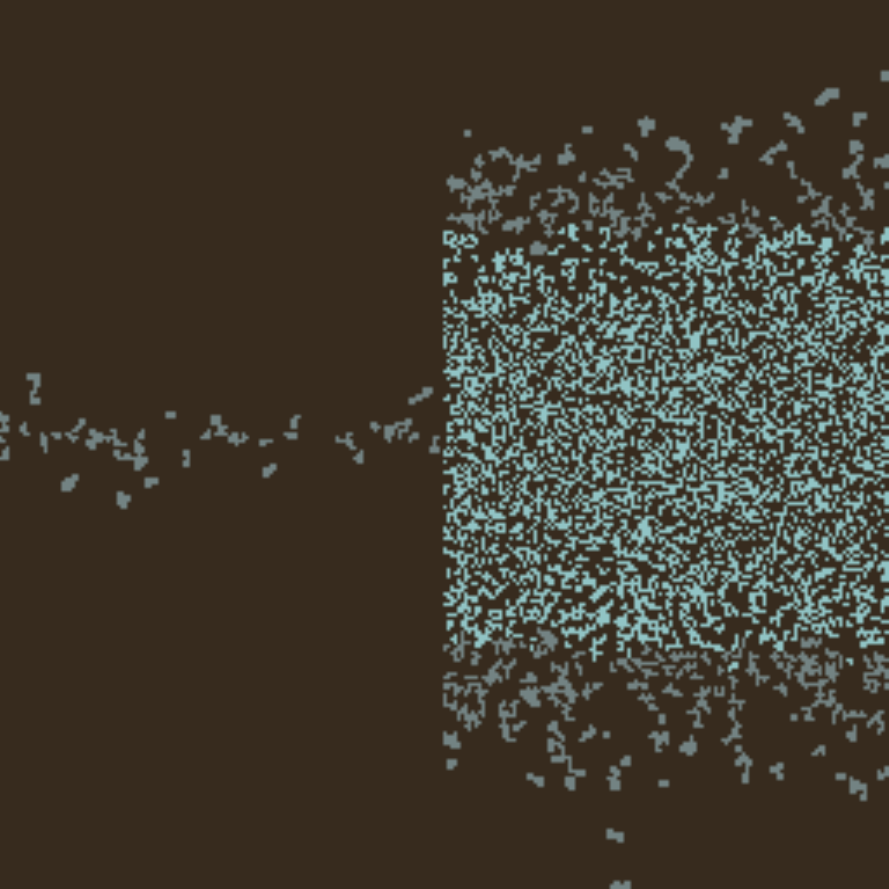} &
\includegraphics[width=\ww, height=\ww]{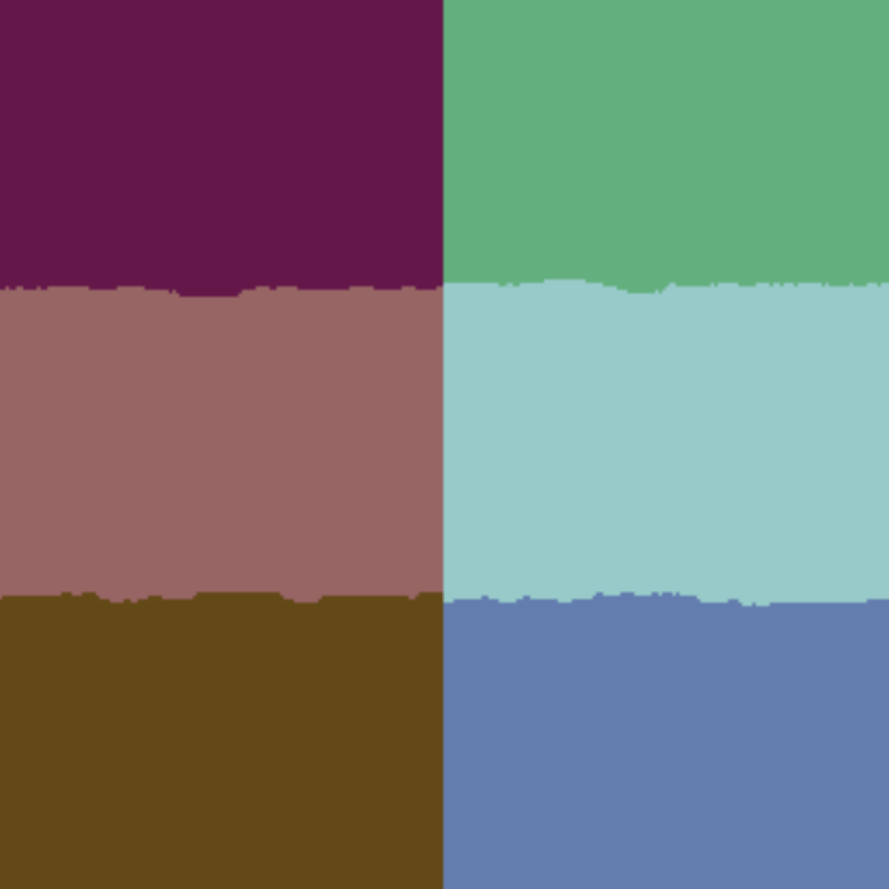} &
\includegraphics[width=\ww, height=\ww]{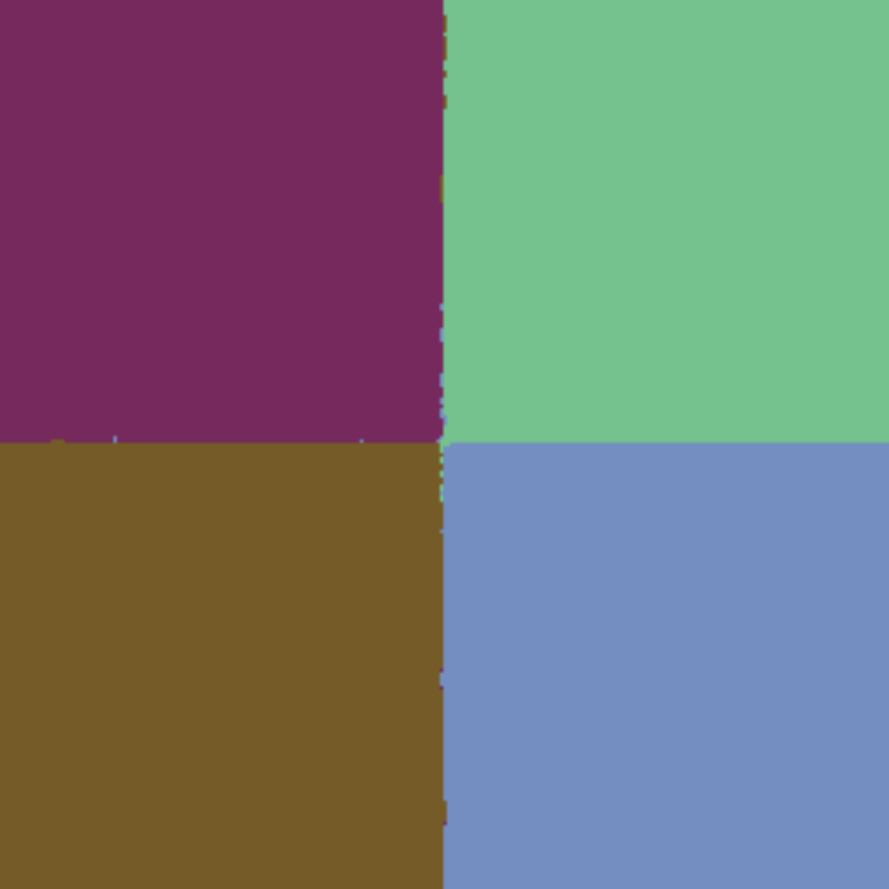} \\
{\small (B) Information } &
{\small (B1) Method \cite{LNZS10}} &
{\small (B2) Method \cite{PCCB09}} &
{\small (B3) Method \cite{SW14}} &
{\small (B4) Ours } \vspace{-0.05in} \\
{\small loss + noise} & & & & \\
\includegraphics[width=\ww, height=\ww]{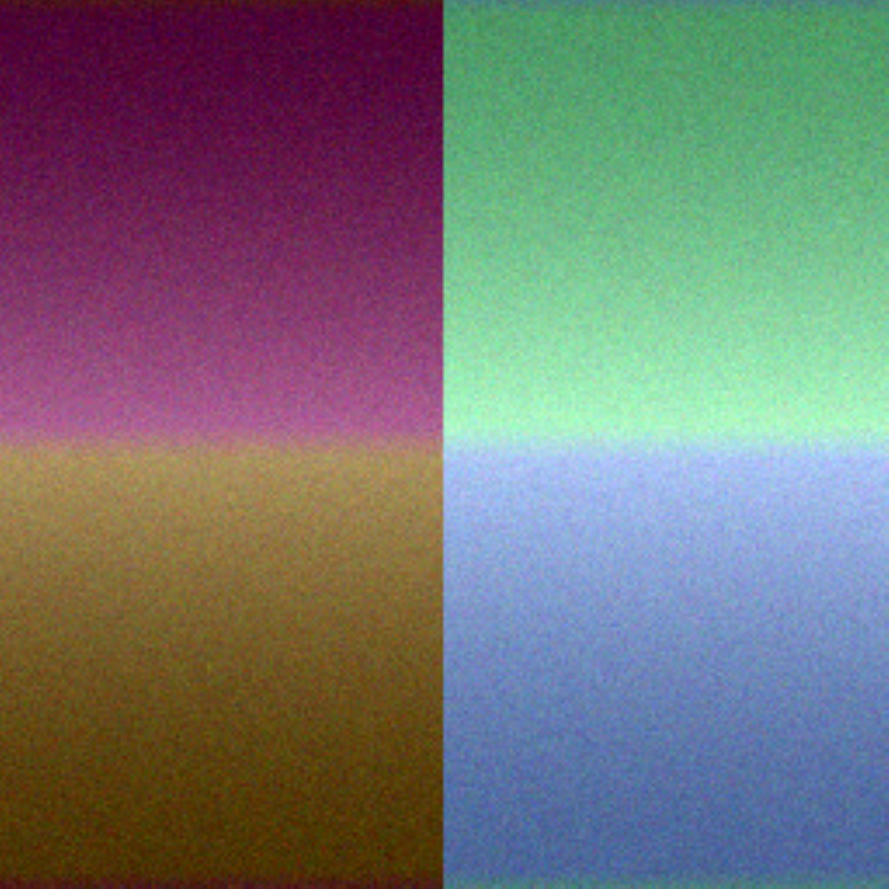} &
\includegraphics[width=\ww, height=\ww]{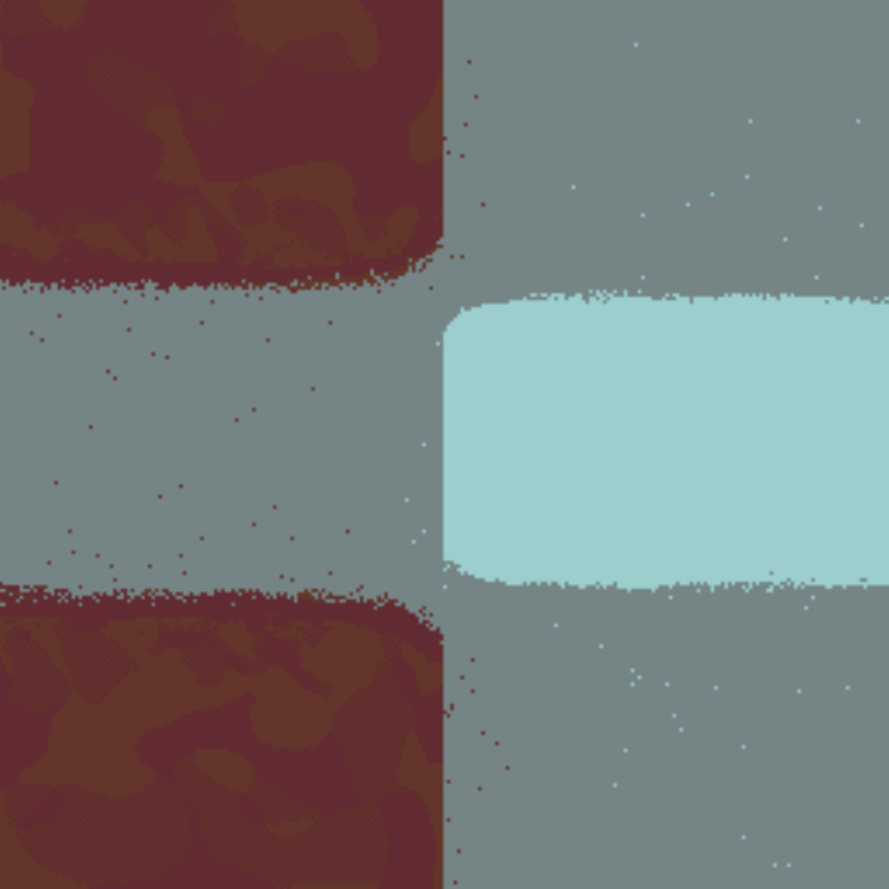} &
\includegraphics[width=\ww, height=\ww]{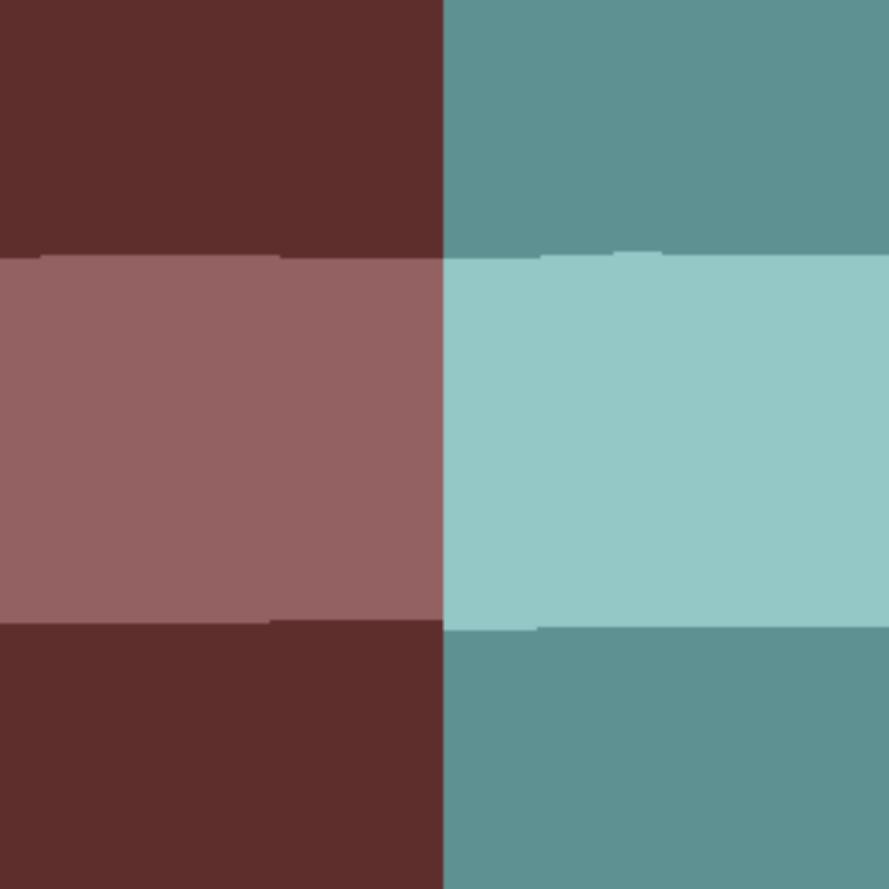} &
\includegraphics[width=\ww, height=\ww]{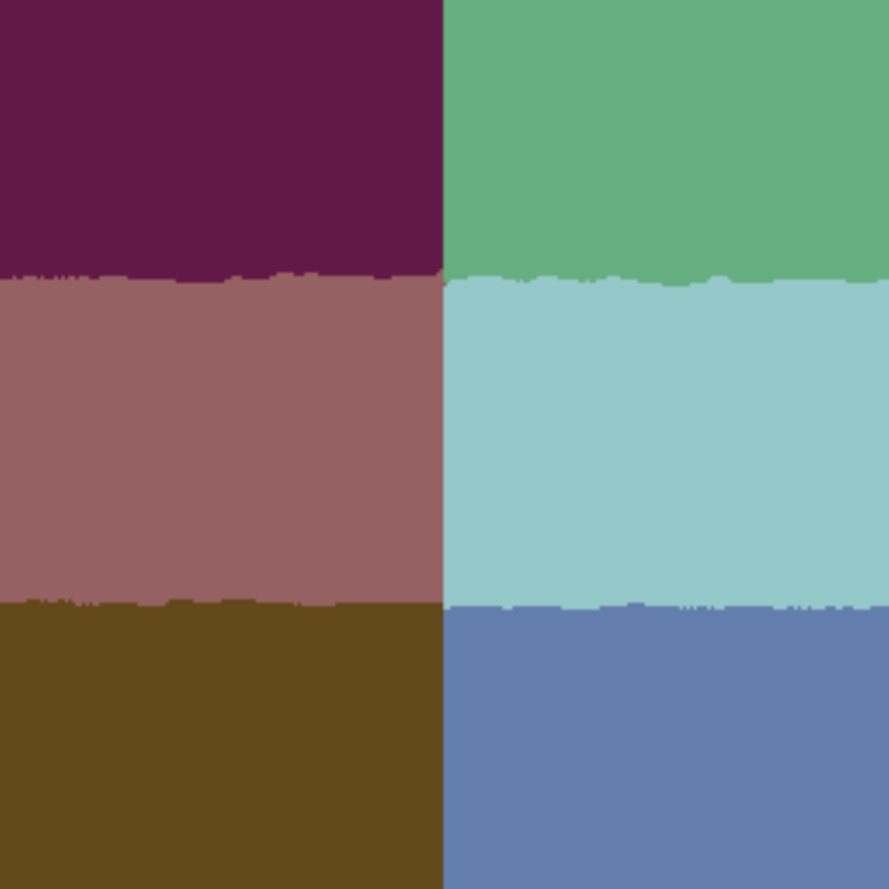} &
\includegraphics[width=\ww, height=\ww]{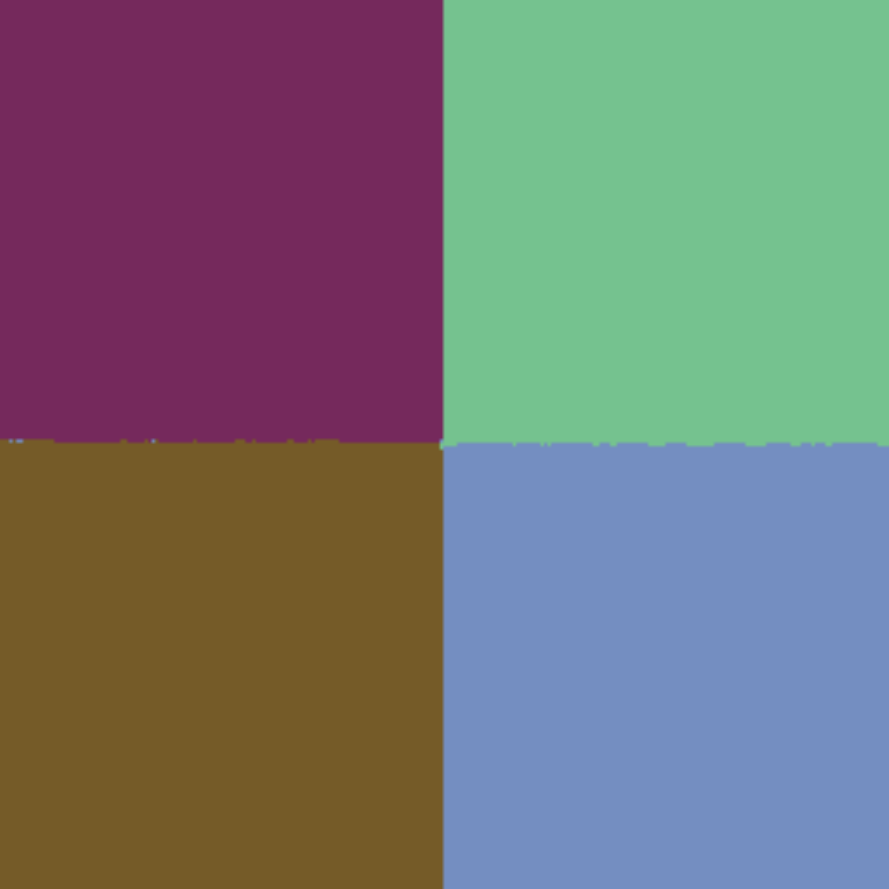} \\
{\small (C) Blur + noise} &
{\small (C1) Method \cite{LNZS10}} &
{\small (C2) Method \cite{PCCB09}} &
{\small (C3) Method \cite{SW14}} &
{\small (C4) Ours }
\end{tabular}
\end{center}
\caption{Four-phase synthetic image segmentation (size: $256\times 256$).
(A): Given Gaussian noisy image with mean 0 and variance 0.001;
(B): Given Gaussian noisy image with $60\%$ information loss;
(C): Given blurry image with Gaussian noise; (A1--A4), (B1--B4) and (C1--C4):
Results of methods \cite{LNZS10}, \cite{PCCB09}, \cite{SW14}, and our SLaT on (A), (B) and (C), respectively. }
\label{fourphase-syn}
\end{figure*}

{\it Example 2. Four-phase segmentation.}
Our next test is on a four-phase synthetic image containing four rectangles
with different colors, see Fig. \ref{fourphase-syn}. The variable illumination
in the figure make the segmentation very challenging. The results
shows that in all cases (noise, information loss
and blur) all three competing methods \cite{LNZS10, PCCB09, SW14} fail
while our method gives extremely good results. Table \ref{tab:time} shows further that
the time cost of our method is the least.

\begin{table*}
\centering
\caption{Comparison of percentage of correct pixels for the 6-phase synthetic image.}
\label{tab:time-syn}
\begin{tabular}{ c| c| c| c|c | c}
\hline
 \multicolumn{2}{c|}{} & {Method \cite{LNZS10}} & {Method \cite{PCCB09}} & {Method \cite{SW14}} & {Our SLaT method} \\ \hline \hline
\multirow{3}*{Fig. \ref{sixphase-color-syn}} & (A) & 70.11\% & {\bf 99.53\%} & 82.55\% & 99.51\% \\ \cline{2-6}
 & (B) & 13.90\% & 16.92\% & 85.04\% & {\bf 99.25\%} \\ \cline{2-6}
 &(C) & 28.08\% & 98.58\% & 74.77\% & {\bf 98.88\%} \\ \hline

 \multicolumn{2}{c|}{Average} & 37.36\% & 71.68\%& 80.79\% & {\bf 99.21}\%\\ \hline

\end{tabular}
\label{tab:acc-syn}
\end{table*}

\begin{table*}
\centering
\caption{Iteration numbers and CPU time in seconds. }
\label{tab:time}
\begin{tabular}{ c| c| r|r|r| r| r|r|c|r } \hline
 \multicolumn{2}{c| }{} & \multicolumn{2}{c| }{Method \cite{LNZS10}} & \multicolumn{2}{c| }{Method \cite{PCCB09}}
 & \multicolumn{2}{c| } {Method \cite{SW14}} & \multicolumn{2}{c } {Our SLaT method} \\
 \hline \multicolumn{2}{c| }{Fig.} & iter. & time & iter. & time & iter. & time
 & iter. for $\{g_i\}_{i=1}^3$ & time
 \\ \hline \hline
 \multirow{3}*{\ref{sixphase-color-syn}}
 &(A) & 200 & 5.03 & 150 & 6.02 & 20 & 4.40 & (92, 86, 98) & {\bf 2.53} \\ \cline{2-10}
&(B) & 200 & 5.65 & 150 & 4.01 & 16 & 3.65 & (98, 95, 106) & {\bf 2.73} \\ \cline{2-10}
 & (C) & 200 & 6.54 & 150 & 4.03 & 17 & 4.18 & (97, 95, 94) & {\bf 2.48} \\ \hline
 \multirow{3}*{\ref{fourphase-syn}}
 & (A) &200 & 13.92 & 150 & 13.89 & 17 & 16.89 & (54, 54, 51) & {\bf 5.47} \\ \cline{2-10}
 & (B) &200 & 13.16 & 150 & 14.32 & 14 & 13.62 & (101, 92, 88) & {\bf 7.74} \\ \cline{2-10}
 & (C) &200 & 17.68 & 150 & 16.37 & 16 & 15.75 & (154, 147, 142) & {\bf 9.89} \\ \hline
 \multirow{3}*{\ref{twophase-color-rose}}
 & (A) &200 & 10.58 & 150 & 7.53 & 19 & 11.62 & (50, 73, 93) & {\bf 5.11} \\ \cline{2-10}
 & (B) &200 & 9.59 & 150 & 7.36 & 20 & 14.64 & (84, 105, 115) & {\bf 6.43} \\ \cline{2-10}
& (C) &200 & 10.39 & 150 & {\bf 7.39} & 19 & 9.76 & (200, 200, 200) & 17.75 \\ \hline
 \multirow{3}*{\ref{fourphase-color-flower}}
 &(A) &200 &44.26 & 150 & 66.01 & 19 & 106.35 & (97, 106, 109) & {\bf 25.13} \\ \cline{2-10}
&(B) & 200 & 52.12 & 150 & 54.76 & 20 & 110.68 & (148, 161, 171) & {\bf 38.30} \\ \cline{2-10}
 & (C) & 200 & 44.51 & 150 & 55.09 & 18 & 101.09 & (116, 125, 124) & {\bf 30.00} \\ \hline
 \multirow{3}*{\ref{twophase-color-pyramid}}
 & (A) &200 & 17.76 & 150 & {\bf 19.02} & 16 & 25.08 & (80, 83, 99) & 20.99 \\ \cline{2-10}
 & (B) &200 & 18.41 & 150 & {\bf 16.45} & 16 & 28.33 & (109, 114, 129) & 22.45 \\ \cline{2-10}
& (C) & 200 & 18.02 & 150 & {\bf 18.21} & 15 & 31.93 & (127, 120, 144) & 30.92 \\ \hline
 \multirow{3}*{\ref{twophase-color-kangaroo}}
 &(A) & 200 & 18.47 & 150 & 19.62 & 15 & 27.56 & (47, 42, 62) & {\bf 10.98} \\ \cline{2-10}
&(B) &200 & 17.35 & 150 & 16.63 & 15 & 26.63 & (86, 85, 93) & {\bf 15.93} \\ \cline{2-10}
 & (C) &200 & 18.07 & 150 & 17.61 & 15 & 23.13 & (48, 48, 52) & {\bf 15.02} \\ \hline
 \multirow{3}*{\ref{threephase-color-china}}
 &(A) &200 & 24.57 & 150 & 31.64 & 20 & 56.28 & (101, 95, 94) & {\bf 27.29} \\ \cline{2-10}
&(B) & 200 & 27.15 & 150 & 28.92 & 21 & 63.02 & (154, 142, 131) & {\bf 27.89} \\ \cline{2-10}
 & (C) & 200 & 26.54 & 150 & {\bf 29.79} & 20 & 55.45 & (161, 147, 141) & 33.79 \\ \hline
 \multirow{3}*{\ref{threephase-color-elephant}}
 &(A) & 200 & 26.62 & 150 & 32.87 & 17 & 87.13 & (35, 35, 36) & {\bf 14.23} \\ \cline{2-10}
&(B) &200 & 24.77 & 150 & 26.39 & 16 & 60.98 & (102, 102, 103) & {\bf 18.99} \\ \cline{2-10}
 & (C) &200 & 25.26 & 150 & 31.16 & 18 & 77.73 & (48, 50, 58) & {\bf 18.15} \\ \hline
 \multirow{3}*{\ref{fourphase-color-man}}
 &(A) & 200 & 32.23 & 150 & 41.91 & 19 & 47.12 & (106, 102, 108) & {\bf 21.93} \\ \cline{2-10}
&(B) & 200 & 34.83 & 150 & 44.70 & 20 & 53.48 & (116, 116, 117) & {\bf 23.95} \\ \cline{2-10}
 & (C) & 200 & 35.01 & 150 & 49.93 &19 & 49.14 & (67, 65, 63) & {\bf 21.04} \\ \hline

 \hline \multicolumn{2}{c| }{Average} & 200 & 22.17 & 150 & 25.25 & 18 & 41.69
 & (99, 99, 104) & {\bf 17.67}\\ \hline
\end{tabular}
\end{table*}

\subsection{Segmentation of Real-world Color Images}

\begin{figure*}[!htb]
\begin{center}
\begin{tabular}{ccccc}
\includegraphics[width=\ww, height=\hhh]{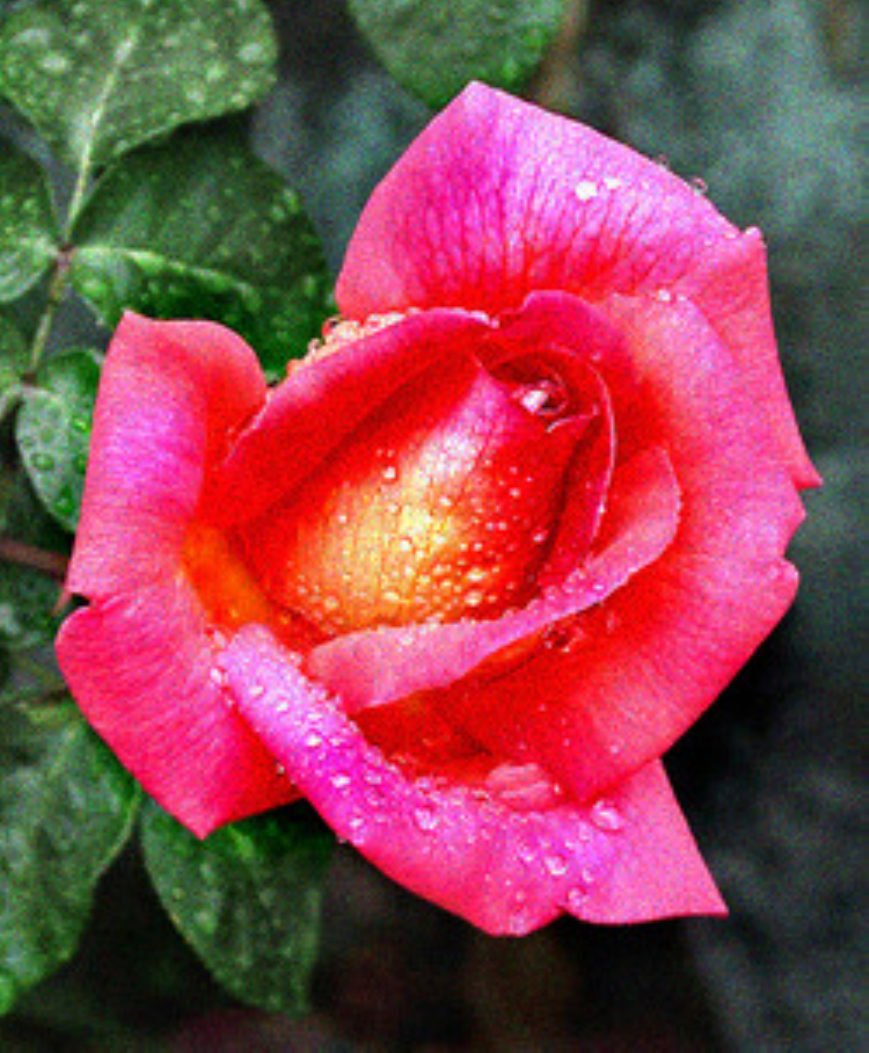} &
\includegraphics[width=\ww, height=\hhh]{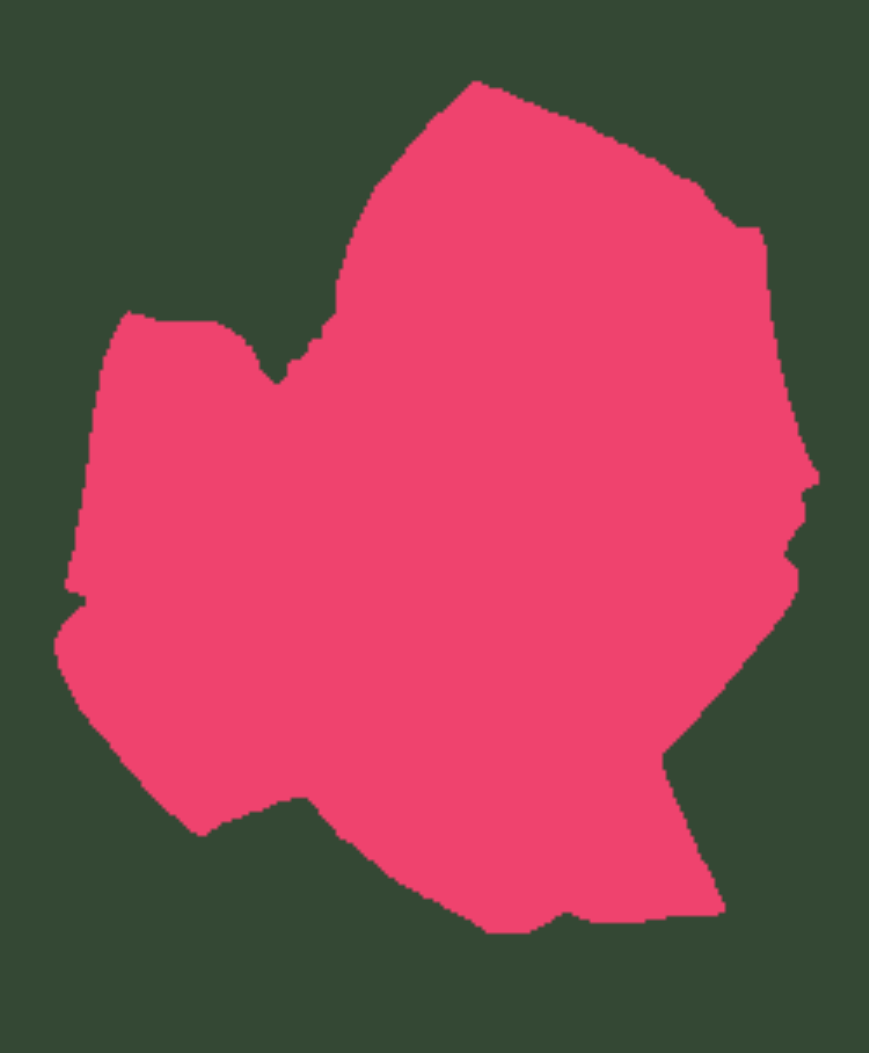} &
\includegraphics[width=\ww, height=\hhh]{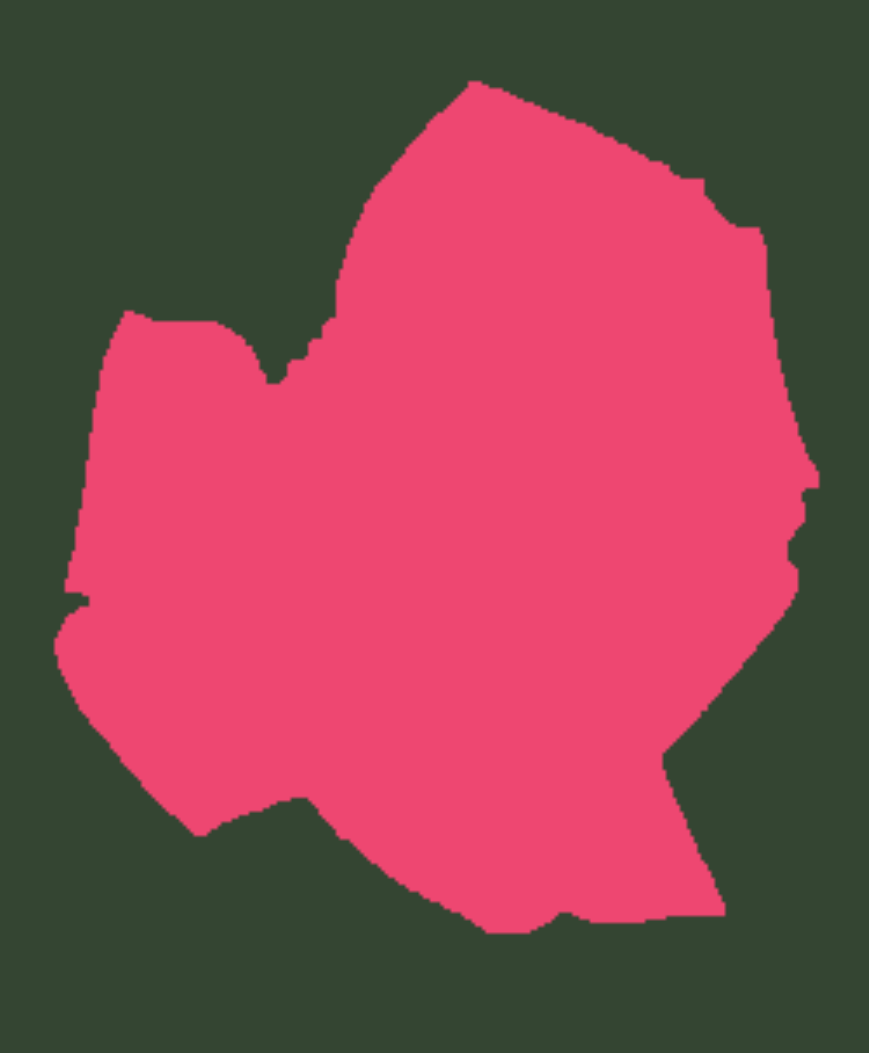} &
\includegraphics[width=\ww, height=\hhh]{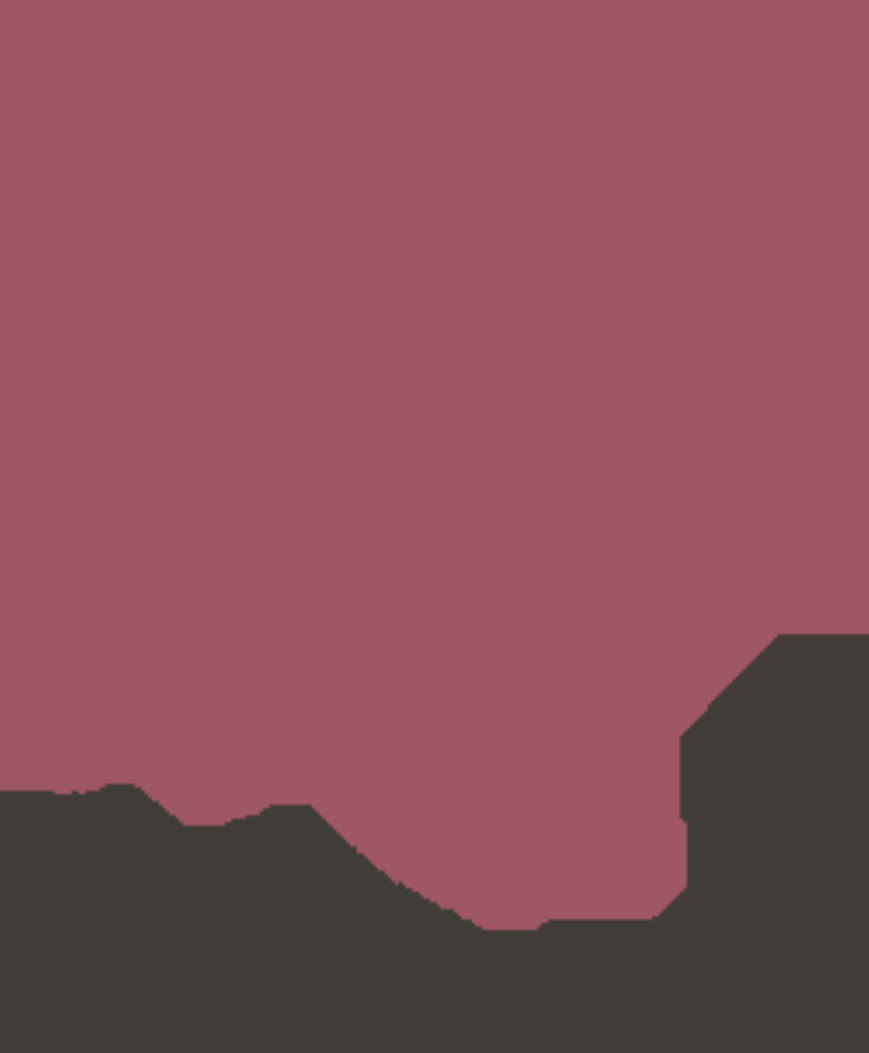} &
\includegraphics[width=\ww, height=\hhh]{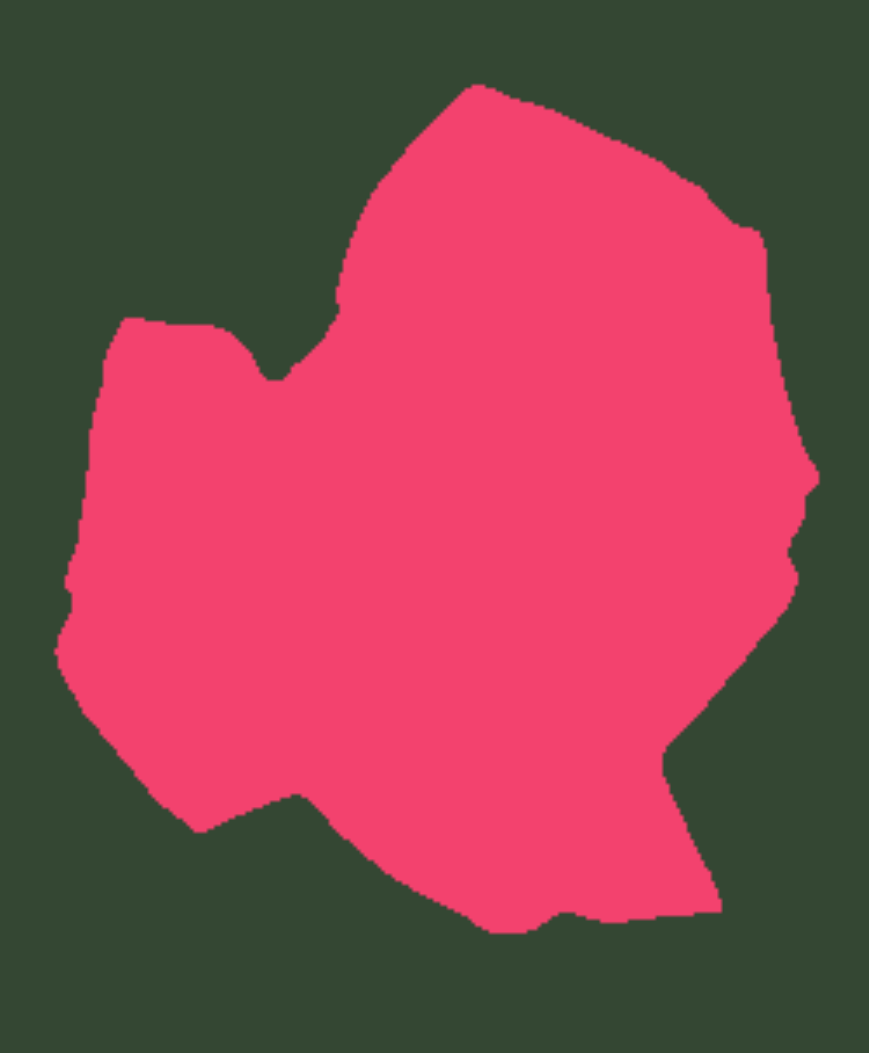} \\
{\small (A) Noisy image} &
{\small (A1) Method \cite{LNZS10}} &
{\small (A2) Method \cite{PCCB09}} &
{\small (A3) Method \cite{SW14}} &
{\small (A4) Ours } \\
\includegraphics[width=\ww, height=\hhh]{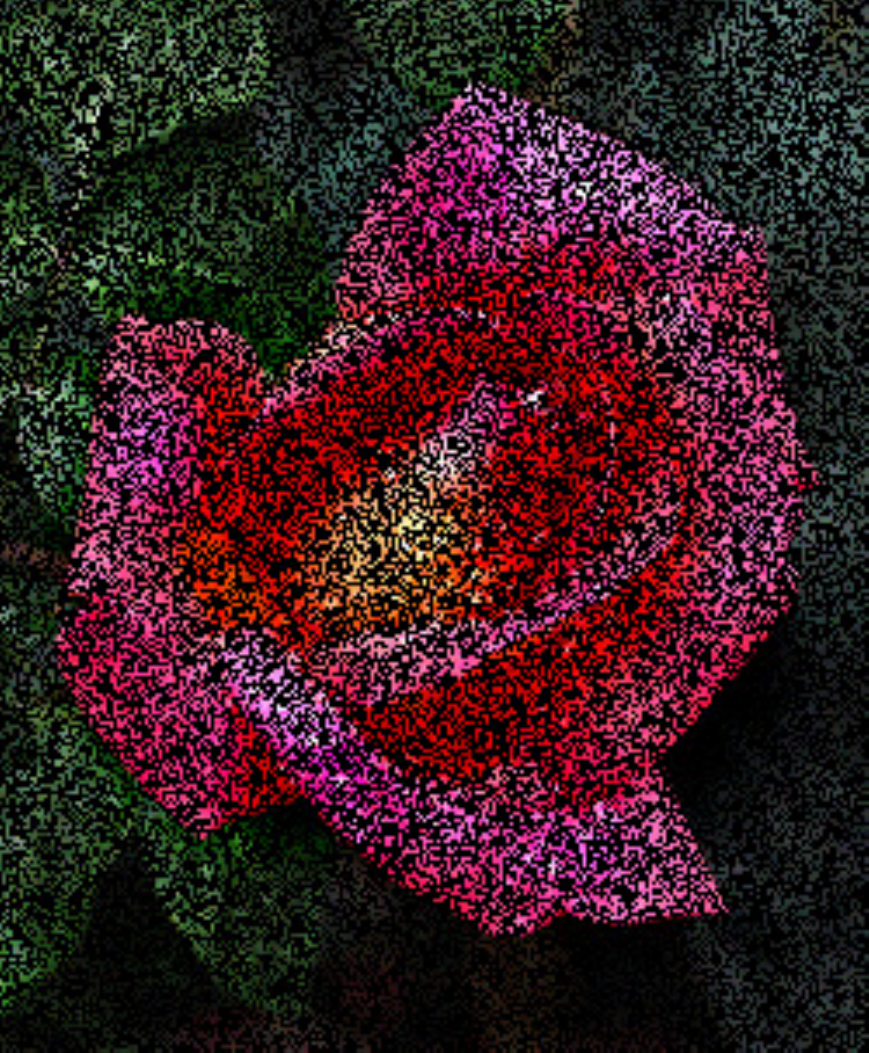} &
\includegraphics[width=\ww, height=\hhh]{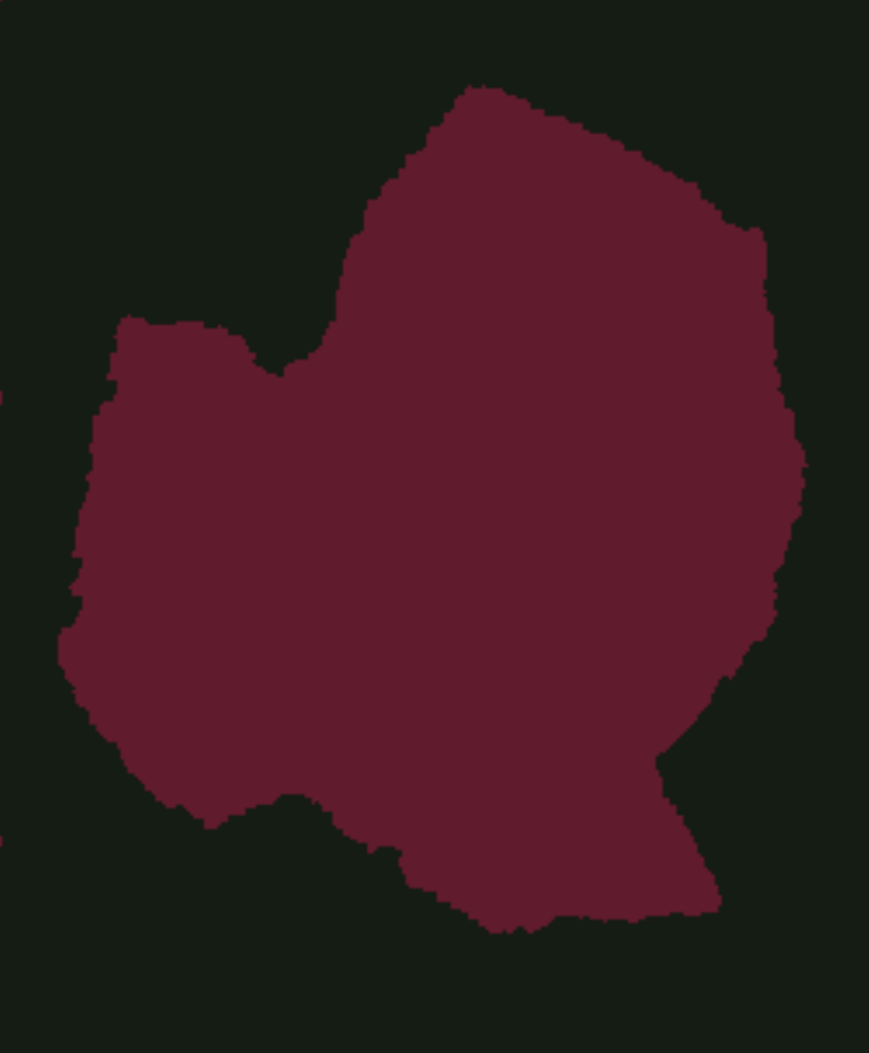} &
\includegraphics[width=\ww, height=\hhh]{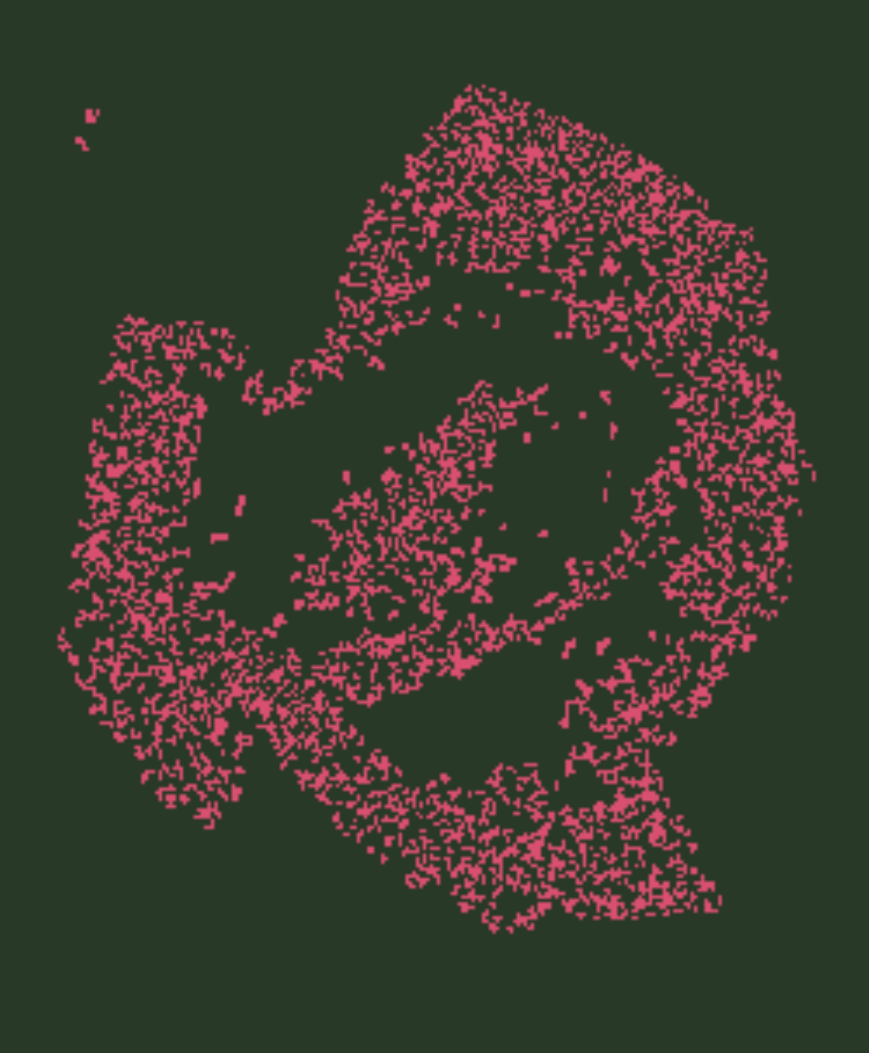} &
\includegraphics[width=\ww, height=\hhh]{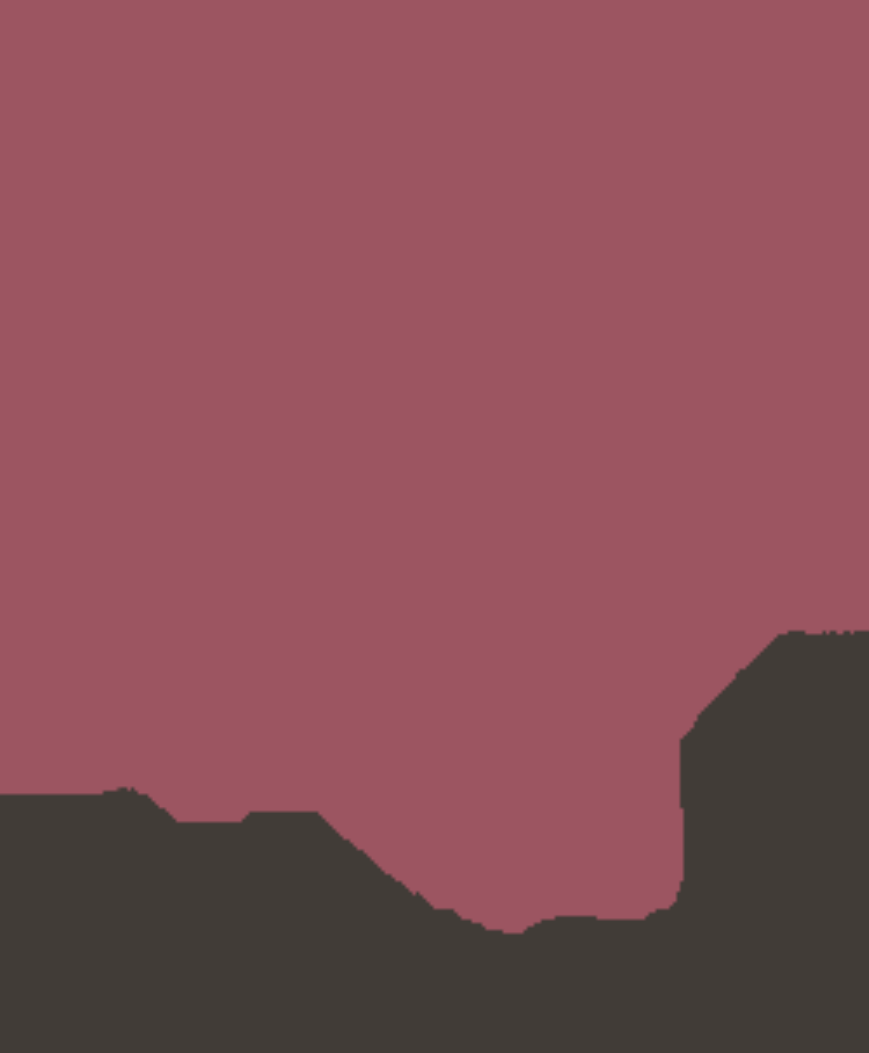} &
\includegraphics[width=\ww, height=\hhh]{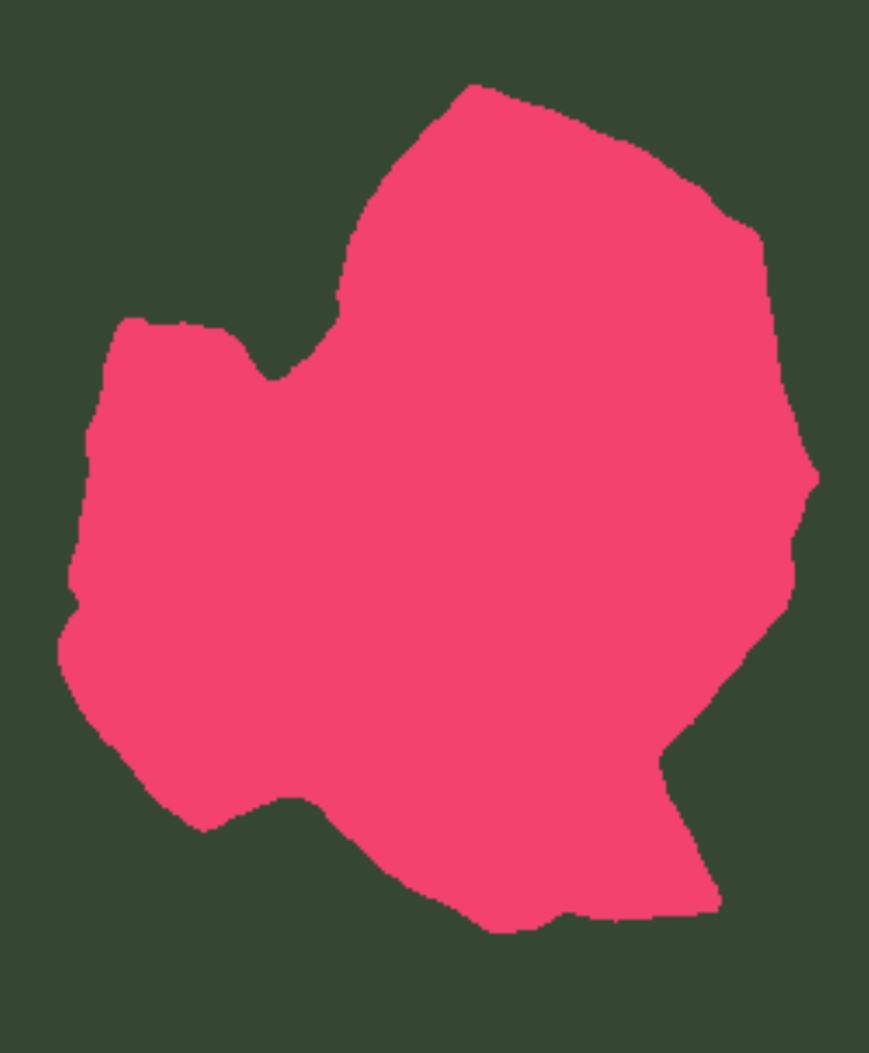} \\
{\small (B) Information} &
{\small (B1) Method \cite{LNZS10}} &
{\small (B2) Method \cite{PCCB09}} &
{\small (B3) Method \cite{SW14}} &
{\small (B4) Ours } \vspace{-0.05in} \\
{\small loss + noise} & & & & \\
\includegraphics[width=\ww, height=\hhh]{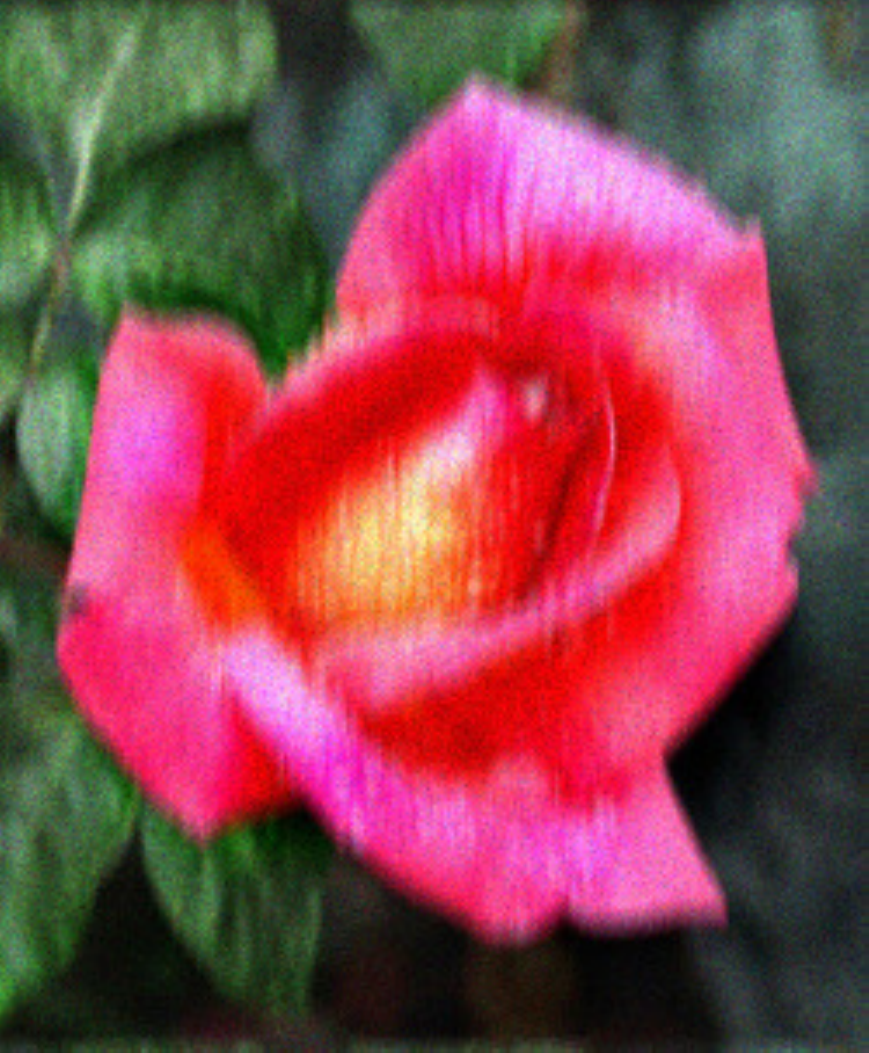} &
\includegraphics[width=\ww, height=\hhh]{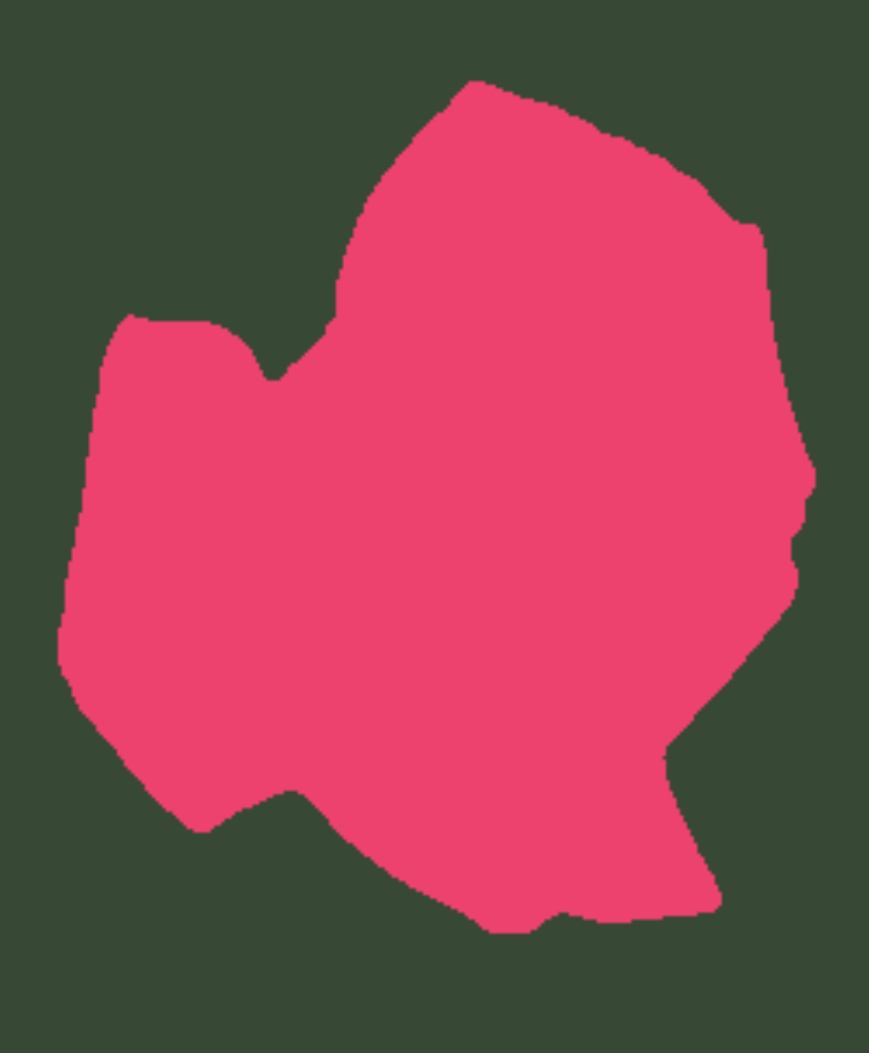} &
\includegraphics[width=\ww, height=\hhh]{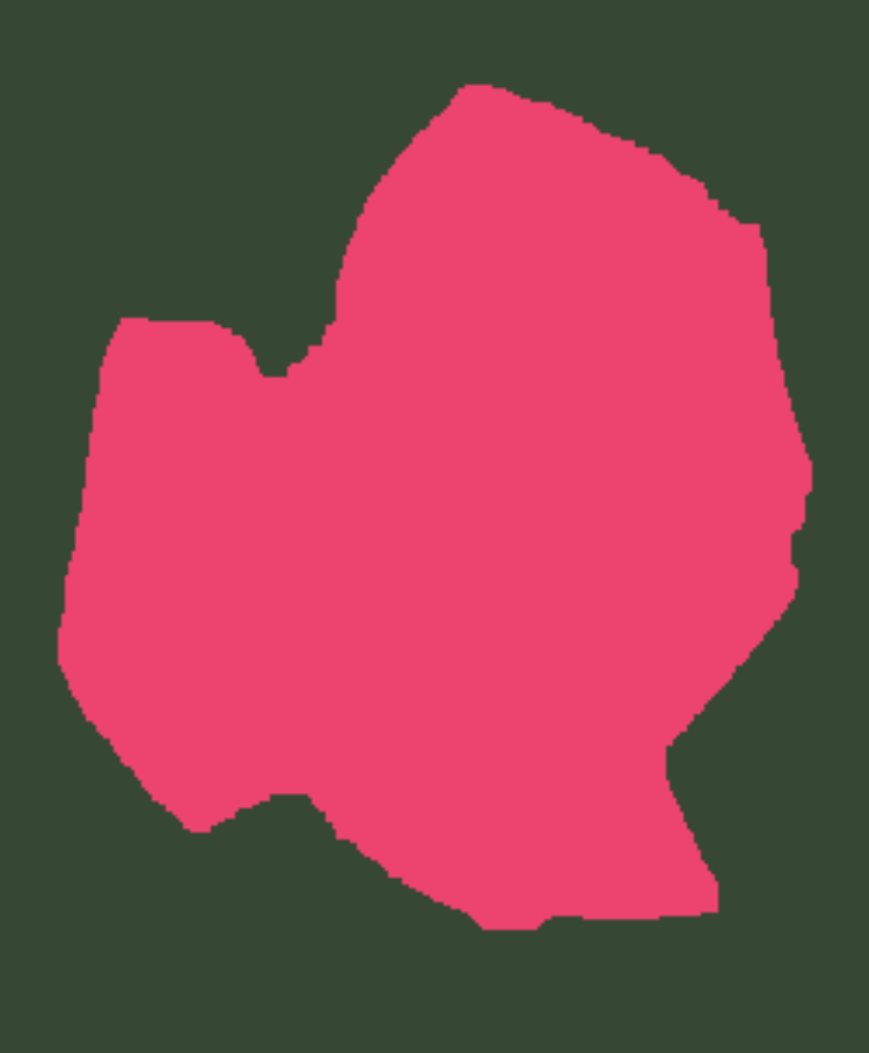} &
\includegraphics[width=\ww, height=\hhh]{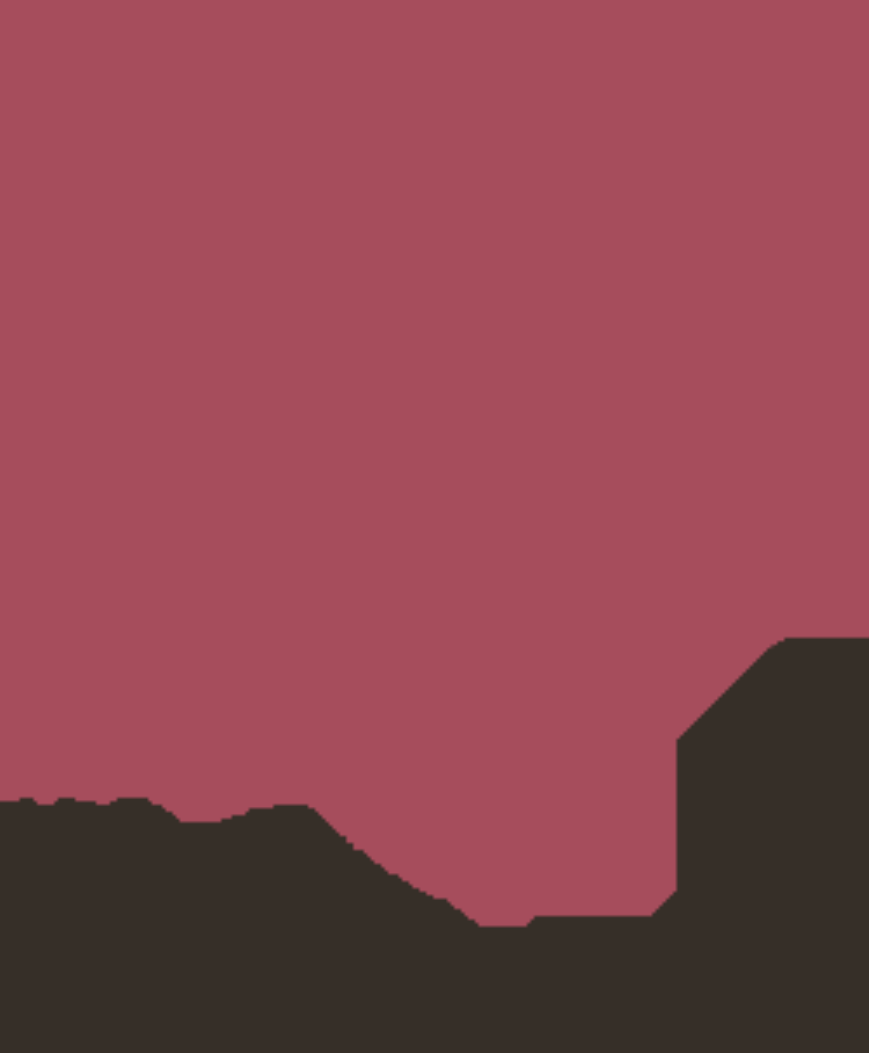} &
\includegraphics[width=\ww, height=\hhh]{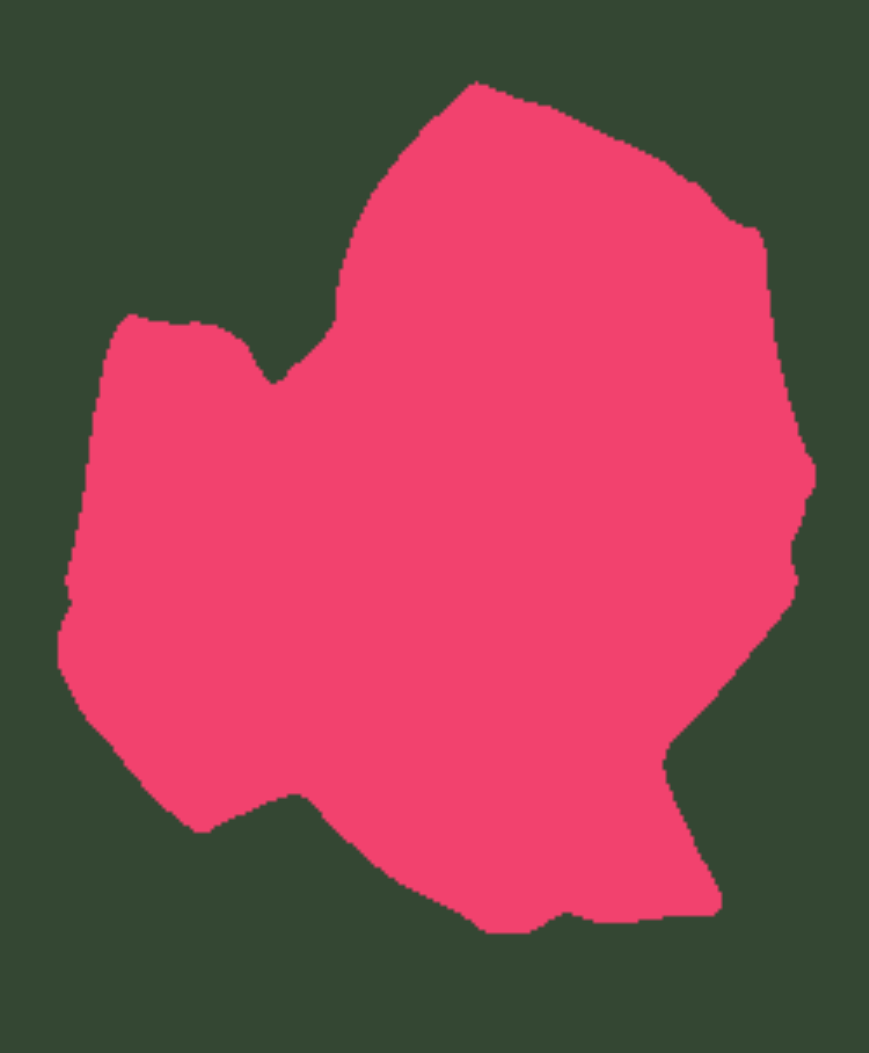} \\
{\small (C) Blur + noise} &
{\small (C1) Method \cite{LNZS10}} &
{\small (C2) Method \cite{PCCB09}} &
{\small (C3) Method \cite{SW14}} &
{\small (C4) Ours }
\end{tabular}
\end{center}
\caption{Two-phase rose segmentation (size: $303\times 250$).
(A): Given Poisson noisy image;
(B): Given Poisson noisy image with $60\%$ information loss;
(C): Given blurry image with Poisson noise; (A1-A4), (B1-B4) and (C1-C4): Results of methods
\cite{LNZS10}, \cite{PCCB09}, \cite{SW14}, and our SLaT on (A), (B) and (C), respectively.
}\label{twophase-color-rose}
\end{figure*}

\begin{figure*}[!htb]
\begin{center}
\begin{tabular}{ccccc}
\includegraphics[width=\ww, height=\hh]{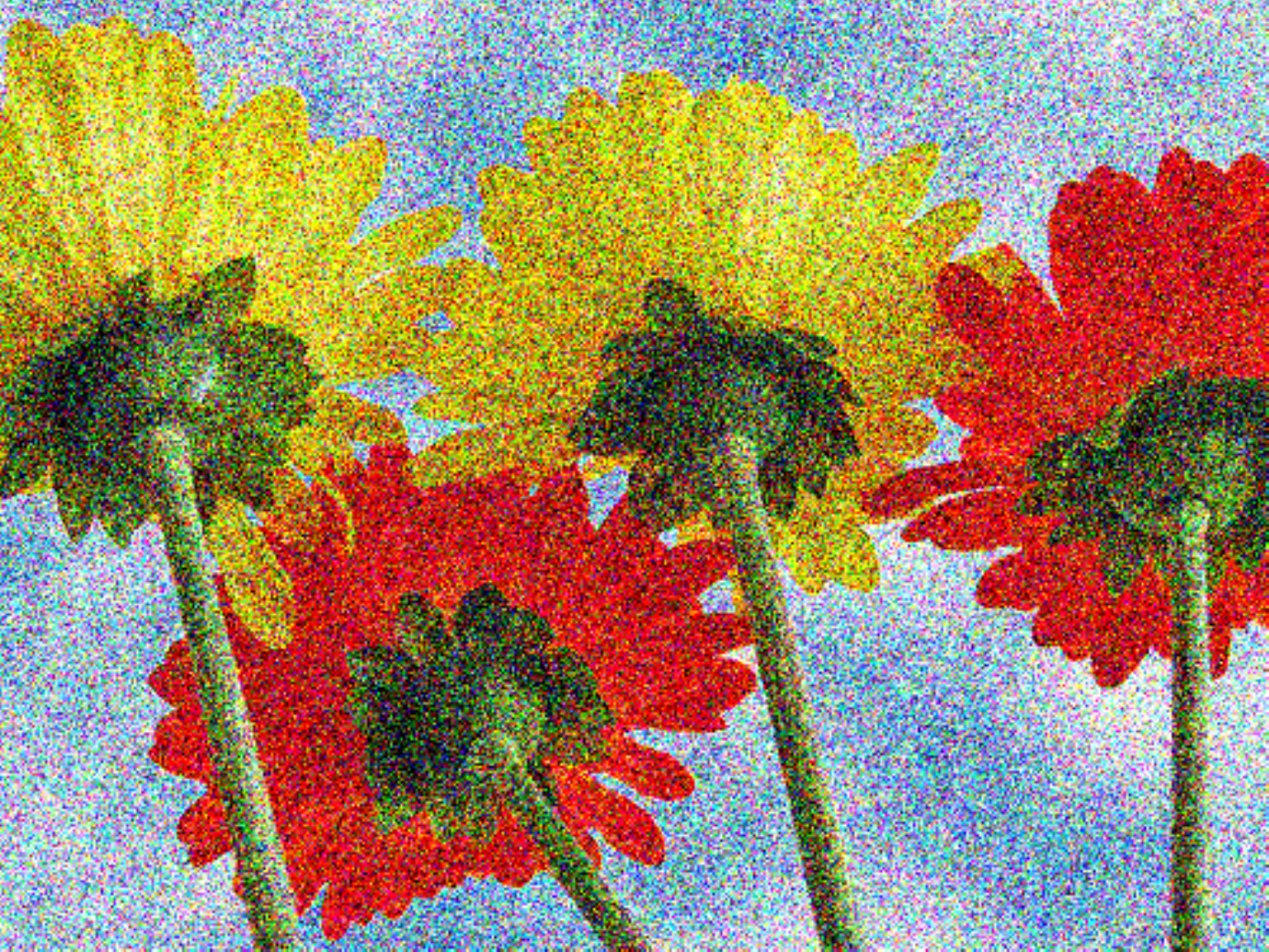} &
\includegraphics[width=\ww, height=\hh]{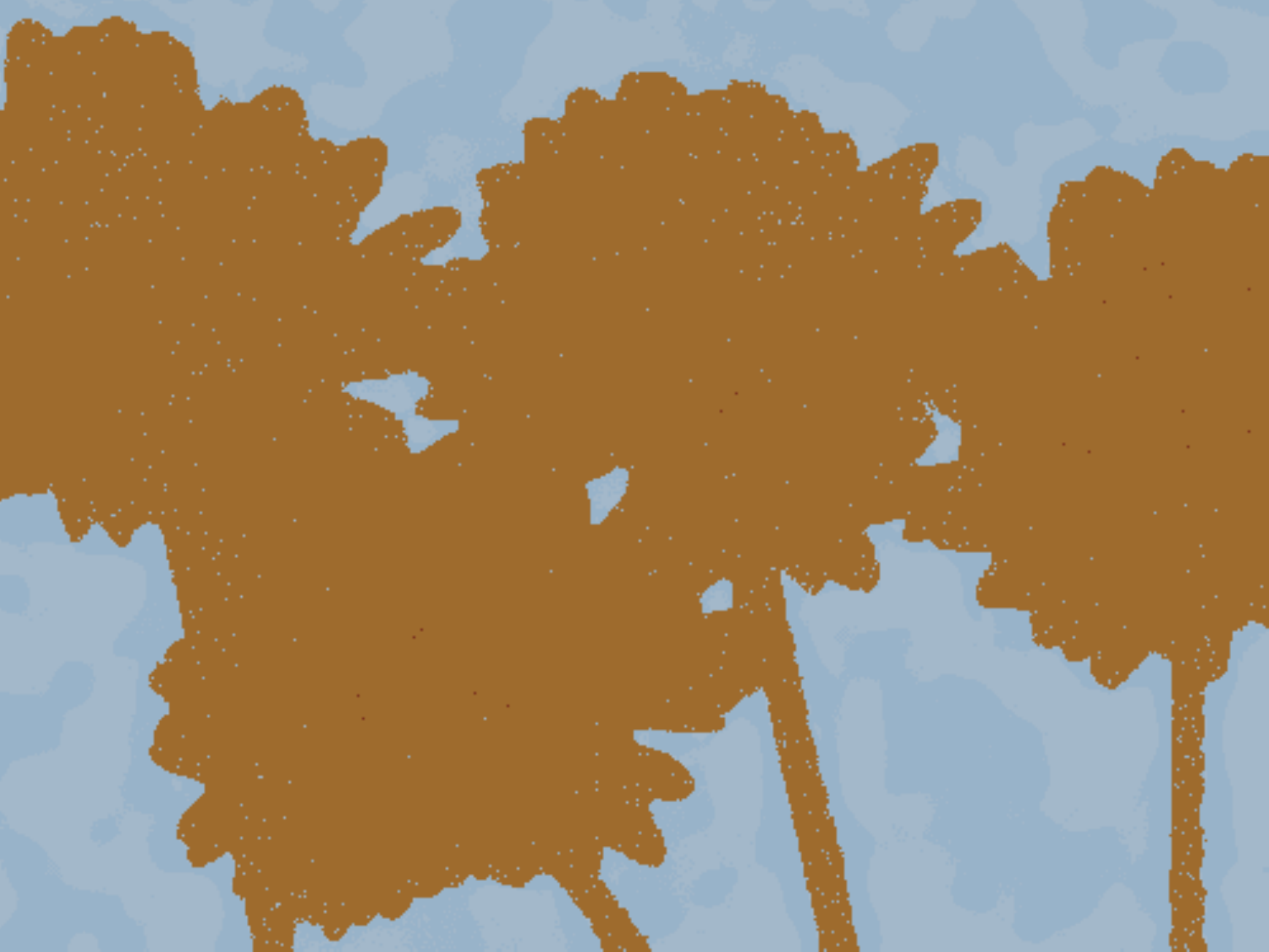} &
\includegraphics[width=\ww, height=\hh]{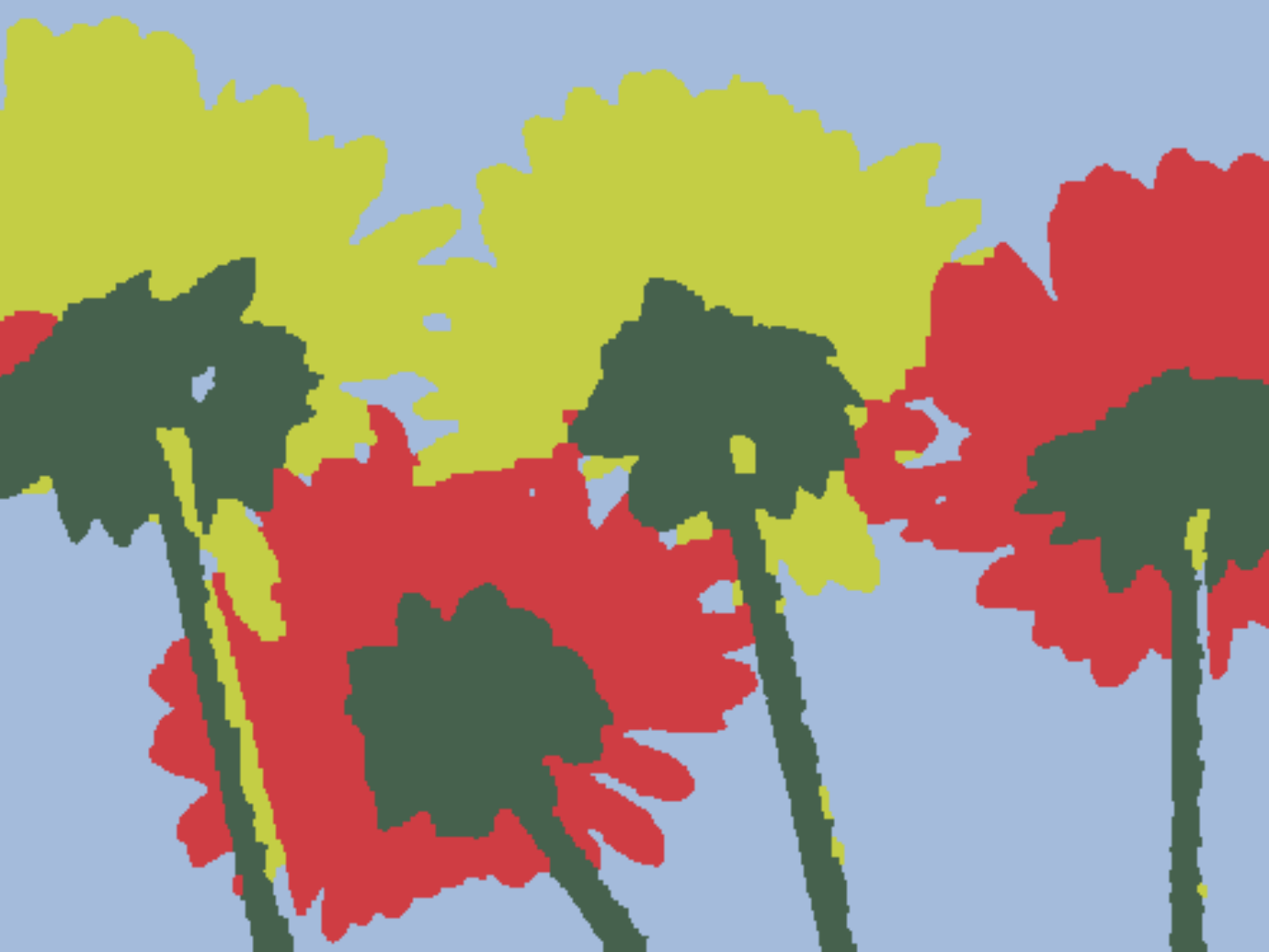} &
\includegraphics[width=\ww, height=\hh]{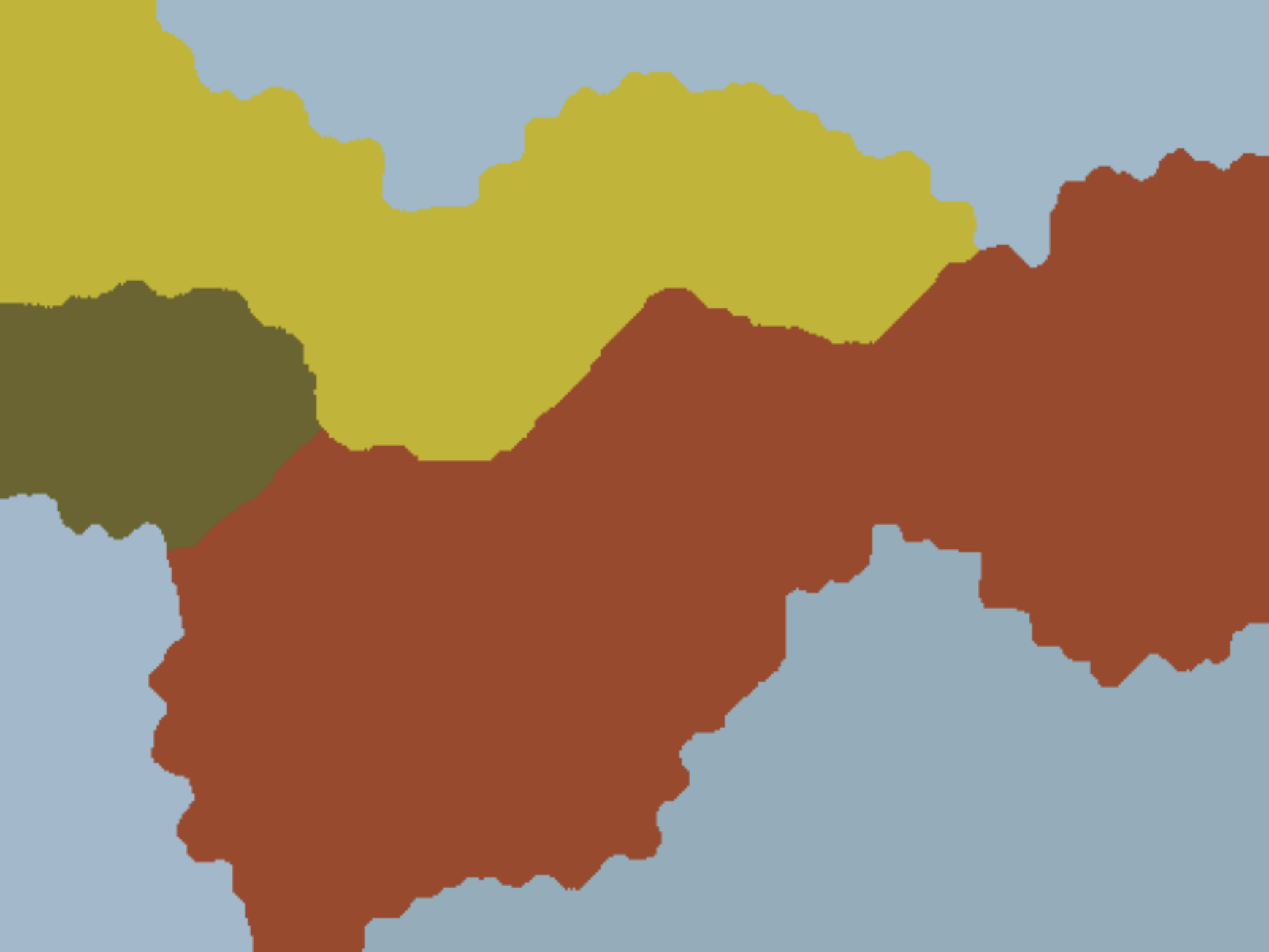} &
\includegraphics[width=\ww, height=\hh]{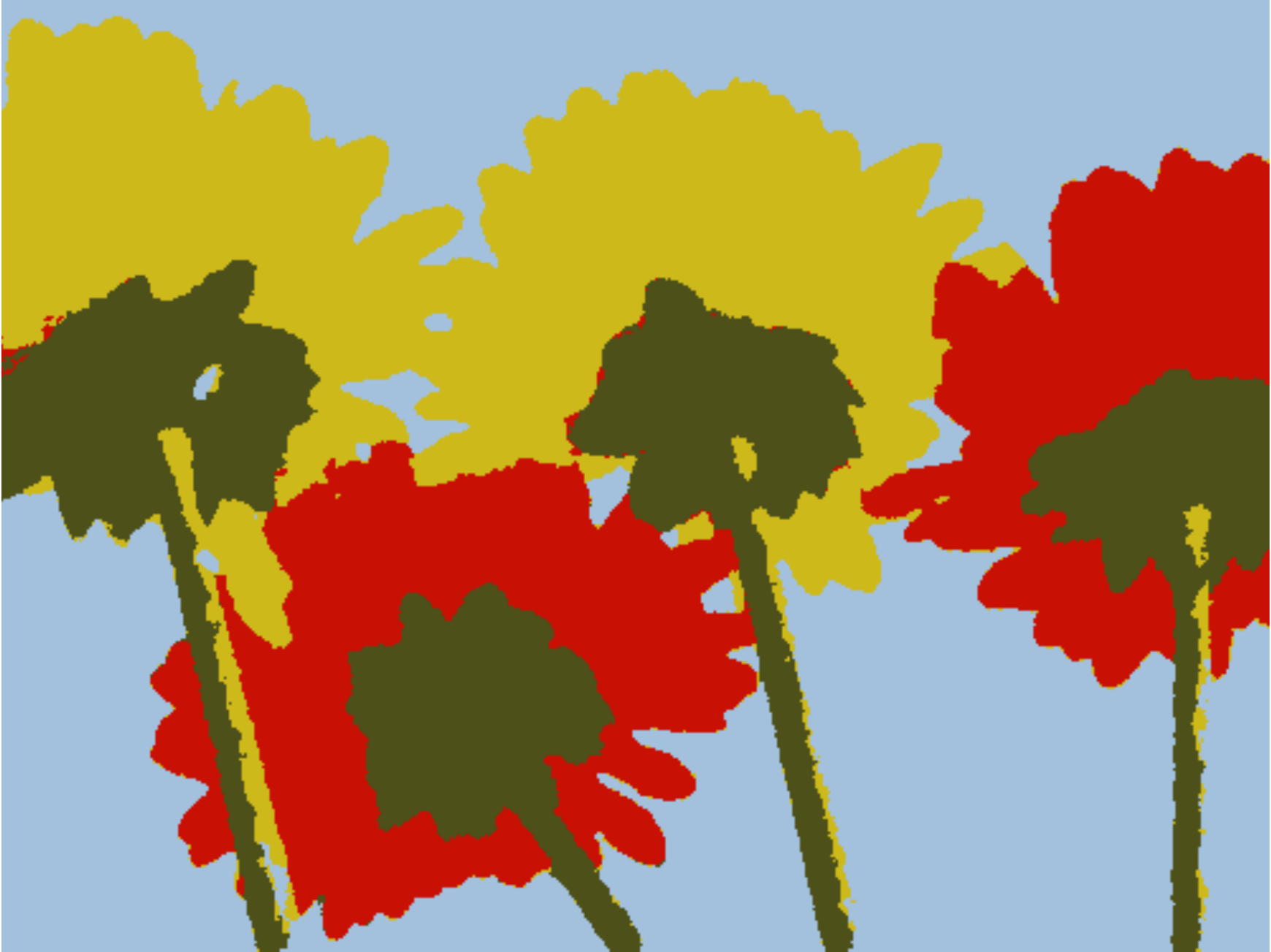} \\
{\small (A) Noisy image} &
{\small (A1) Method \cite{LNZS10}} &
{\small (A2) Method \cite{PCCB09}} &
{\small (A3) Method \cite{SW14}} &
{\small (A4) Ours } \\
\includegraphics[width=\ww, height=\hh]{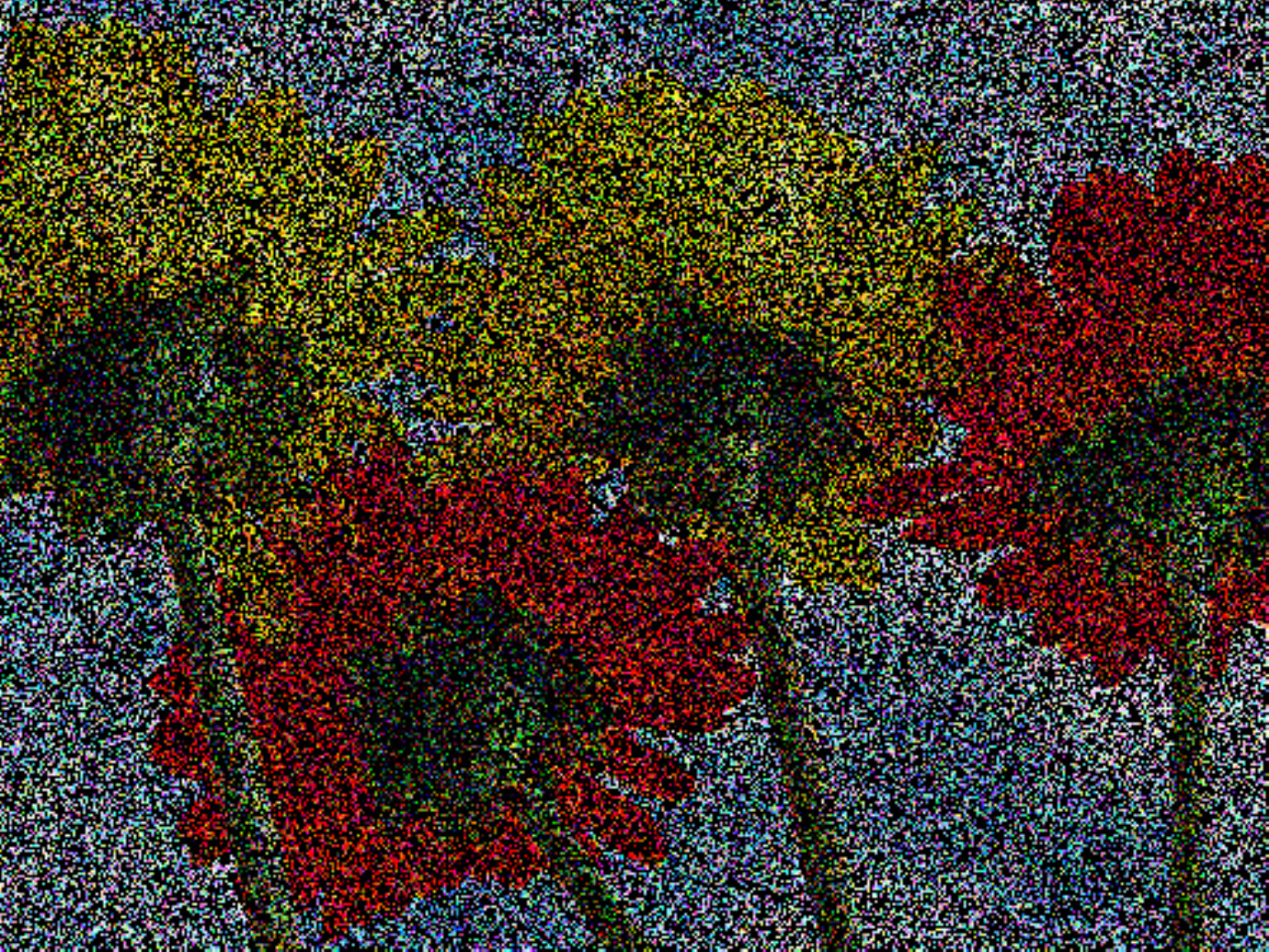} &
\includegraphics[width=\ww, height=\hh]{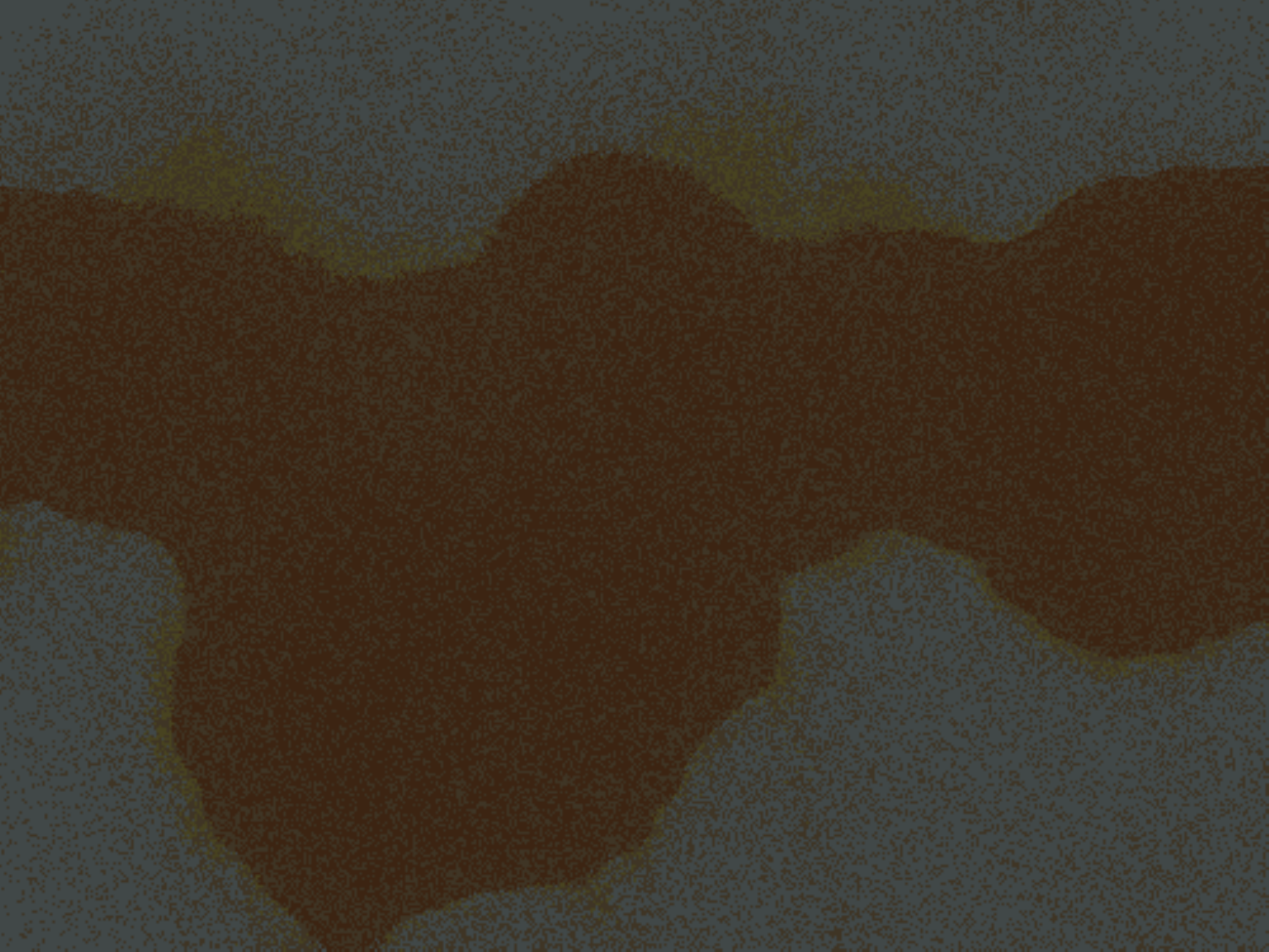} &
\includegraphics[width=\ww, height=\hh]{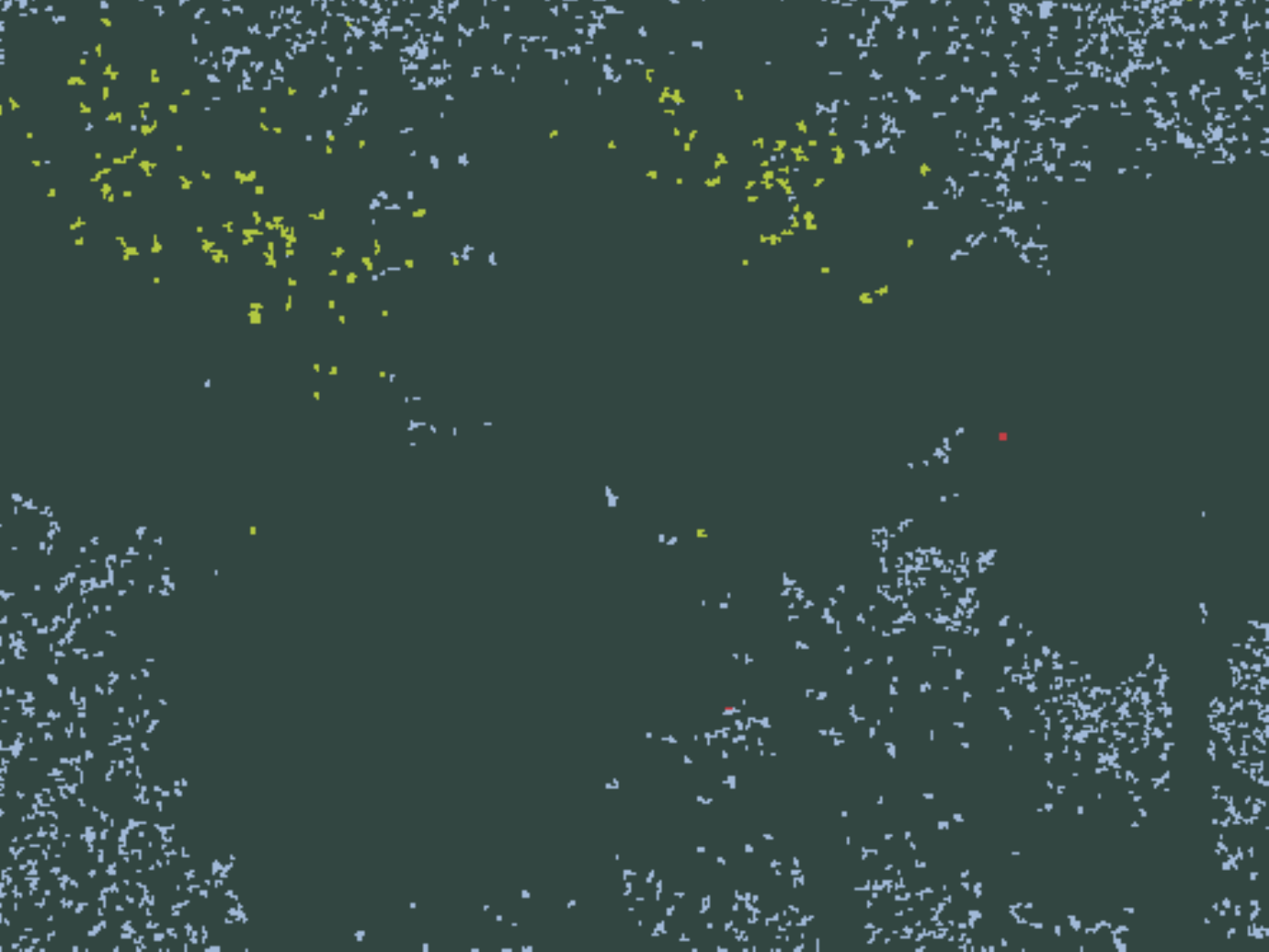} &
\includegraphics[width=\ww, height=\hh]{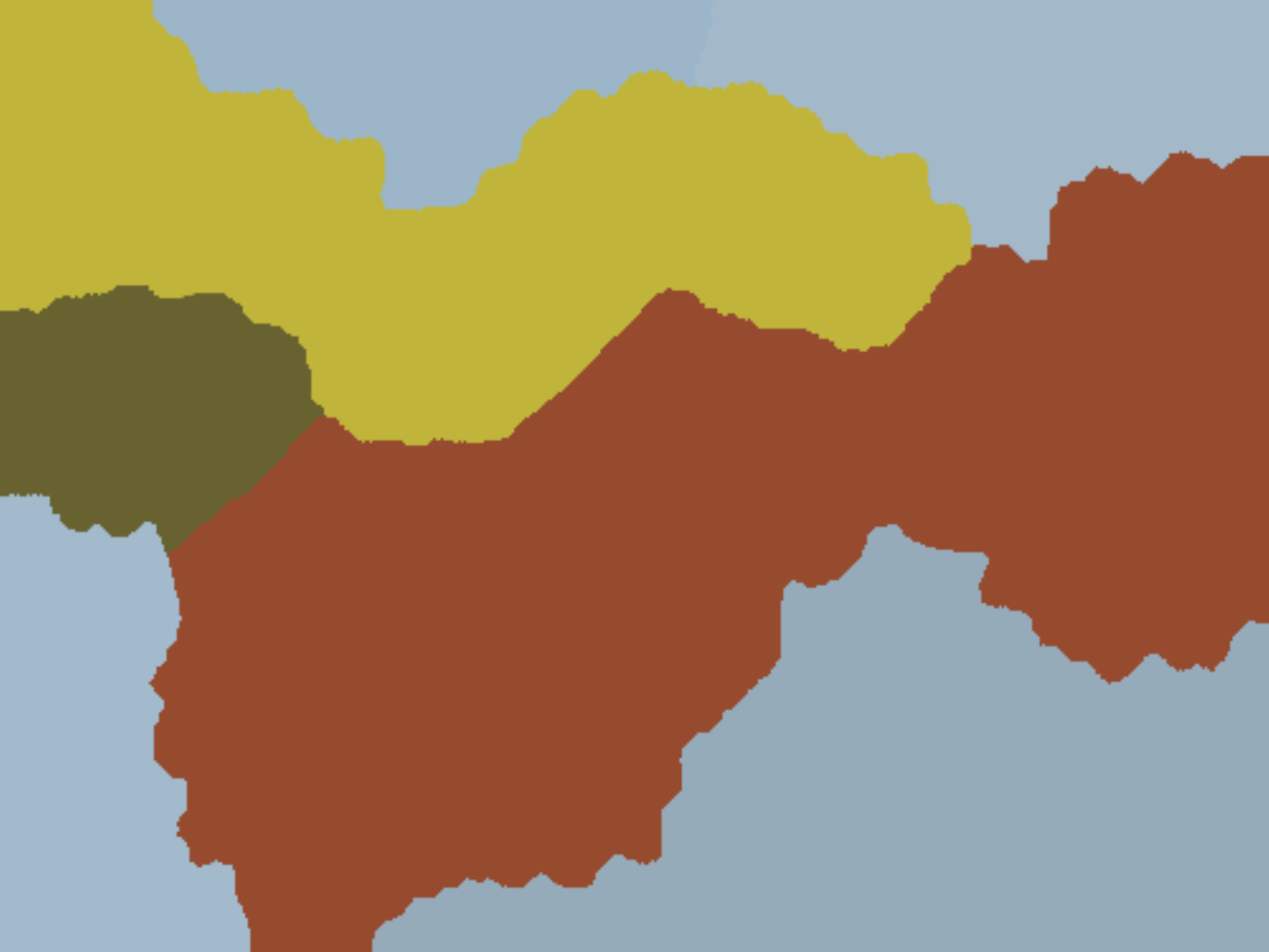} &
\includegraphics[width=\ww, height=\hh]{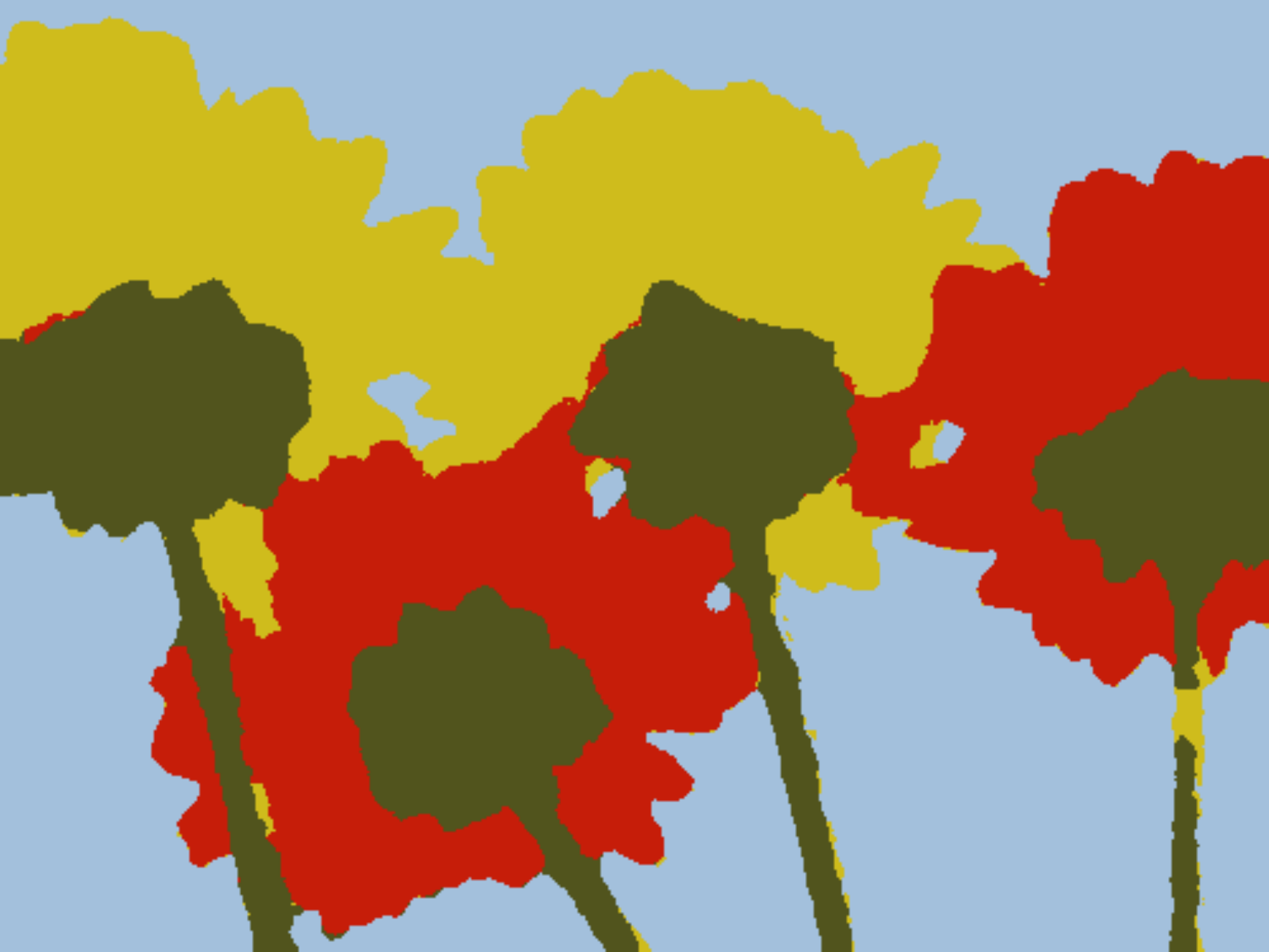} \\
{\small (B) Information} &
{\small (B1) Method \cite{LNZS10}} &
{\small (B2) Method \cite{PCCB09}} &
{\small (B3) Method \cite{SW14}} &
{\small (B4) Ours } \vspace{-0.05in} \\
{\small loss + noise} & & & & \\
\includegraphics[width=\ww, height=\hh]{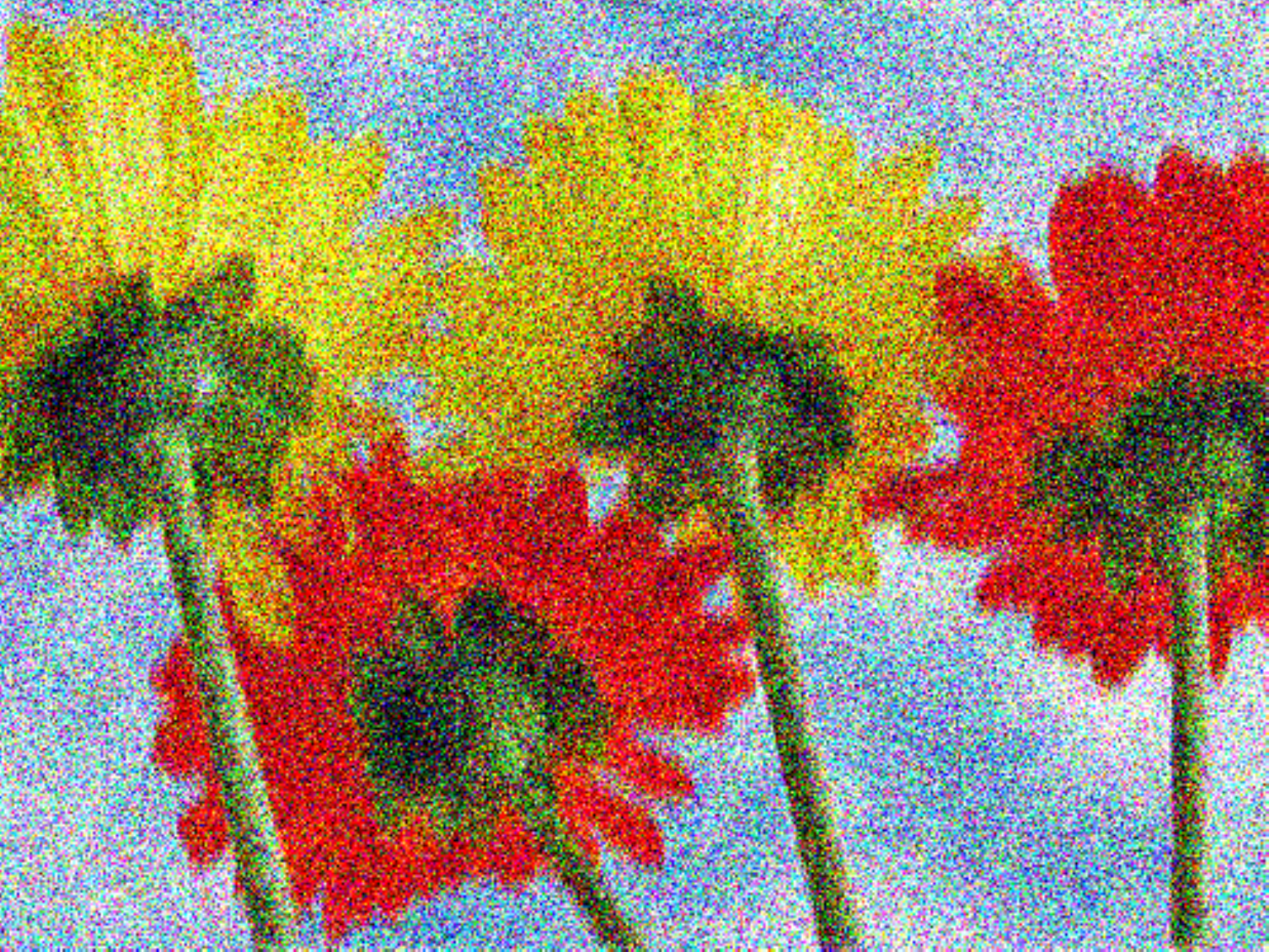} &
\includegraphics[width=\ww, height=\hh]{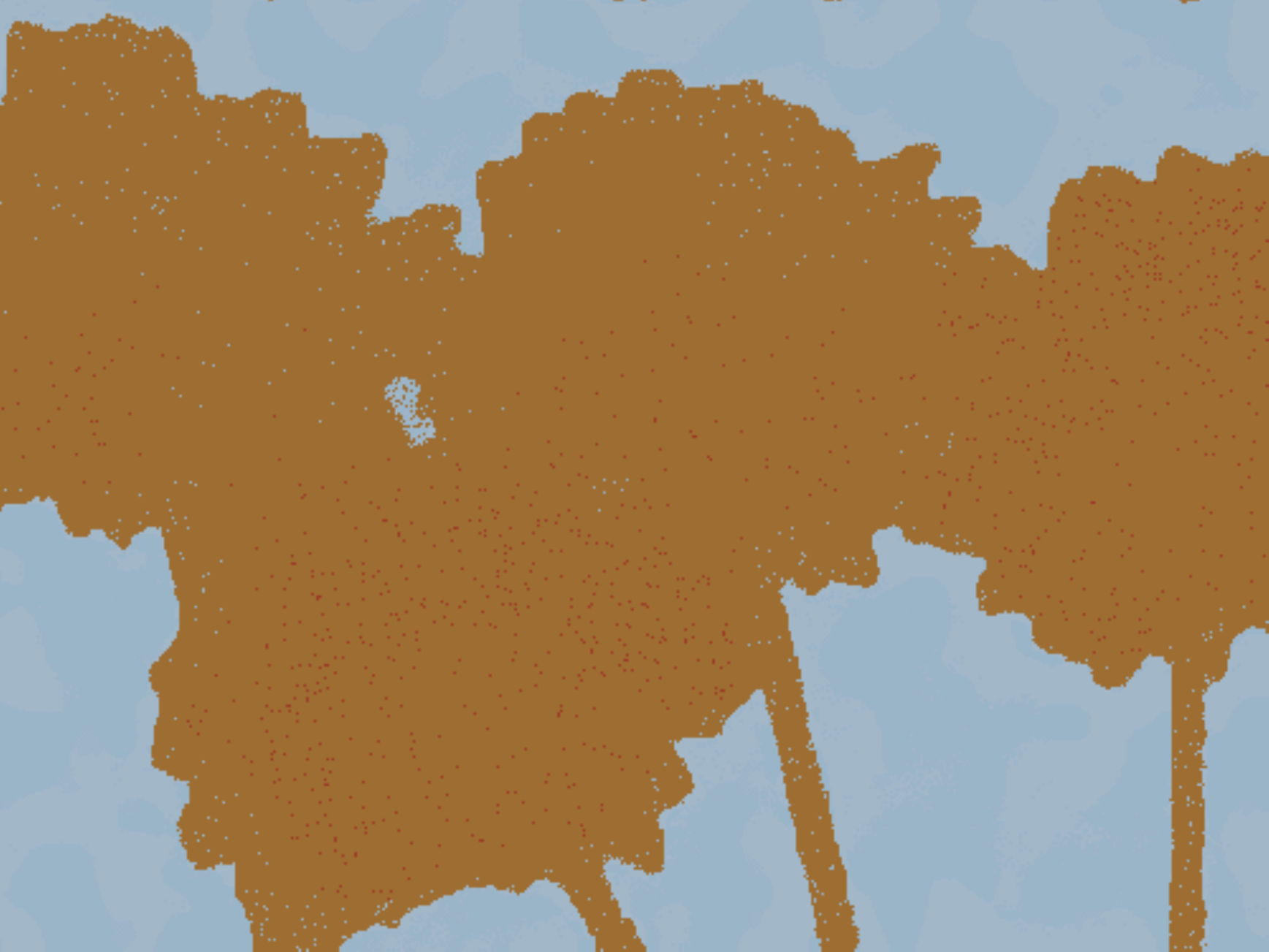} &
\includegraphics[width=\ww, height=\hh]{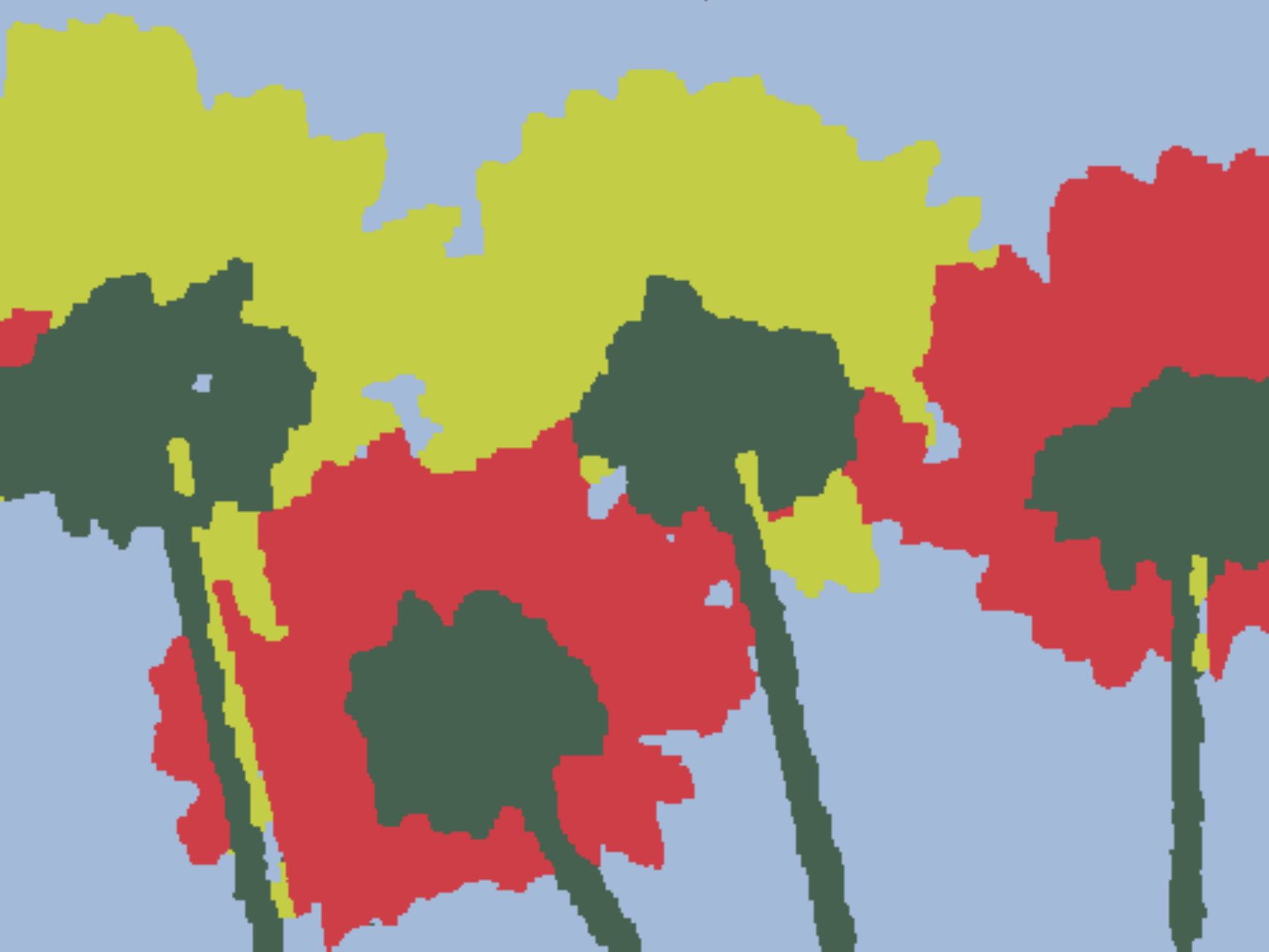} &
\includegraphics[width=\ww, height=\hh]{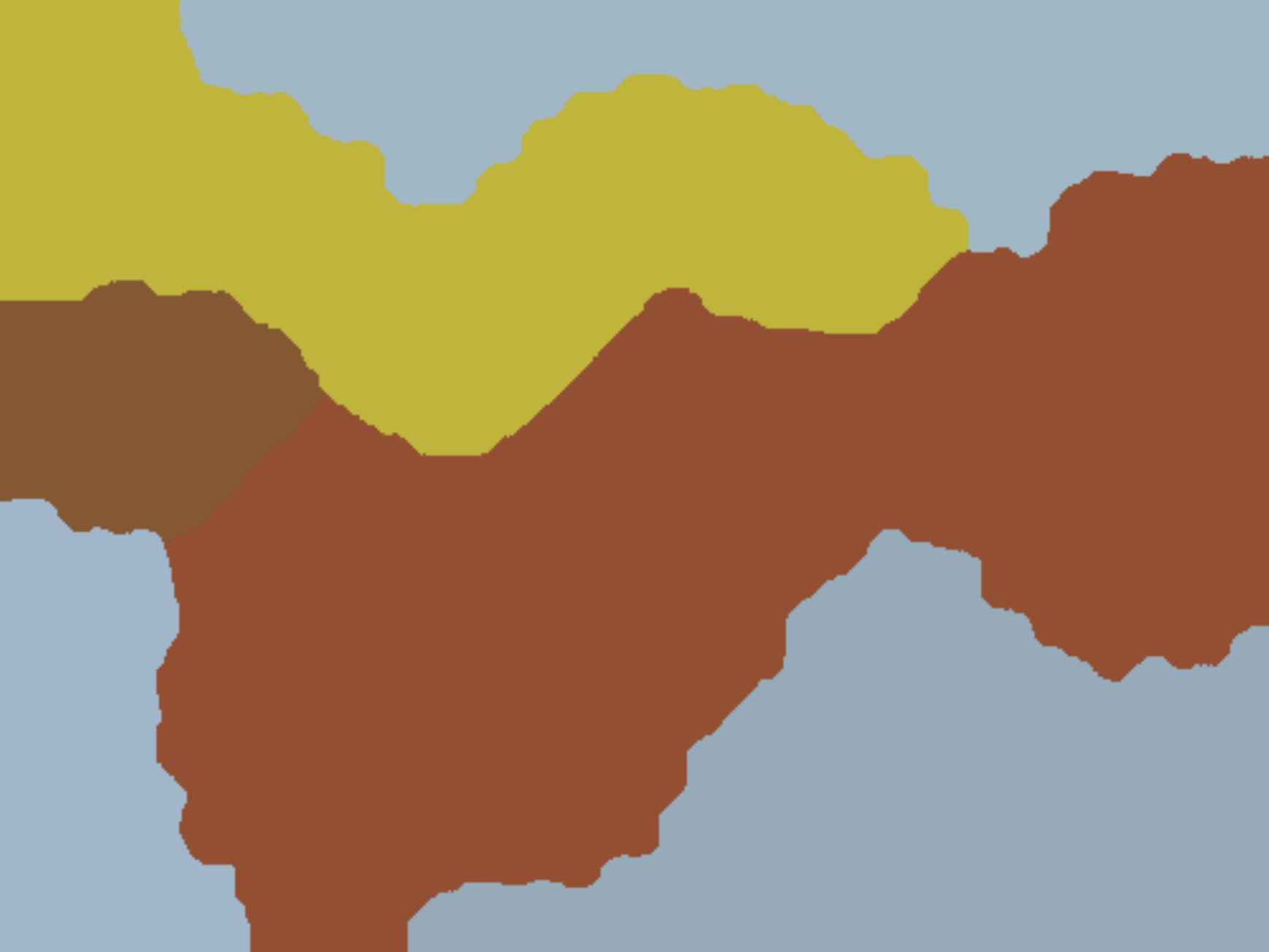} &
\includegraphics[width=\ww, height=\hh]{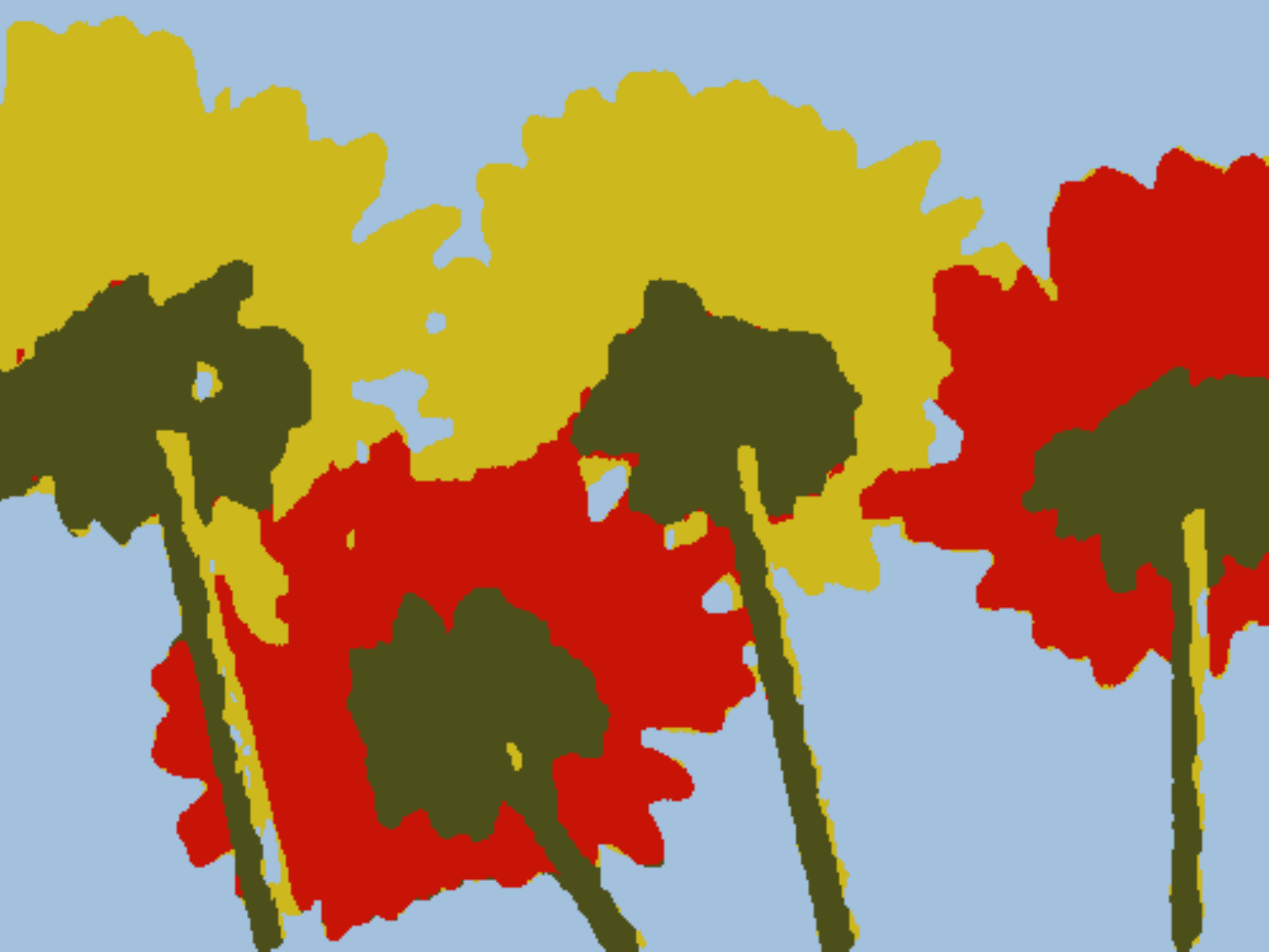} \\
{\small (C) Blur + noise} &
{\small (C1) Method \cite{LNZS10}} &
{\small (C2) Method \cite{PCCB09}} &
{\small (C3) Method \cite{SW14}} &
{\small (C4) Ours }
\end{tabular}
\end{center}
\caption{Four-phase sunflower segmentation (size: $375\times 500$).
(A): Given Gaussian noisy image with mean 0 and variance 0.1;
(B): Given Gaussian noisy image with $60\%$ information loss;
(C): Given blurry image with Gaussian noise;
(A1-A4), (B1-B4) and (C1-C4): Results of methods \cite{LNZS10},
\cite{PCCB09}, \cite{SW14}, and our SLaT on (A), (B) and (C), respectively.
}\label{fourphase-color-flower}
\end{figure*}

\begin{figure*}[!htb]
\begin{center}
\begin{tabular}{ccccc}
\includegraphics[width=\ww, height=\hh]{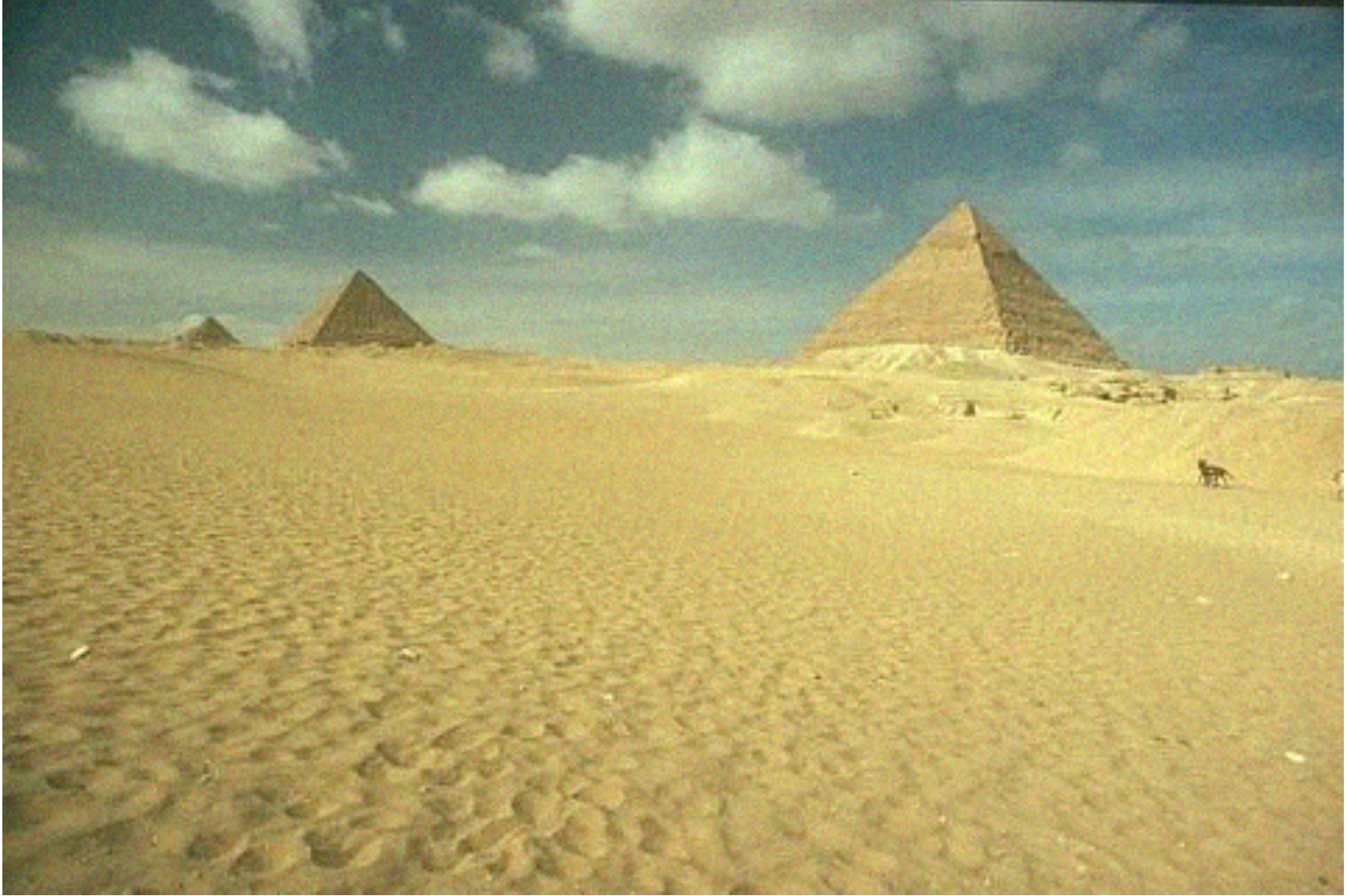} &
\includegraphics[width=\ww, height=\hh]{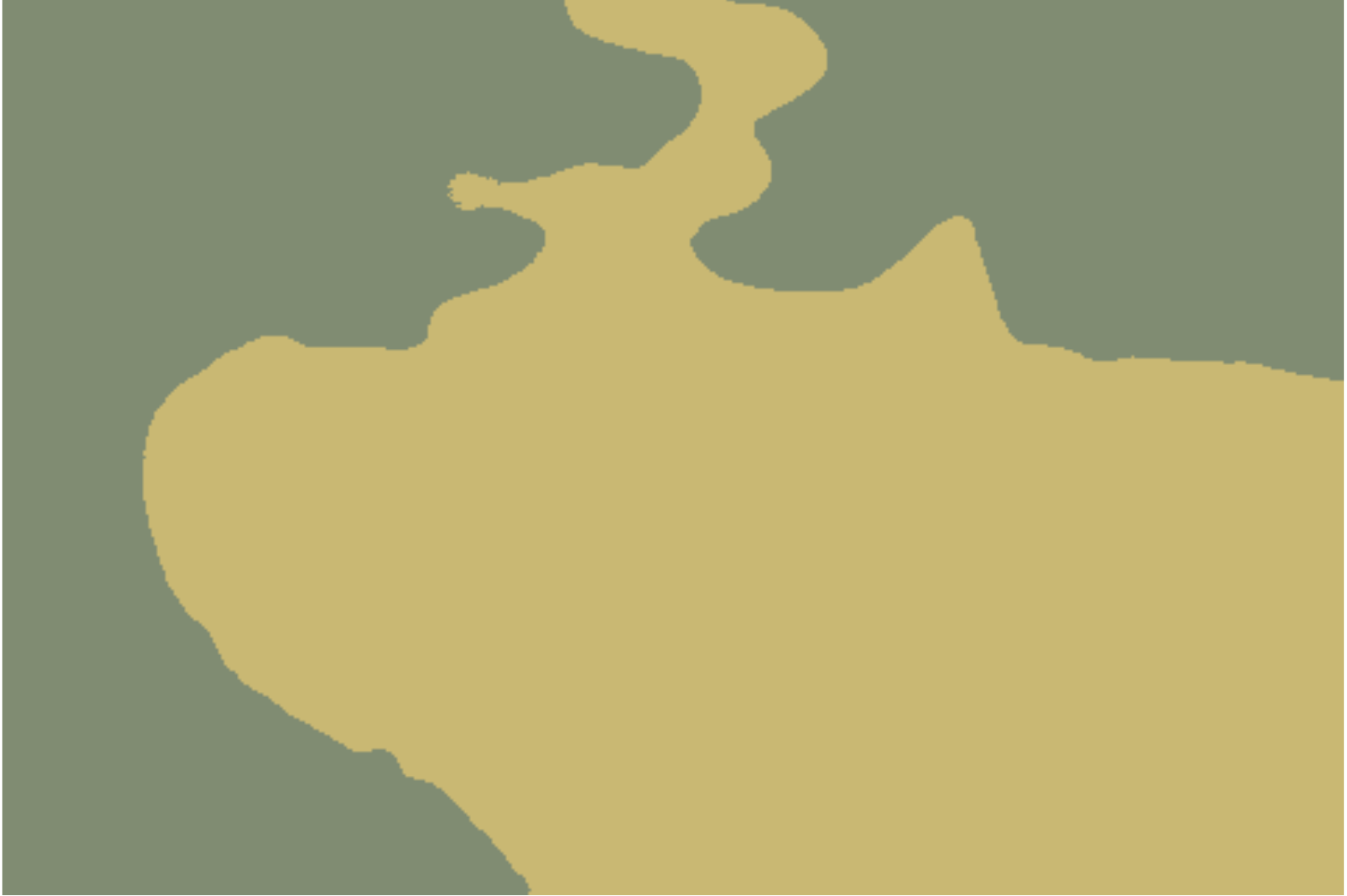} &
\includegraphics[width=\ww, height=\hh]{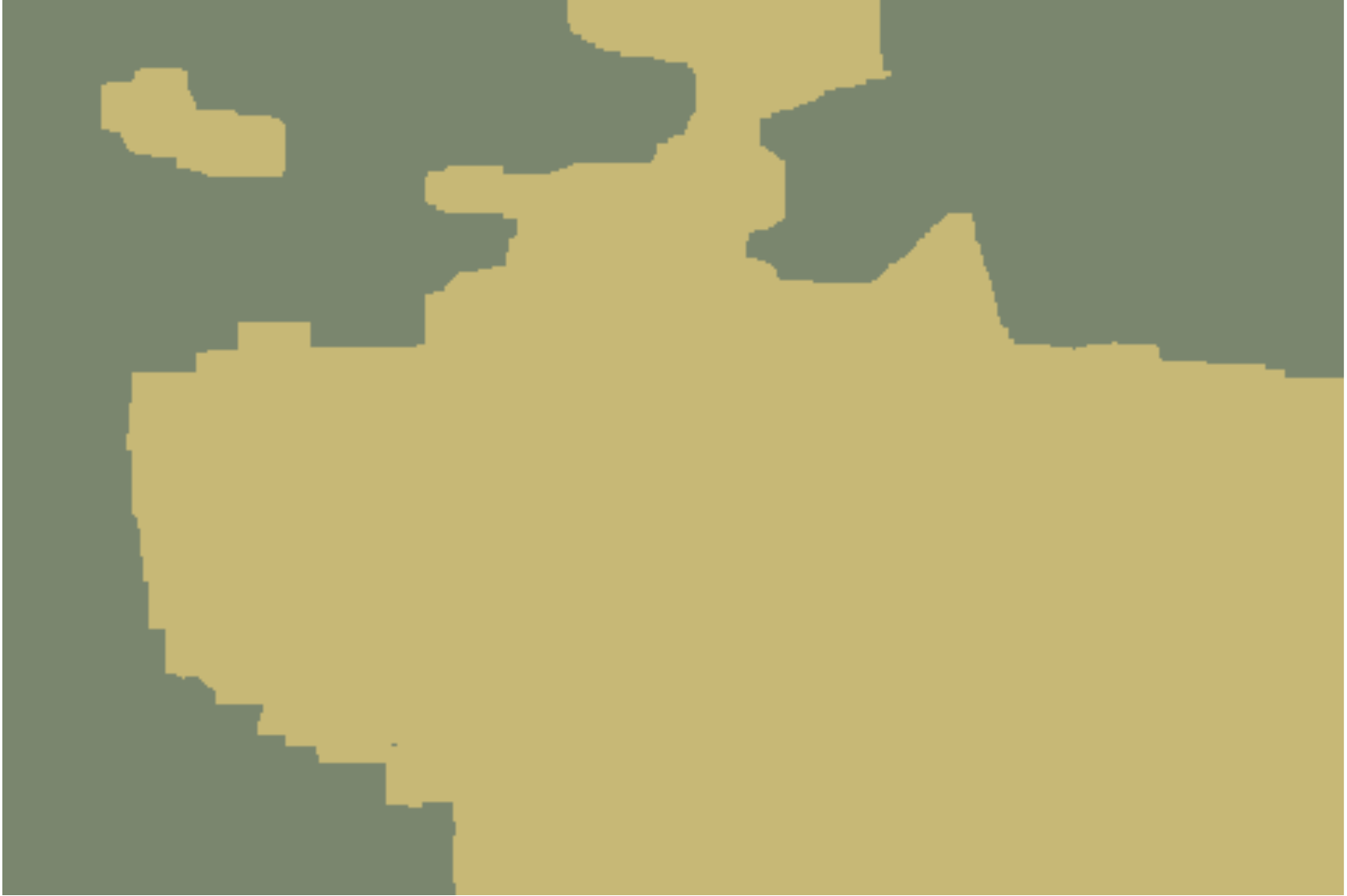} &
\includegraphics[width=\ww, height=\hh]{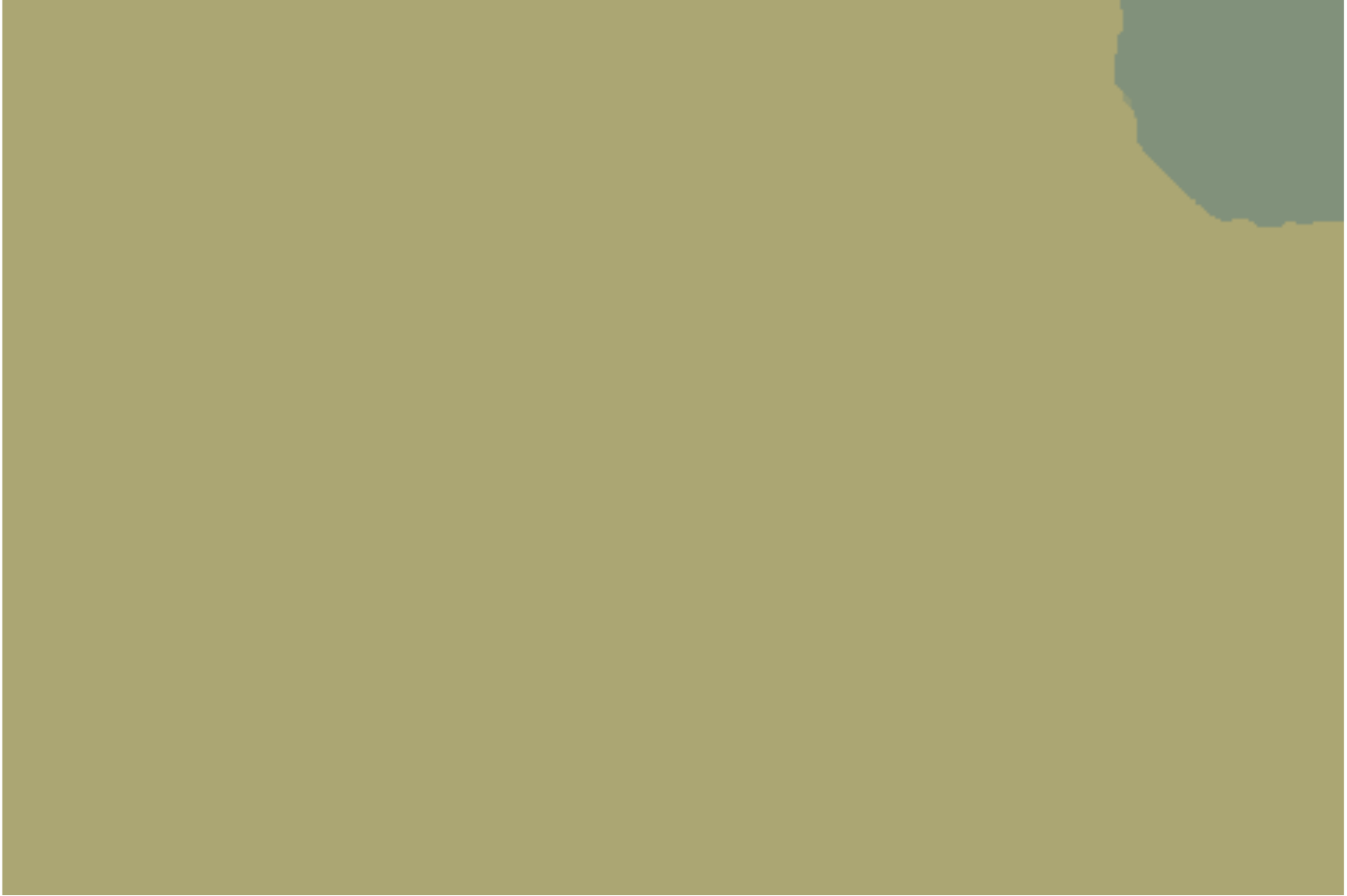} &
\includegraphics[width=\ww, height=\hh]{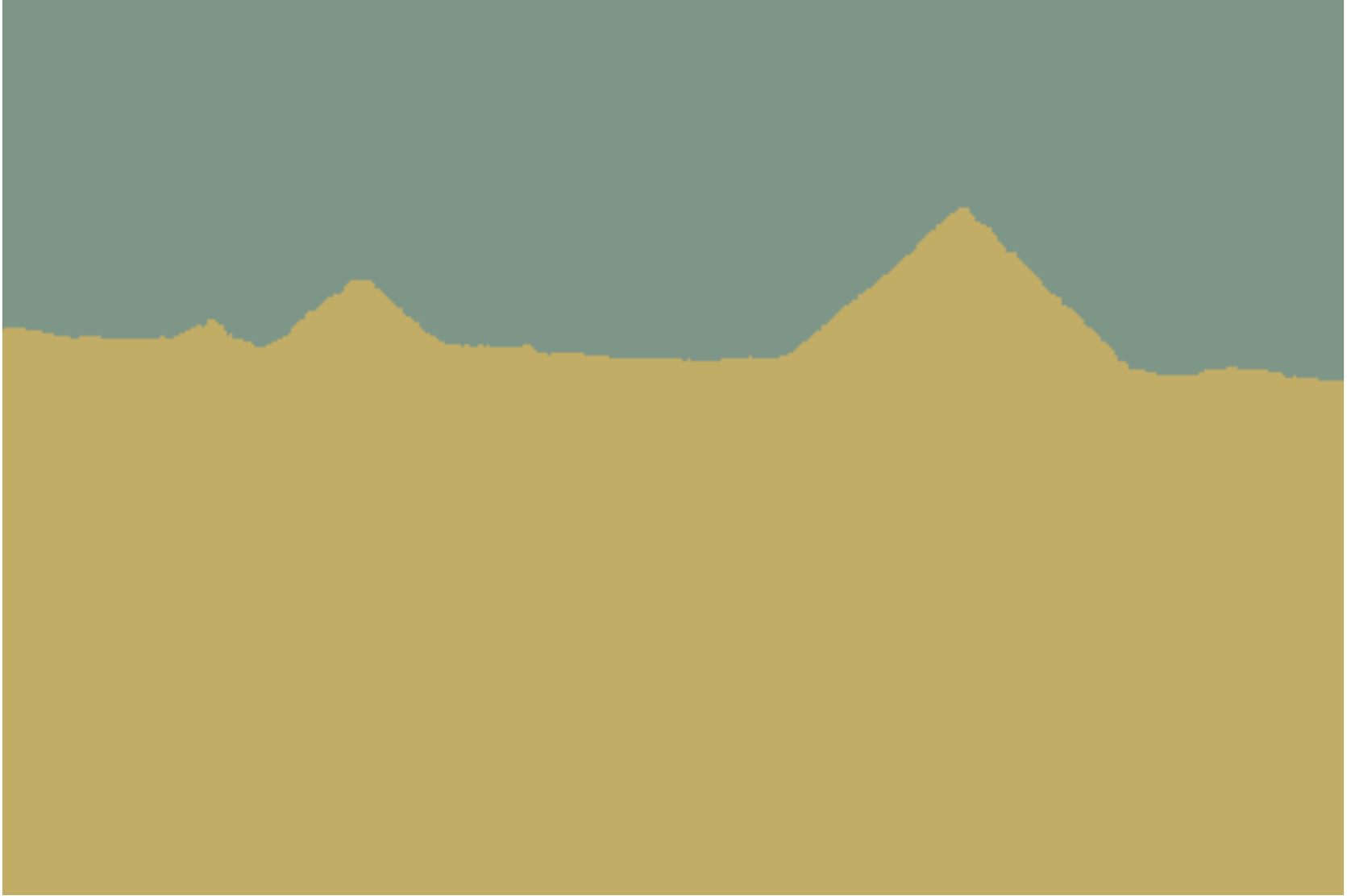} \\
{\small (A) Noisy image} &
{\small (A1) Method \cite{LNZS10}} &
{\small (A2) Method \cite{PCCB09}} &
{\small (A3) Method \cite{SW14}} &
{\small (A4) Ours } \\
\includegraphics[width=\ww, height=\hh]{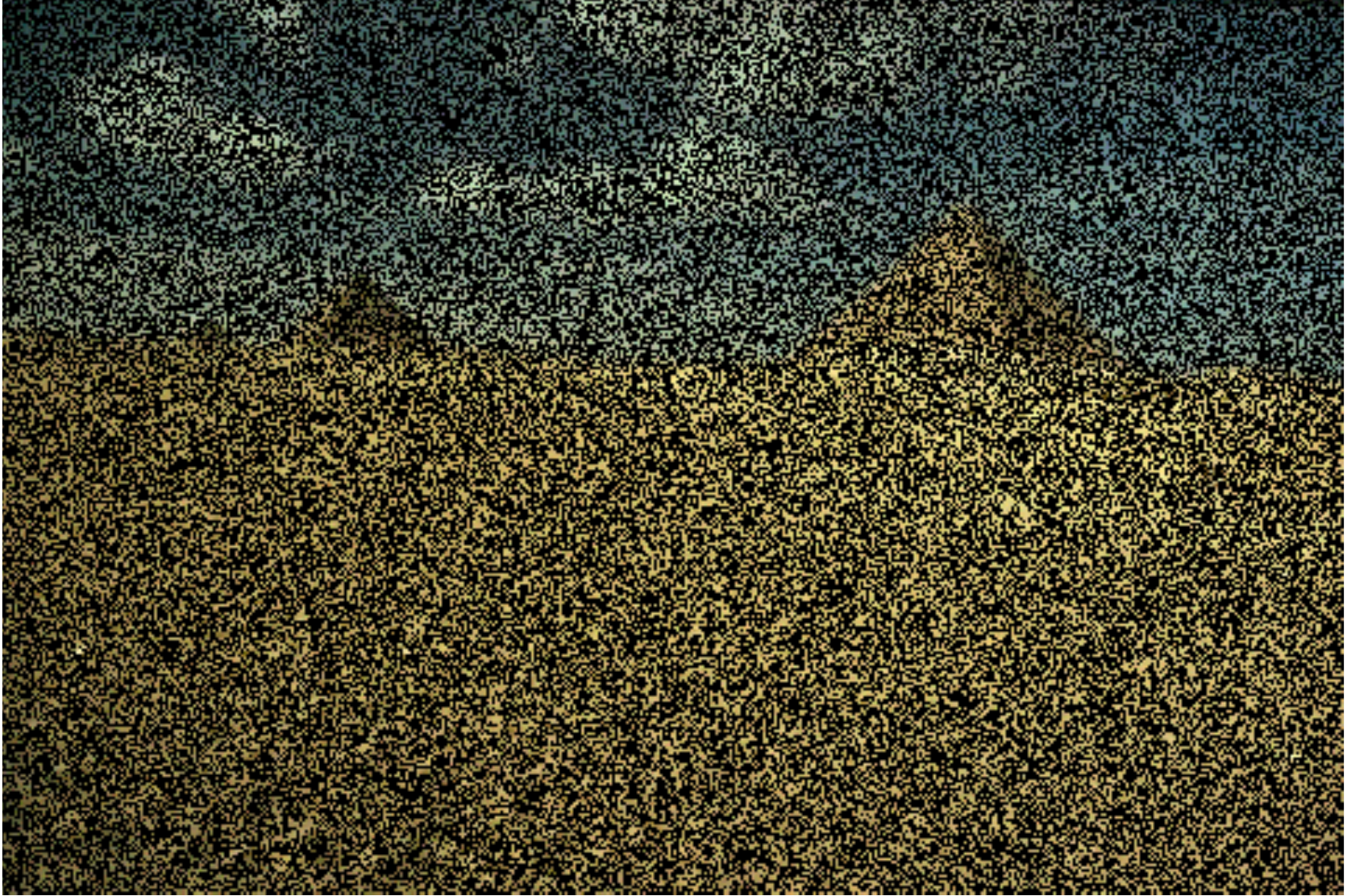} &
\includegraphics[width=\ww, height=\hh]{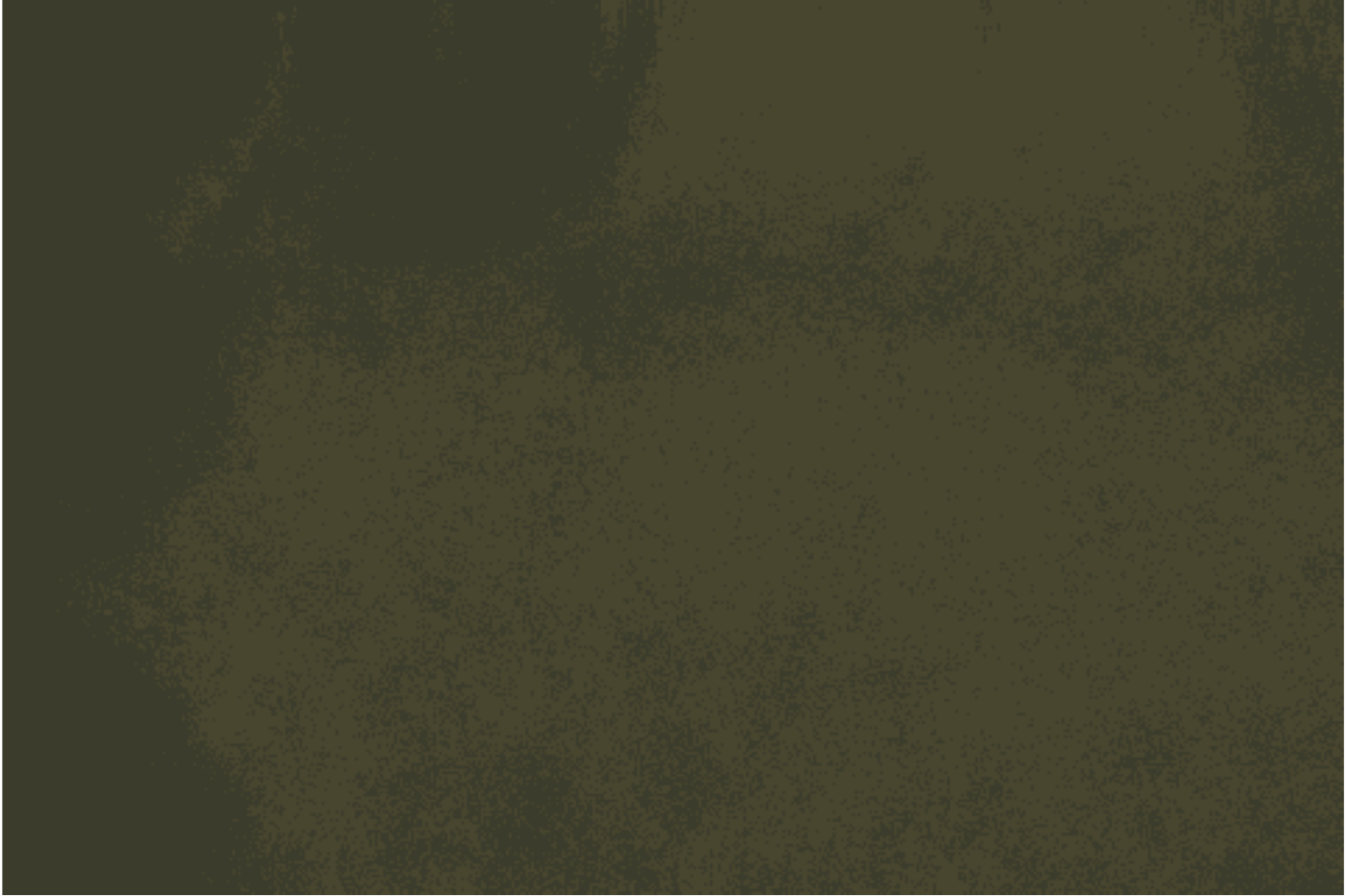} &
\includegraphics[width=\ww, height=\hh]{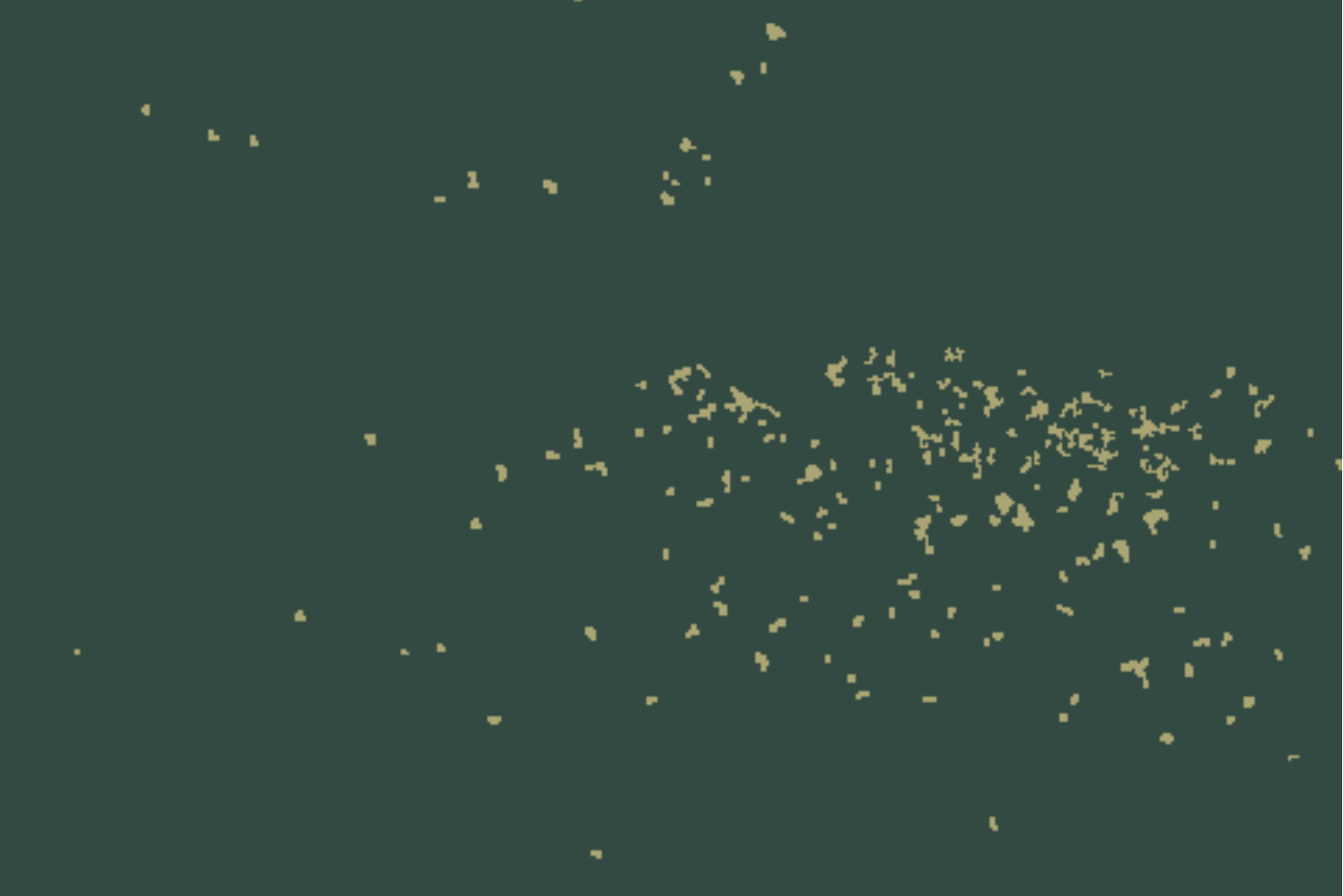} &
\includegraphics[width=\ww, height=\hh]{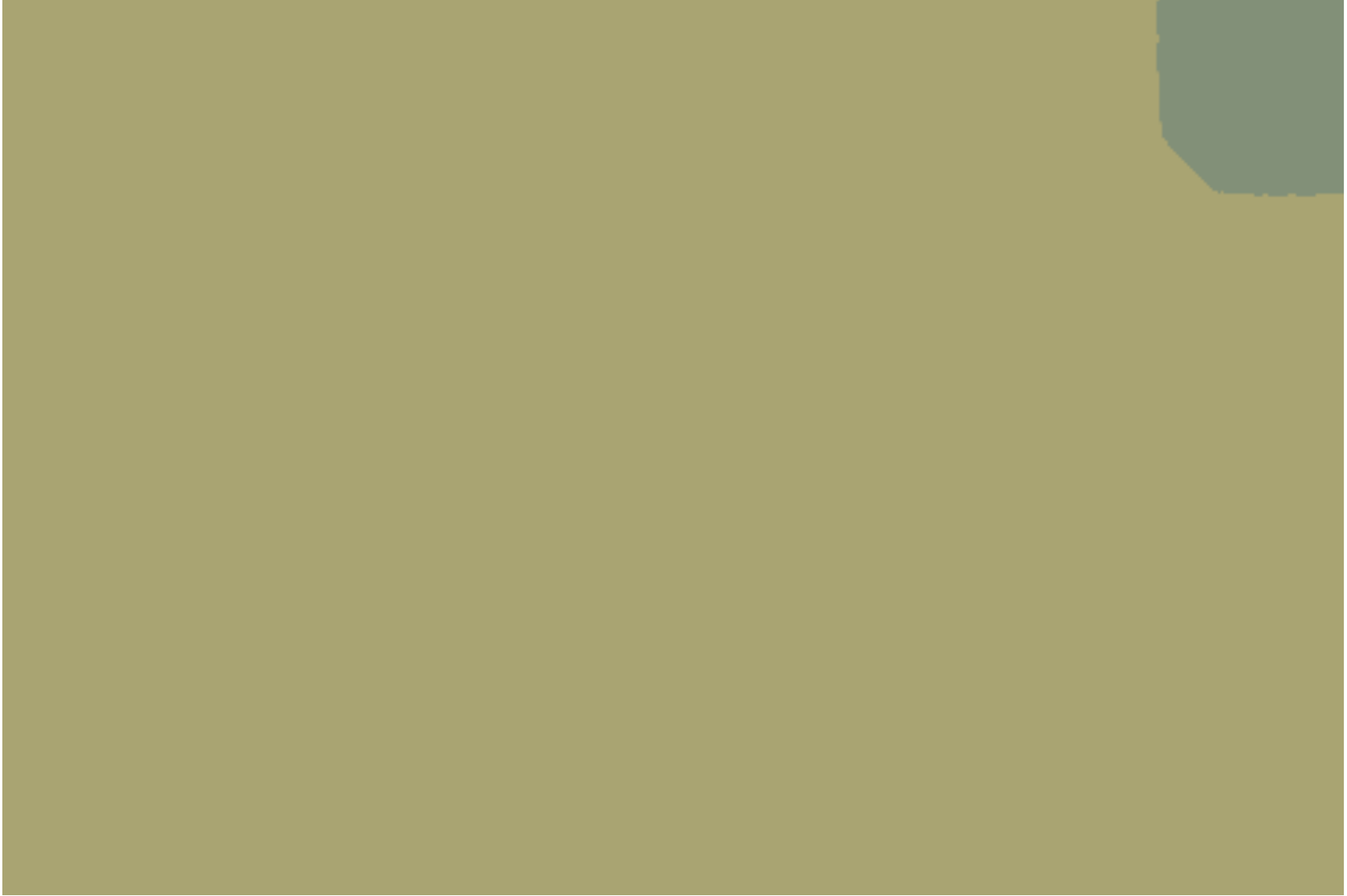} &
\includegraphics[width=\ww, height=\hh]{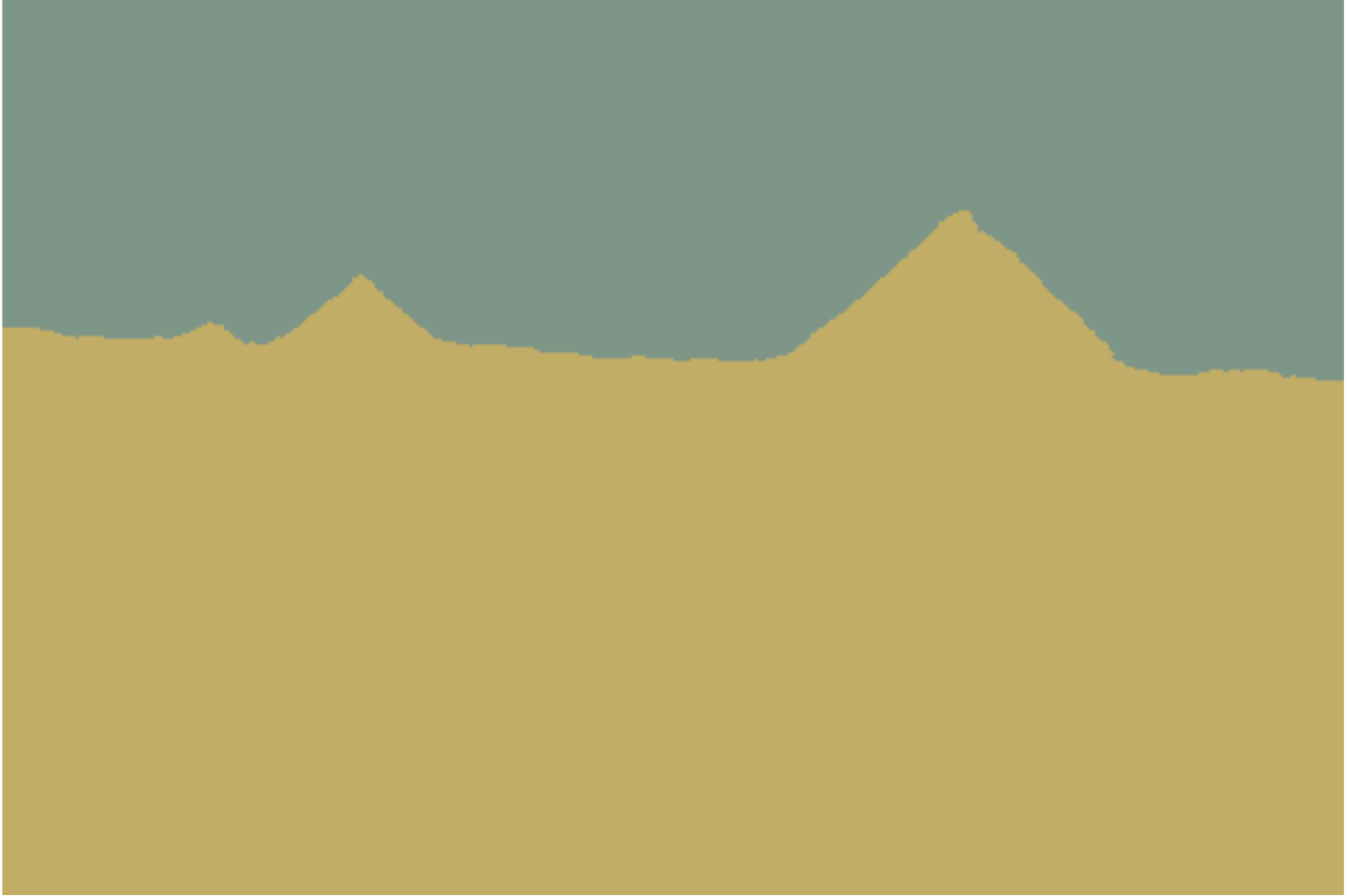} \\
{\small (B) Information } &
{\small (B1) Method \cite{LNZS10}} &
{\small (B2) Method \cite{PCCB09}} &
{\small (B3) Method \cite{SW14}} &
{\small (B4) Ours } \vspace{-0.05in} \\
{\small loss + noise} & & & & \\
\includegraphics[width=\ww, height=\hh]{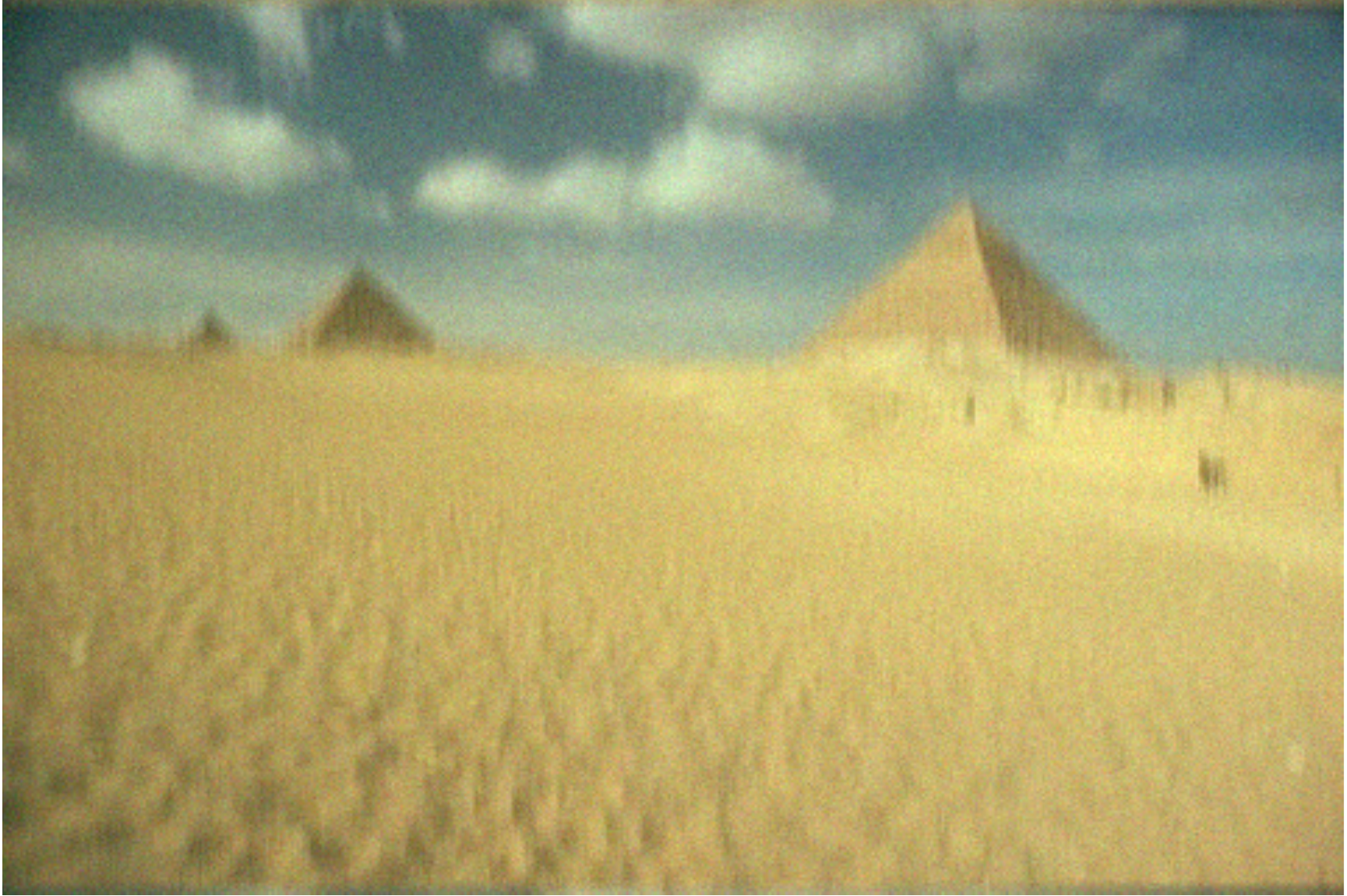} &
\includegraphics[width=\ww, height=\hh]{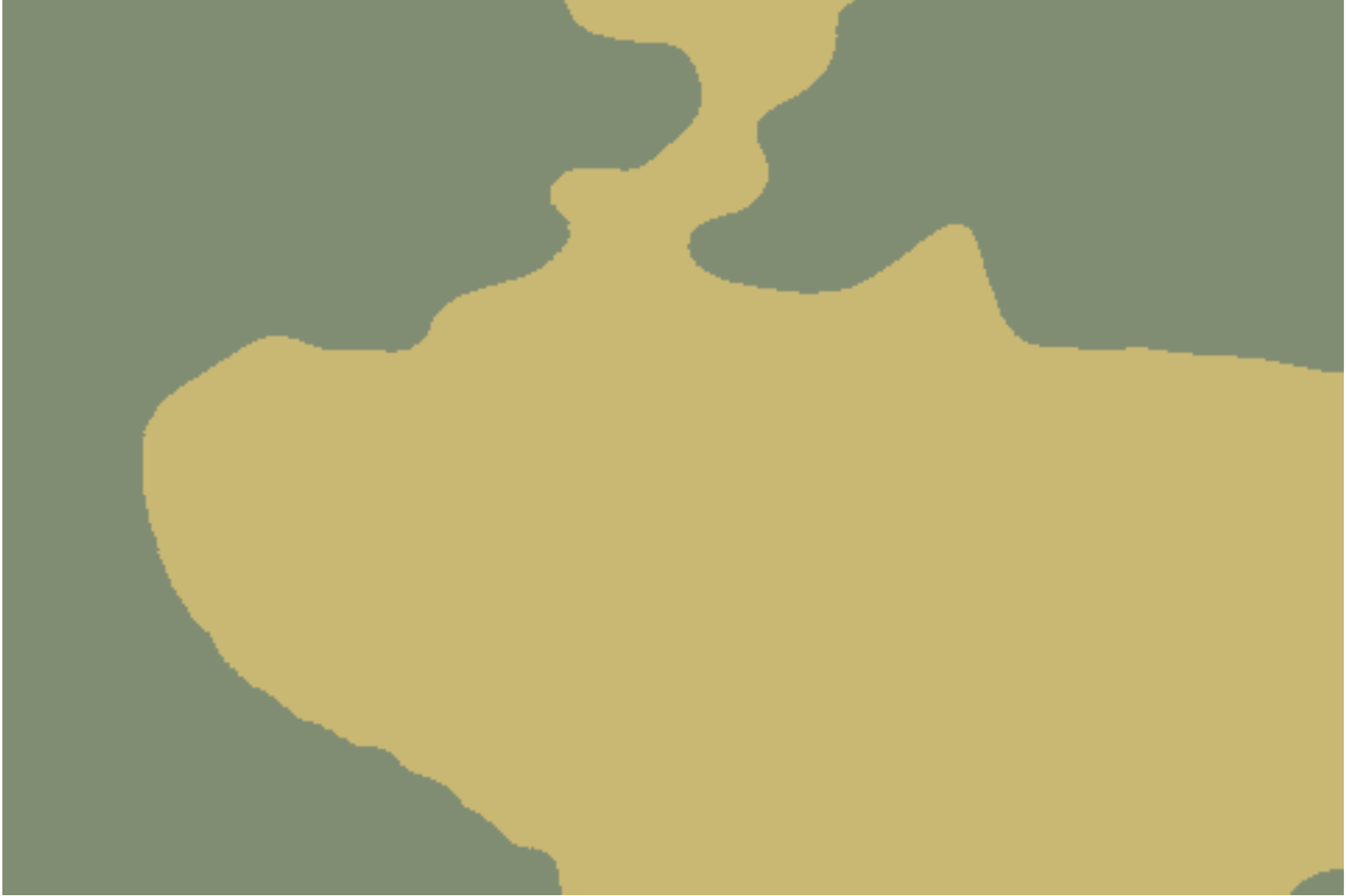} &
\includegraphics[width=\ww, height=\hh]{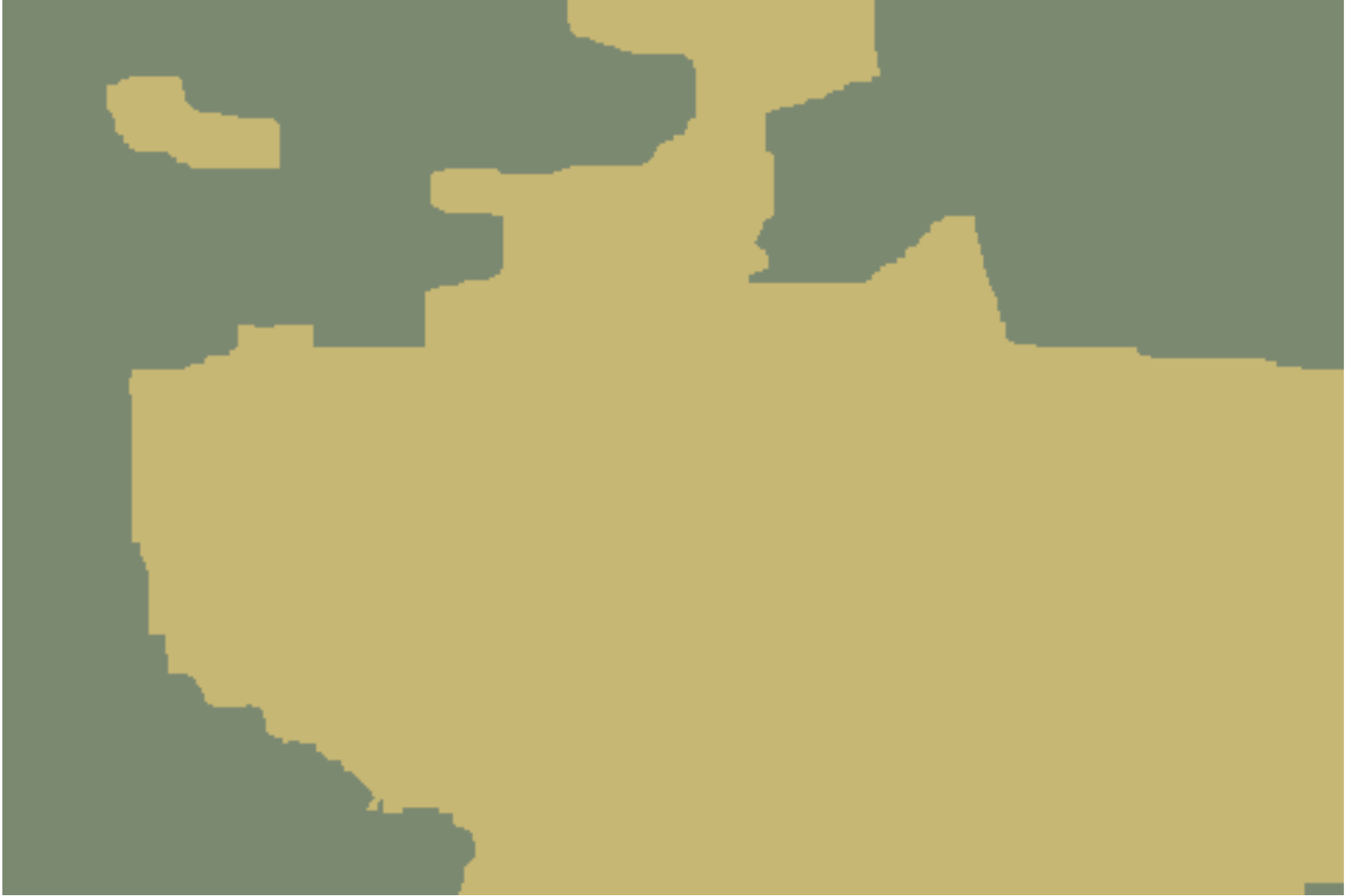} &
\includegraphics[width=\ww, height=\hh]{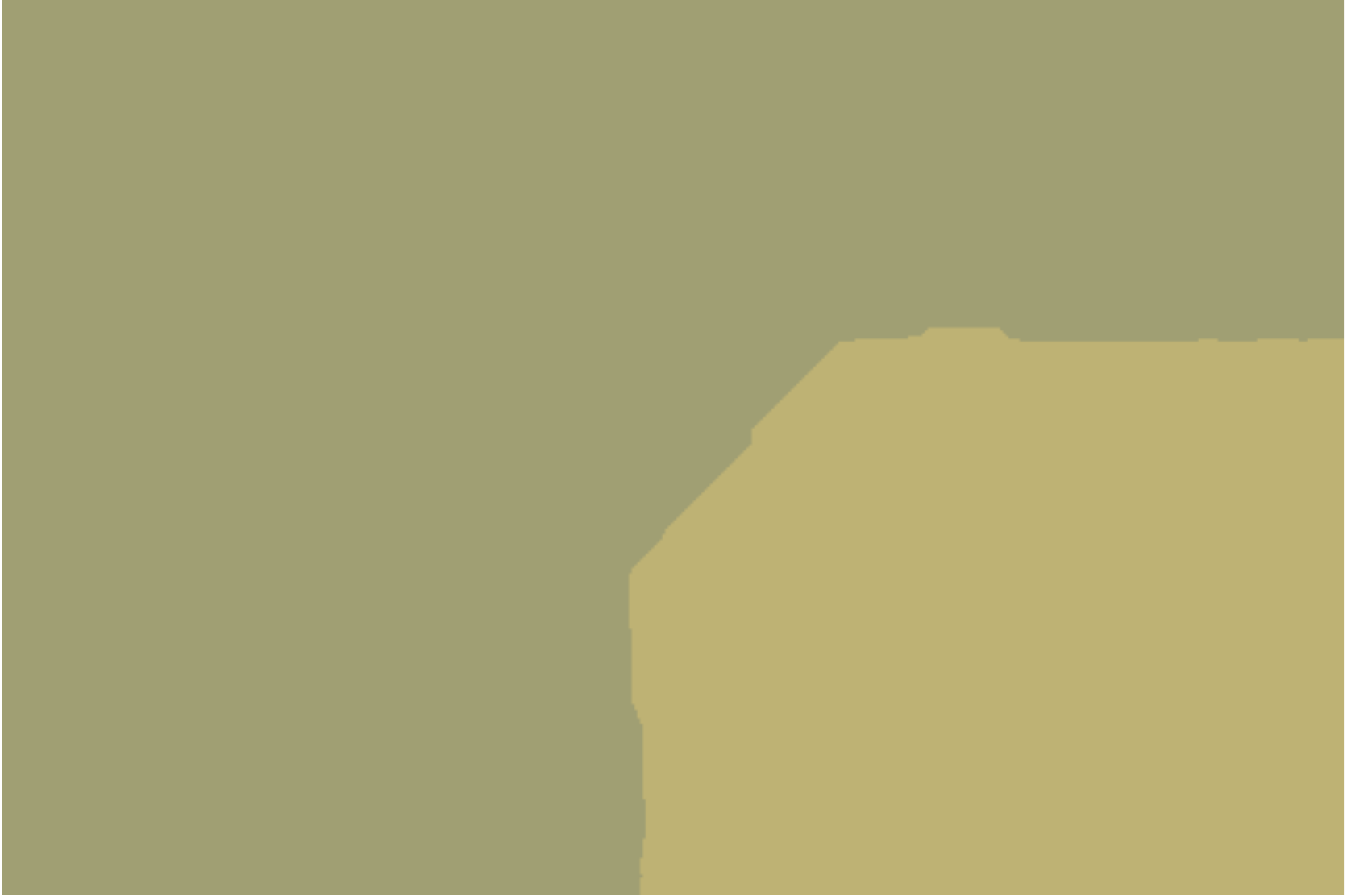} &
\includegraphics[width=\ww, height=\hh]{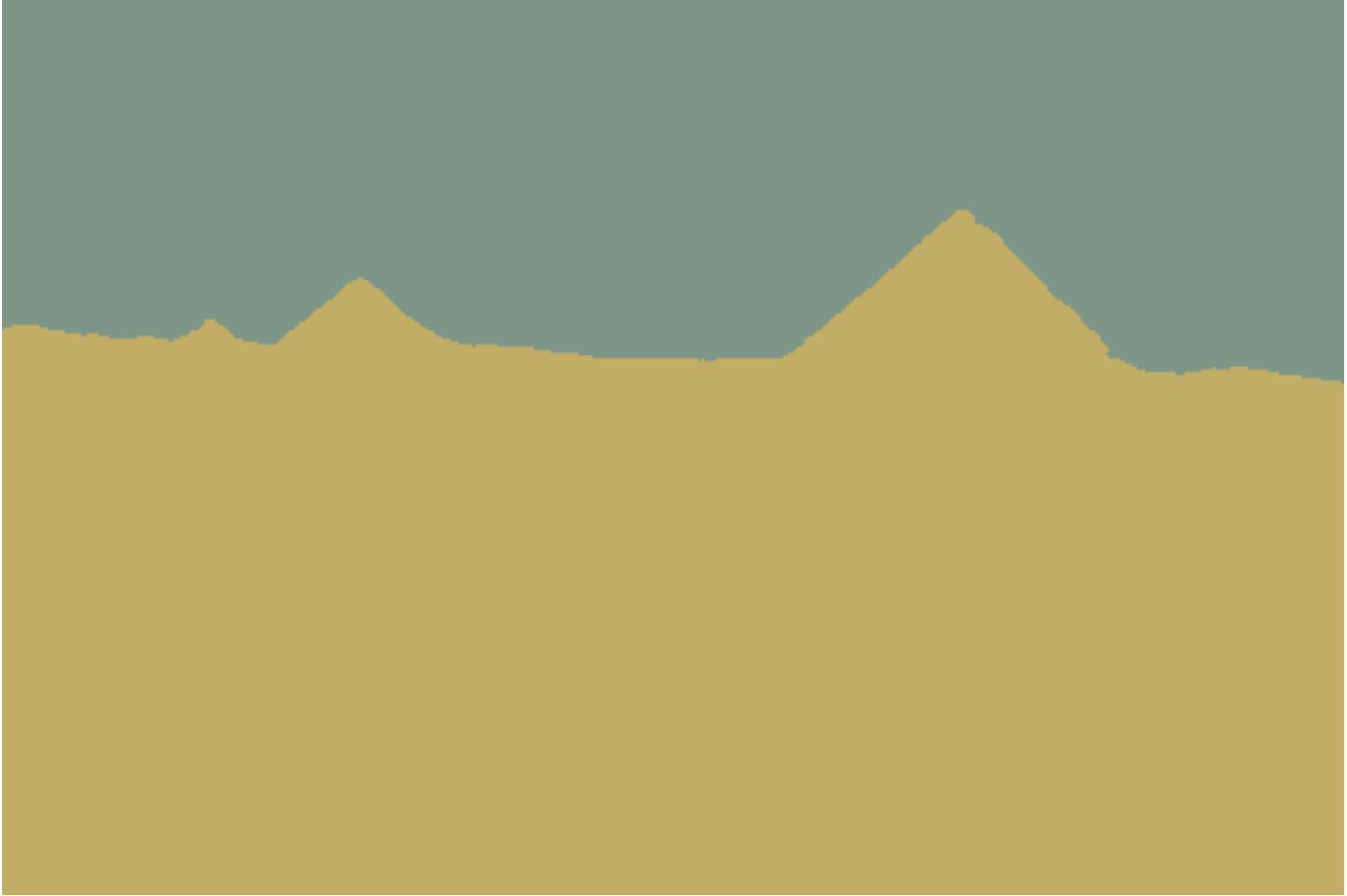} \\
{\small (C) Blur + noise} &
{\small (C1) Method \cite{LNZS10}} &
{\small (C2) Method \cite{PCCB09}} &
{\small (C3) Method \cite{SW14}} &
{\small (C4) Ours }
\end{tabular}
\end{center}
\caption{Two-phase pyramid segmentation (size: $321\times 481$).
(A): Given Gaussian noisy image with mean 0 and variance 0.001;
(B): Given Gaussian noisy image with $60\%$ information loss;
(C): Given blurry image with Gaussian noise;
(A1-A4), (B1-B4) and (C1-C4): Results of methods \cite{LNZS10},
\cite{PCCB09}, \cite{SW14}, and our SLaT on (A), (B) and (C), respectively.
}\label{twophase-color-pyramid}
\end{figure*}

\begin{figure*}[!htb]
\begin{center}
\begin{tabular}{ccccc}
\includegraphics[width=\ww, height=\hh]{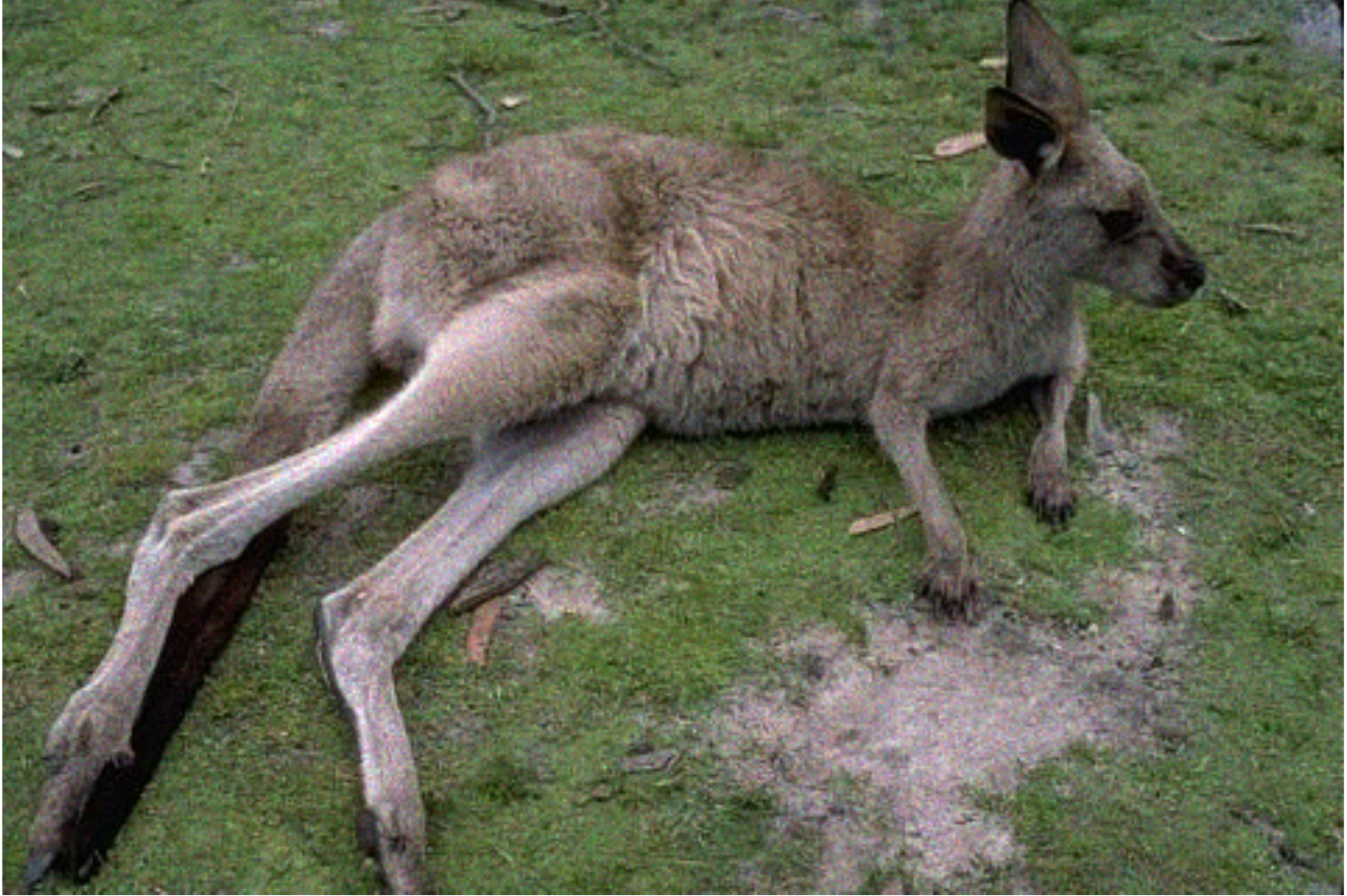} &
\includegraphics[width=\ww, height=\hh]{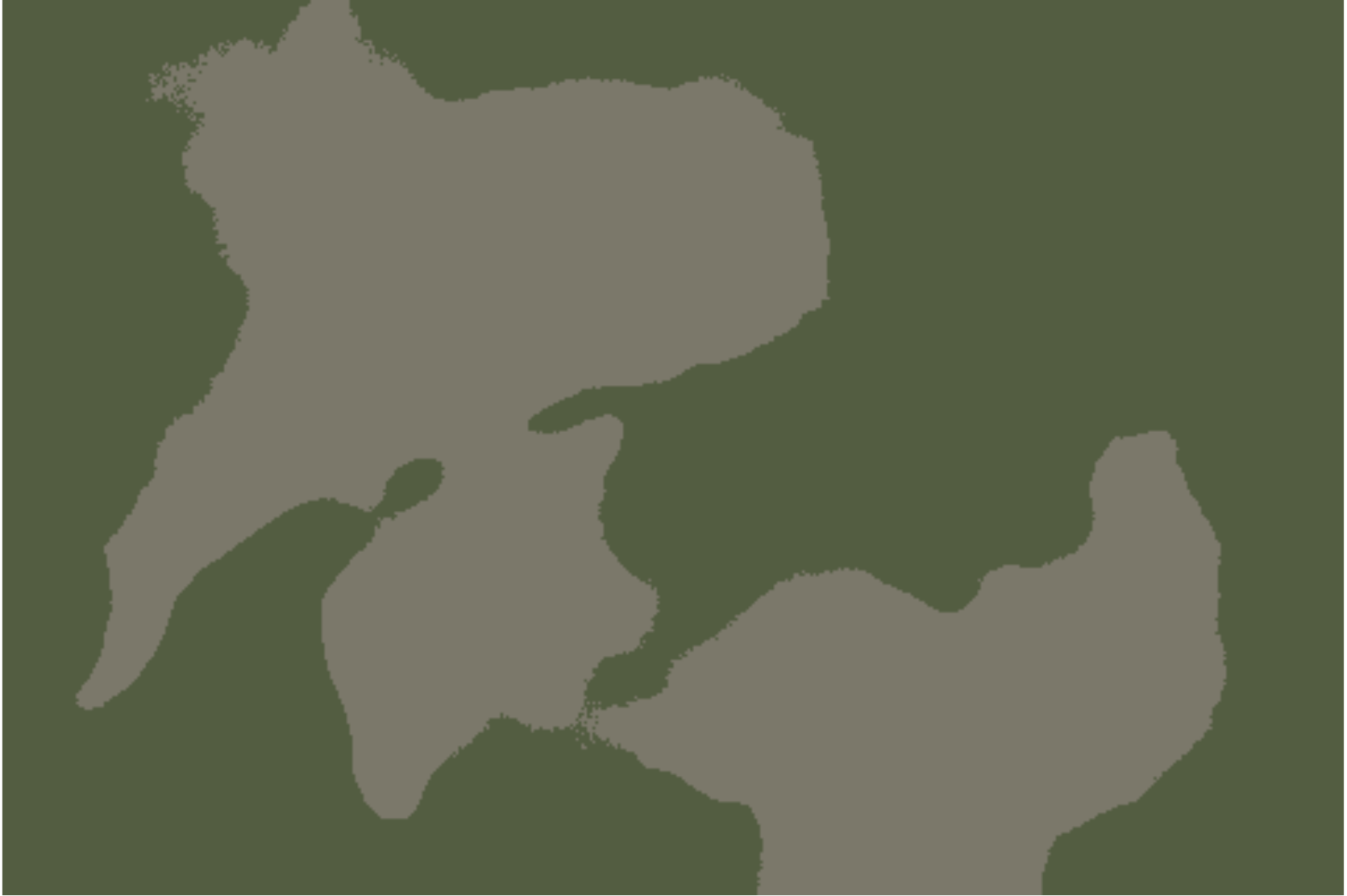} &
\includegraphics[width=\ww, height=\hh]{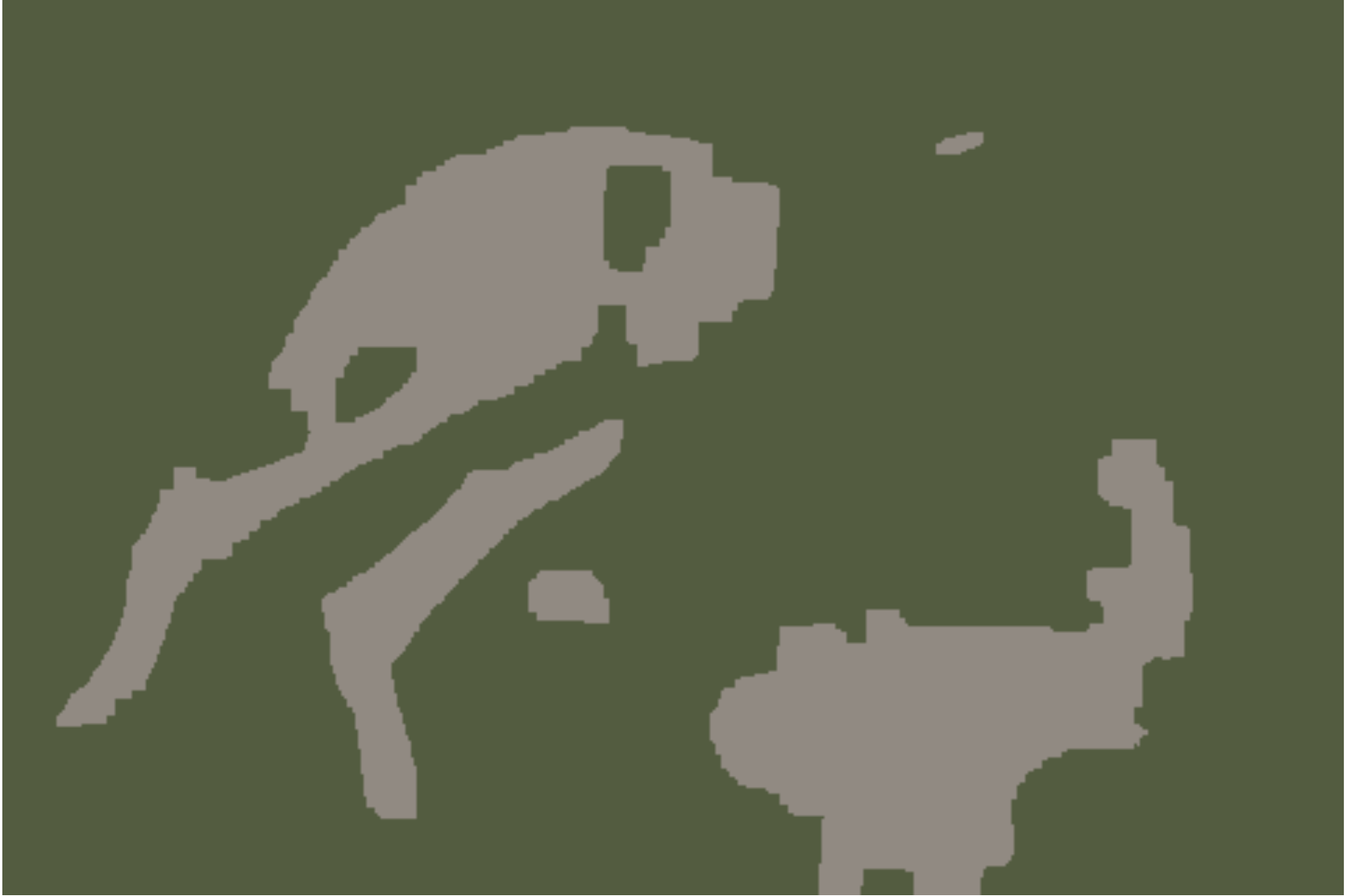} &
\includegraphics[width=\ww, height=\hh]{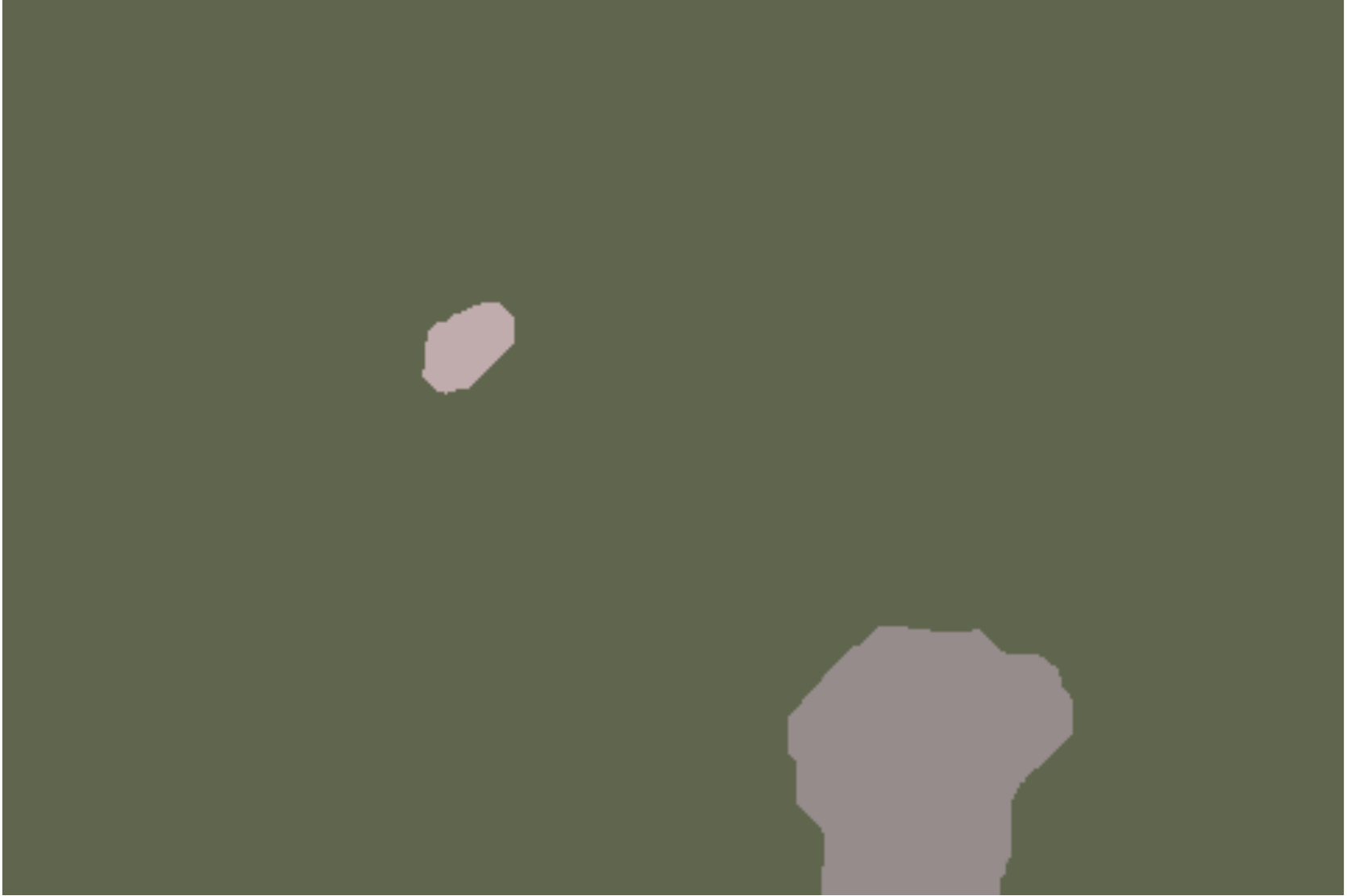} &
\includegraphics[width=\ww, height=\hh]{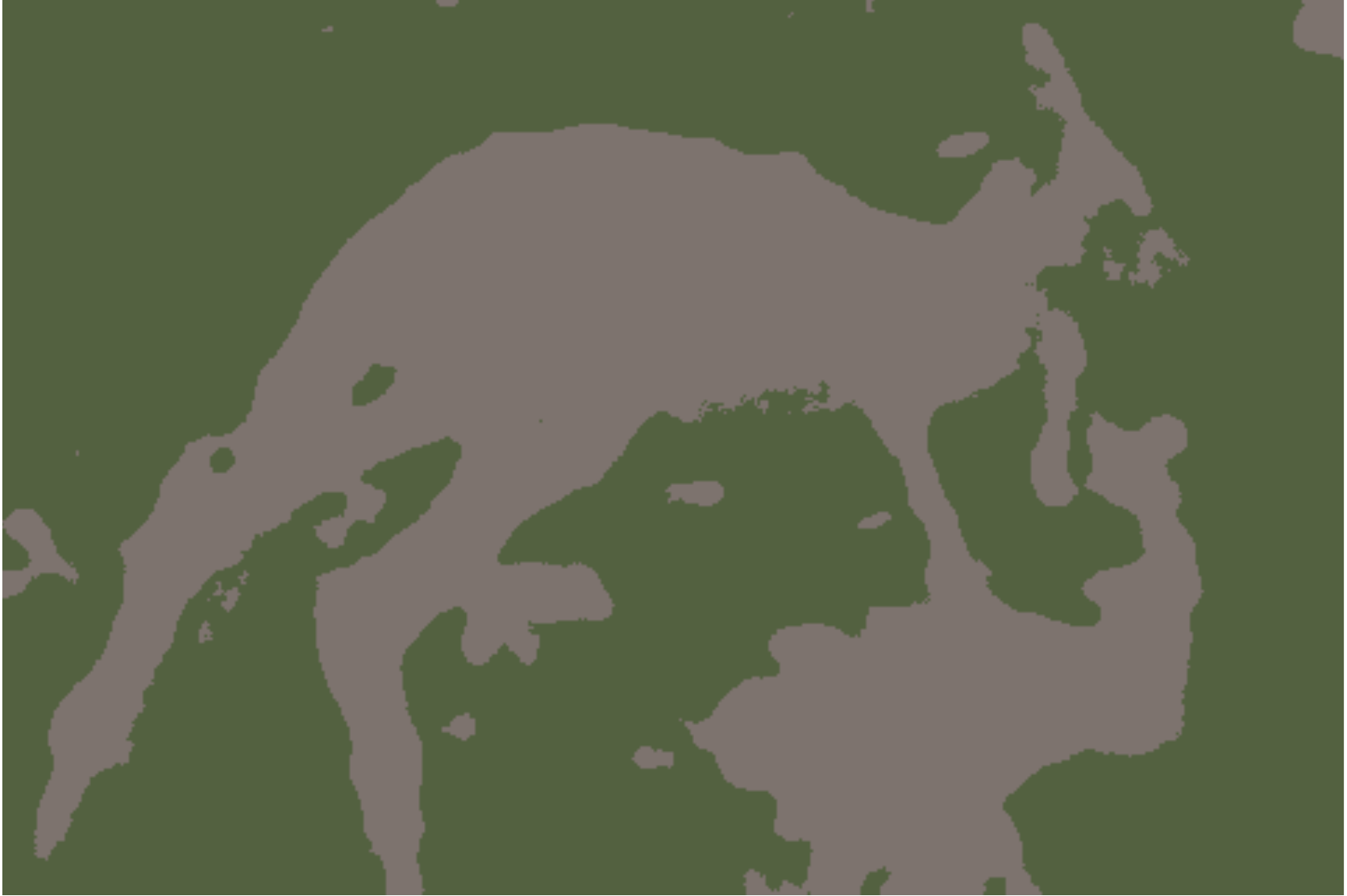} \\
{\small (A) Noisy image} &
{\small (A1) Method \cite{LNZS10}} &
{\small (A2) Method \cite{PCCB09}} &
{\small (A3) Method \cite{SW14}} &
{\small (A4) Ours } \\
\includegraphics[width=\ww, height=\hh]{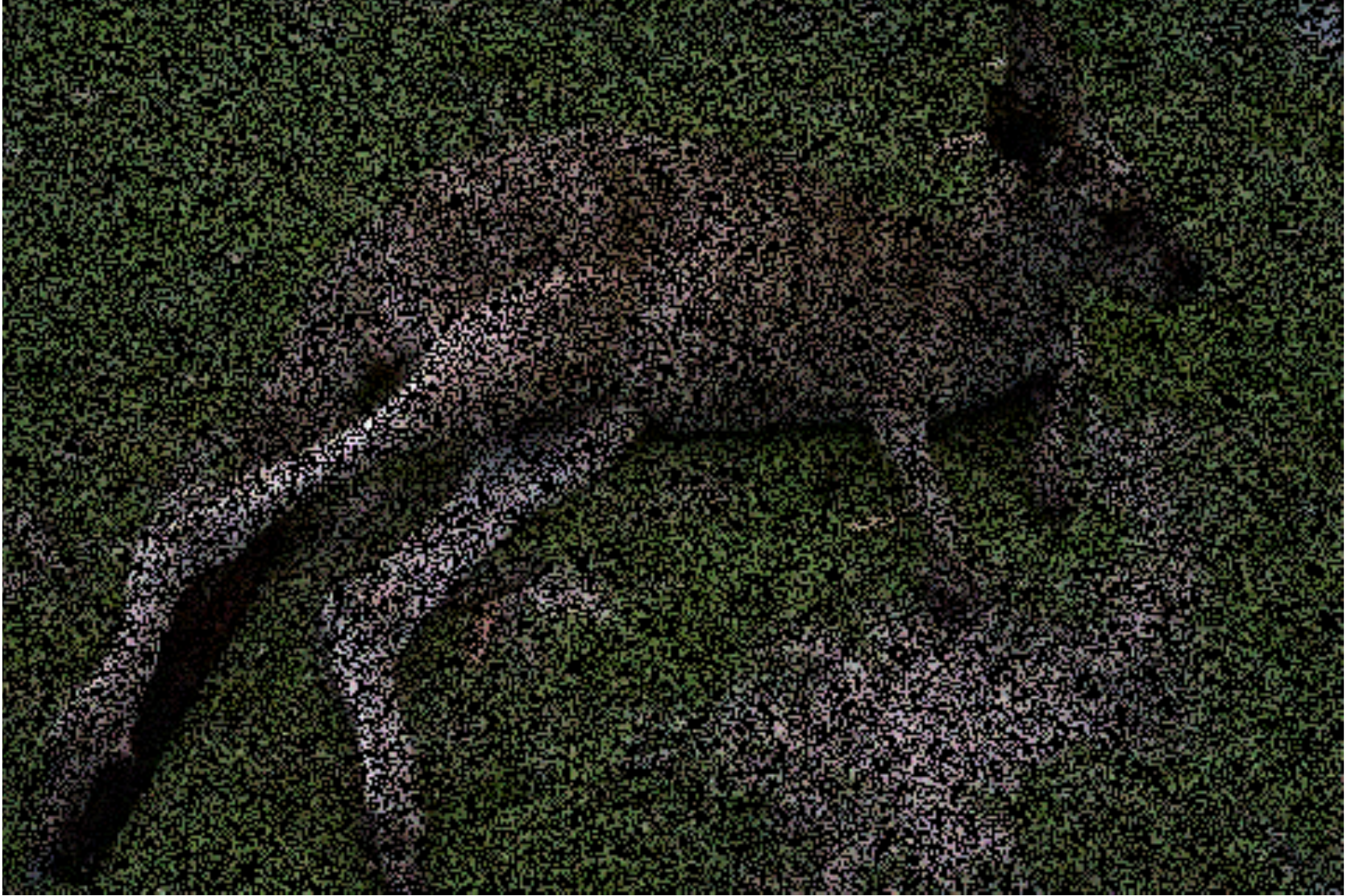} &
\includegraphics[width=\ww, height=\hh]{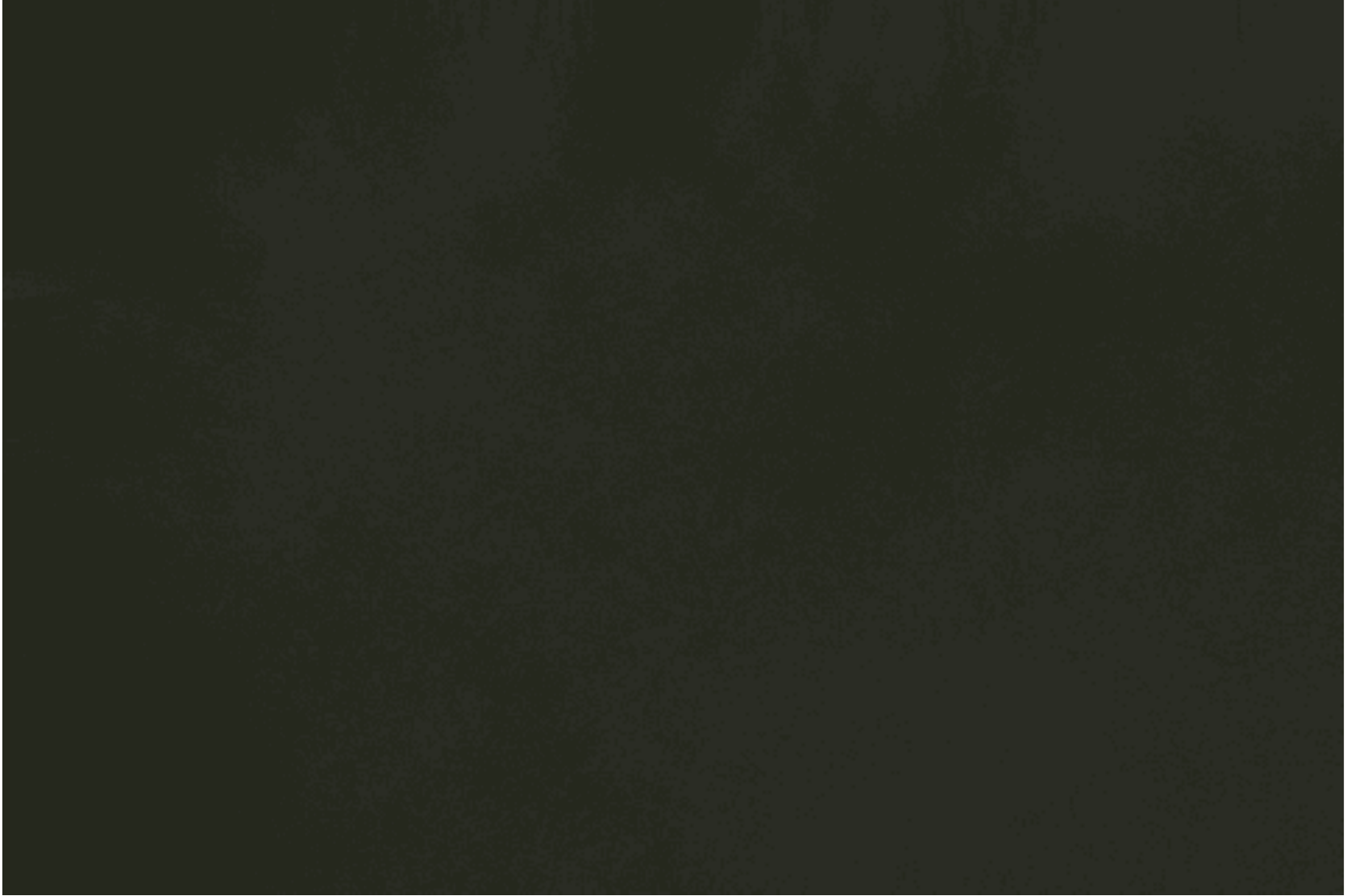} &
\includegraphics[width=\ww, height=\hh]{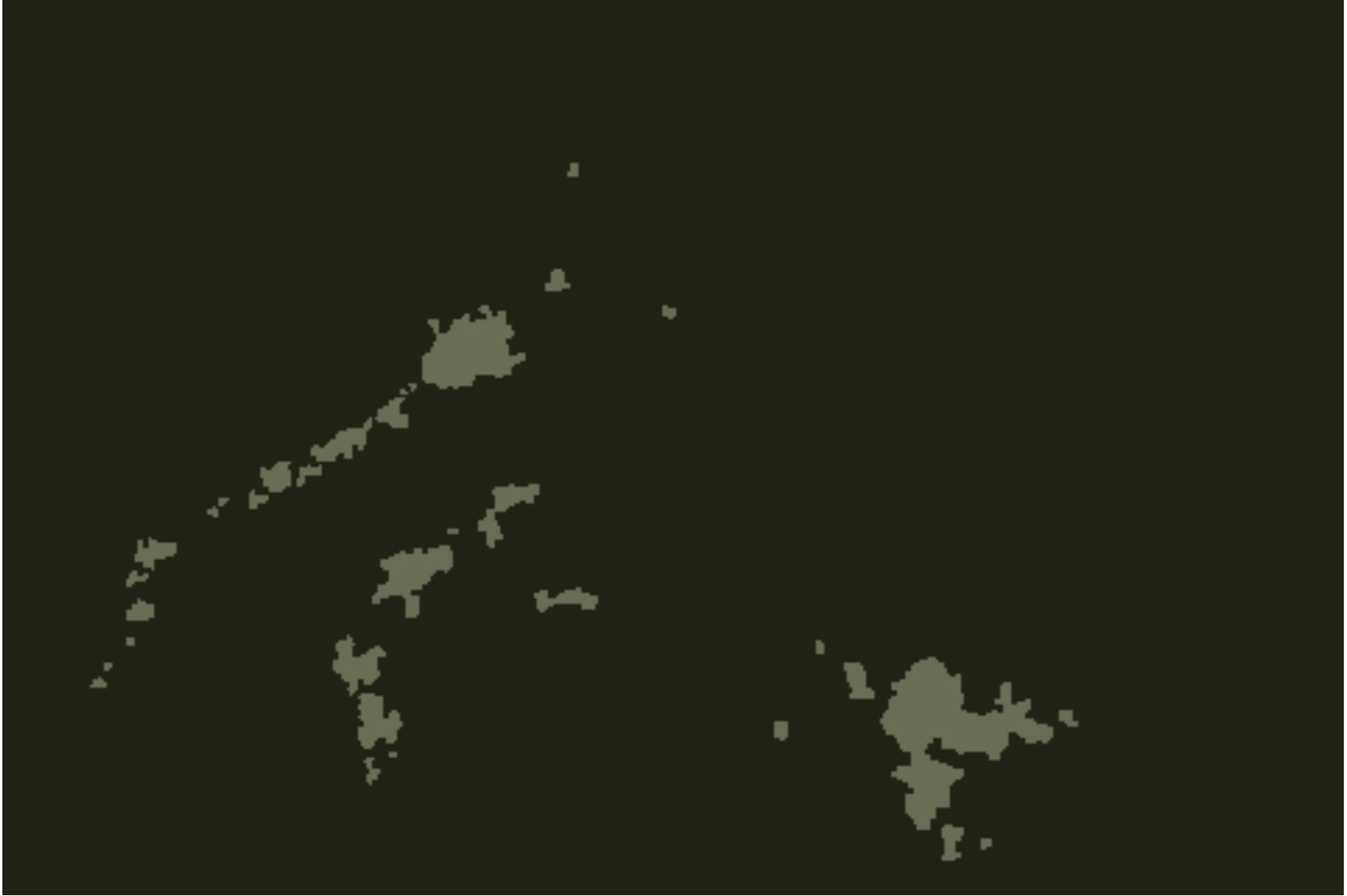} &
\includegraphics[width=\ww, height=\hh]{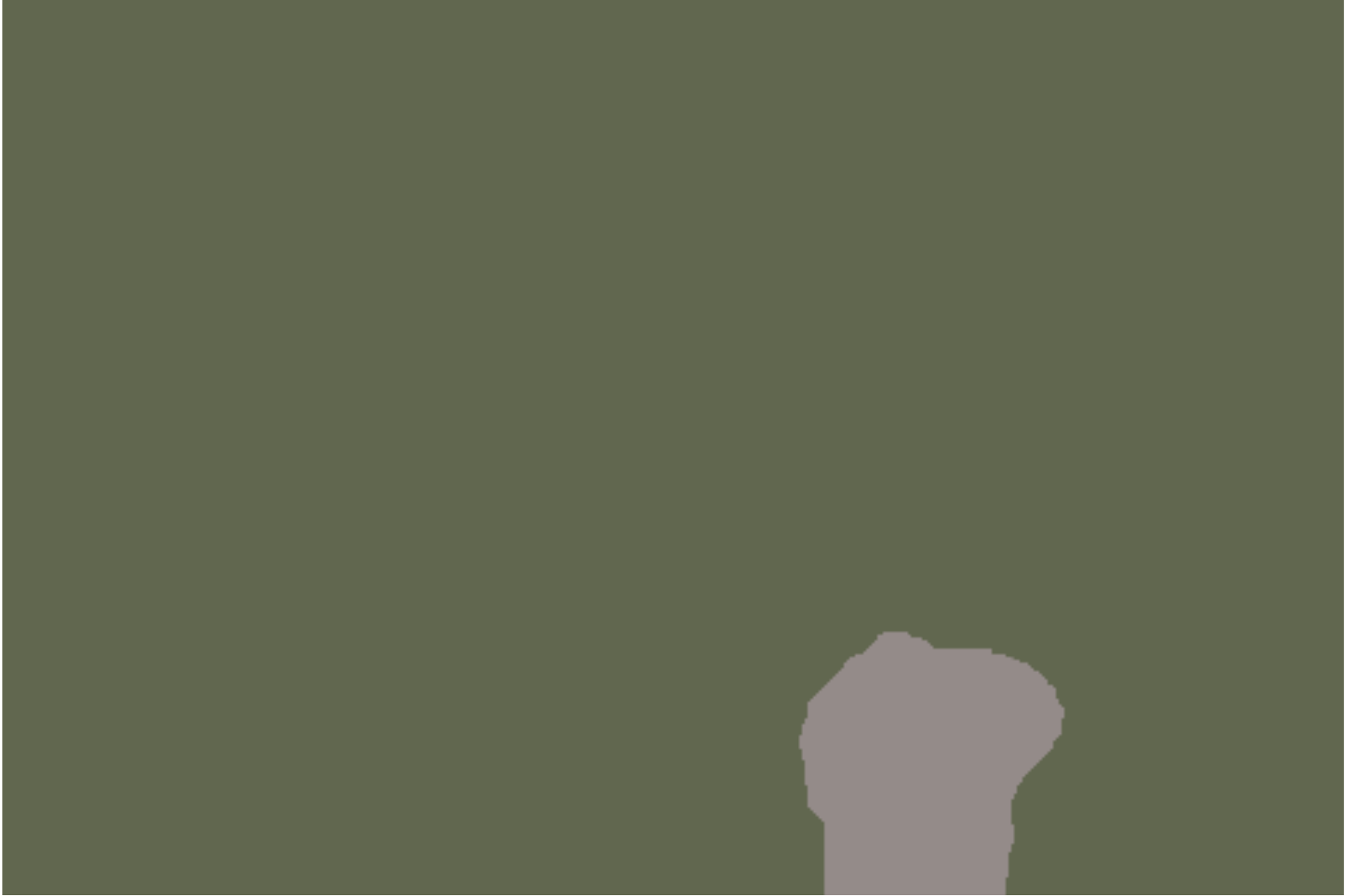} &
\includegraphics[width=\ww, height=\hh]{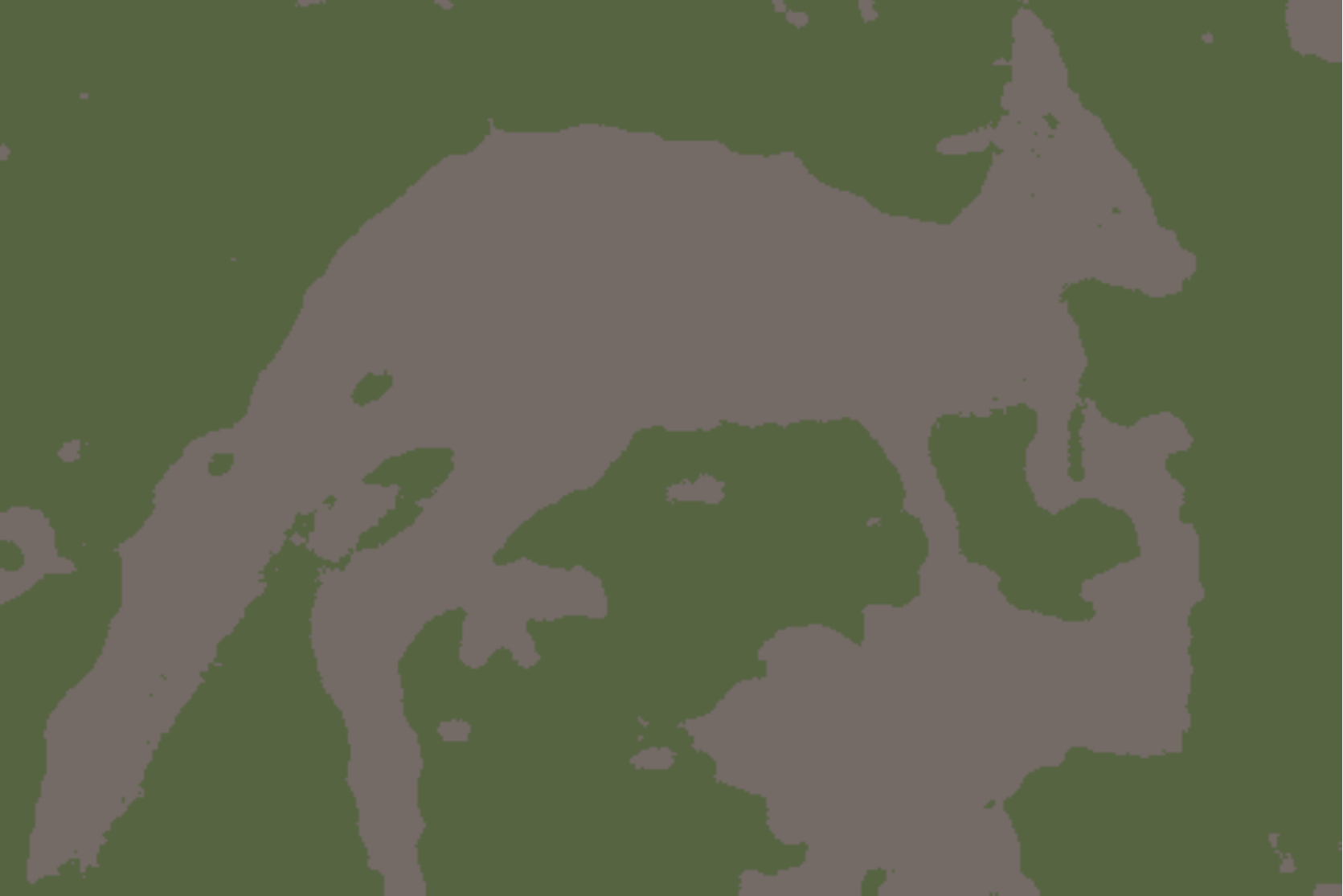} \\
{\small (B) Information} &
{\small (B1) Method \cite{LNZS10}} &
{\small (B2) Method \cite{PCCB09}} &
{\small (B3) Method \cite{SW14}} &
{\small (B4) Ours } \vspace{-0.05in} \\
{\small loss + noise} & & & & \\
\includegraphics[width=\ww, height=\hh]{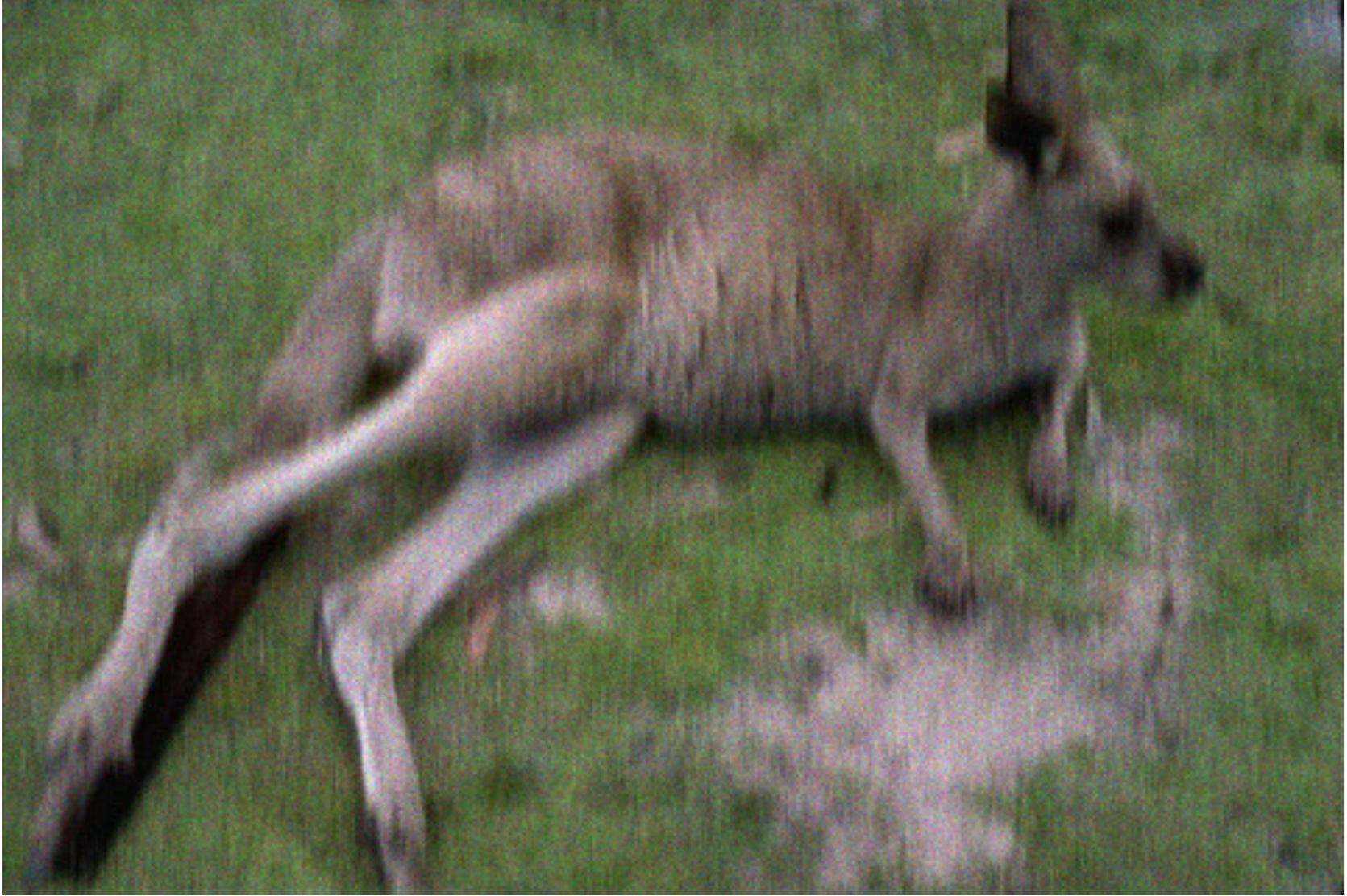} &
\includegraphics[width=\ww, height=\hh]{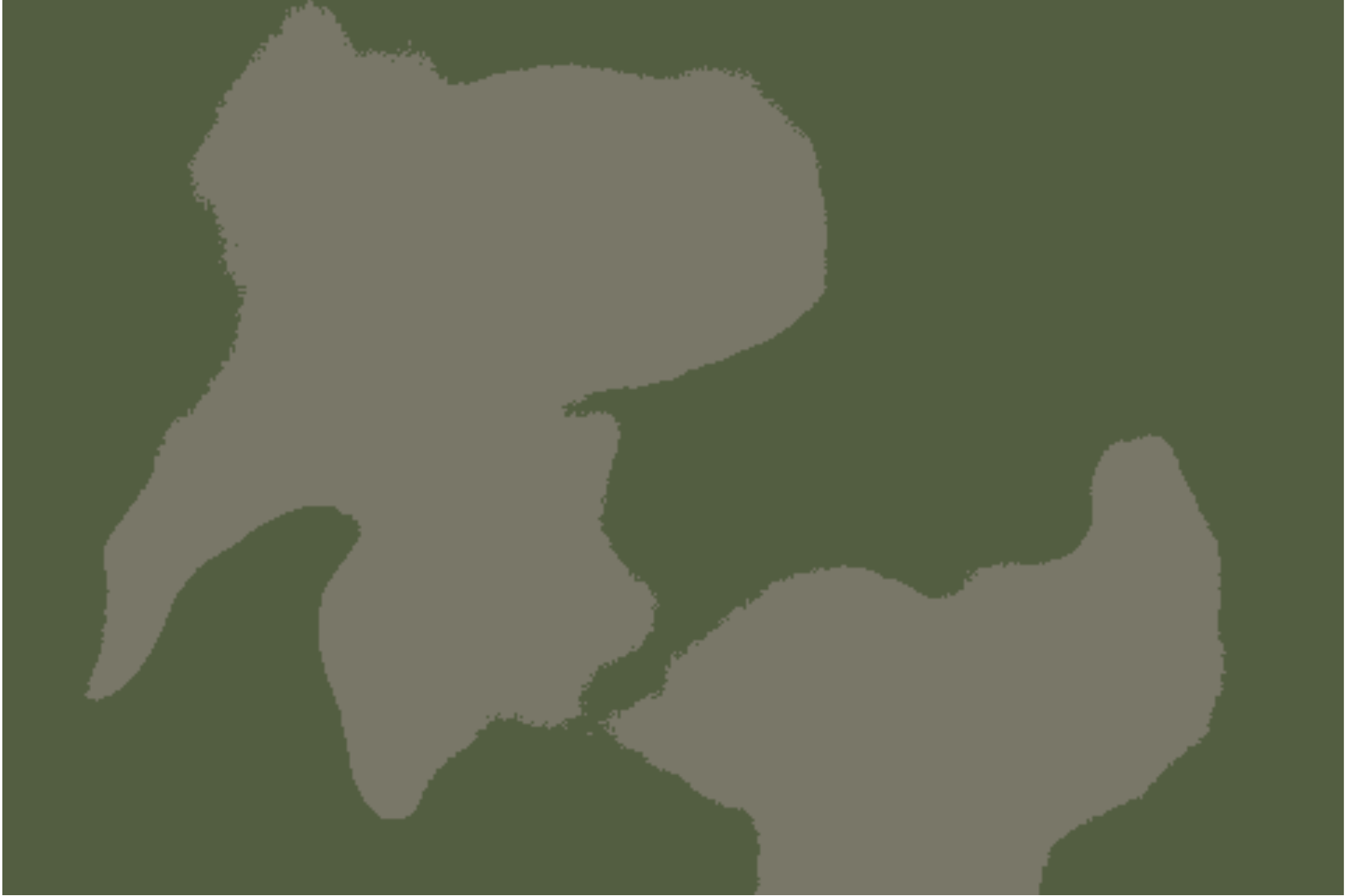} &
\includegraphics[width=\ww, height=\hh]{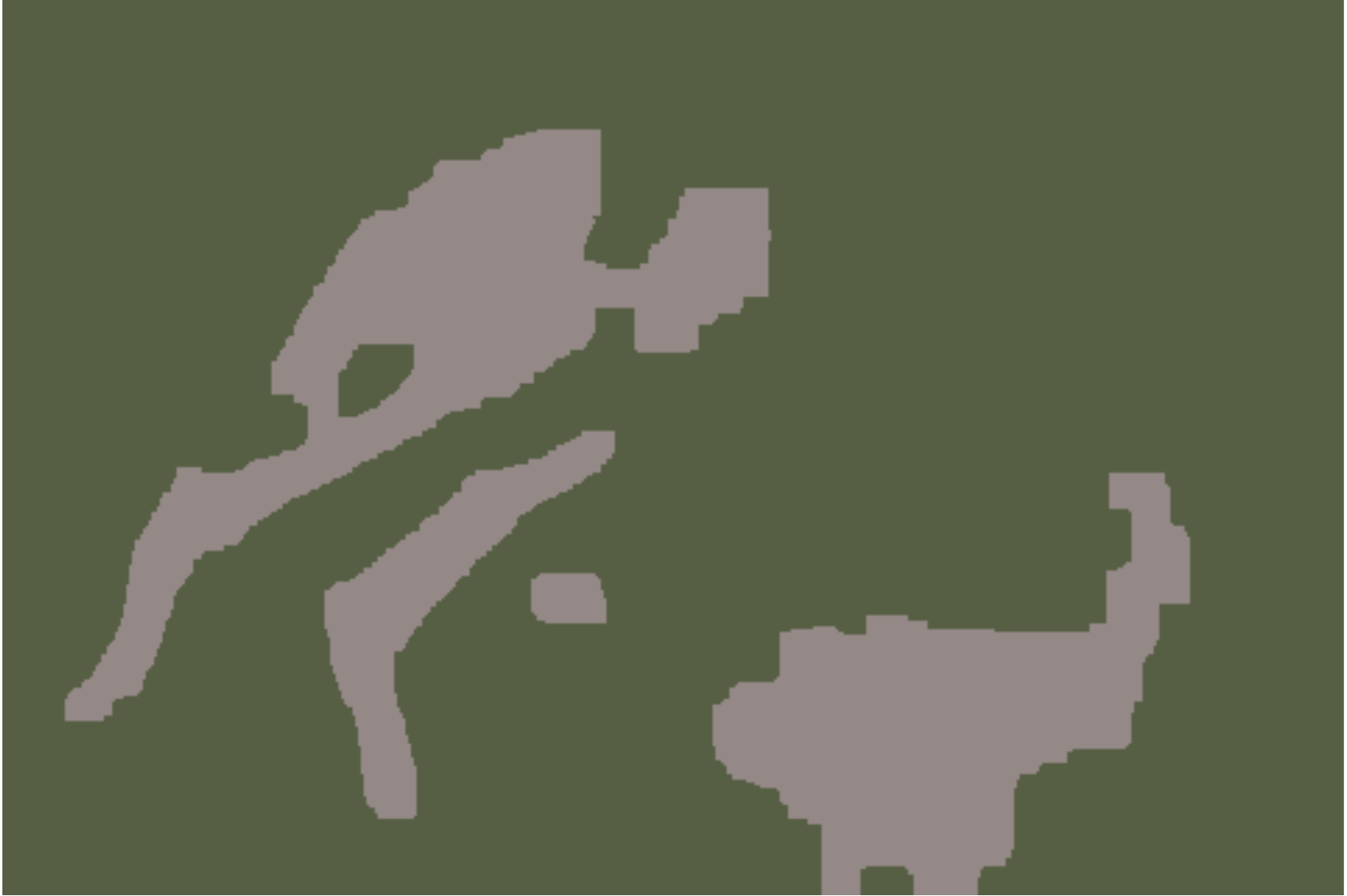} &
\includegraphics[width=\ww, height=\hh]{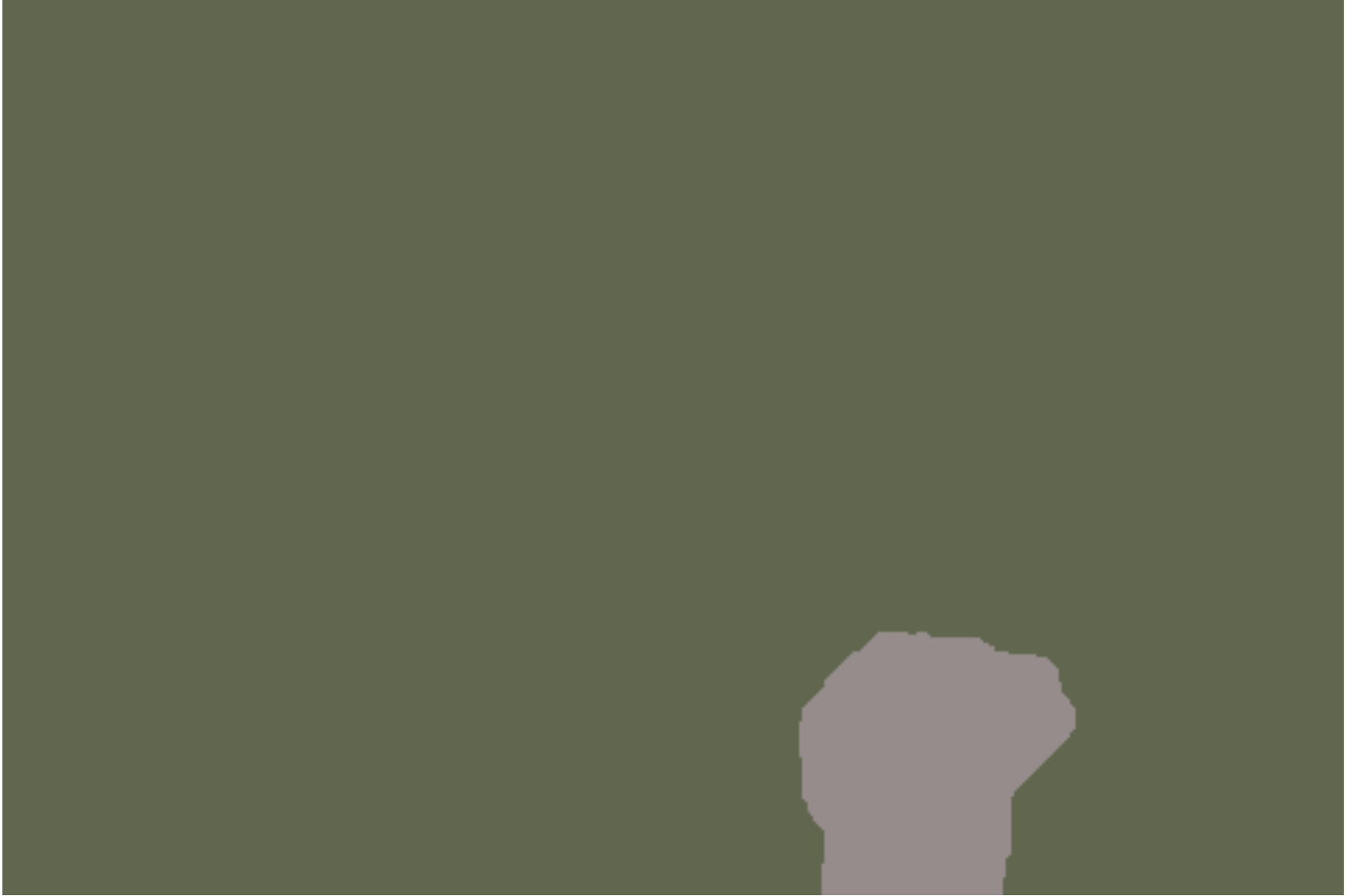} &
\includegraphics[width=\ww, height=\hh]{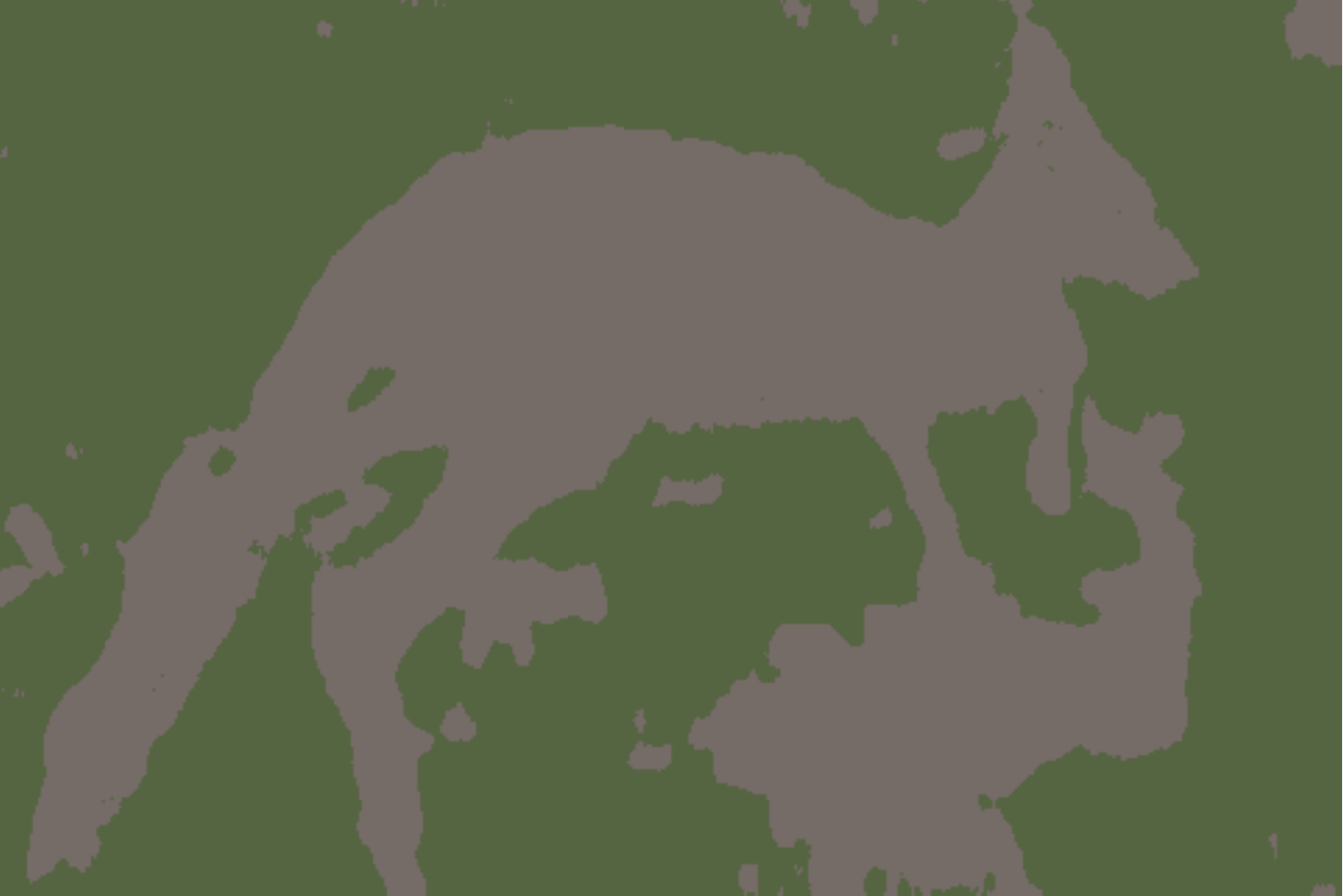} \\
{\small (C) Blur + noise} &
{\small (C1) Method \cite{LNZS10}} &
{\small (C2) Method \cite{PCCB09}} &
{\small (C3) Method \cite{SW14}} &
{\small (C4) Ours }
\end{tabular}
\end{center}
\caption{Two-phase kangaroo segmentation (size: $321\times 481$).
(A): Given Poisson noisy image;
(B): Given Poisson noisy image with $60\%$ information loss;
(C): Given blurry image with Poisson noise;
(A1-A4), (B1-B4) and (C1-C4): Results of methods \cite{LNZS10},
\cite{PCCB09}, \cite{SW14}, and our SLaT on (A), (B) and (C), respectively.
}\label{twophase-color-kangaroo}
\end{figure*}

\begin{figure*}[!htb]
\begin{center}
\begin{tabular}{ccccc}
\includegraphics[width=\ww, height=\hhh]{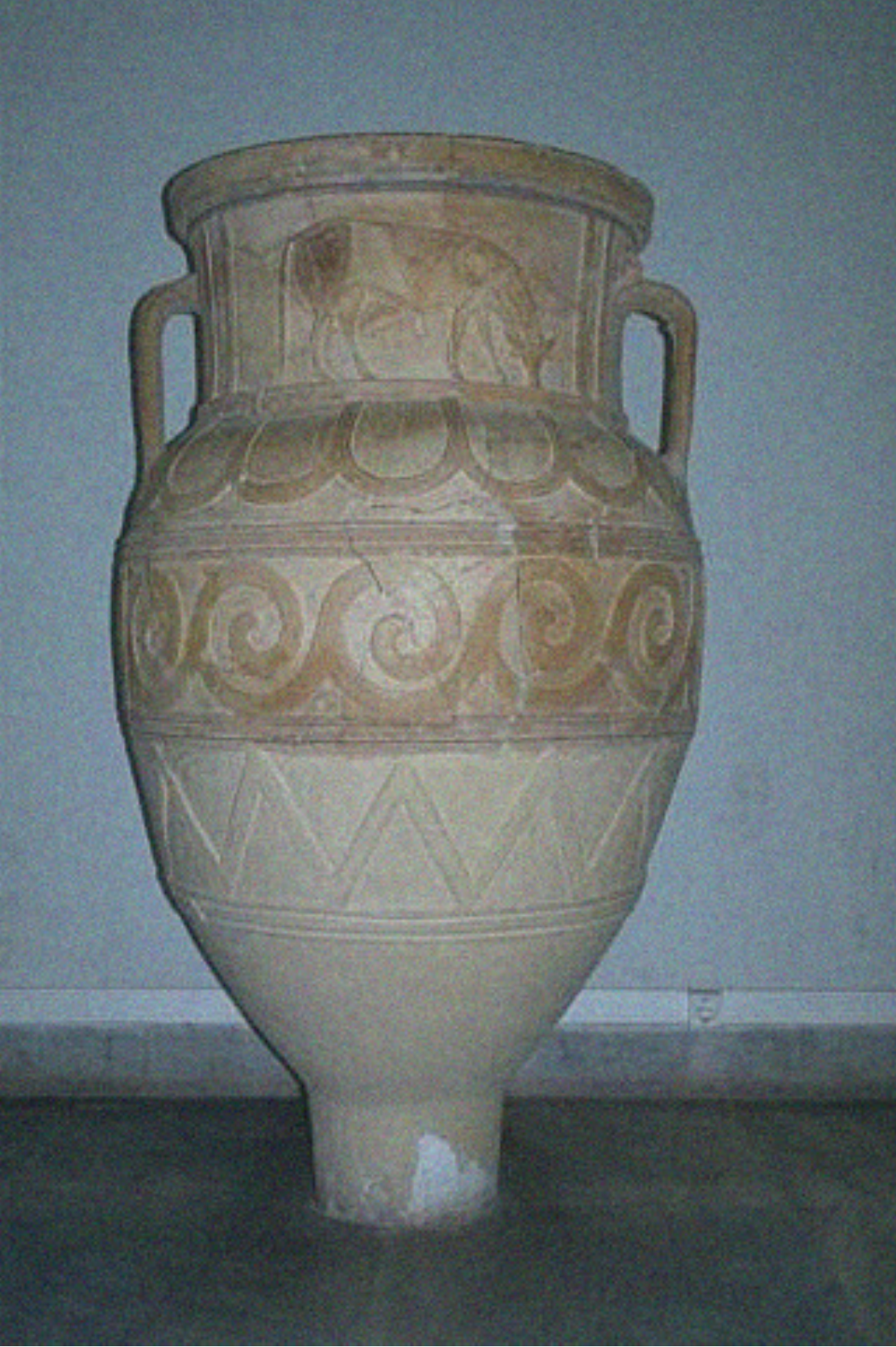} &
\includegraphics[width=\ww, height=\hhh]{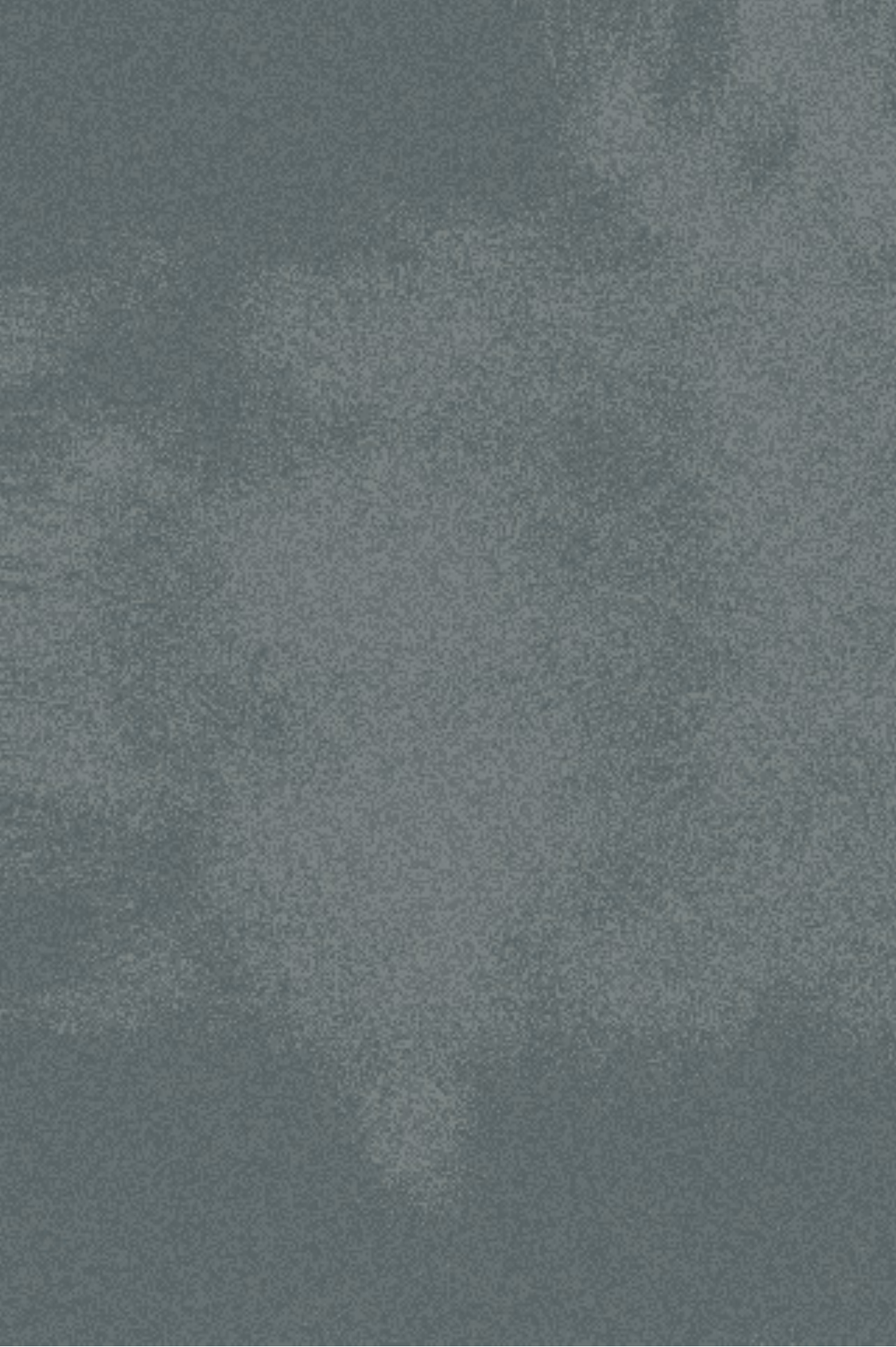} &
\includegraphics[width=\ww, height=\hhh]{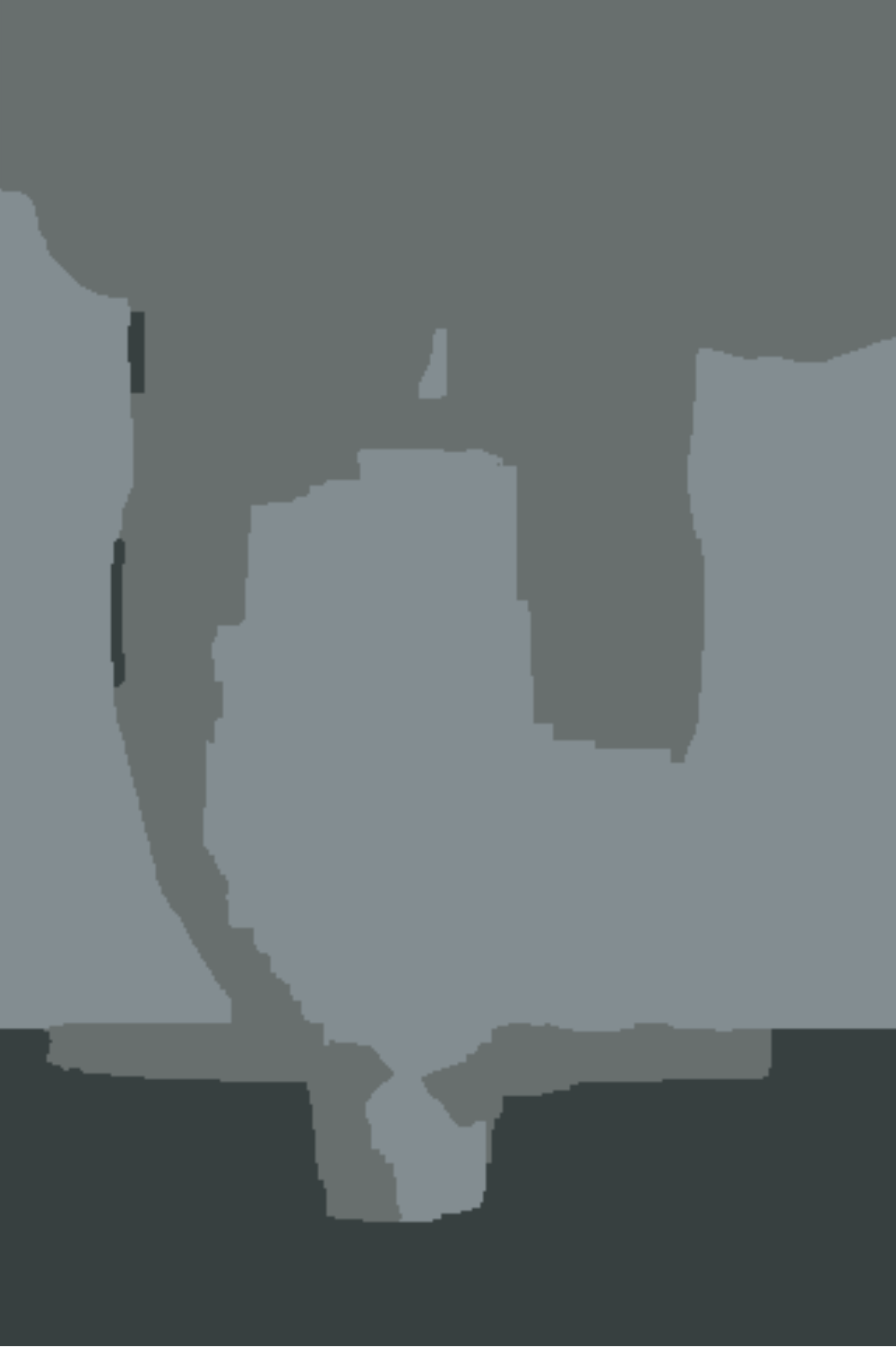} &
\includegraphics[width=\ww, height=\hhh]{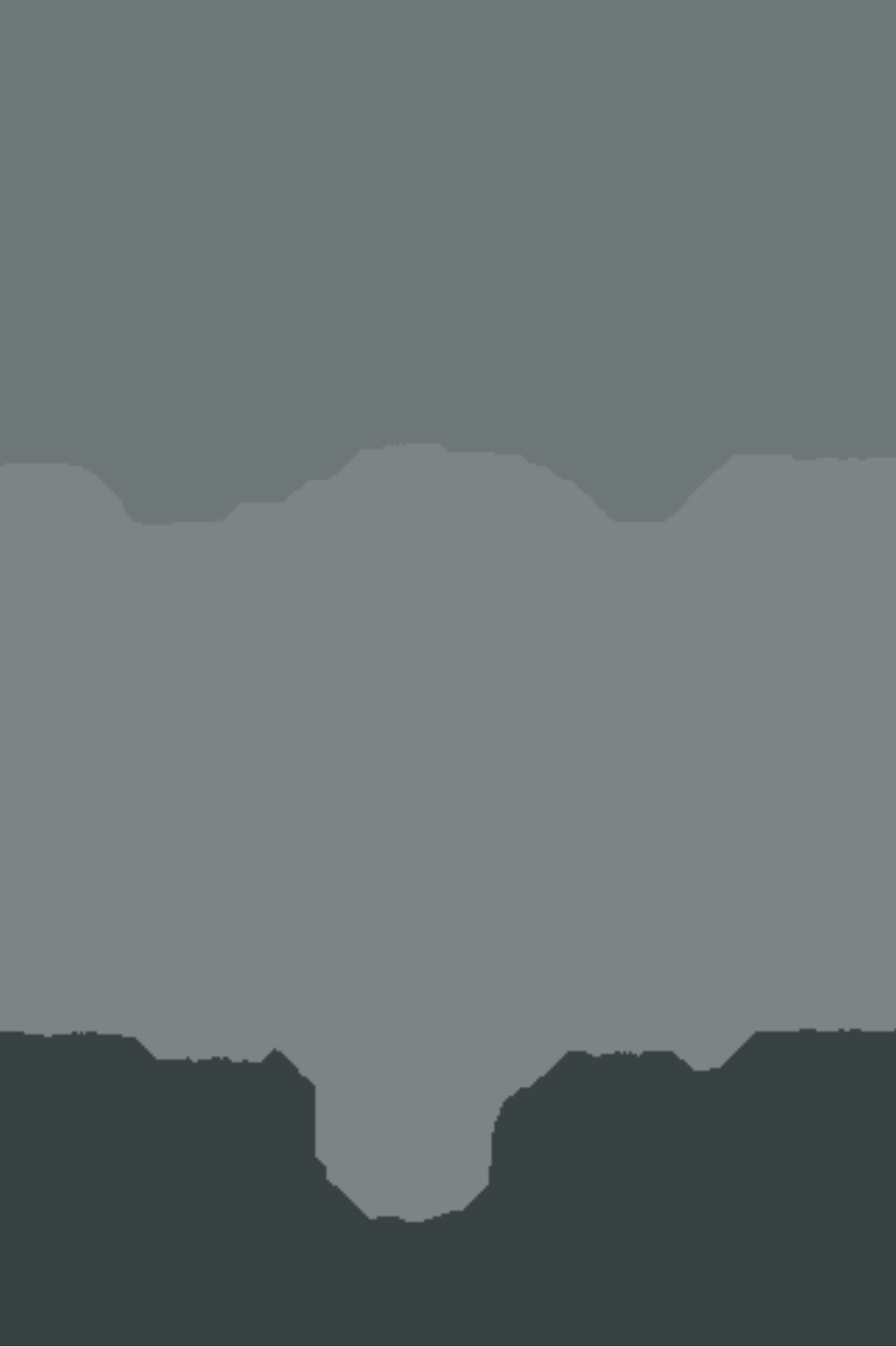} &
\includegraphics[width=\ww, height=\hhh]{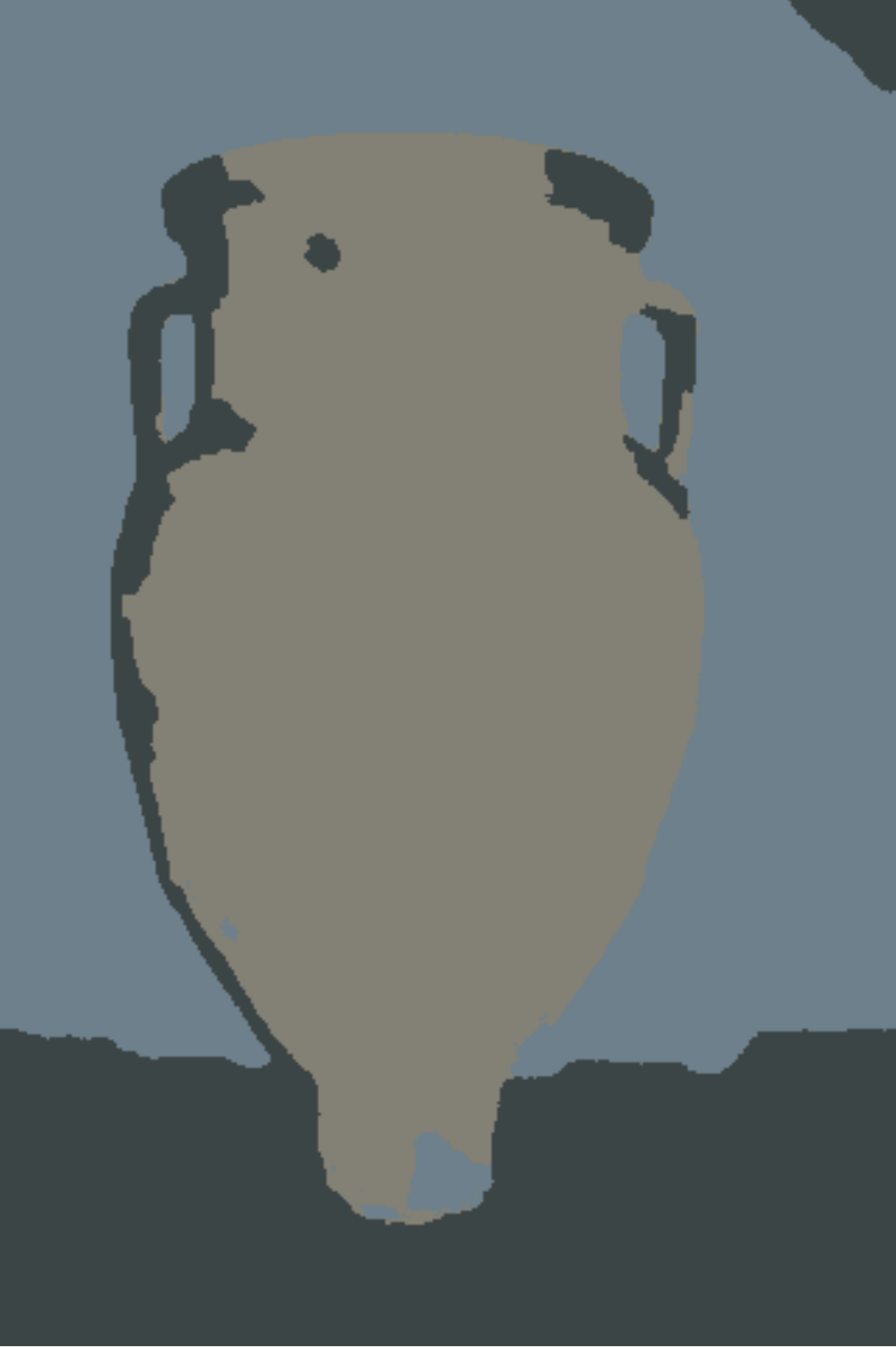} \\
{\small (A) Noisy image} &
{\small (A1) Method \cite{LNZS10}} &
{\small (A2) Method \cite{PCCB09}} &
{\small (A3) Method \cite{SW14}} &
{\small (A4) Ours } \\
\includegraphics[width=\ww, height=\hhh]{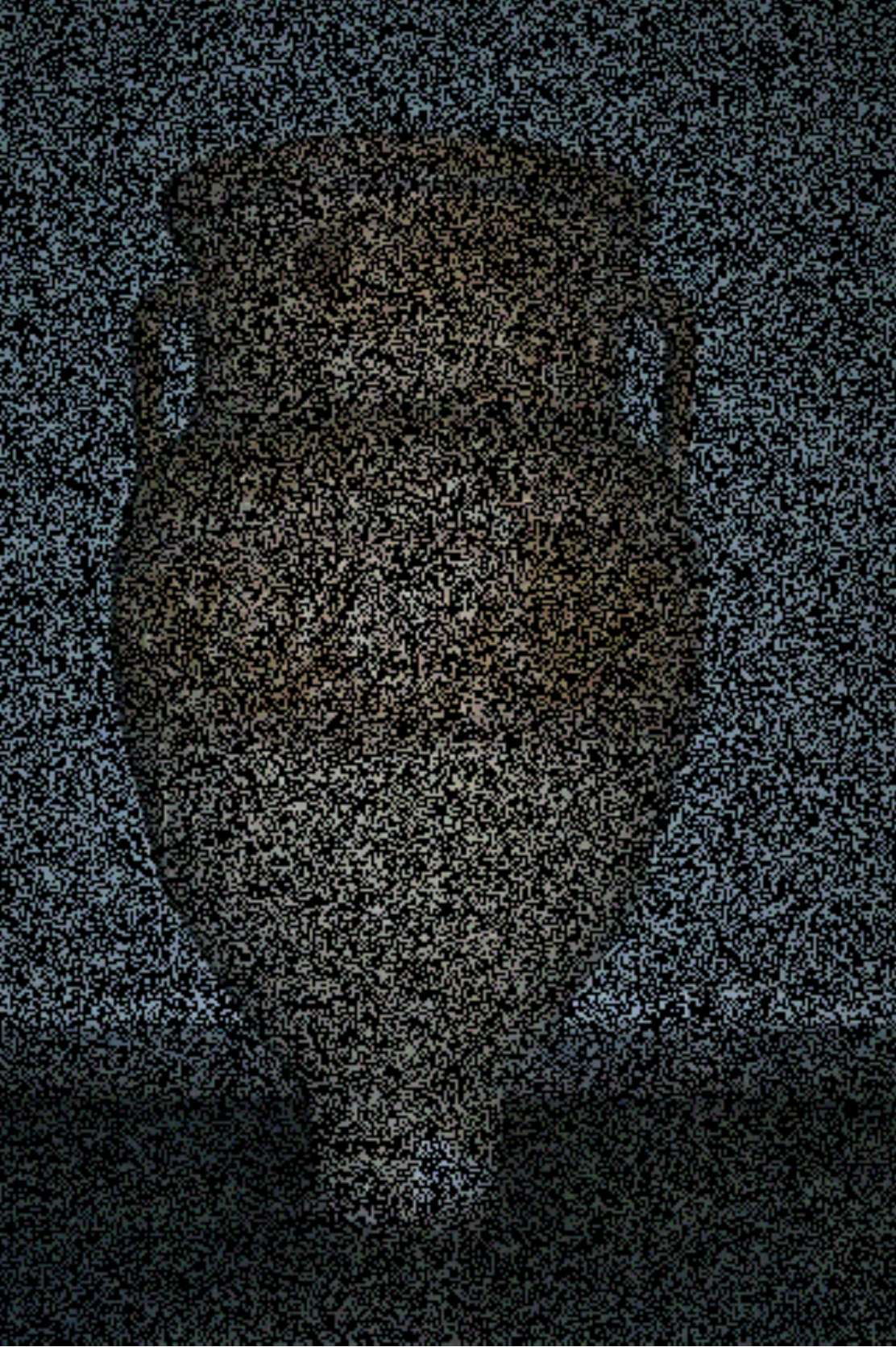} &
\includegraphics[width=\ww, height=\hhh]{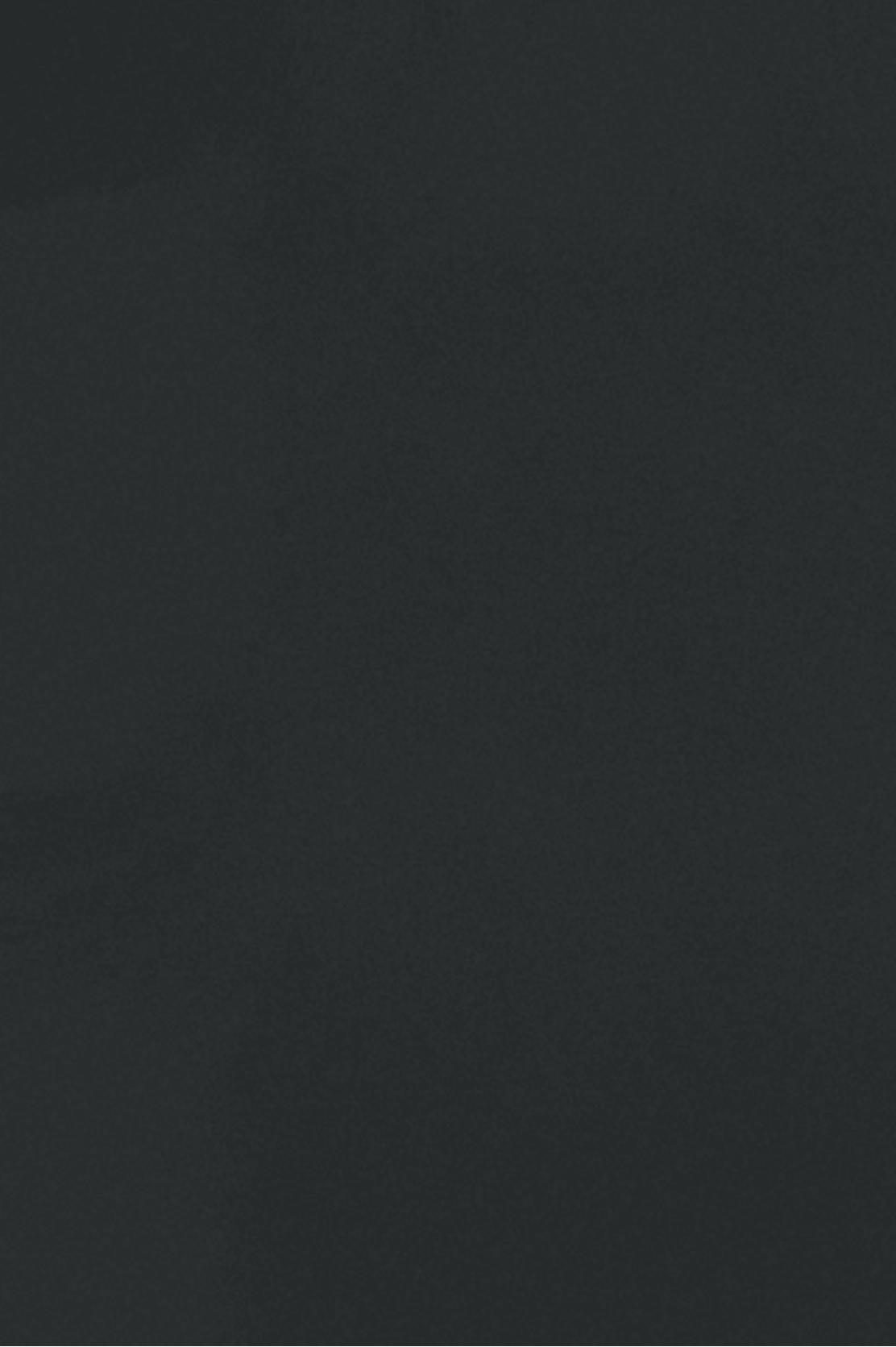} &
\includegraphics[width=\ww, height=\hhh]{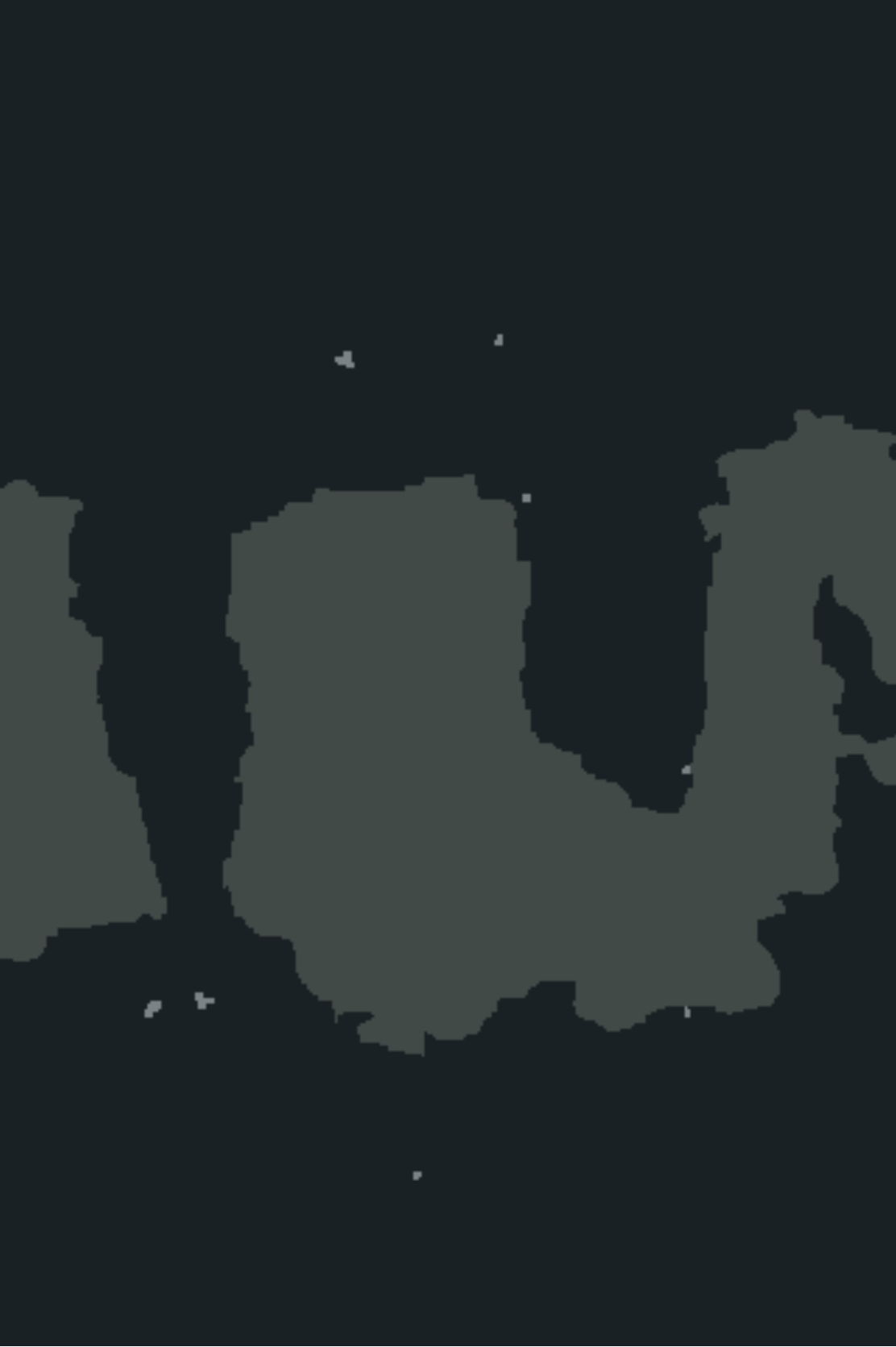} &
\includegraphics[width=\ww, height=\hhh]{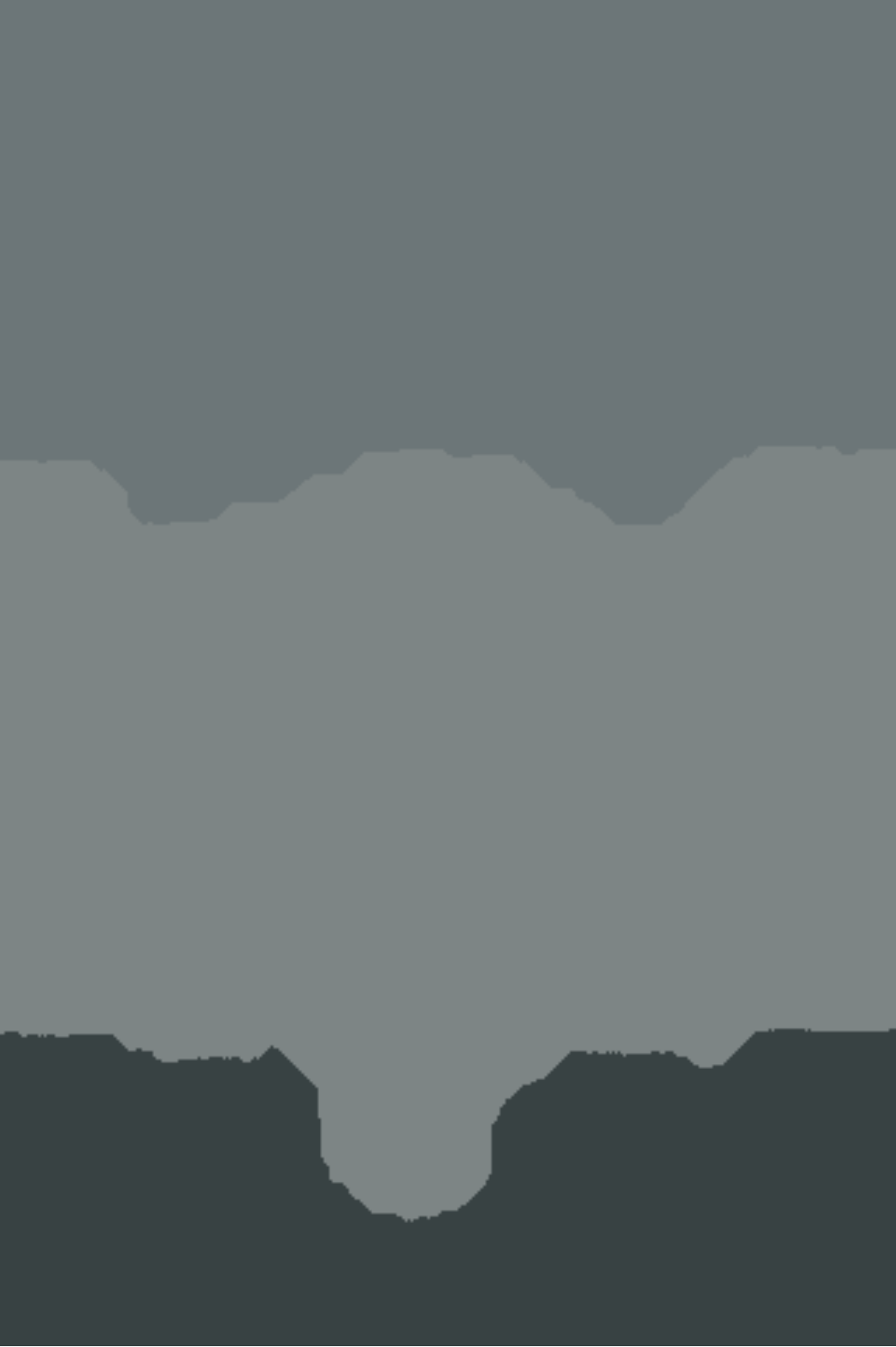} &
\includegraphics[width=\ww, height=\hhh]{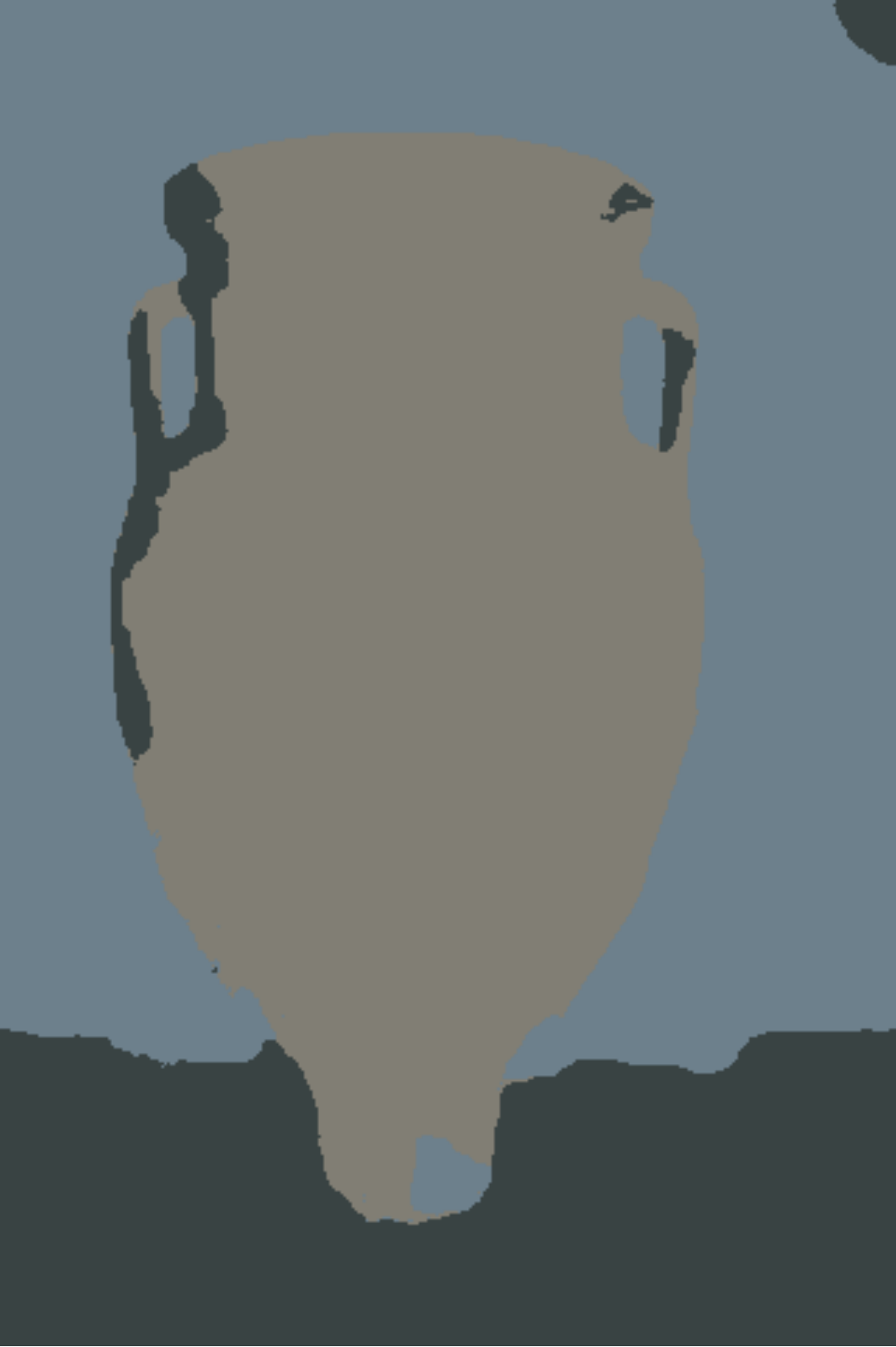} \\
{\small (B) Information} &
{\small (B1) Method \cite{LNZS10}} &
{\small (B2) Method \cite{PCCB09}} &
{\small (B3) Method \cite{SW14}} &
{\small (B4) Ours } \vspace{-0.05in} \\
{\small loss + noise} & & & & \\
\includegraphics[width=\ww, height=\hhh]{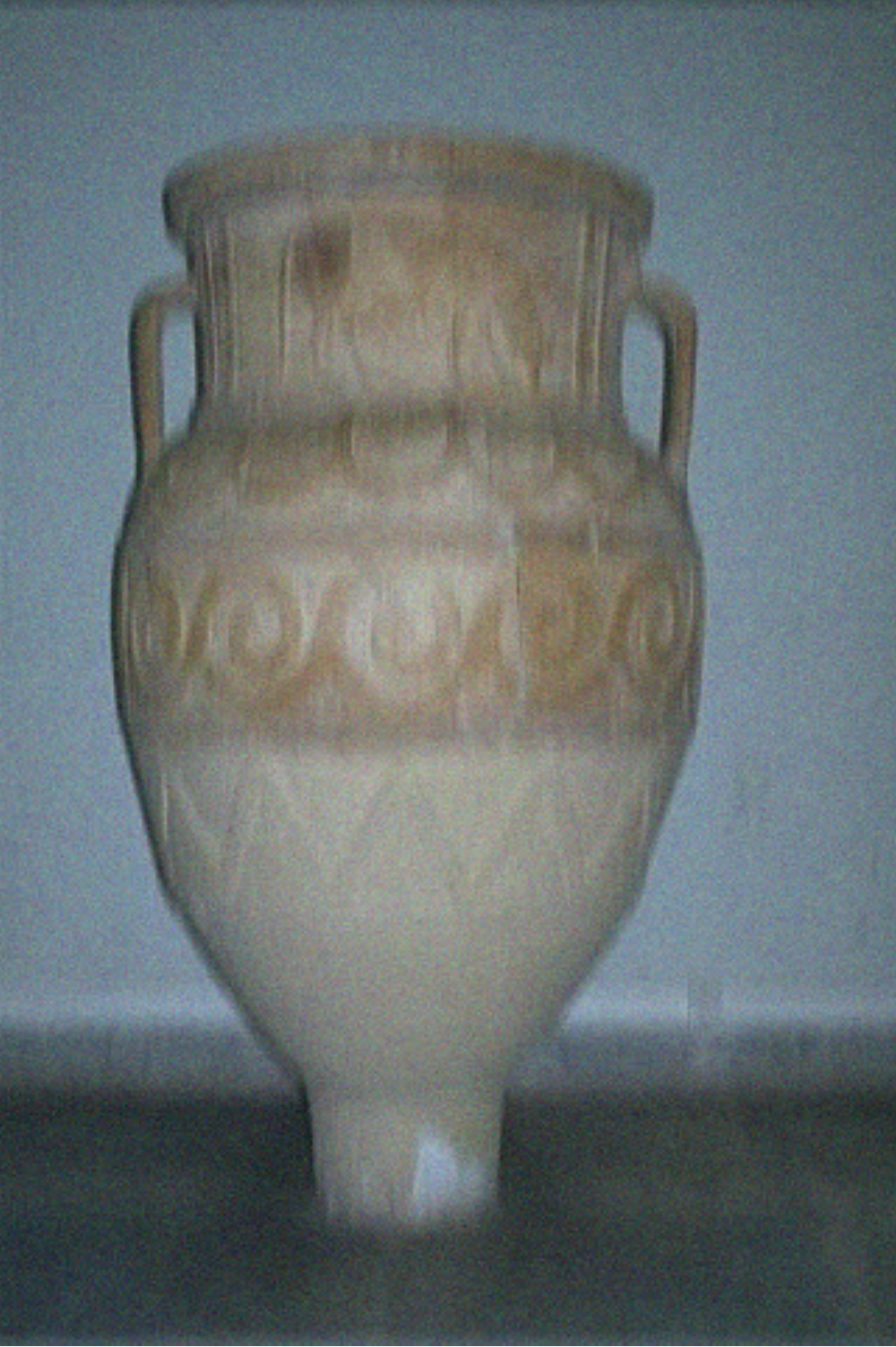} &
\includegraphics[width=\ww, height=\hhh]{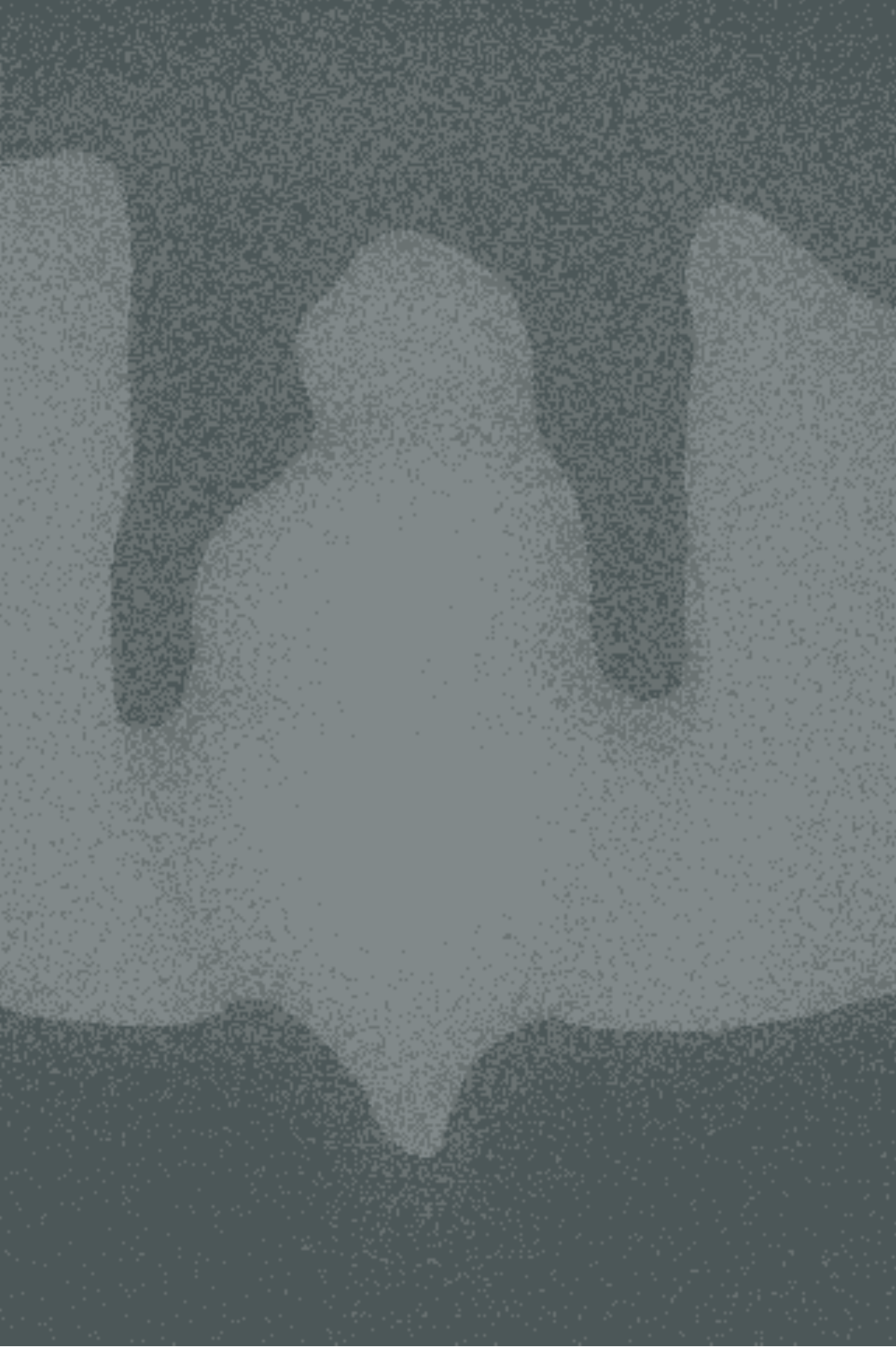} &
\includegraphics[width=\ww, height=\hhh]{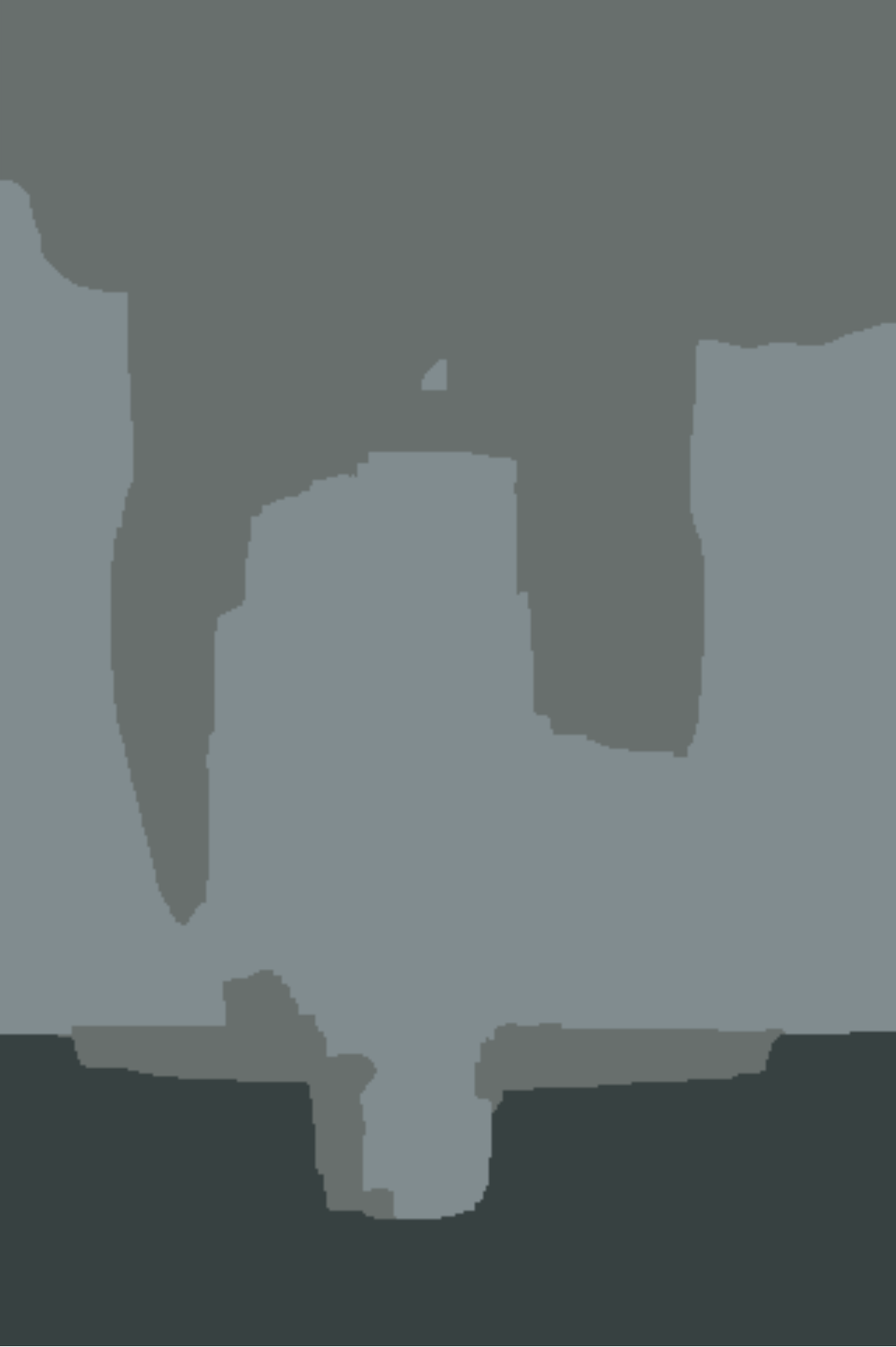} &
\includegraphics[width=\ww, height=\hhh]{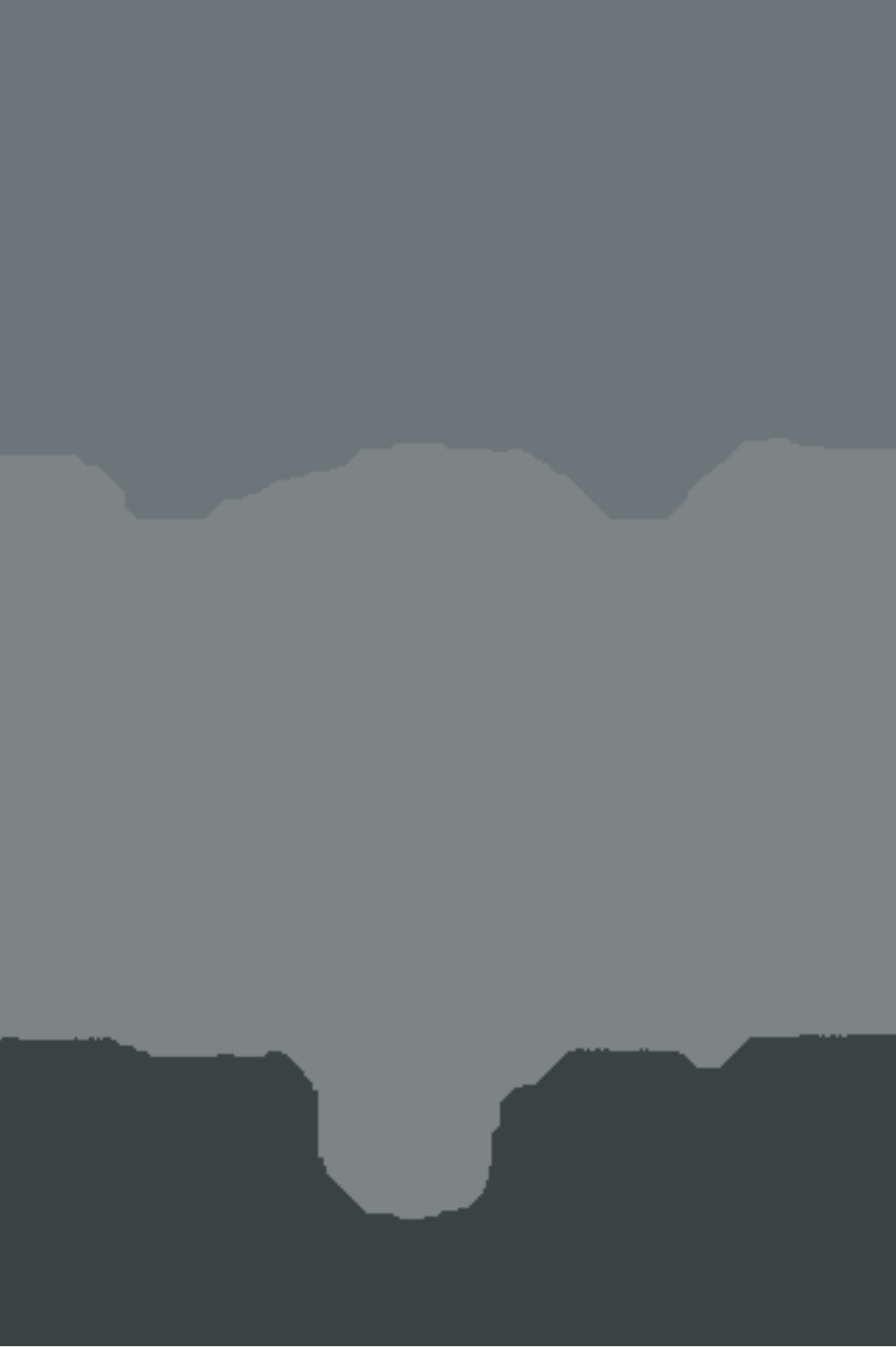} &
\includegraphics[width=\ww, height=\hhh]{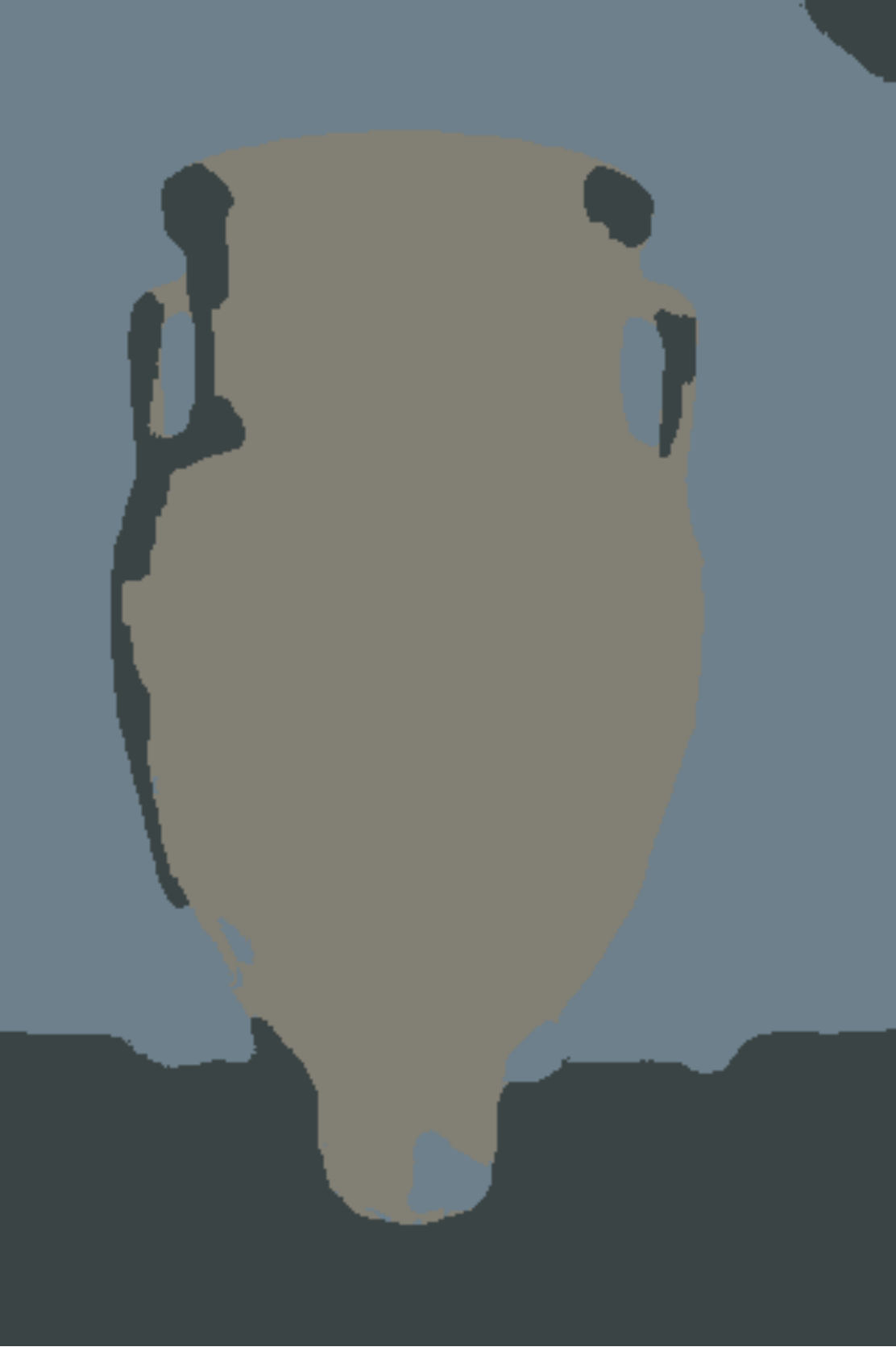} \\
{\small (C) Blur + noise} &
{\small (C1) Method \cite{LNZS10}} &
{\small (C2) Method \cite{PCCB09}} &
{\small (C3) Method \cite{SW14}} &
{\small (C4) Ours }
\end{tabular}
\end{center}
\caption{Three-phase vase segmentation (size: $481\times 321$).
(A): Given Gaussian noisy image with mean 0 and noise 0.001;
(B): Given Gaussian noisy image with $60\%$ information loss;
(C): Given blurry image with Gaussian noise;
(A1-A4), (B1-B4) and (C1-C4): Results of methods \cite{LNZS10},
\cite{PCCB09}, \cite{SW14}, and our SLaT on (A), (B) and (C), respectively.
}\label{threephase-color-china}
\end{figure*}

\begin{figure*}[!htb]
\begin{center}
\begin{tabular}{ccccc}
\includegraphics[width=\ww, height=\hh]{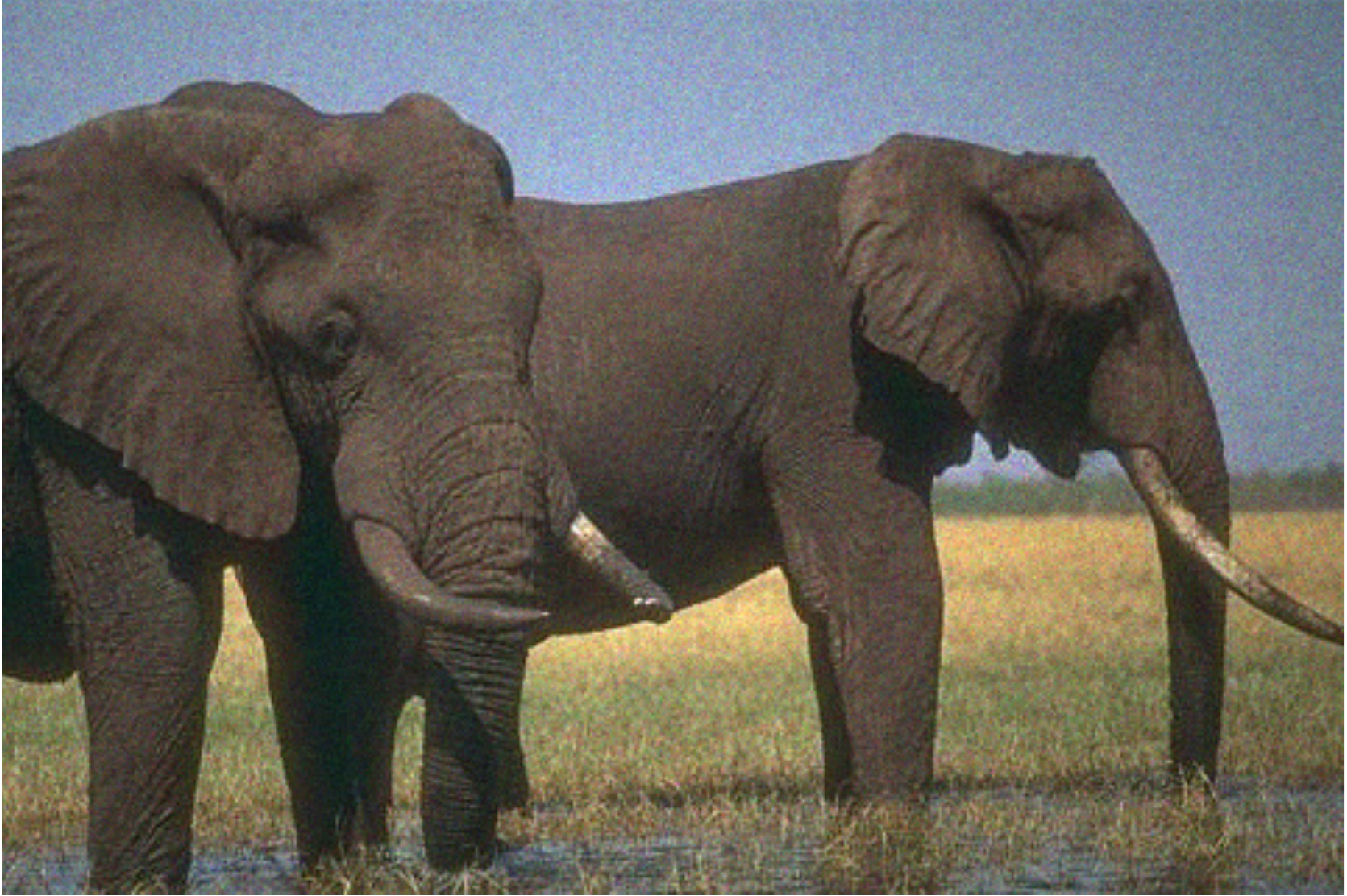} &
\includegraphics[width=\ww, height=\hh]{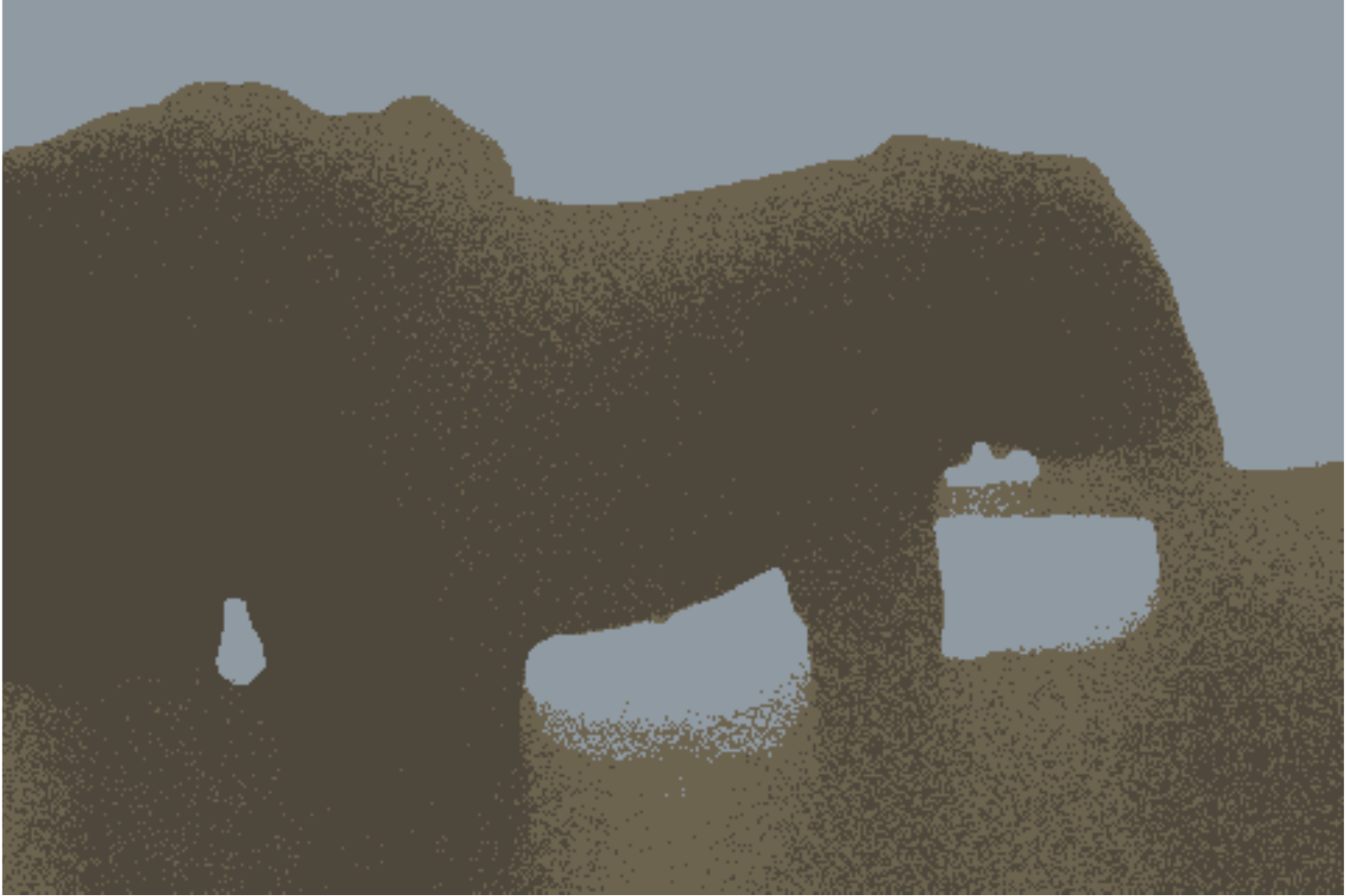} &
\includegraphics[width=\ww, height=\hh]{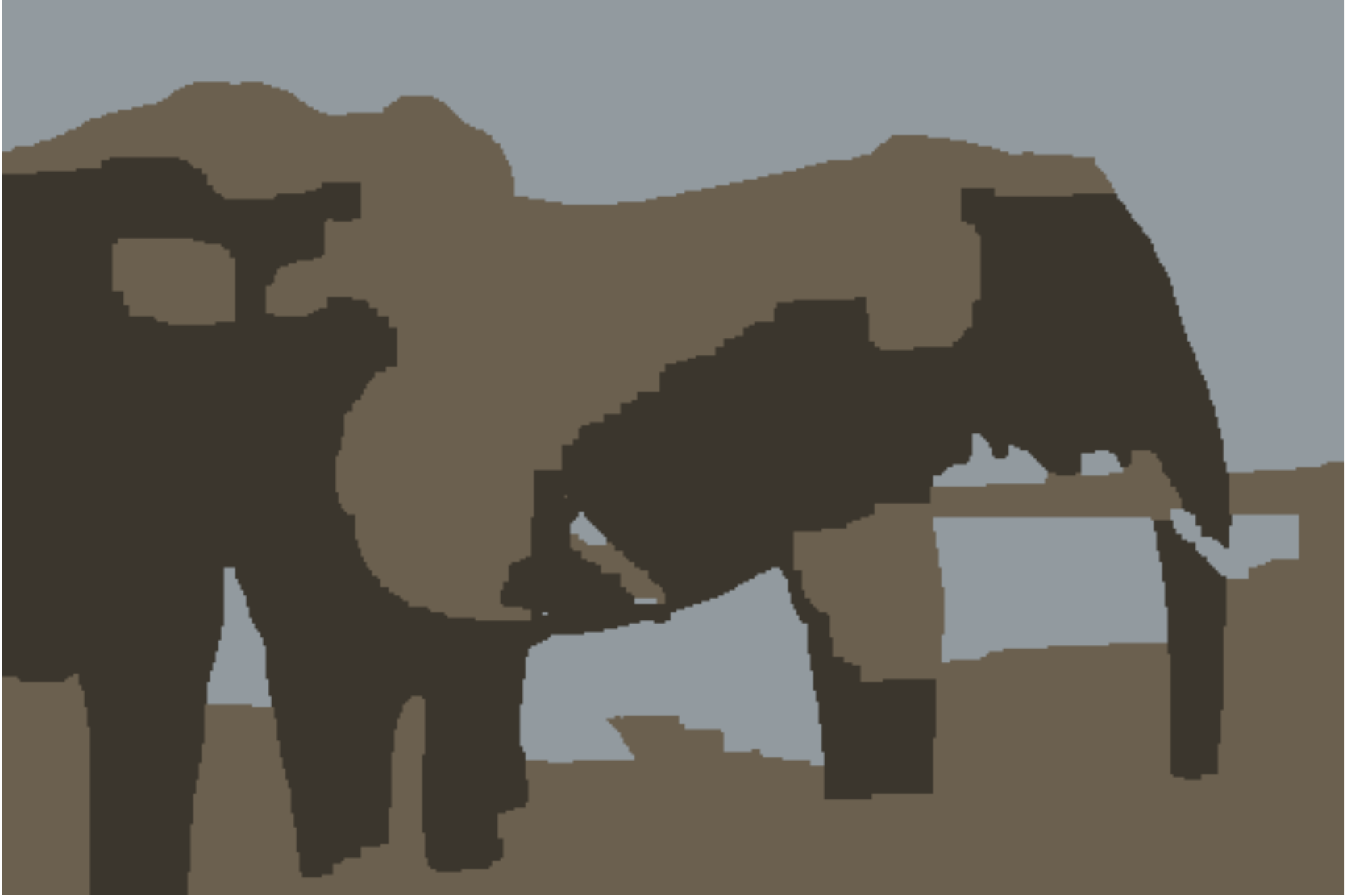} &
\includegraphics[width=\ww, height=\hh]{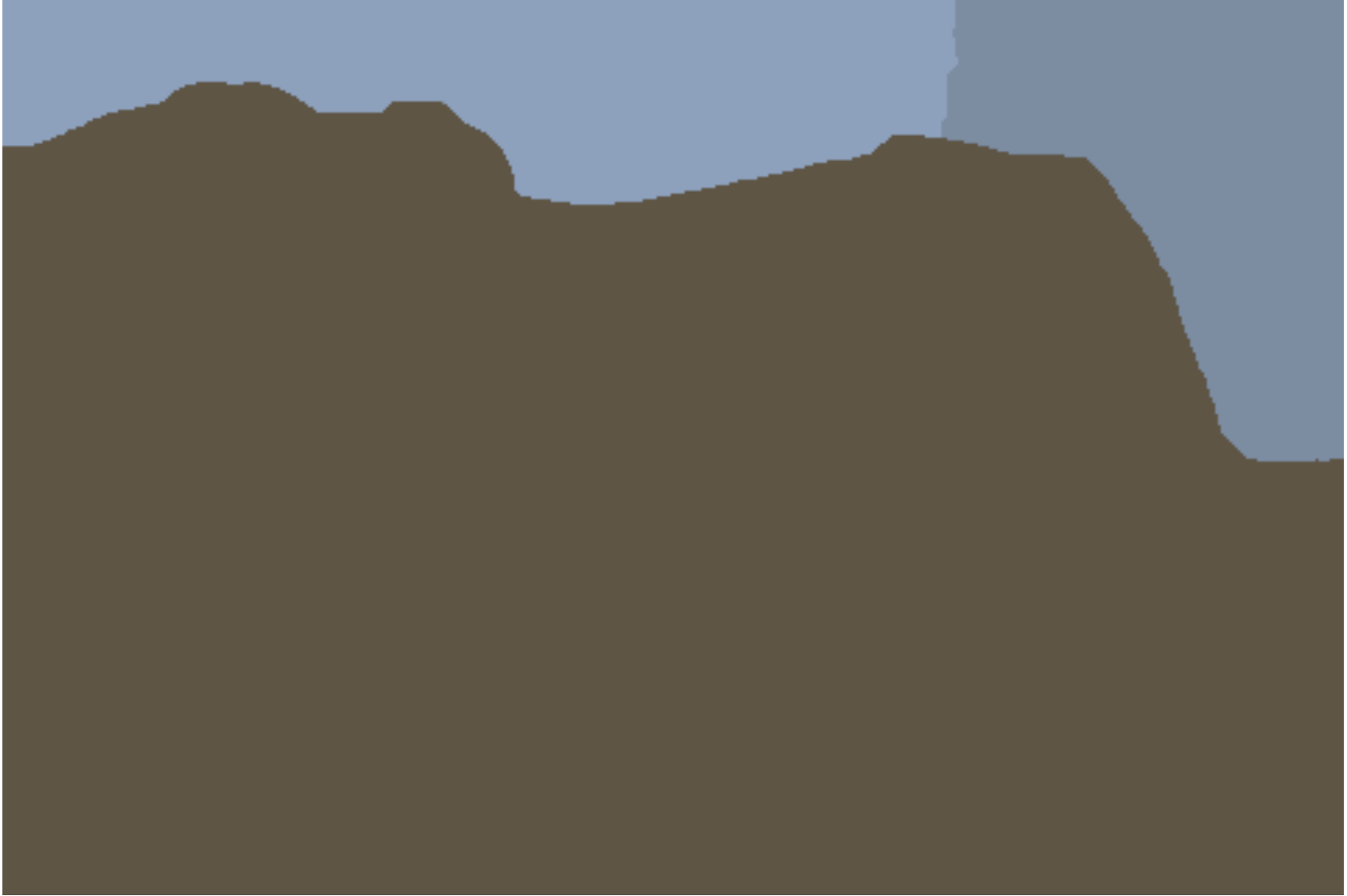} &
\includegraphics[width=\ww, height=\hh]{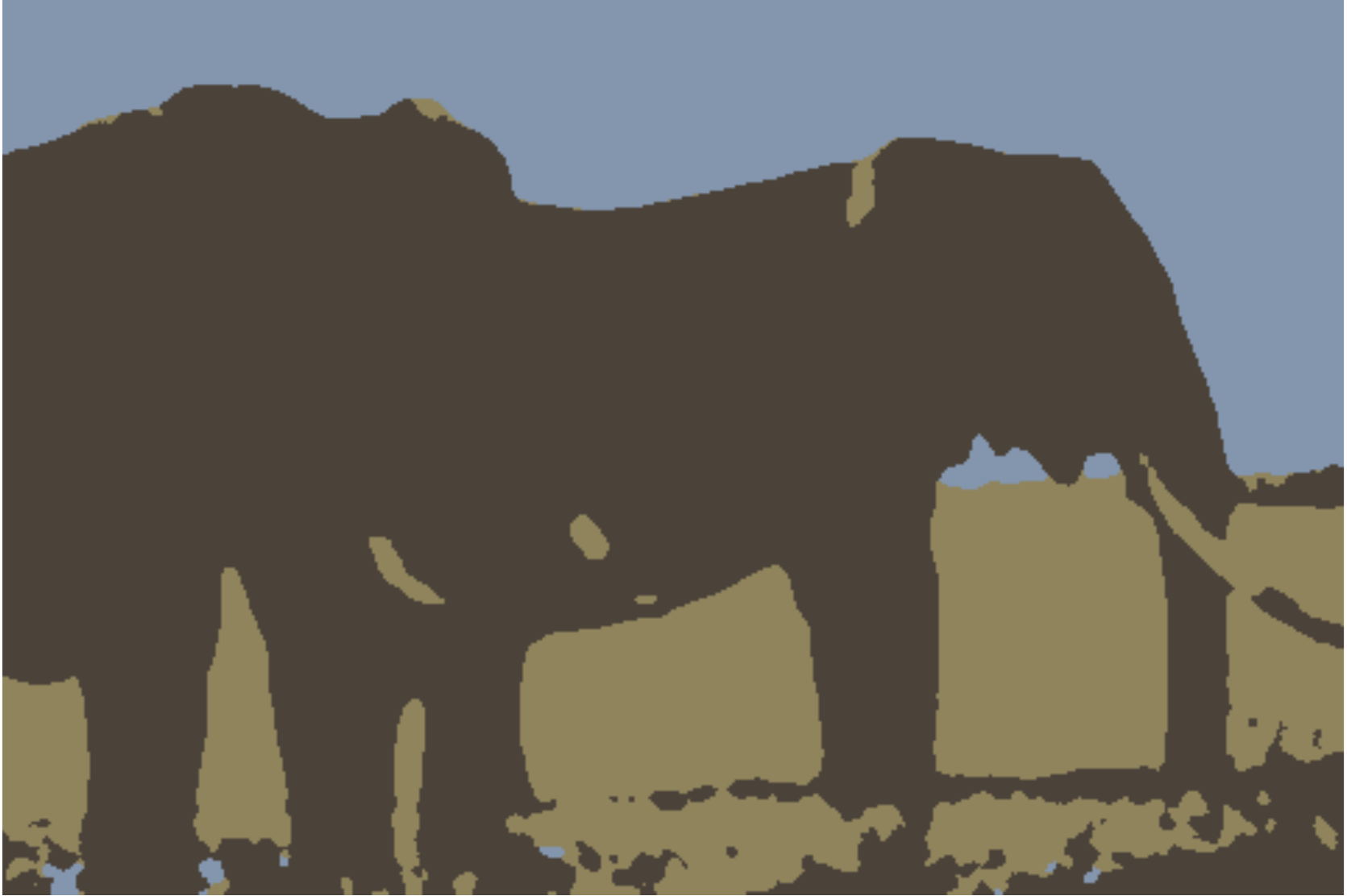} \\
{\small (A) Noisy image} &
{\small (A1) Method \cite{LNZS10}} &
{\small (A2) Method \cite{PCCB09}} &
{\small (A3) Method \cite{SW14}} &
{\small (A4) Ours } \\
\includegraphics[width=\ww, height=\hh]{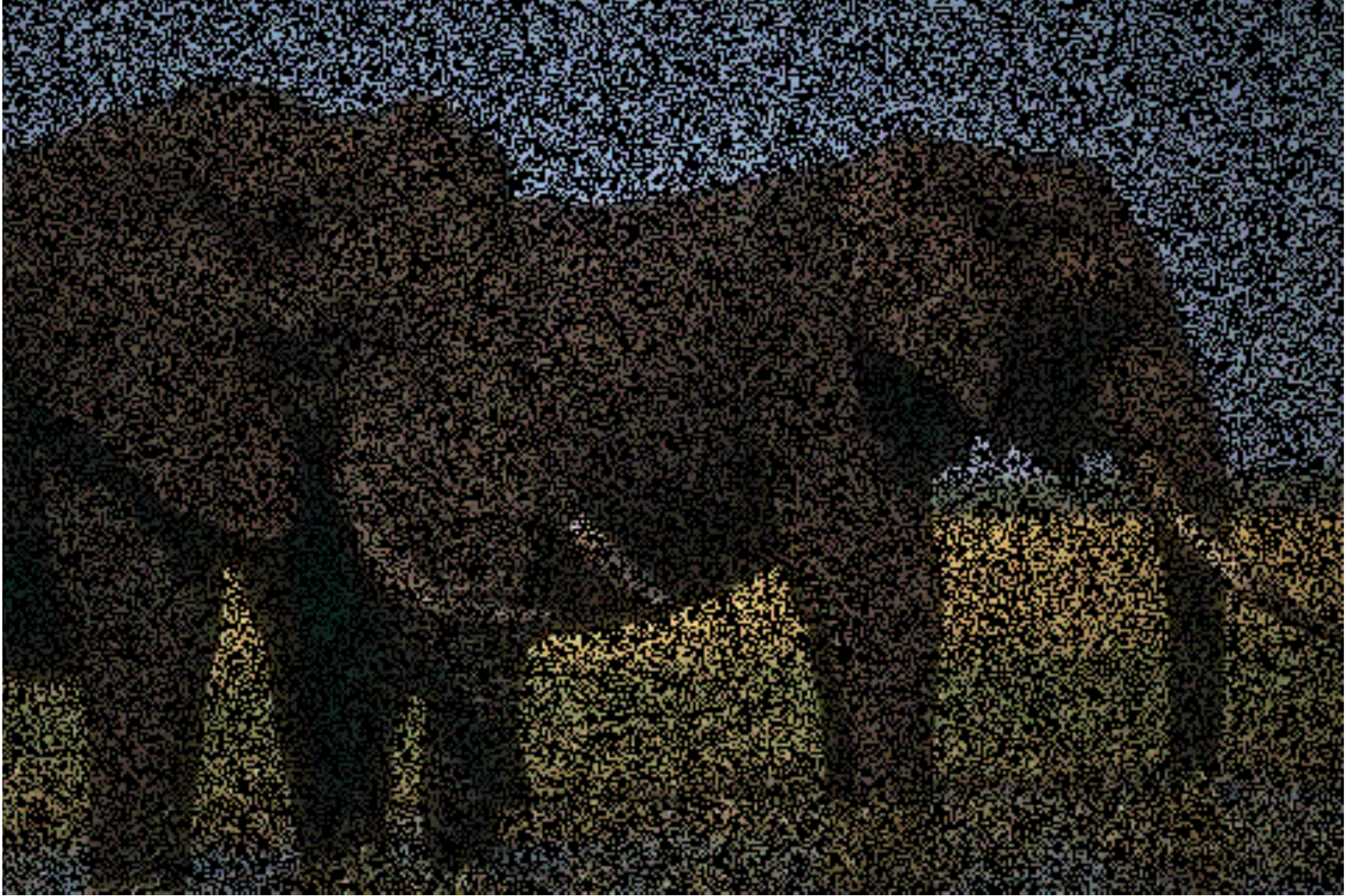} &
\includegraphics[width=\ww, height=\hh]{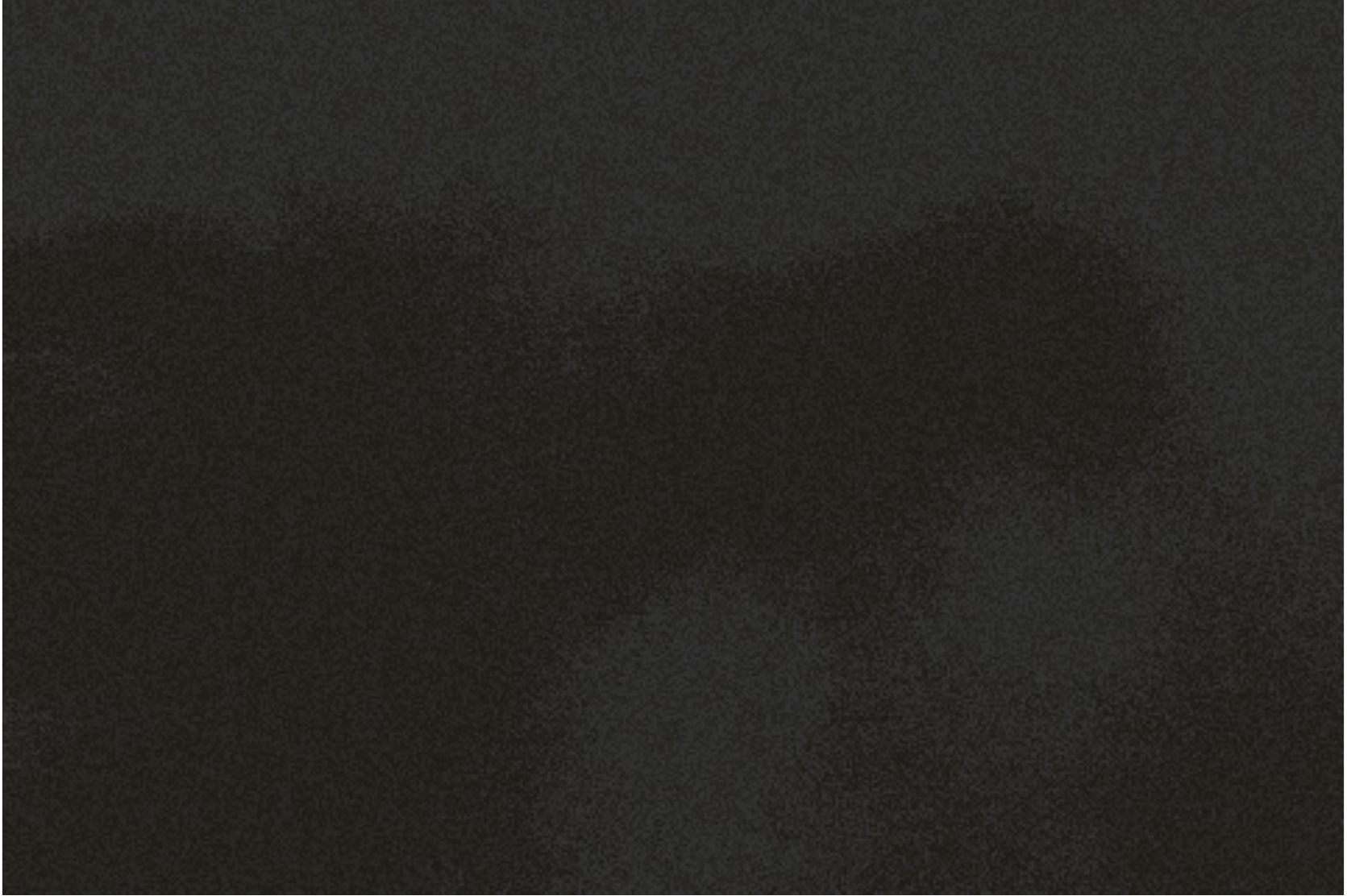} &
\includegraphics[width=\ww, height=\hh]{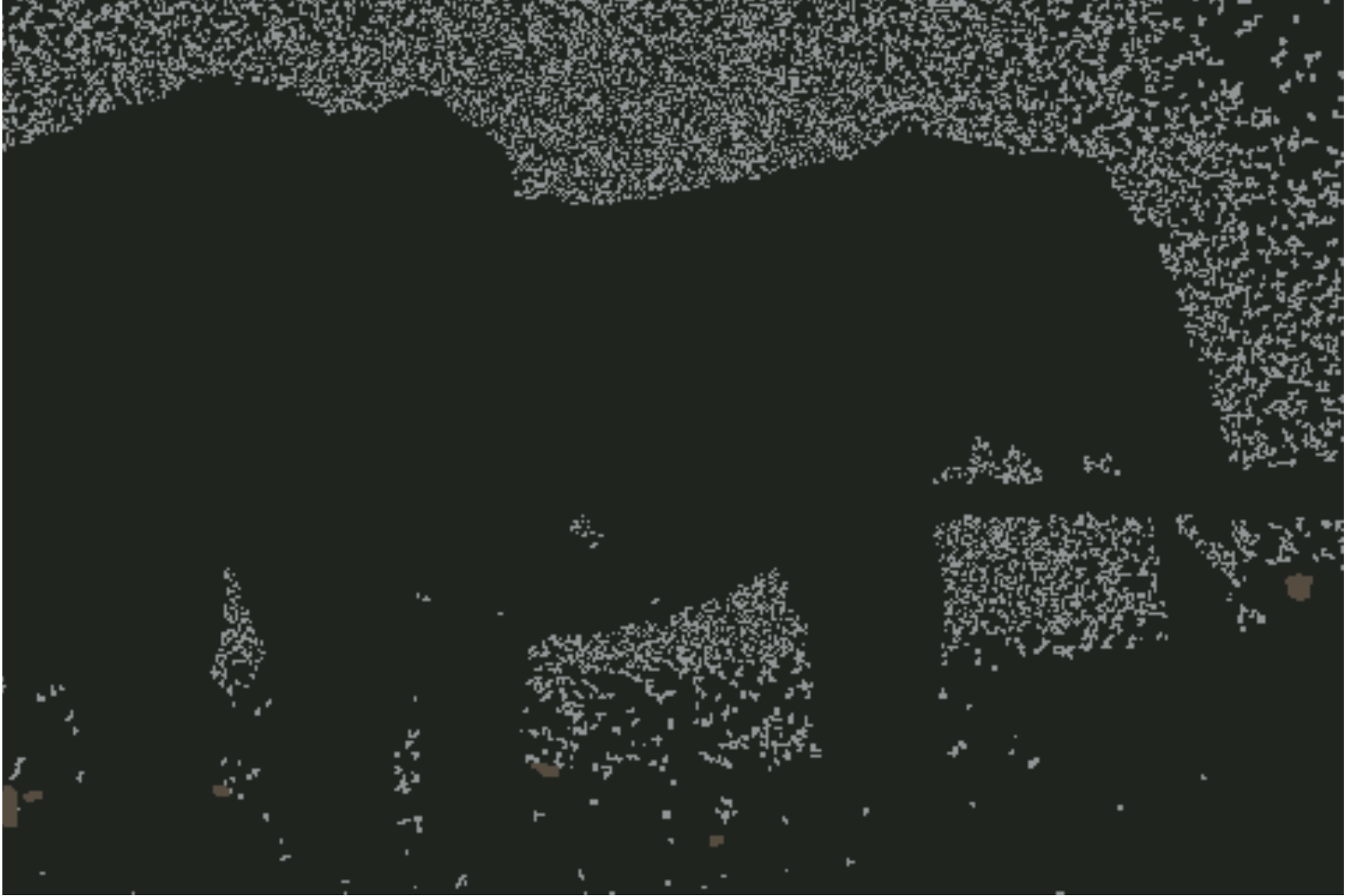} &
\includegraphics[width=\ww, height=\hh]{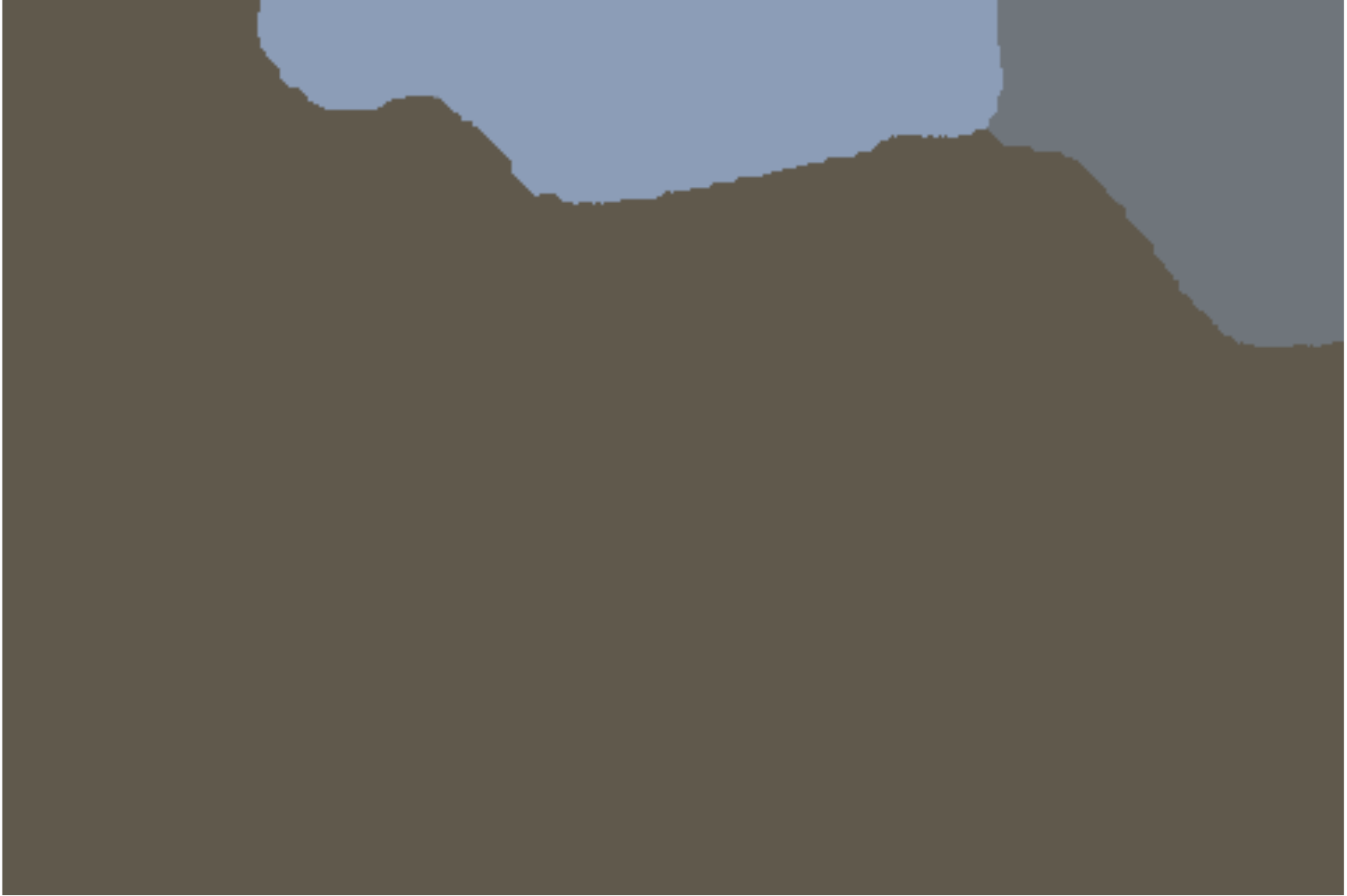} &
\includegraphics[width=\ww, height=\hh]{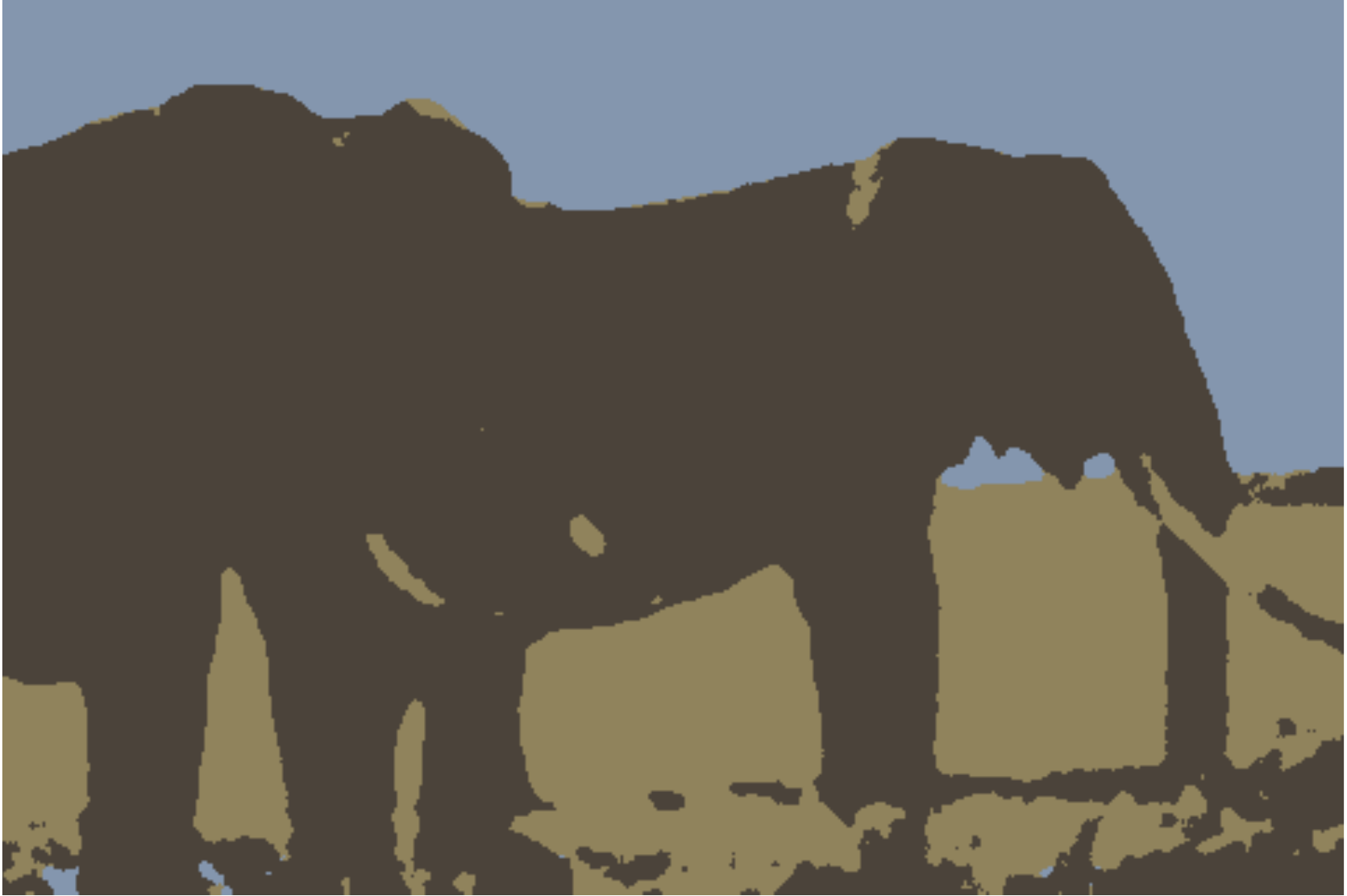} \\
{\small (B) Information} &
{\small (B1) Method \cite{LNZS10}} &
{\small (B2) Method \cite{PCCB09}} &
{\small (B3) Method \cite{SW14}} &
{\small (B4) Ours } \vspace{-0.05in} \\
{\small loss + noise} & & & & \\
\includegraphics[width=\ww, height=\hh]{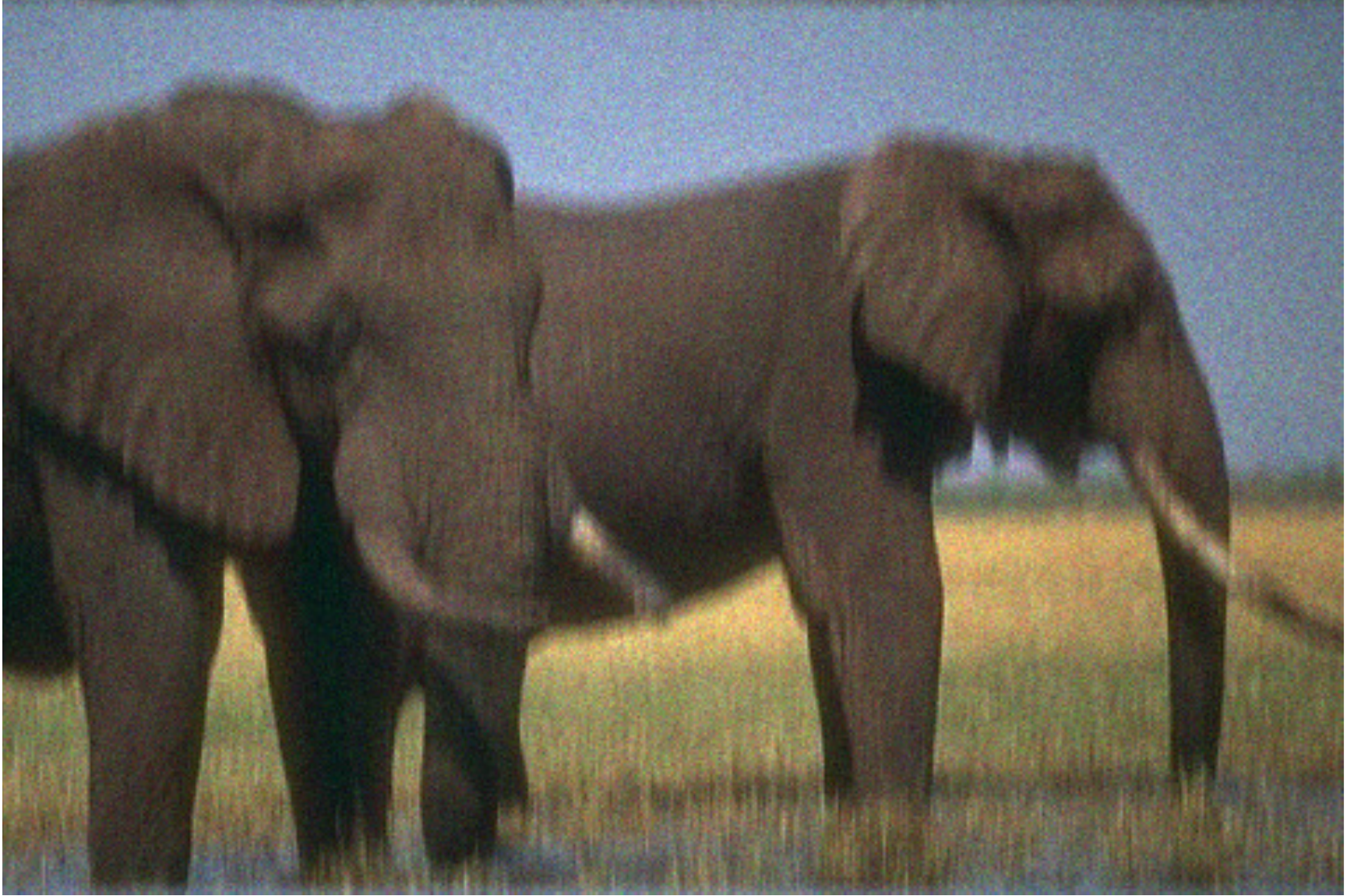} &
\includegraphics[width=\ww, height=\hh]{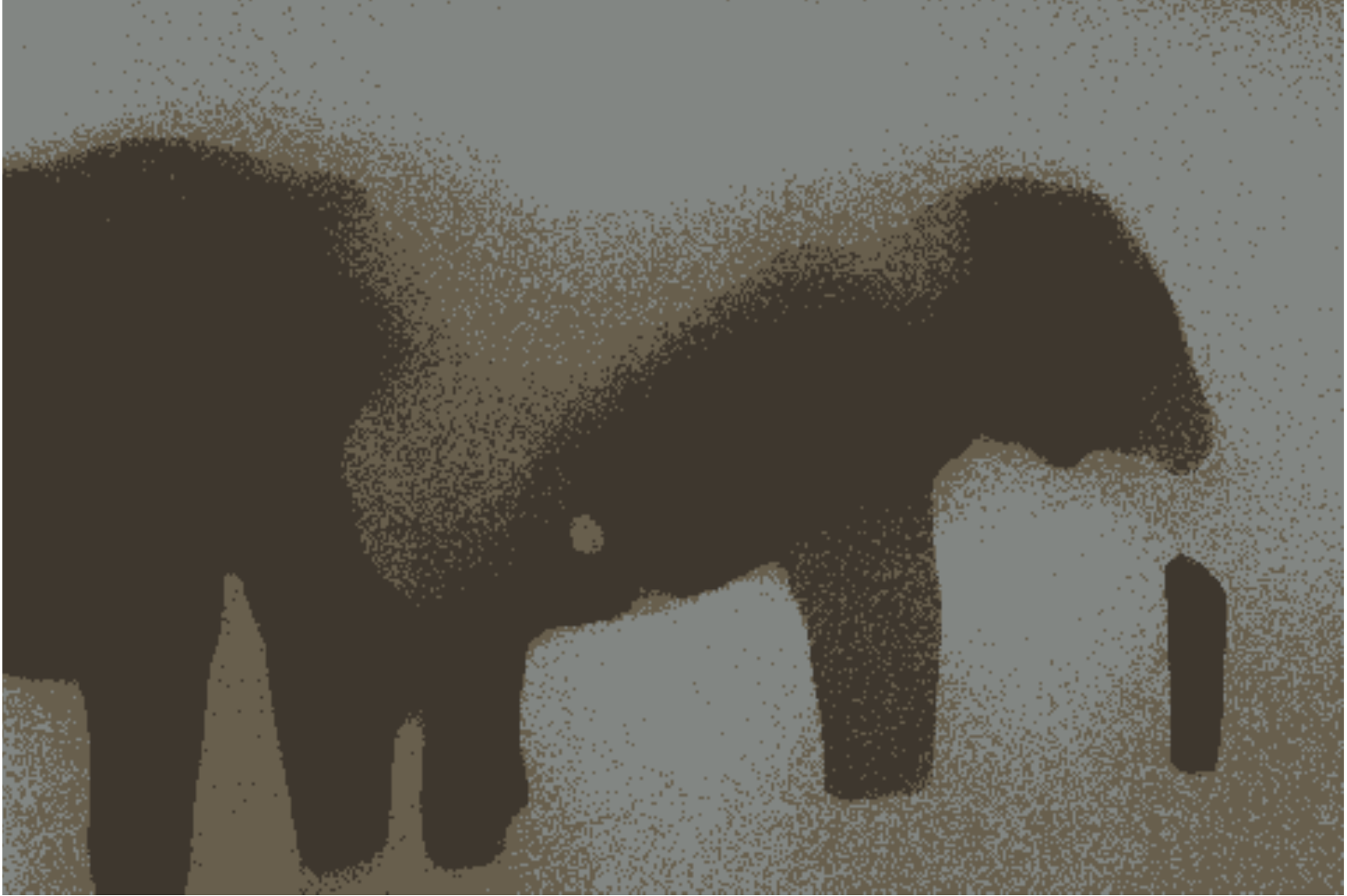} &
\includegraphics[width=\ww, height=\hh]{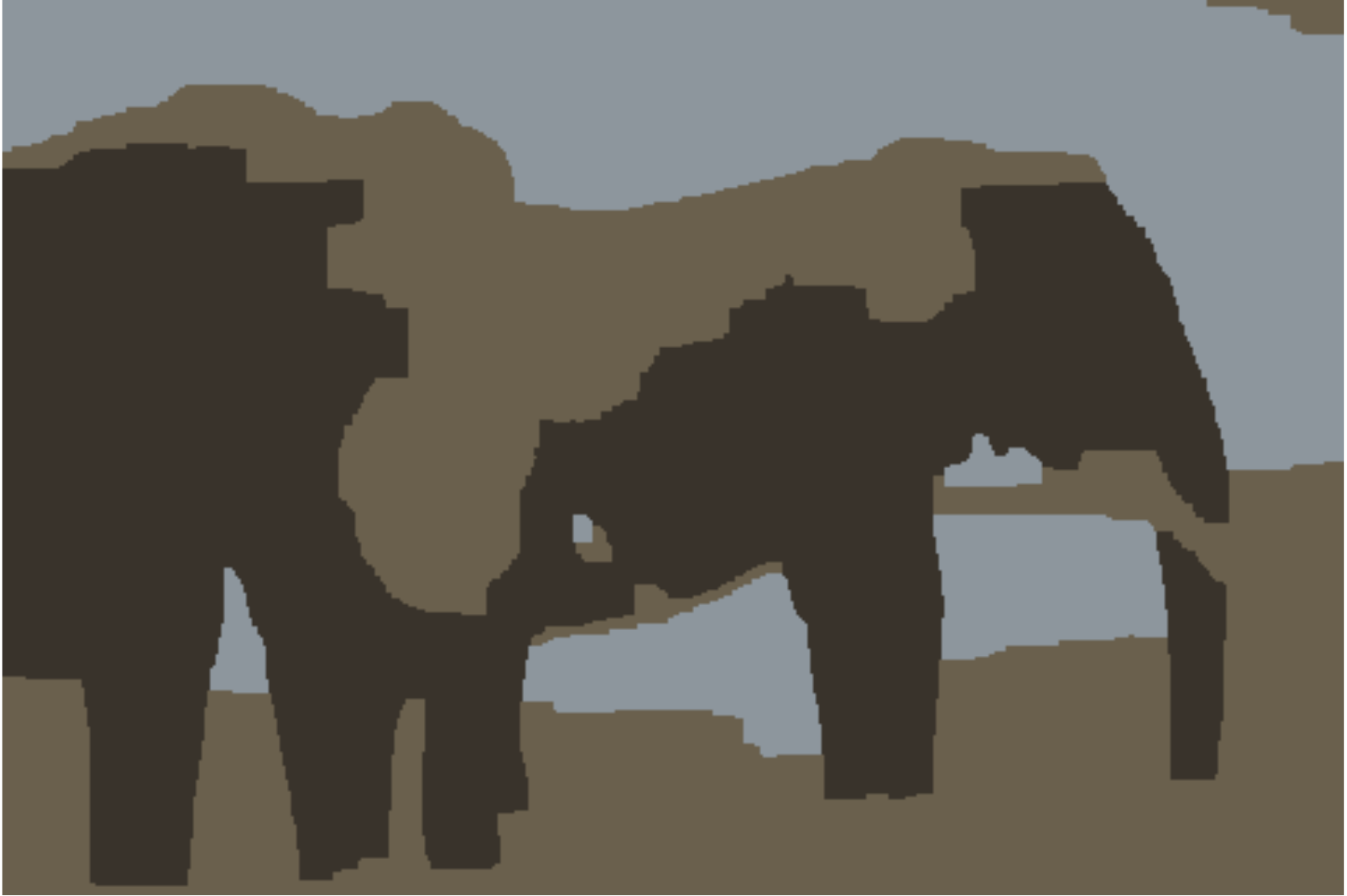} &
\includegraphics[width=\ww, height=\hh]{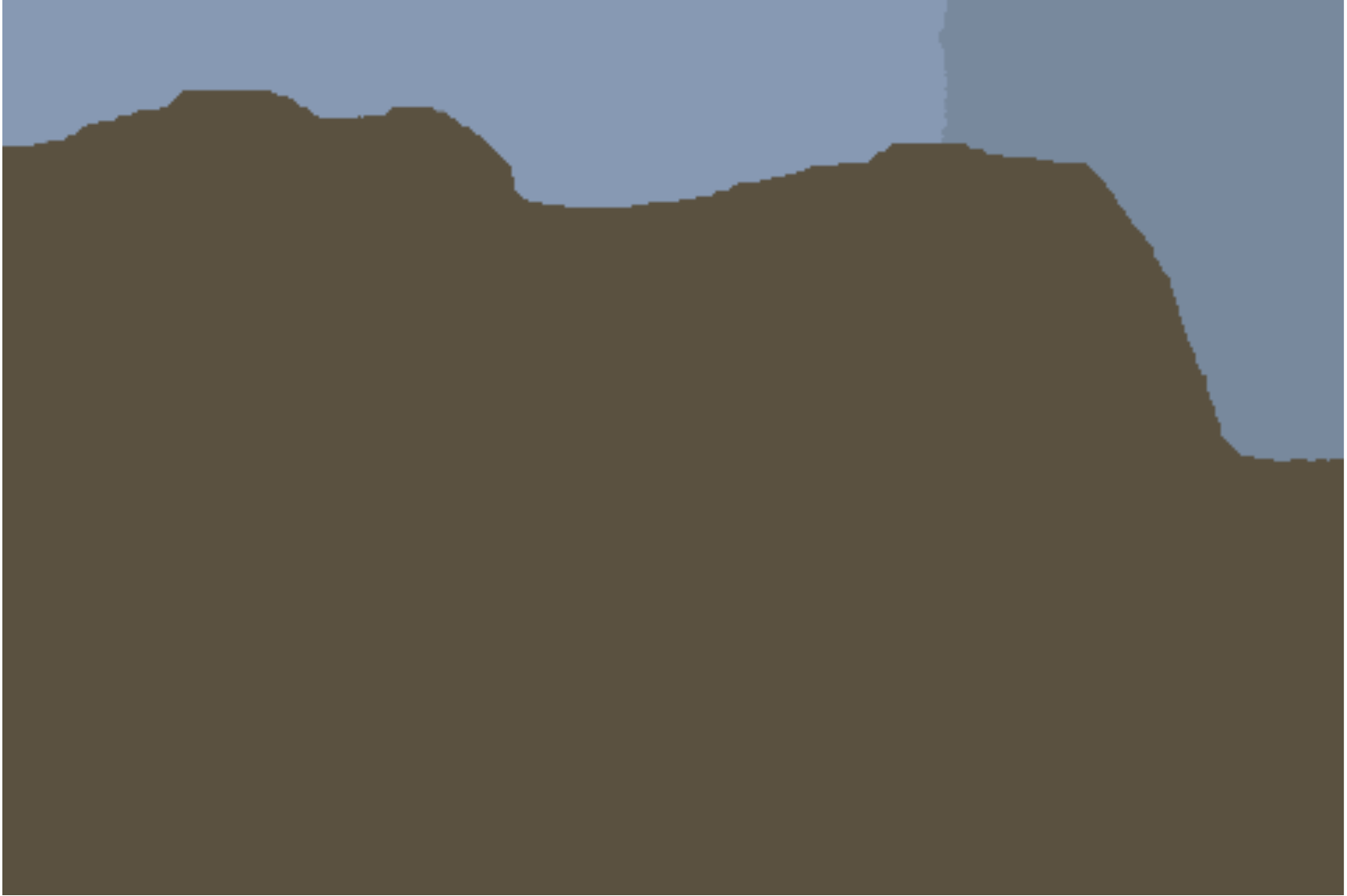} &
\includegraphics[width=\ww, height=\hh]{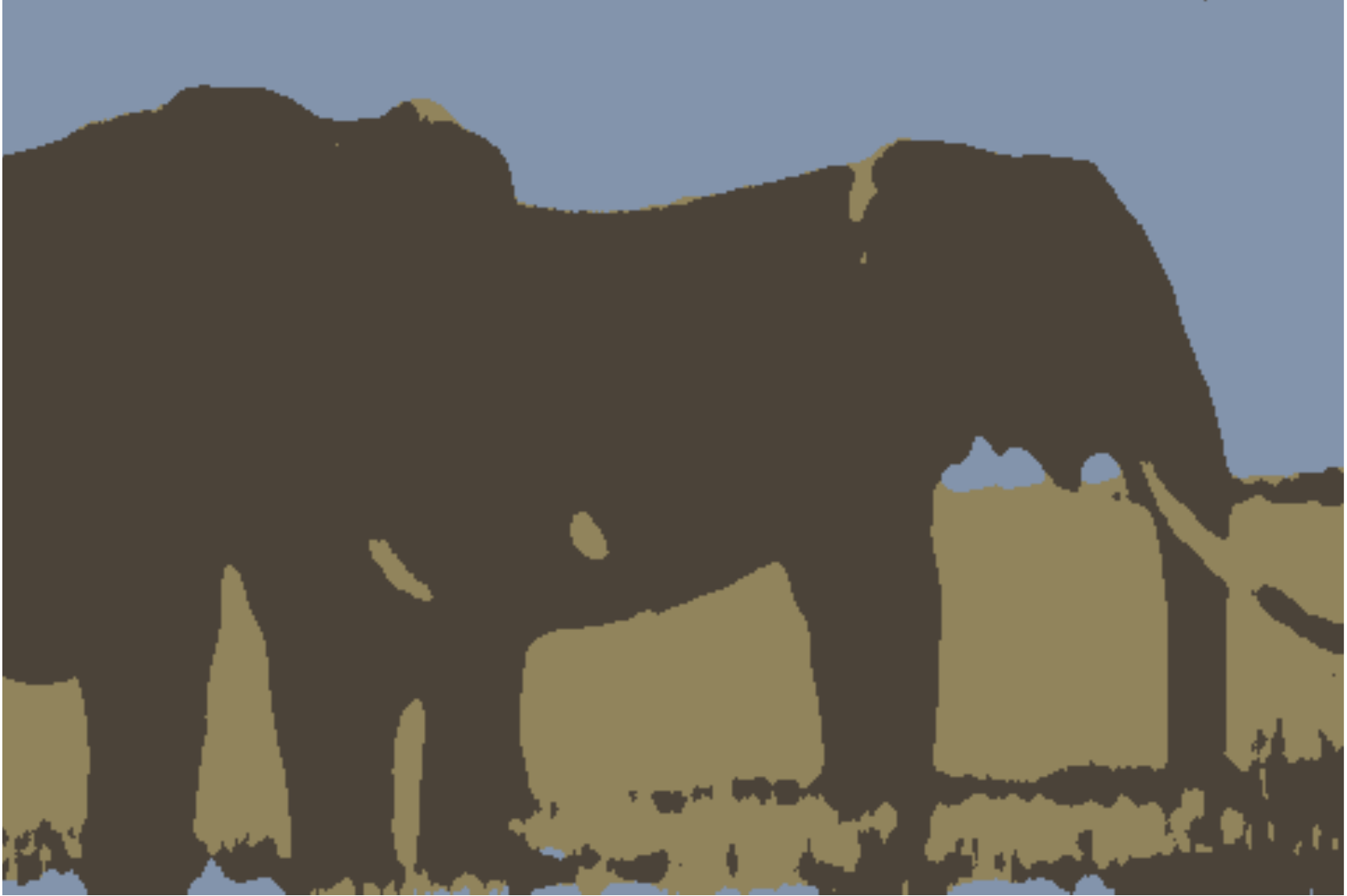} \\
{\small (C) Blur + noise} &
{\small (C1) Method \cite{LNZS10}} &
{\small (C2) Method \cite{PCCB09}} &
{\small (C3) Method \cite{SW14}} &
{\small (C4) Ours }
\end{tabular}
\end{center}
\caption{Three-phase elephant segmentation (size: $321\times 481$).
(A): Given Poisson noisy image;
(B): Given Poisson noisy image with $60\%$ information loss;
(C): Given blurry image with Poisson noise;
(A1-A4), (B1-B4) and (C1-C4): Results of methods \cite{LNZS10},
\cite{PCCB09}, \cite{SW14}, and our SLaT on (A), (B) and (C), respectively.
}\label{threephase-color-elephant}
\end{figure*}

\begin{figure*}[!htb]
\begin{center}
\begin{tabular}{ccccc}
\includegraphics[width=\ww, height=\hh]{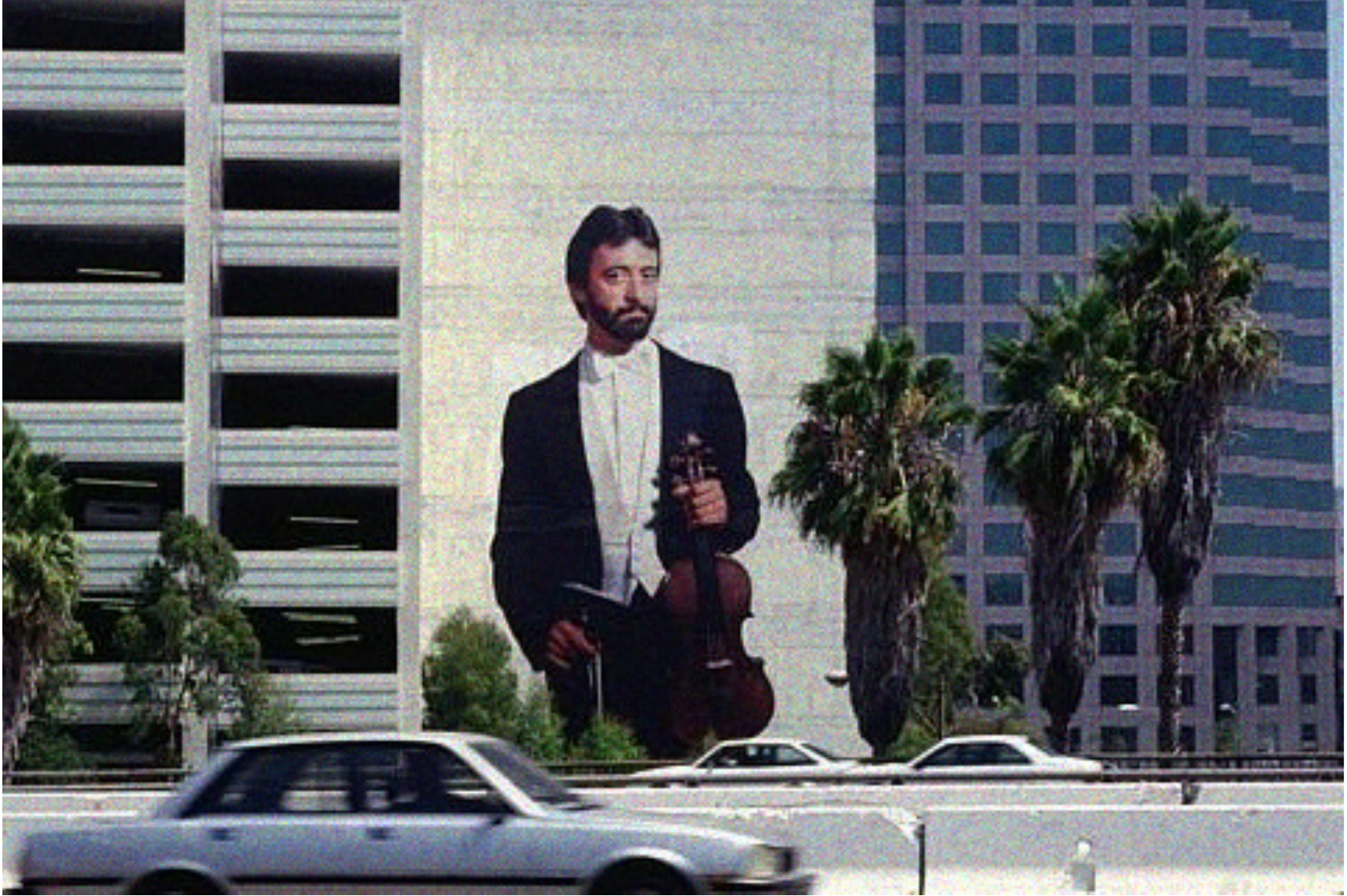} &
\includegraphics[width=\ww, height=\hh]{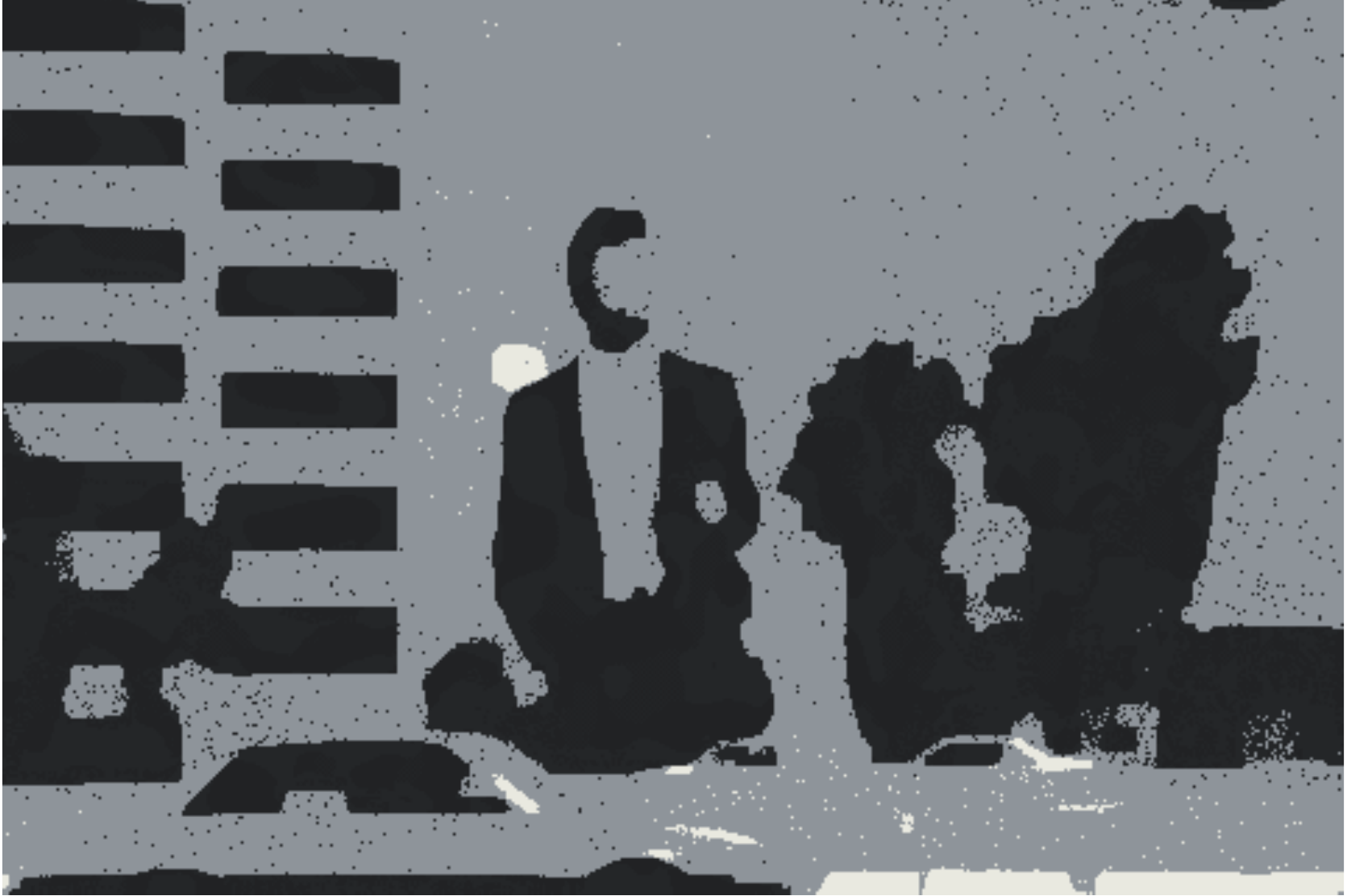} &
\includegraphics[width=\ww, height=\hh]{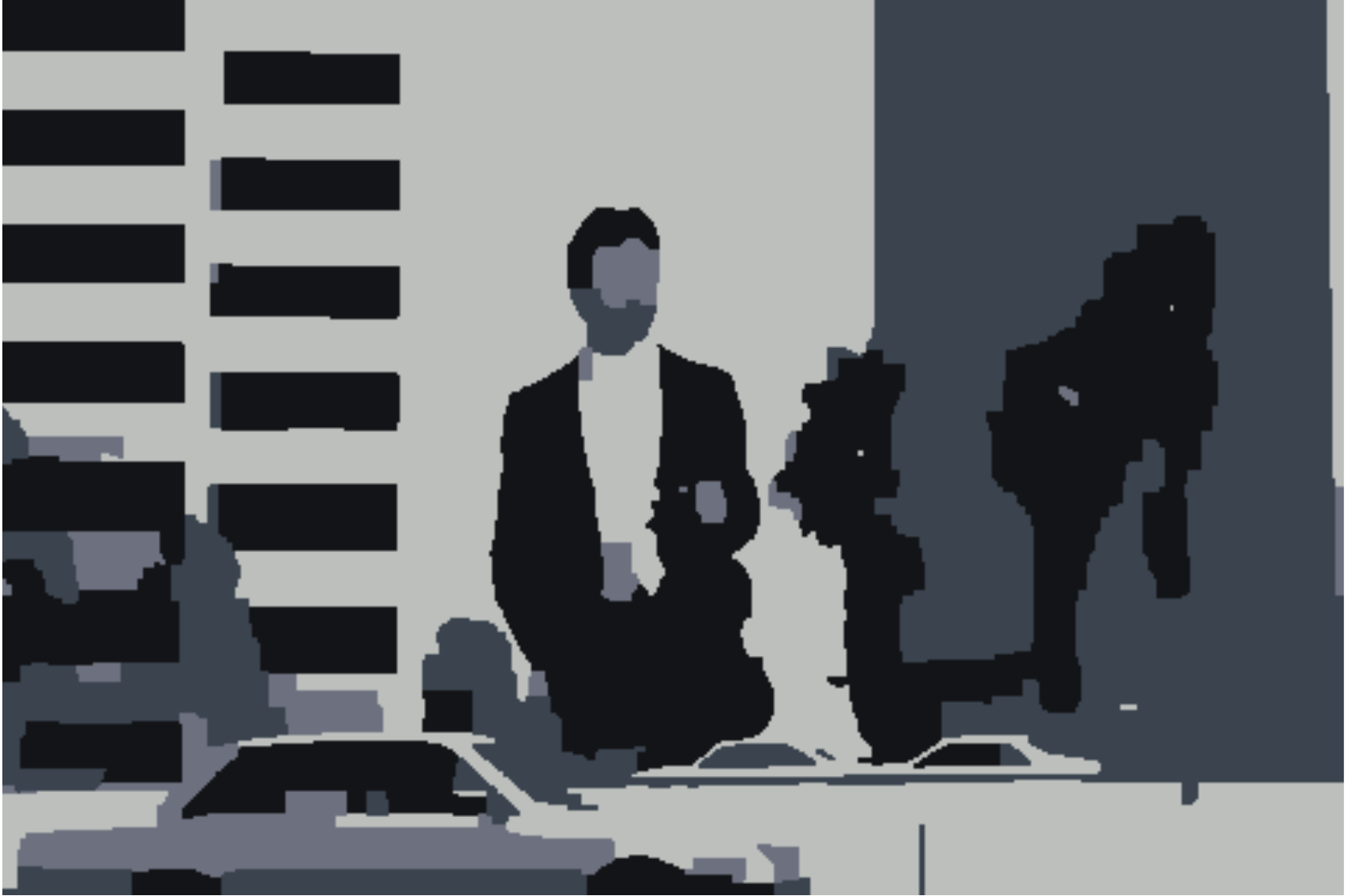} &
\includegraphics[width=\ww, height=\hh]{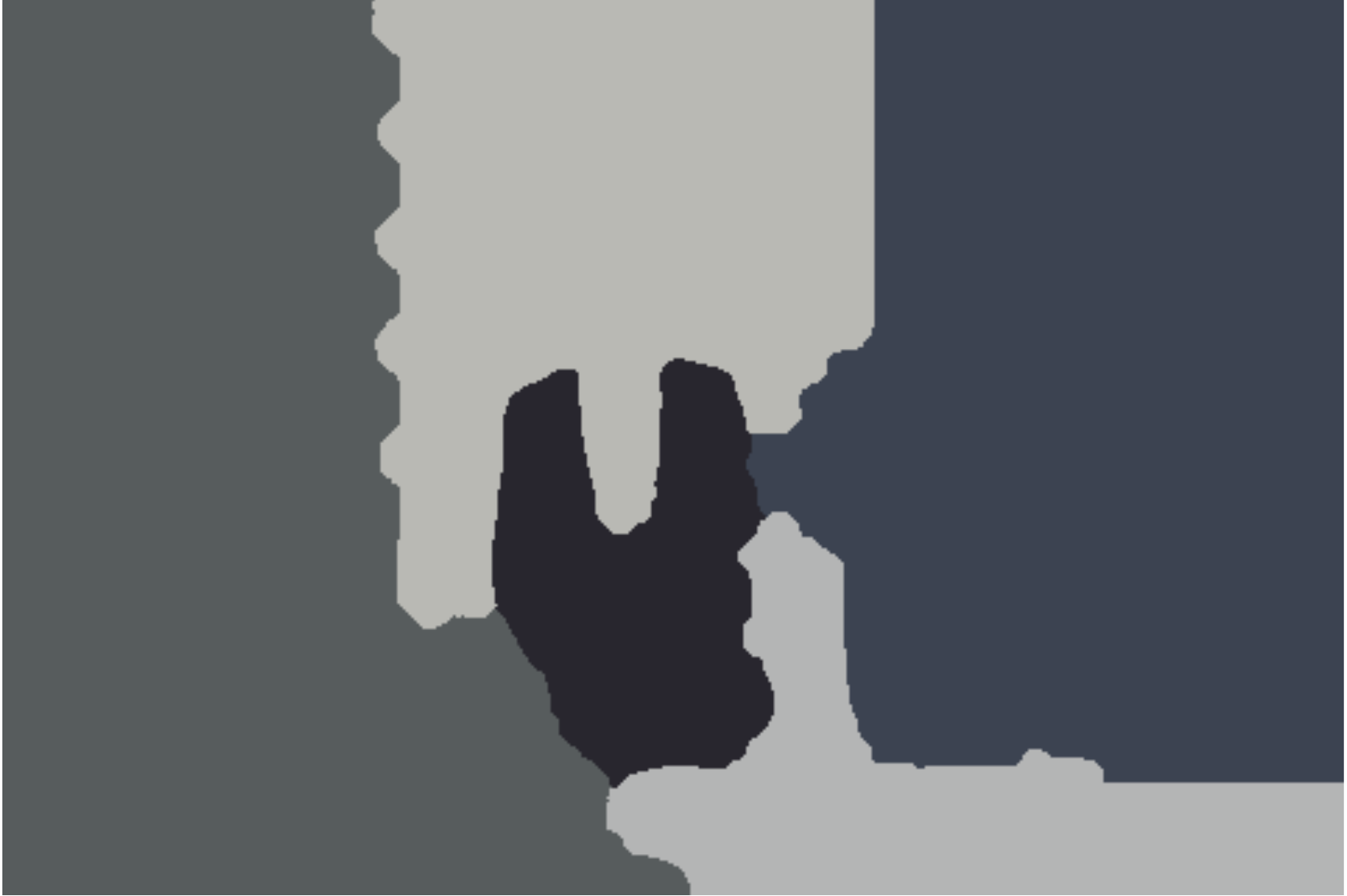} &
\includegraphics[width=\ww, height=\hh]{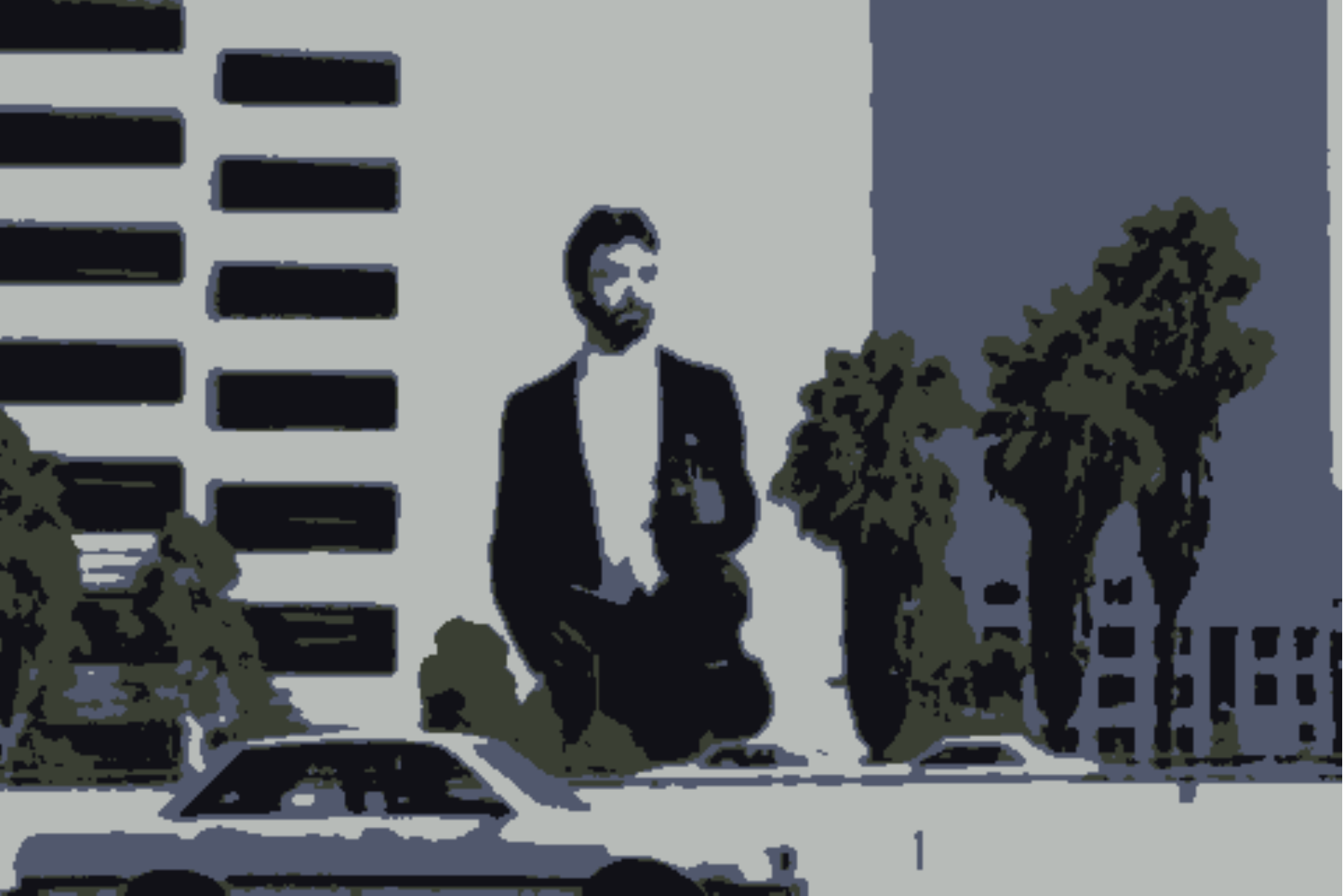} \\
{\small (A) Noisy image} &
{\small (A1) Method \cite{LNZS10}} &
{\small (A2) Method \cite{PCCB09}} &
{\small (A3) Method \cite{SW14}} &
{\small (A4) Ours } \\
\includegraphics[width=\ww, height=\hh]{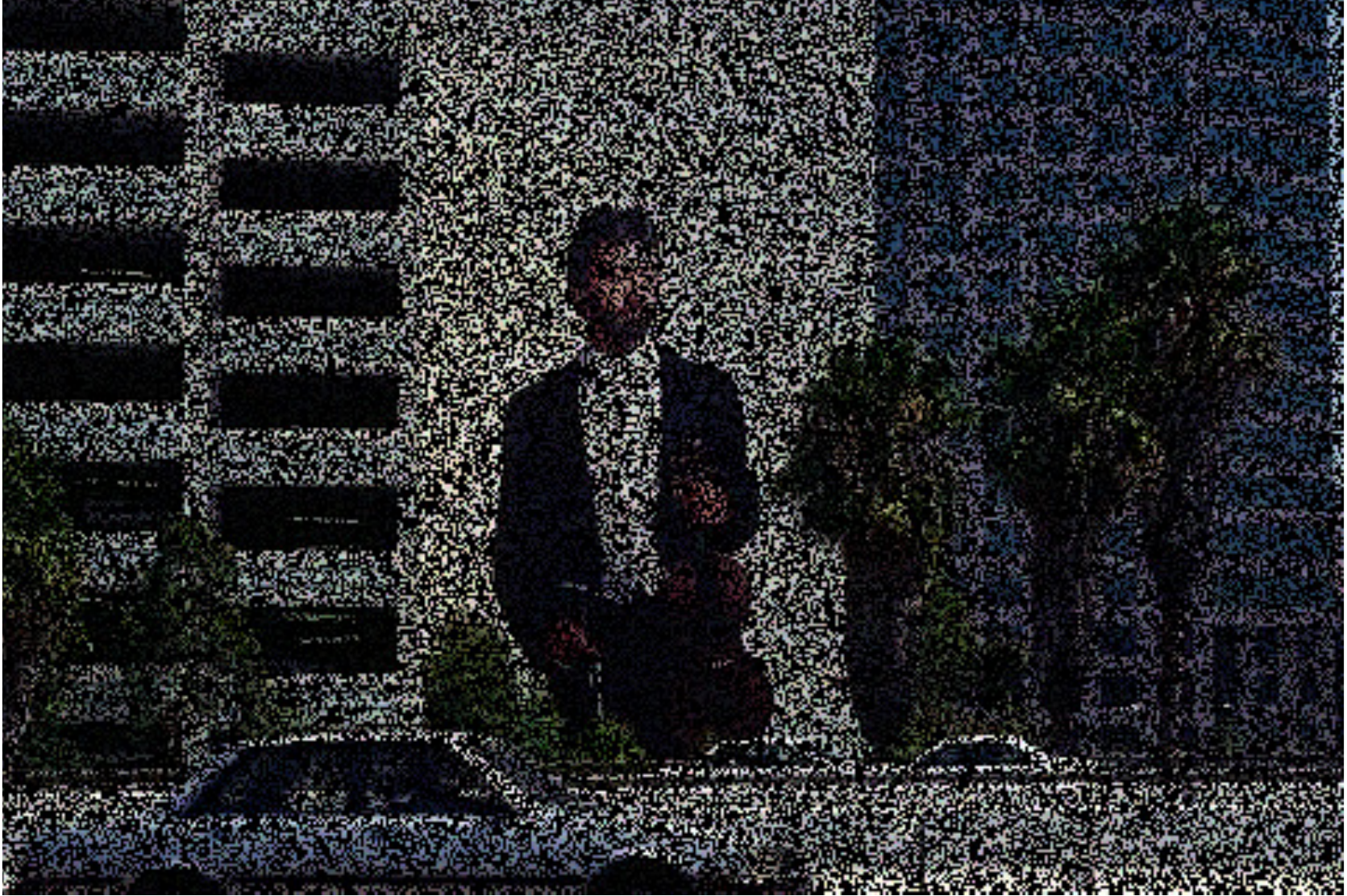} &
\includegraphics[width=\ww, height=\hh]{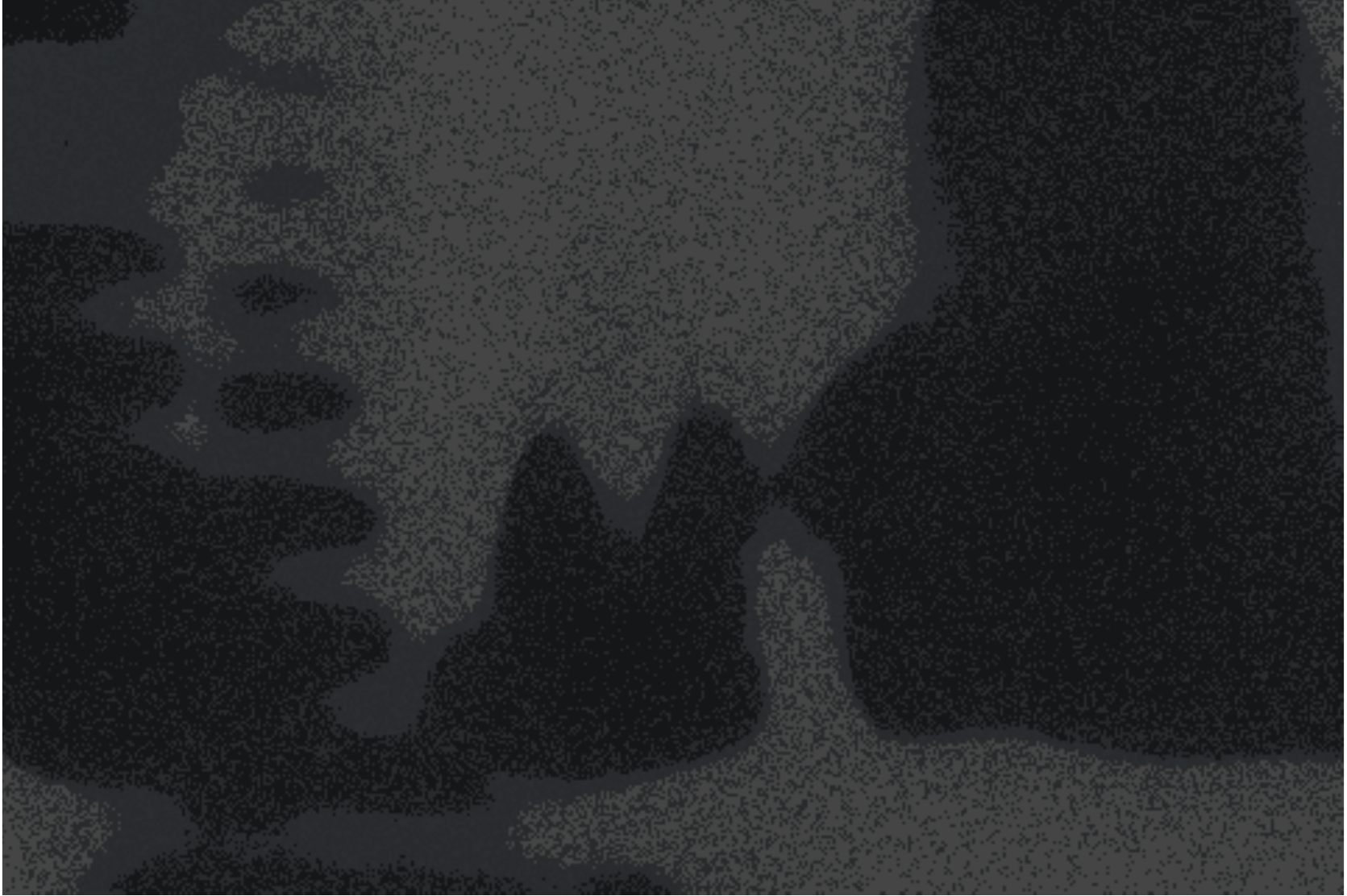} &
\includegraphics[width=\ww, height=\hh]{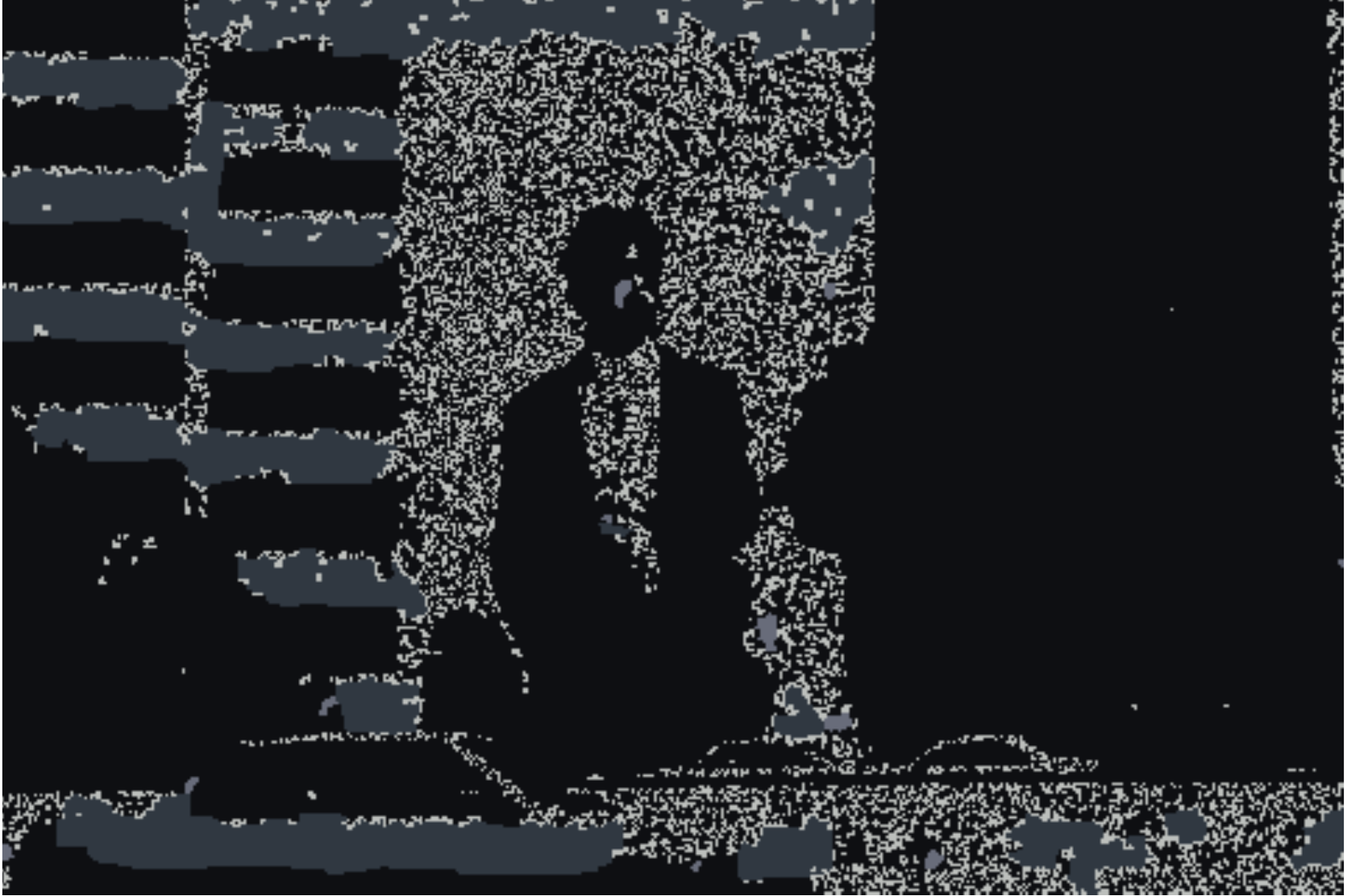} &
\includegraphics[width=\ww, height=\hh]{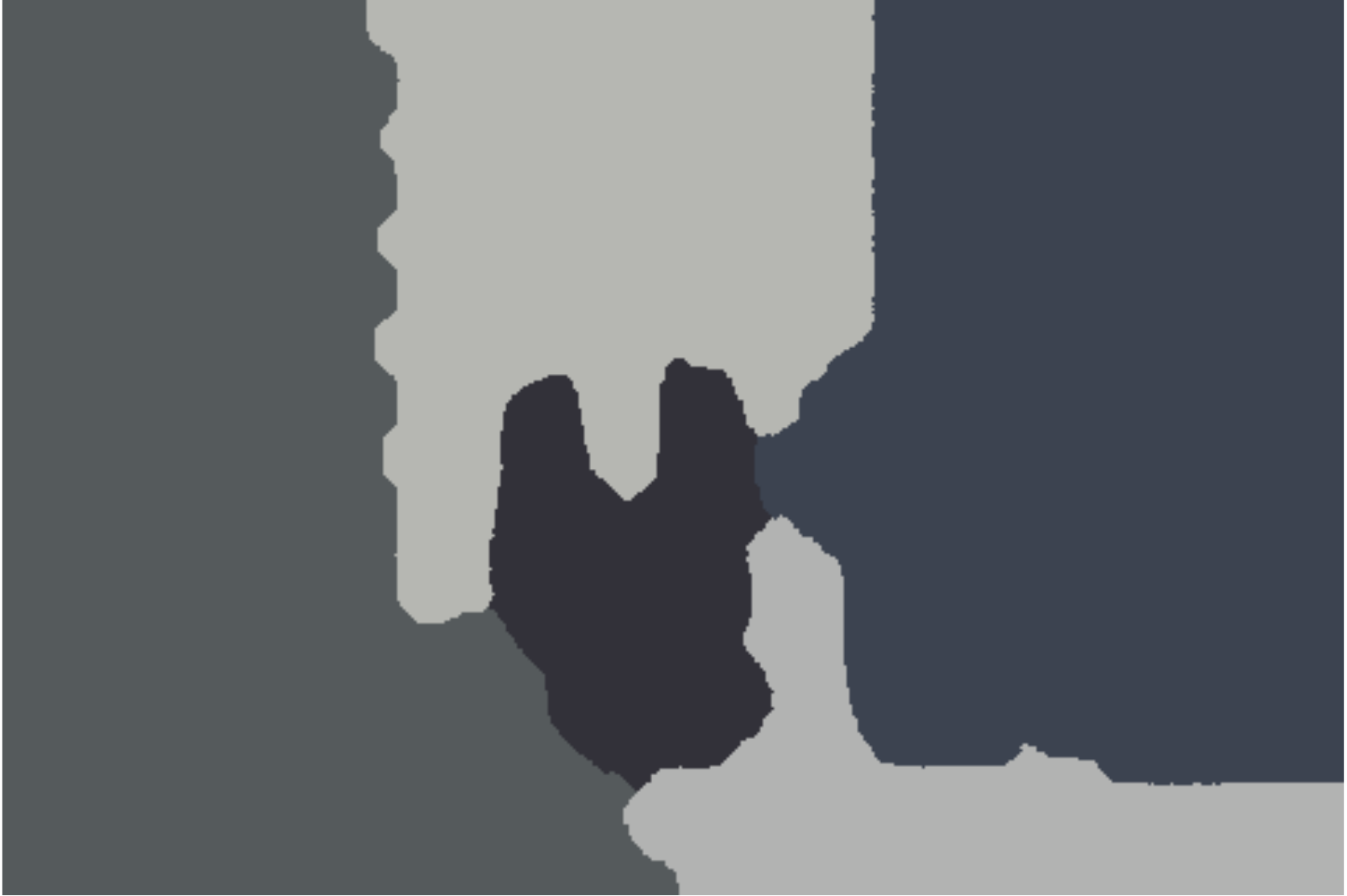} &
\includegraphics[width=\ww, height=\hh]{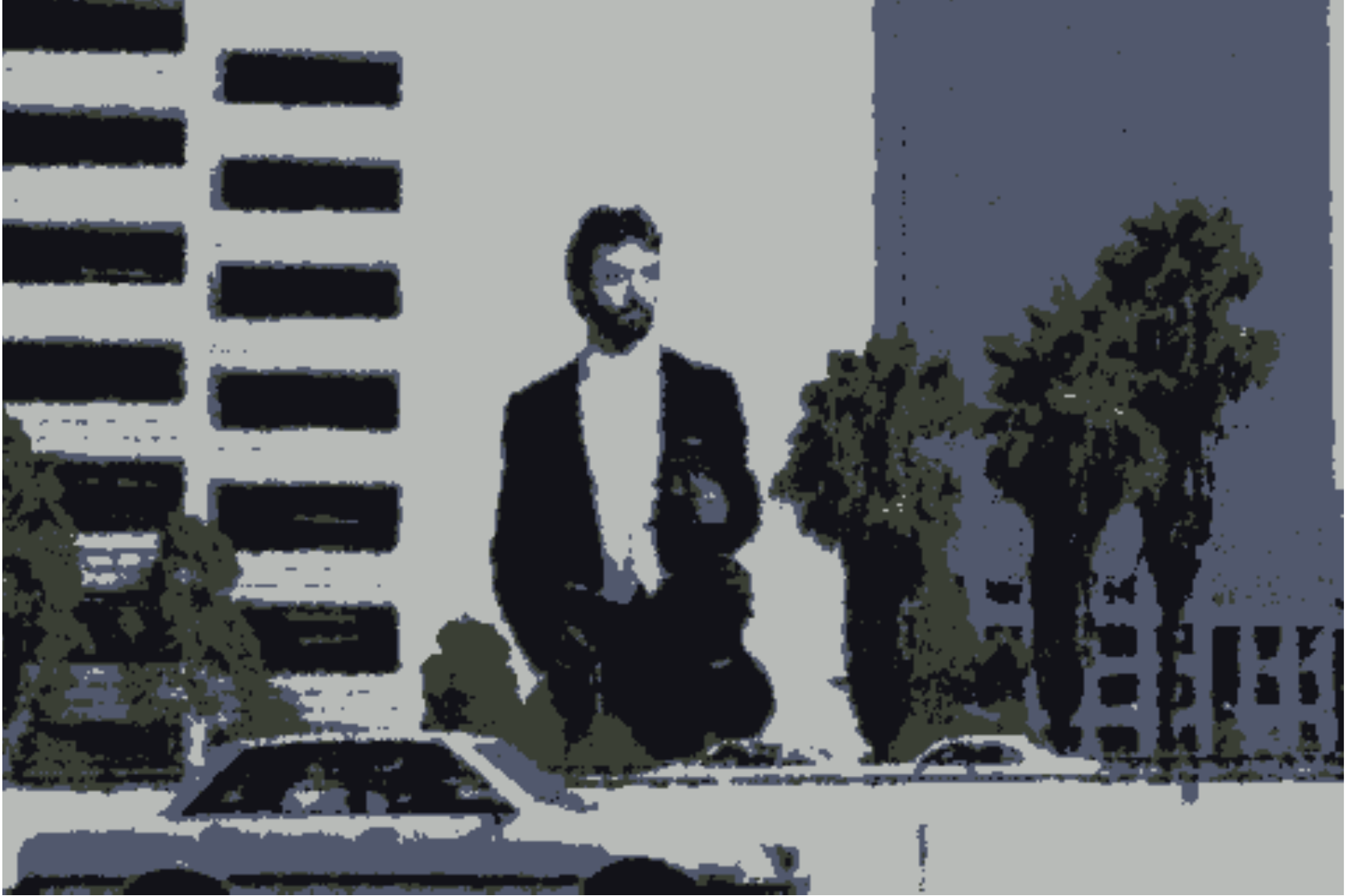} \\
{\small (B) Information } &
{\small (B1) Method \cite{LNZS10}} &
{\small (B2) Method \cite{PCCB09}} &
{\small (B3) Method \cite{SW14}} &
{\small (B4) Ours } \vspace{-0.05in} \\
{\small loss + noise} & & & & \\
\includegraphics[width=\ww, height=\hh]{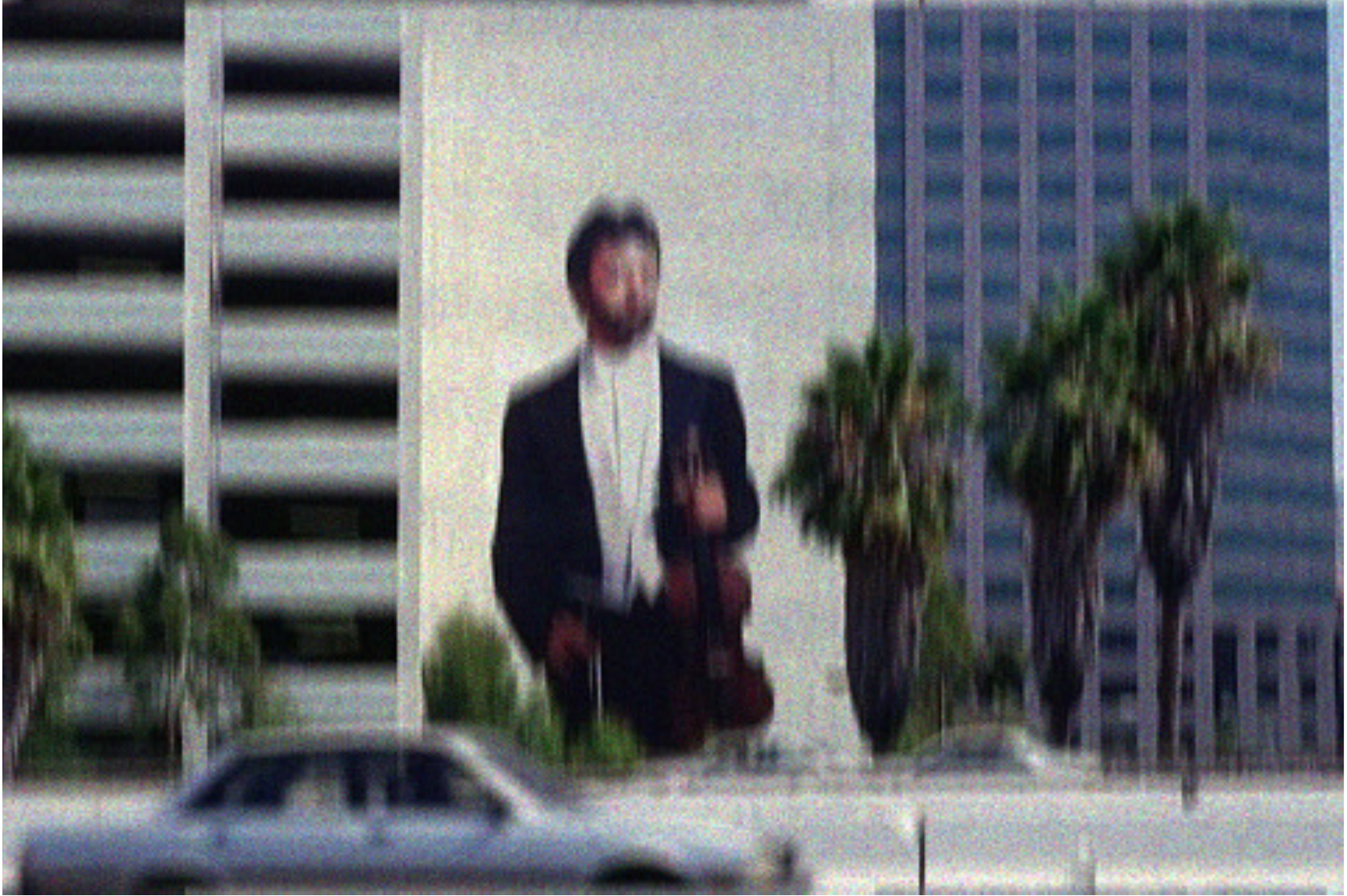} &
\includegraphics[width=\ww, height=\hh]{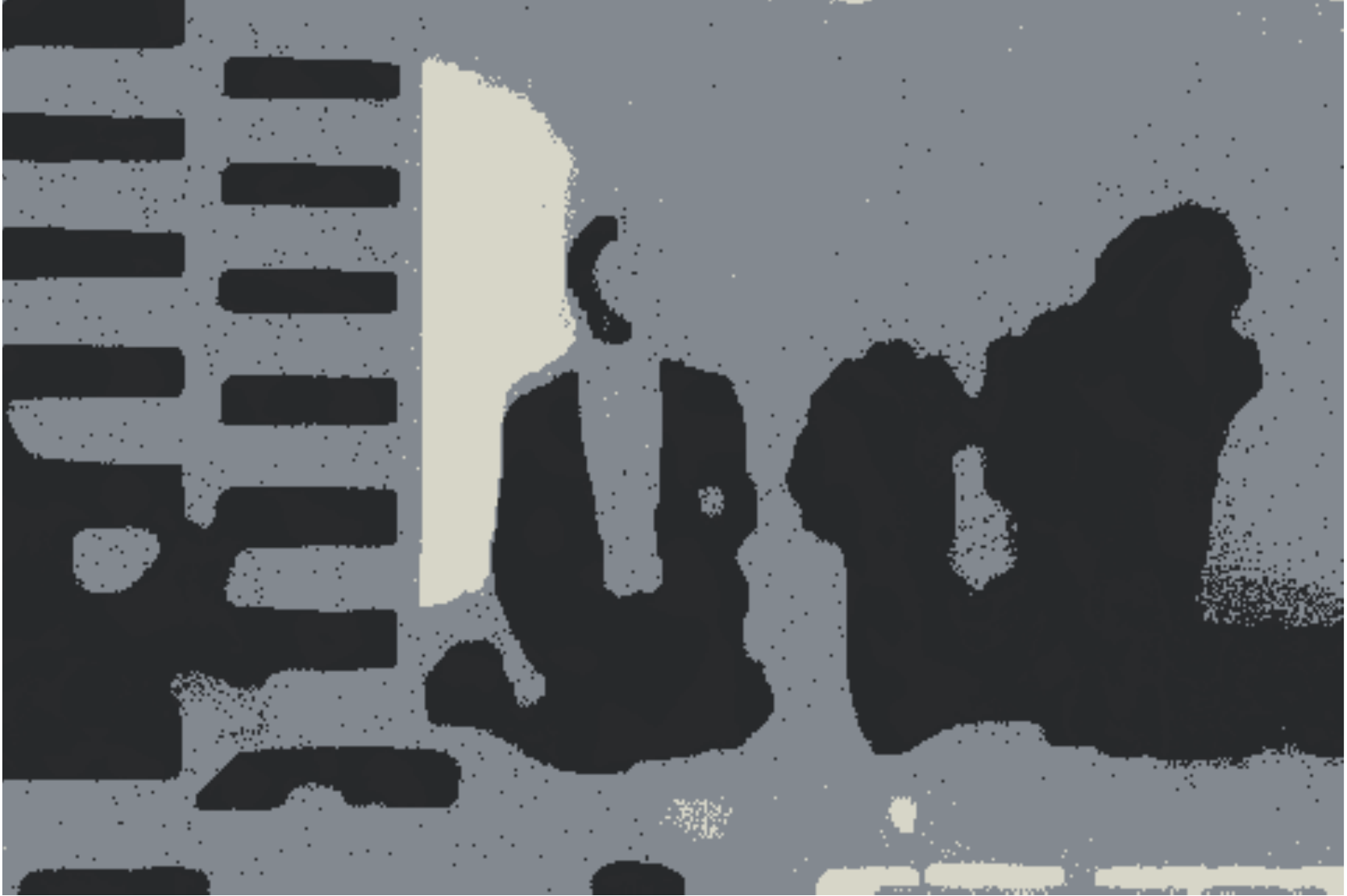} &
\includegraphics[width=\ww, height=\hh]{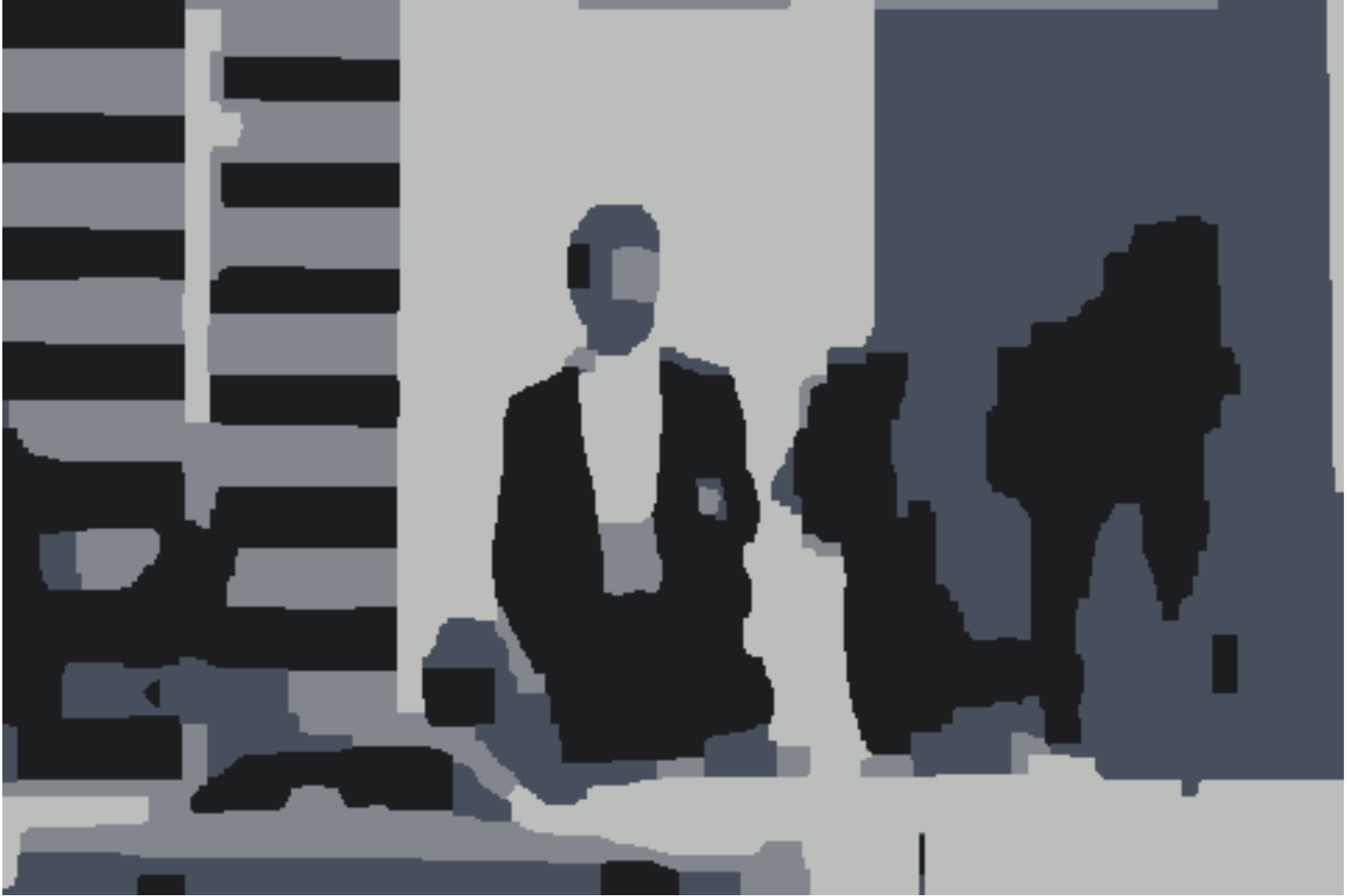} &
\includegraphics[width=\ww, height=\hh]{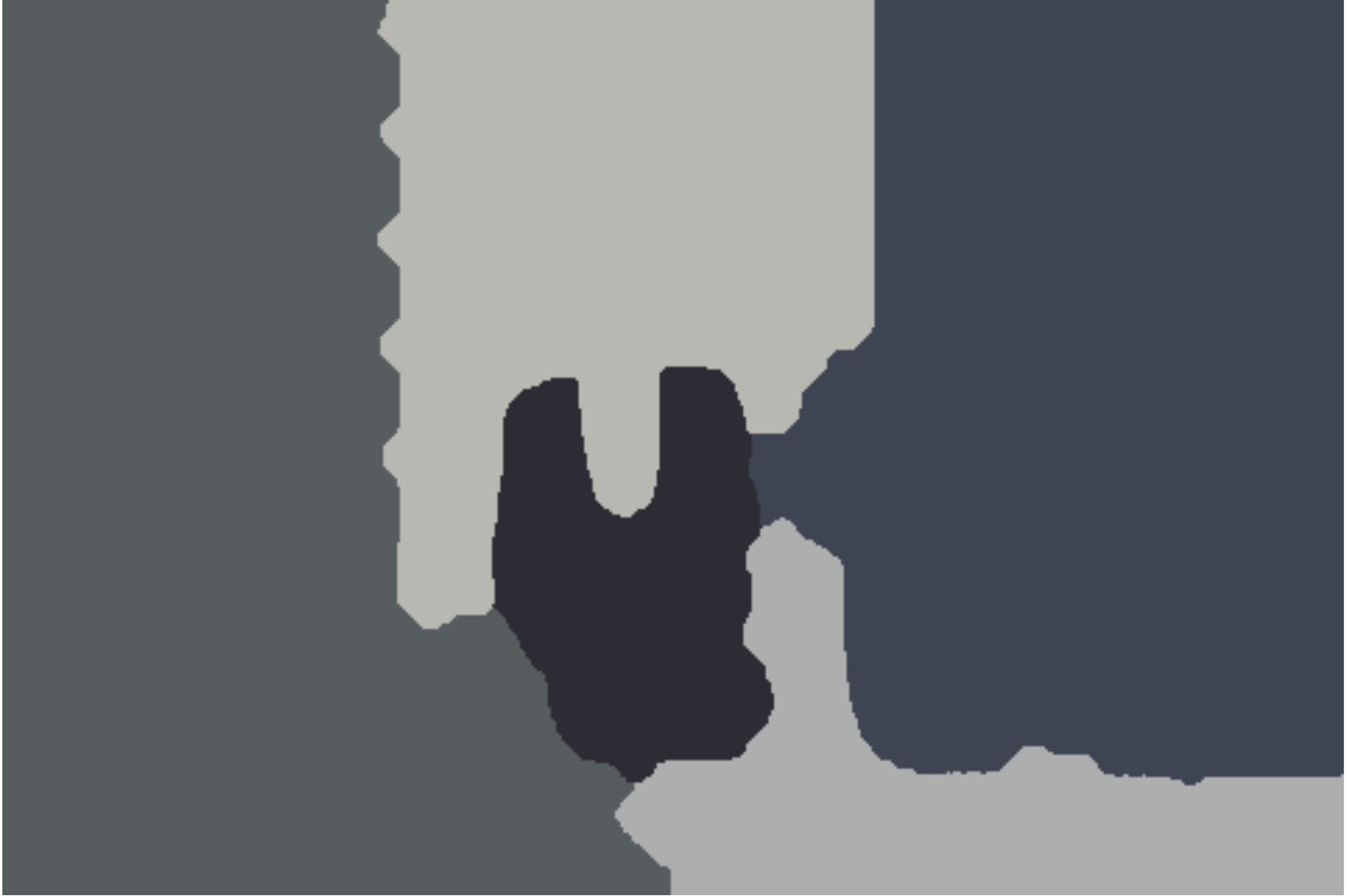} &
\includegraphics[width=\ww, height=\hh]{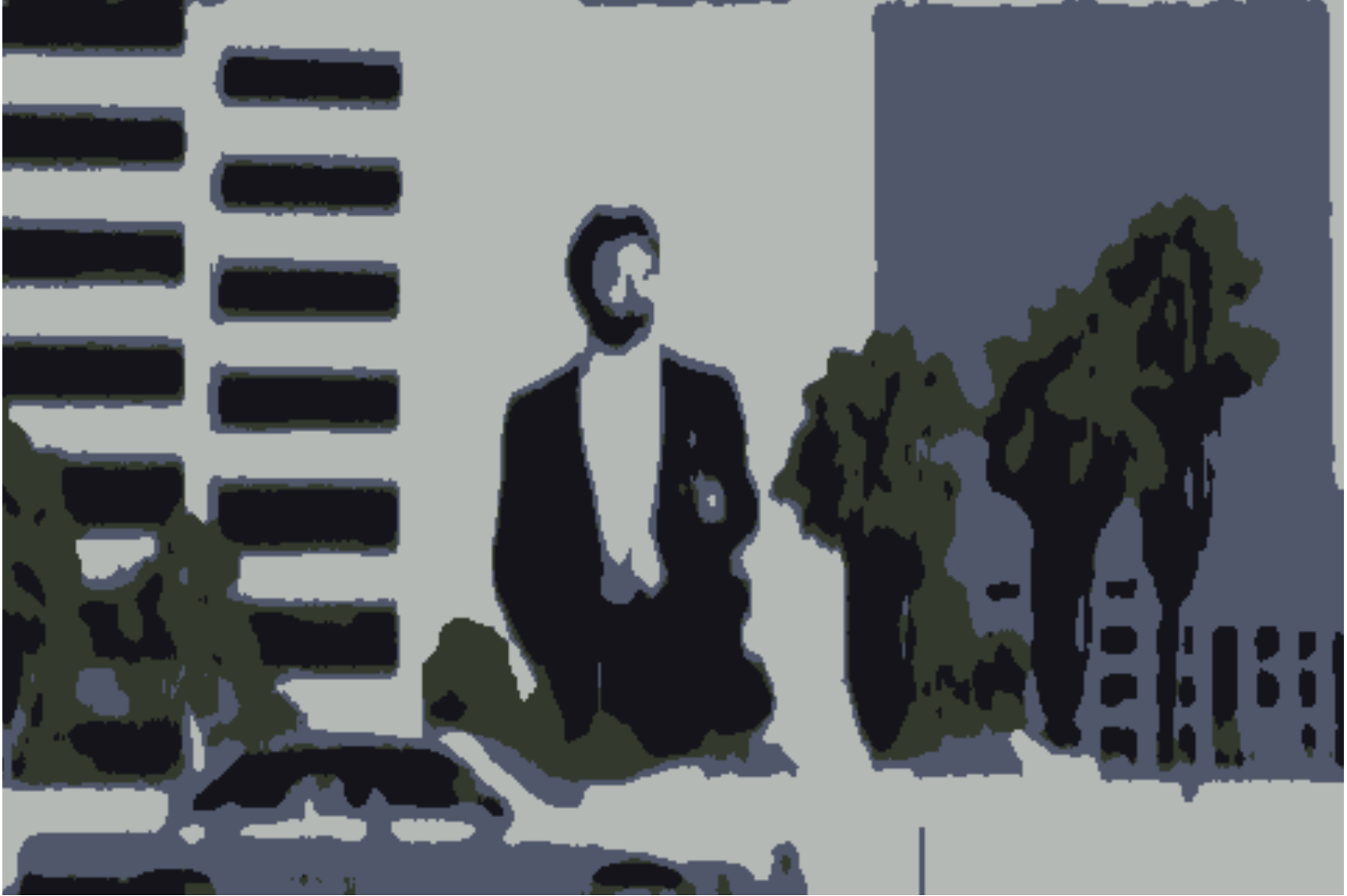} \\
{\small (C) Blur + noise} &
{\small (C1) Method \cite{LNZS10}} &
{\small (C2) Method \cite{PCCB09}} &
{\small (C3) Method \cite{SW14}} &
{\small (C4) Ours }
\end{tabular}
\end{center}
\caption{Four-phase man segmentation (size: $321\times 481$).
(A): Given Poisson noisy image;
(B): Given Poisson noisy image with $60\%$ information loss;
(C): Given blurry image with Poisson noise;
(A1-A4), (B1-B4) and (C1-C4): Results of methods \cite{LNZS10},
\cite{PCCB09}, \cite{SW14}, and our SLaT on (A), (B) and (C), respectively.
}\label{fourphase-color-man}
\end{figure*}

In this section, we compare our method with the three competing
methods for 7 real-world color images in two-phase and multiphase
segmentations, see Figs. \ref{twophase-color-rose}--\ref{fourphase-color-man}.
We see from the figures that our method is far superior than
those by the competing methods. The timing of the methods given in
Table \ref{tab:time} shows that our method in most of the cases gives the least timing.
Again, we emphasize that our method is easily parallelizable.
 All presented experiments clearly show that all goals
listed in Introduction are fulfilled.

\section{Conclusions}\label{sec:conclusions}
In this paper we proposed a three-stage image segmentation method for color images.
At the first stage of our method, a convex variational model is used in parallel
on each channel of the color image to obtain a smooth color image.
Then in the second stage we transform this smooth image to a secondary color space so as
to obtain additional information of the image in the less-correlated
color space. In the last stage, multichannel thresholding is used to
threshold the combined image from the two color spaces.
The new three-stage method, named SLaT for Smoothing, Lifting and Thresholding,
has the ability to segment images corrupted by noise,
blur, or when some pixel information is lost. Experimental results
on RGB images coupled with Lab secondary color space demonstrate that
our method gives much better segmentation results than some state-of-the-art segmentation models
both in terms of quality and CPU time cost. Our future work includes
finding an automatical way to determine $\lambda$ and possibly an
improved model \eqref{model-1st-extend-n} that can better promote geometry.


\section*{Acknowledgment}
The authors thank G. Steidl and M. Bertalm\'{\i}o for constructive discussions.


\section*{Appendix I: Proof of Theorem \ref{thm-eu}}

First consider $\Phi(f_i, g_i) = (f_i-{\cal A} g_i)^2$.
Using \eqref{no}, $E(g_i)$ defined in (\ref{model-1st-extend-n}) can be rewritten as
\begin{equation}
E(g_i)=\frac{\lambda}{2} \int_{\Omega}(\omega_i\cdot f_i -(\omega_i{\cal A} )g_i)^2 dx
+  \frac{\mu}{2} \int_{\Omega}|\nabla g_i|^2 dx
+ \int_{\Omega} |\nabla g_i| dx.
\end{equation}
Noticing that $\omega_i\cdot f_i\in L^2(\Omega)$ and that $(\omega_i{\cal A}) :L^2(\Omega)\rightarrow L^2(\Omega)$
is linear and bounded, the statement follows from \cite[Theorem 2.4.]{CCZ13}.

Next consider that $\Phi(f_i, g_i) = {\cal A}g_i - f_i \log ({\cal A}g_i)$. Then
\begin{equation}
E(g_i)=\frac{\lambda}{2} \int_{\Omega} \omega_i\cdot ({\cal A}g_i) - (\omega_i\cdot f_i) \log ({\cal A}g_i) dx
+  \frac{\mu}{2} \| \nabla g_i\|^2_{L^2(\Omega)} + \| \nabla g_i\|_{L^2(\Omega)}.
\end{equation}
I) Existence: Since $W^{1,2}(\Omega)$ is a reflective Banach space and $E(g_i)$ is convex lower semi-continuous, by \cite[Proposition 1.2]{ET}
we need to prove that $E(g_i)$
is coercive on $W^{1,2}(\Omega)$, i.e. that
$E(g_i)\rightarrow+\infty$ as $\|g_i\|_{ W^{1,2}(\Omega) }:=\|g_i\|_{L^2(\Omega)} + \|\nabla g_i\|_{L^2(\Omega)}\rightarrow+\infty$.

The function ${\cal A} g_i\mapsto ({\cal A} g_i - f \log {\cal A} g_i)$ is strictly convex with
a minimizer pointwisely satisfying ${\cal A} g_i = f \in[0,1]$, hence $\Phi(f_i, g_i) \geq0$.
Thus $\| \nabla g_i\|_{L^2(\Omega)} $ is upper bounded by $E(g_i)>0$ for any $g_i\in W^{1,2}(\Omega)$ and $f\neq0$.
Using the Poincar\'e inequality, see \cite{E}, we have:
\begin{equation}\label{boun_g_gt}
\|g_i-g_{i_{\Omega}}\|_{L^2(\Omega)} \leq C_1\|\nabla g_i\|_{L^2(\Omega)}
\leq C_1 E(g_i),
\end{equation}
where $C_1>0$ is a constant and $g_{i_{\Omega}}=\frac{1}{|\Omega|}\int_{\Omega} g_idx$.
Let us set $C_2:= \left(1-\frac{1}{e} \| f_i\|_\infty \right)$.
We have $C_2>0$ because $ \| f_i\|_\infty\leq 1$.
Recall the fact that $\frac{t}{e}\ge \log t$ for any $t>0$ which
can be easily verified by showing that $t/e-\log t$ is convex
for $t>0$ with minimum at $e$. Hence we have
\begin{equation*}
 \omega_i\cdot \Phi(f_i, g_i) \geq(\omega_i {\cal A}) \;g_i- \frac{1}{e} (\omega_i\cdot f_i) {\cal A}g_i
\\= \omega_i\cdot (1-\frac{1}{e} f_i) {\cal A}g_i \geq C_2 \; (\omega_i {\cal A}) \;g_i
\end{equation*}
which should be understood pointwisely. Hence,
\begin{equation} \label{boun_gt}
\|(\omega_i {\cal A}) \;g_i\|_{L^1(\Omega)} 
\leq\frac{2}{C_2\lambda}E(g_i). \end{equation}
Let $C_3:=\|(\omega_i{\cal A})\mathbf{1}\|_{L^1(\Omega)} $ where $\mathbf{1}(x)=1$ for any $x\in\Omega$.
Using $\mathrm{Ker} (\nabla) =\{u\in L^2(\Omega) : u=c\;1~ \mathrm{a.e.} \ {\rm for} \ x\in\Omega, c\in \mathbb{R}\} $
together with the assumption
${\rm Ker} (\omega_i {\cal A} )\bigcap {\rm Ker} (\nabla) = \{0 \}$
one has $C_3>0$. Using \eqref{boun_gt} together with the fact
that $g_{i_{\Omega}} >0$ yields

\begin{equation*}
 |g_{i_{\Omega}} | \; \|(\omega_i {\cal A})\mathbf{1}\|_{L^1(\Omega)} = |g_{i_{\Omega}} | \; C_3\\=
 \|\omega_i\cdot ({\cal A}\mathbf{1} g_{i_{\Omega}} ) \|_{L^1(\Omega)}\leq
\frac{2}{C_2\lambda} E(g_i),
\end{equation*}
and thus
\[
|g_{i_{\Omega}} | \leq \frac{2}{C_2C_3\lambda} E(g_i).
\]
Applying the triangular inequality in \eqref{boun_g_gt} gives
$\|g_i\|_{L^2(\Omega)}-|g_{i_{\Omega}}| \leq C_1\|\nabla g_i\|_{L^2(\Omega)}$.
Hence
\[ \|g_i\|_{L^2(\Omega)}\leq |g_{i_{\Omega}}| + C_1\|\nabla g_i\|_{L^2(\Omega)}
\leq \left(\frac{2}{C_2C_3\lambda} + 1 \right) E(g_i).
\]
Comparing with \eqref{boun_g_gt} yet again shows that we have obtained
\[\|g_i\|_{ W^{1,2}(\Omega) }=\|g_i\|_{L^2(\Omega)} + \|\nabla g_i\|_{L^2(\Omega)}
\leq \left(\frac{2}{C_2C_3\lambda} + 1 +C_1\right) E(g_i).
\]
Therefore, $E$ is coercive.

II) Uniqueness: Suppose $\bar g_{i_1}$ and $\bar g_{i_2}$ are both minimizers of $E(g_i)$.
The convexity of $E$ and the strict convexity of ${\cal A} g_i\mapsto({\cal A} g_i - f \log {\cal A} g_i)$
entail ${\cal A} \bar g_{i_1} = {\cal A} \bar g_{i_2}$ on $\Omega_0^i$ and $\nabla \bar g_{i_1} = \nabla \bar g_{i_2}$.
Further, the assumption on ${\rm Ker} (\omega_i {\cal A} )\bigcap {\rm Ker} (\nabla) $ shows that
$\bar g_{i_1}=\bar g_{i_2}$.

\ifCLASSOPTIONcaptionsoff
 \newpage
\fi



%

\begin{IEEEbiography}{Xiaohao Cai} received the M.S. degree in mathematics from Zhejiang
University, China, in 2008, and the Ph.D. degree in mathematics from
The Chinese University of Hong Kong, Hong Kong in 2012.

He is currently a postdoctoral researcher in the Department of Plant
Sciences, and Department of Applied Mathematics and Theoretical
Physics, University of Cambridge. His research interests include image
processing, numerical analysis and their applications in processing of
digital image, video, biomedical imaging, remote sensing data, just to
name a few.
\end{IEEEbiography}
%
%
%
\begin{IEEEbiography}{Raymond Chan} was born in 1958 in Hong Kong.
He received his B.Sc. degree in Mathematics from
the Chinese University of Hong Kong and his M.Sc.
and Ph.D. degree in Applied Mathematics from New
York University. He is now a Chair Professor in the
Department of Mathematics, The Chinese University
of Hong Kong. His research interests include numerical linear algebra and image processing problem.
\end{IEEEbiography}
\begin{IEEEbiography}{Mila Nikolova} received the Ph.D.
degree from the Universit\'{e} de Paris-Sud, France, in
1995.

She got a Habilitation to direct research in 2006.
Currently, she is a Research Director with the
National Center for Scientific Research (CNRS),
Cachan Cedex, France. She performs her duty as a
full-time Researcher at the Centre de Math\'{e}matiques
et de Leurs Applications (CMLA), the 8536 UMR
of CNRS, in the Department of Mathematics of ENS
de Cachan, France. Her research interests are in
mathematical Image and signal reconstruction, inverse problems, regularization
and variational methods and the properties of their solutions, and scientific
computing.

\end{IEEEbiography}
\begin{IEEEbiography}{Tieyong Zeng} received the B.S. degree from Peking
University, Beijing, China, in 2000, the M.S. degree
from Ecole Polytechnique, Paris, France, in 2004,
and the Ph.D. degree from the University of Paris
XIII, Paris, in 2007. Before joining Hong Kong Baptist University, Kowloon Tong, China,
as an Assistant Professor, he was a Post-Doctoral Researcher
with the Centre de Math\'{e}matiques et de Leurs
Applications, ENS de Cachan, Cachan, France. His
research interests are image processing, statistical
learning, and scientific computing.

\end{IEEEbiography}




\end{document}